\documentclass[10pt,journal,compsoc]{IEEEtran}
\usepackage{epsfig}
\usepackage{amsmath}
\usepackage{amssymb}
\usepackage{algorithm}
\usepackage{algorithmic}
\usepackage{multirow}
\usepackage{url}
\usepackage[tight,footnotesize]{subfigure}
\usepackage{amsthm}
\usepackage{mathrsfs}
\usepackage[pagebackref=false,breaklinks=true,letterpaper=true,colorlinks,citecolor=blue,bookmarks=true]{hyperref}
\usepackage{enumerate}

\newtheorem{definition}{Definition}
\newtheorem{theorem}{Theorem}
\newtheorem{lemma}{Lemma}

\newtheorem{corollary}{Corollary}
\newtheorem{assumption}{Assumption}

\begin{document}

\title{VR-SGD: A Simple Stochastic Variance Reduction Method for Machine Learning}

\author{Fanhua~Shang,~\IEEEmembership{Member,~IEEE,}
        Kaiwen~Zhou,
        Hongying~Liu,
        James~Cheng,
        Ivor~W.~Tsang,\\
        Lijun~Zhang,~\IEEEmembership{Member,~IEEE,}
        Dacheng~Tao,~\IEEEmembership{Fellow,~IEEE,}
        and~Licheng~Jiao,~\IEEEmembership{Fellow,~IEEE}
\IEEEcompsocitemizethanks{\IEEEcompsocthanksitem F.\ Shang, H.\ Liu (Corresponding author) and L.\ Jiao are with the Key Laboratory of Intelligent Perception and Image Understanding of Ministry of Education, School of Artificial Intelligence, Xidian University, China. E-mails: \{fhshang, hyliu\}@xidian.edu.cn, lchjiao@mail.xidian.edu.cn.\protect
\IEEEcompsocthanksitem K.\ Zhou and J.\ Cheng are with the Department of Computer Science and Engineering, The Chinese University of Hong Kong, Hong Kong. E-mails: \{kwzhou, jcheng\}@cse.cuhk.edu.hk.\protect
\IEEEcompsocthanksitem I.W.\ Tsang is with the Centre for Artificial Intelligence, University of Technology Sydney, Ultimo, NSW 2007, Australia. E-mail: Ivor.Tsang@uts.edu.au.\protect
\IEEEcompsocthanksitem L.\ Zhang is with the National Key Laboratory for Novel Software Technology, Nanjing University, Nanjing 210023, China. E-mail: zhanglj@lamda.nju.edu.cn.\protect
\IEEEcompsocthanksitem D.\ Tao is with the UBTECH Sydney Artificial Intelligence Centre and the School of Information Technologies, the Faculty of Engineering and Information Technologies, the University of Sydney, 6 Cleveland St, Darlington, NSW 2008, Australia. E-mail: dacheng.tao@sydney.edu.au.}
\thanks{Manuscript received March 22, 2018.}}

\markboth{IEEE TRANSACTIONS ON KNOWLEDGE AND DATA ENGINEERING}
{Shell \MakeLowercase{\textit{et al.}}: Bare Demo of IEEEtran.cls for Computer Society Journals}

\IEEEtitleabstractindextext{
\begin{abstract}
In this paper, we propose a simple variant of the original SVRG, called variance reduced stochastic gradient descent (VR-SGD). Unlike the choices of snapshot and starting points in SVRG and its proximal variant, Prox-SVRG, the two vectors of VR-SGD are set to the average and last iterate of the previous epoch, respectively. The settings allow us to use much larger learning rates, and also make our convergence analysis more challenging. We also design two different update rules for smooth and non-smooth objective functions, respectively, which means that VR-SGD can tackle non-smooth and/or non-strongly convex problems directly without any reduction techniques. Moreover, we analyze the convergence properties of VR-SGD for strongly convex problems, which show that VR-SGD attains linear convergence. Different from most algorithms that have no convergence guarantees for non-strongly convex problems, we also provide the convergence guarantees of VR-SGD for this case, and empirically verify that VR-SGD with varying learning rates achieves similar performance to its momentum accelerated variant that has the optimal convergence rate $\mathcal{O}(1/T^2)$. Finally, we apply VR-SGD to solve various machine learning problems, such as convex and non-convex empirical risk minimization, and leading eigenvalue computation. Experimental results show that VR-SGD converges significantly faster than SVRG and Prox-SVRG, and usually outperforms state-of-the-art accelerated methods, e.g., Katyusha.
\end{abstract}

}
\maketitle

\section{Introduction}
\IEEEPARstart{I}{n} this paper, we focus on the following composite optimization problem:
\begin{equation}\label{equ01}
\min_{x\in\mathbb{R}^{d}}\; F(x)\stackrel{\rm{def}}{=}\frac{1}{n}\sum^{n}_{i=1}f_{i}(x)+g(x)
\end{equation}
where $f(x)\!=\!\frac{1}{n}\!\sum^{n}_{i=1}\!f_{i}(x)$, $f_{i}(x)\!:\!\mathbb{R}^{d}\!\rightarrow\!\mathbb{R},\,i\!=\!1,\ldots,n$ are the smooth functions, and $g(x)$ is a relatively simple (but possibly non-differentiable) convex function (referred to as a regularizer). The formulation (\ref{equ01}) arises in many places in machine learning, signal processing, data science, statistics and operations research, such as \emph{regularized empirical risk minimization} (ERM). For instance, one popular choice of the component function $f_{i}(\cdot)$ in binary classification problems is the logistic loss, i.e., $f_{i}(x)\!=\!\log(1+\exp(-b_{i}a^{T}_{i}x))$, where $\{(a_{1},b_{1}),\ldots,(a_{n},b_{n})\}$ is a collection of training examples, and $b_{i}\!\!\in\!\{\pm1\}$. Some popular choices for the regularizer include the $\ell_{2}$-norm regularizer (i.e., $g(x)\!=\!(\lambda/2)\|x\|^{2}$), the $\ell_{1}$-norm regularizer (i.e., $g(x)\!=\!\lambda\|x\|_{1}$), and the elastic-net regularizer (i.e., $g(x)\!=\!(\lambda_{1}/2)\|x\|^{2}\!+\!\lambda_{2}\|x\|_{1}$). Some other applications include deep neural networks \cite{johnson:svrg,krizhevsky:deep,zhang:easgd,zhu:vrnc,reddi:svrnc}, group Lasso \cite{liu:avrrg}, sparse learning and coding \cite{li:svrg,zhang:sl,qu:svrg,paquette:catalyst}, non-negative matrix factorization \cite{kasai:nmf}, phase retrieval \cite{duchi:ssgd}, matrix completion \cite{recht:psgd,wang:svrgd}, conditional random fields \cite{schmidt:crf}, generalized eigen-decomposition and canonical correlation analysis \cite{zhu:cca}, and eigenvector computation \cite{shamir:pca,garber:svd} such as principal component analysis (PCA) and singular value decomposition (SVD).

\subsection{Stochastic Gradient Descent}
We are especially interested in developing efficient algorithms to solve Problem (\ref{equ01}) involving the sum of a large number of component functions. The standard and effective method for solving (\ref{equ01}) is the (proximal) gradient descent (GD) method, including Nesterov's accelerated gradient descent (AGD) \cite{nesterov:fast,nesterov:co} and accelerated proximal gradient (APG) \cite{teng:apg,beck:fista}. For the \emph{smooth} problem (\ref{equ01}), GD takes the following update rule: starting with $x_{0}$, and for any $k\!\geq\!0$
\begin{equation}\label{equ02}
x_{k+1}=x_{k}-\eta_{k}\!\left[\frac{1}{n}\sum^{n}_{i=1}\nabla\! f_{i}(x_{k})+\nabla\! g(x_{k})\right]
\end{equation}
where $\eta_{k}\!>\!0$ is commonly referred to as the learning rate in machine learning or step-size in optimization. When $g(\cdot)$ is \emph{non-smooth} (e.g., the $\ell_{1}$-norm regularizer), we typically introduce the following proximal operator to replace (\ref{equ02}),
\begin{equation}\label{equ03}
\!\!x_{k+\!1}\!=\textup{Prox}^{g}_{\eta_{k}}(y_{k}):=\mathop{\arg\min}_{x\in\mathbb{R}^{d}}\left\{\frac{1}{2\eta_{k}}\|x\!-\!y_{k}\|^{2}\!+\!g(x)\right\}
\end{equation}
where $y_{k}\!=\!x_{k}\!-\!(\eta_{k}/n)\sum^{n}_{i=1}\!\nabla\! f_{i}(x_{k}).$ GD has been proven to achieve linear convergence for \emph{strongly convex} problems, and both AGD and APG attain the optimal convergence rate $\mathcal{O}(1/T^2)$ for \emph{non-strongly convex} problems, where $T$ denotes the number of iterations. However, the per-iteration cost of all the batch (or deterministic) methods is $O(nd)$, which is expensive for very large $n$.

Instead of evaluating the full gradient of $f(\cdot)$ at each iteration, an efficient alternative is the stochastic (or incremental) gradient descent (SGD) method \cite{robbins:sgd}. SGD only evaluates the gradient of a single component function at each iteration, and has much \emph{lower} per-iteration cost, $O(d)$. Thus, SGD has been successfully applied to many large-scale learning problems \cite{zhang:sgd,hu:sgd,bubeck:sgd}, especially training for deep learning models \cite{krizhevsky:deep,zhang:easgd,silver:go}, and its update rule is
\begin{equation}\label{equ04}
x_{k+1}=x_{k}-\eta_{k}[\nabla\!f_{i_{k}}\!(x_{k})+\nabla\!g(x_{k})]
\end{equation}
where $\eta_{k}\!\propto\!1/\sqrt{k}$, and the index $i_{k}$ can be chosen uniformly at random from $\{1,2,\ldots,n\}$. Although the expectation of the stochastic gradient estimator $\nabla\!f_{i_{k}}\!(x_{k})$ is an \emph{unbiased} estimation for $\nabla\!f(x_{k})$, i.e., $\mathbb{E}[\nabla\!f_{i_{k}}\!(x_{k})]\!=\!\nabla\! f(x_{k})$, the variance of $\nabla\!f_{i_{k}}\!(x_{k})$ may be large due to the variance of random sampling~\cite{johnson:svrg}. Thus, stochastic gradient estimators are also called ``noisy gradients", and we need to gradually reduce its step size, which leads to slow convergence. In particular, even under the strongly convex (SC)condition, standard SGD attains a slower \emph{sub-linear} convergence rate $\mathcal{O}(1/T)$~\cite{rakhlin:sgd}.

\subsection{Accelerated SGD}
\label{sec12}
Recently, many SGD methods with \emph{variance reduction} have been proposed, such as stochastic average gradient (SAG) \cite{roux:sag}, stochastic variance reduced gradient (SVRG) \cite{johnson:svrg}, stochastic dual coordinate ascent (SDCA) \cite{shalev-Shwartz:sdca}, SAGA \cite{defazio:saga}, stochastic primal-dual coordinate (SPDC) \cite{zhang:spdc}, and their proximal variants, such as Prox-SAG \cite{schmidt:sag}, Prox-SVRG \cite{xiao:prox-svrg} and Prox-SDCA \cite{shalev-Shwartz:prox-sdca}. These accelerated SGD methods can use a constant learning rate $\eta$ instead of diminishing step sizes for SGD, and fall into the following three categories: \emph{primal} methods such as SVRG and SAGA, \emph{dual} methods such as SDCA, and \emph{primal-dual} methods such as SPDC. In essence, many of the primal methods use the full gradient at the snapshot $\widetilde{x}$ or the average gradient to progressively reduce the variance of stochastic gradient estimators, as well as the dual and primal-dual methods, \emph{which leads to a revolution in the area of first-order optimization} \cite{shang:fsvrg}. Thus, they are also known as the hybrid gradient descent method \cite{zhang:svrg} or semi-stochastic gradient descent method \cite{koneeny:mini}. In particular, under the strongly convex condition, most of the accelerated SGD methods enjoy a linear convergence rate (also known as a geometric or exponential rate) and the oracle complexity of $\mathcal{O}\!\left((n\!+\!{L}/{\mu})\log({1}/{\epsilon})\right)$ to obtain an $\epsilon$-suboptimal solution, where each $f_{i}(\cdot)$ is $L$-smooth, and $F(\cdot)$ is $\mu$-strongly convex. The complexity bound shows that they converge faster than accelerated deterministic methods, whose oracle complexity is $\mathcal{O}(n\sqrt{L/\mu}\log({1}/{\epsilon}))$ \cite{nitanda:svrg,lin:vrsg}.

SVRG \cite{johnson:svrg} and its proximal variant, Prox-SVRG \cite{xiao:prox-svrg}, are particularly attractive because of their \emph{low storage} requirement compared with other methods such as SAG, SAGA and SDCA, which require storage of all the gradients of component functions or dual variables. At the beginning of the $s$-th epoch in SVRG, the full gradient $\nabla\! f(\widetilde{x}^{s-\!1})$ is computed at the snapshot $\widetilde{x}^{s-\!1}$, which is updated periodically.
\begin{definition}\label{def1}
The stochastic variance reduced gradient estimator is independently introduced in \cite{johnson:svrg,zhang:svrg} as follows:
\begin{equation}\label{equ051}
\widetilde{\nabla}\! f_{i^{s}_{k}}\!(x^{s}_{k})=\nabla\! f_{i^{s}_{k}}\!(x^{s}_{k})-\nabla\! f_{i^{s}_{k}}\!(\widetilde{x}^{s-\!1})+\nabla\!
f(\widetilde{x}^{s-\!1}),
\end{equation}
where $s$ is the epoch that iteration $k$ belongs to.
\end{definition}
It is not hard to verify that the variance of the SVRG estimator $\widetilde{\nabla}\! f_{i^{s}_{k}}\!(x^{s}_{k})$ (i.e., $\mathbb{E}\|\widetilde{\nabla}\! f_{i^{s}_{k}}\!(x^{s}_{k})\!-\!\nabla\!f(x^{s}_{k})\|^2$) can be much smaller than that of the SGD estimator $\nabla\! f_{i_{k}}\!(x^{s}_{k})$ (i.e., $\mathbb{E}\|\nabla\! f_{i_{k}}\!(x^{s}_{k})\!-\!\nabla\!f(x^{s}_{k})\|^2$). Theoretically, for \emph{non-strongly convex} (Non-SC) problems, the variance reduced methods converge slower than the accelerated batch methods such as FISTA~\cite{beck:fista}, i.e., $\mathcal{O}(1/T)$ vs.\ $\mathcal{O}(1/T^2)$.

More recently, many \emph{acceleration} techniques were proposed to further speed up the stochastic variance reduced methods mentioned above. These techniques mainly include the Nesterov's acceleration techniques in~\cite{hu:sgd,nitanda:svrg,lin:vrsg,frostig:sgd,lan:rpdg}, reducing the number of gradient calculations in early iterations \cite{shang:fsvrg,babanezhad:vrsg,zhu:univr}, the projection-free property of the conditional gradient method (also known as the Frank-Wolfe algorithm~\cite{frank:cg}) as in~\cite{hazan:svrf}, the stochastic sufficient decrease technique~\cite{shang:vrsgd}, and the momentum acceleration tricks in~\cite{shang:fsvrg,zhu:Katyusha,hien:asmd}. \cite{lin:vrsg} proposed an accelerating Catalyst framework and achieved the oracle complexity of $\mathcal{O}((n\!+\!\!\sqrt{n{L}/{\mu}})\log({L}/{\mu})\log({1}/{\epsilon}))$ for strongly convex problems. \cite{zhu:Katyusha} and \cite{zhou:fsvrg} proved that the accelerated methods can attain the oracle complexity of $\mathcal{O}(n\log(1/\epsilon)\!+\!\sqrt{nL/\epsilon})$ for non-strongly convex problems. The overall complexity matches the theoretical upper bound provided in \cite{woodworth:bound}. Katyusha~\cite{zhu:Katyusha}, point-SAGA~\cite{defazio:sagab} and MiG~\cite{zhou:fsvrg} achieve the best-known oracle complexity of $\mathcal{O}((n\!+\!\sqrt{nL/\mu})\log(1/\epsilon))$ for strongly convex problems, which is identical to the upper complexity bound in~\cite{woodworth:bound}. Hence, Katyusha and MiG are the \emph{best-known} stochastic optimization method for both \emph{SC} and \emph{Non-SC} problems, as pointed out in \cite{woodworth:bound}. However, selecting the best values for the parameters in the accelerated methods (e.g., the momentum parameter) is still an open problem. In particular, most of accelerated stochastic variance reduction methods, including Katyusha, require at least one auxiliary variable and one momentum parameter, which lead to complicated algorithm design and high per-iteration complexity, especially for very high-dimensional and sparse data.

\subsection{Our Contributions}
From the above discussions, we can see that most of the accelerated stochastic variance reduction methods such as~\cite{shang:fsvrg,nitanda:svrg,zhu:univr,hazan:svrf,shang:vrsgd,zhu:Katyusha,liu:svrg,lei:svrg} and applications such as~\cite{li:svrg,qu:svrg,paquette:catalyst,wang:svrgd,shamir:pca,garber:svd,liu:szvr,liu:scsg,liu:sco} are based on the SVRG method~\cite{johnson:svrg}. Thus, any key improvement on SVRG is very important for the research of stochastic optimization. In this paper, we propose a simple variant of the original SVRG~\cite{johnson:svrg}, called \emph{variance reduced stochastic gradient descent} (\emph{VR-SGD}). The snapshot point and starting point of each epoch in VR-SGD are set to the \emph{average} and \emph{last iterate} of the previous epoch, respectively. This is different from the settings of SVRG and Prox-SVRG~\cite{xiao:prox-svrg}, where the two points of the former are set to be the last iterate, and those of the latter are set to be the average of the previous epoch. This difference makes the convergence analysis of VR-SGD significantly more \emph{challenging} than that of SVRG and Prox-SVRG. Our empirical results show that the performance of VR-SGD is significantly better than its counterparts, SVRG and Prox-SVRG. Impressively, VR-SGD with varying learning rates achieves better or at least comparable performance with accelerated methods, such as Catalyst~\cite{lin:vrsg} and Katyusha~\cite{zhu:Katyusha}. The main contributions of this paper are summarized below.
\begin{itemize}
\item The snapshot and starting points of VR-SGD are set to two different vectors, i.e., $\widetilde{x}^{s}\!=\!\frac{1}{m}\!\sum^{m}_{k=1}x^{s}_{k}$ (Option I) or $\widetilde{x}^{s}\!=\!\frac{1}{m-\!1}\!\sum^{m-\!1}_{k=1}x^{s}_{k}$ (Option II), and $x^{s+\!1}_{0}\!=\!x^{s}_{m}$. In particular, we find that the settings of VR-SGD allow us to take much \emph{larger} learning rates than SVRG, e.g., $1/L$ vs.\ $1/(10L)$, and thus significantly speed up its convergence in practice. Moreover, VR-SGD has an advantage over SVRG in terms of \emph{robustness} of learning rate selection.
\item Unlike \emph{proximal} stochastic gradient methods, e.g., Prox-SVRG and Katyusha, which have a unified update rule for the two cases of \emph{smooth} and \emph{non-smooth} objectives (see Section~\ref{sec22} for details), VR-SGD employs two different update rules for the two cases, respectively, as in (\ref{equ21}) and (\ref{equ22}) below. Empirical results show that gradient update rules as in (\ref{equ21}) for \emph{smooth} optimization problems are better choices than proximal update formulas as in \eqref{equ14}.
\item We provide the convergence guarantees of VR-SGD for solving smooth/non-smooth and \emph{non-strongly convex} (or general convex) functions. In comparison, SVRG and Prox-SVRG do not have any convergence guarantees, as shown in Table~\ref{tab1}.
\item Moreover, we also present a momentum accelerated variant of VR-SGD, discuss their equivalent relationship, and empirically verify that they achieve similar performance to their variant that attains the optimal convergence rate $\mathcal{O}(1/T^2)$.
\item Finally, we theoretically analyze the convergence properties of VR-SGD with Option I or Option II for smooth/non-smooth and \emph{strongly convex} functions, which show that VR-SGD attains \emph{linear} convergence.
\end{itemize}

\begin{table}[t]
\centering
\small
\caption{Comparison of convergence rates of VR-SGD and its counterparts.}
\label{tab1}
\setlength{\tabcolsep}{2pt}
\renewcommand\arraystretch{1.15}
\begin{tabular}{lccc}
\hline
  & SVRG~\cite{johnson:svrg}   & Prox-SVRG~\cite{xiao:prox-svrg} & VR-SGD\\
\hline
SC, smooth  & linear rate & unknown & linear rate\\
SC, non-smooth  & unknown & linear rate & linear rate\\
Non-SC, smooth  & unknown & unknown & $\mathcal{O}(1/T)$\\
Non-SC, non-smooth&  unknown & unknown & $\mathcal{O}(1/T)$\\
\hline
\end{tabular}
\end{table}

\section{Preliminary and Related Work}
Throughout this paper, we use $\|\!\cdot\!\|$ to denote the $\ell_{2}$-norm (also known as the standard Euclidean norm), and $\|\!\cdot\!\|_{1}$ is the $\ell_{1}$-norm, i.e., $\|x\|_{1}\!=\!\sum^{d}_{i=1}\!|x_{i}|$. $\nabla\!f(\cdot)$ denotes the full gradient of $f(\cdot)$ if it is differentiable, or $\partial\!f(\cdot)$ the subgradient if $f(\cdot)$ is only Lipschitz continuous. For each epoch $s\!\in\![S]$ and inner iteration $k\!\in\!\{0,1,\ldots,m\!-\!1\}$, $i^{s}_{k}\!\in\![n]$ is the random chosen index. We mostly focus on the case of Problem~\eqref{equ01} when each $f_{i}(\cdot)$ is $L$-smooth\footnote{In fact, we can extend the theoretical results for the case, when the gradients of all component functions have the same Lipschitz constant $L$, to the more general case, when some component functions $f_{i}(\cdot)$ have different degrees of smoothness.}, and $F(\cdot)$ is $\mu$-strongly convex. The two common assumptions are defined as follows.

\subsection{Basic Assumptions}
\begin{assumption}[Smoothness]\label{assum1}
Each $f_{i}(\cdot)$ is $L$-smooth, that is, there exists a constant $L\!>\!0$ such that for all $x,y\!\in\!\mathbb{R}^{d}$,
\begin{equation}\label{equ11}
\|\nabla\! f_{i}(x)-\nabla\! f_{i}(y)\|\leq L\|x-y\|.
\end{equation}
\end{assumption}

\begin{assumption}[Strong Convexity]\label{assum2}
$F(x)$ is $\mu$-strongly convex, i.e., there exists a constant $\mu\!>\!0$ such that for all $x,y\!\in\! \mathbb{R}^{d}$,
\begin{equation}\label{equ12}
F(y)\geq F(x)+\langle\nabla\!F(x),\,y-x\rangle+\frac{\mu}{2}\|x-y\|^{2}.
\end{equation}
\end{assumption}
Note that when $g(\cdot)$ is \emph{non-smooth}, the inequality in~\eqref{equ12} needs to be revised by simply replacing the gradient $\nabla\!F(x)$ with an arbitrary sub-gradient of $F(\cdot)$ at $x$. In contrast, for a non-strongly convex or \emph{general convex} function, the inequality in~\eqref{equ12} can always be satisfied with $\mu\!=\!0$.

\begin{algorithm}[t]
\caption{SVRG (Option I) and Prox-SVRG (Option II)}
\label{alg1}
\renewcommand{\algorithmicrequire}{\textbf{Input:}}
\renewcommand{\algorithmicensure}{\textbf{Initialize:}}
\renewcommand{\algorithmicoutput}{\textbf{Output:}}
\renewcommand{\baselinestretch}{1.5}
\begin{algorithmic}[1]
\REQUIRE The number of epochs $S$, the number of iterations $m$ per epoch, and the learning rate $\eta$.\\
\ENSURE $\widetilde{x}^{0}$.\\
\FOR{$s=1,2,\ldots,S$}
\STATE {$\widetilde{\mu}^{s}=\frac{1}{n}\!\sum^{n}_{i=1}\!\nabla\!f_{i}(\widetilde{x}^{s-1})$, \;$x^{s}_{0}=\widetilde{x}^{s-1}$;}
\FOR{$k=0,1,\ldots,m-1$}
\STATE {Pick $i^{s}_{k}$ uniformly at random from $[n]$;}
\STATE {$\widetilde{\nabla}\! f_{i^{s}_{k}}(x^{s}_{k})=\nabla\! f_{i^{s}_{k}}(x^{s}_{k})-\nabla\! f_{i^{s}_{k}}(\widetilde{x}^{s-1})+\widetilde{\mu}^{s}$;}
\STATE {Option I: $x^{s}_{k+1}\!=x^{s}_{k}\!-\eta\!\left[\widetilde{\nabla}\!f_{i^{s}_{k}}(x^{s}_{k})+\!\nabla\! g(x^{s}_{k})\right]$,\\
\qquad\;\;\;\,or $x^{s}_{k+1}\!=\textrm{Prox}^{g}_{\eta}\!\left(x^{s}_{k}\!-\eta\widetilde{\nabla}\!f_{i^{s}_{k}}(x^{s}_{k})\right)$;\\}
\STATE {Option II: \\$\!\!\!\!\!\!x^{s}_{k+\!1}\!\!=\!\arg\min_{y\in\mathbb{R}^{d}}\!\left\{\!g(y)\!+\!y^{T}\widetilde{\nabla}\!f_{i^{s}_{k}}\!(x^{s}_{k})\!+\!\frac{1}{2\eta}\|y\!-\!x^{s}_{k}\|^2\!\right\}$;\!}
\ENDFOR
\STATE {Option I:\, $\widetilde{x}^{s}=x^{s}_{m}$;  \hfill $/\!/$\:\emph{Last iterate for snapshot $\widetilde{x}$}}
\STATE {Option II:\,$\widetilde{x}^{s}\!=\!\frac{1}{m}\!\sum^{m}_{k=1}\!x^{s}_{k}$; \hfill $/\!/$\:\emph{Iterate averaging for $\widetilde{x}$}}
\ENDFOR
\OUTPUT {$\widetilde{x}^{S}$}
\end{algorithmic}
\end{algorithm}

\subsection{Related Work}
\label{sec22}
To speed up standard and proximal SGD, many stochastic variance reduced methods~\cite{roux:sag,shalev-Shwartz:sdca,defazio:saga,zhang:svrg} have been proposed for some special cases of Problem \eqref{equ01}. In the case when each $f_{i}(x)$ is $L$-smooth, $f(x)$ is $\mu$-strongly convex, and $g(x)\!\equiv\!0$, Roux \emph{et al.} \cite{roux:sag} proposed a stochastic average gradient (SAG) method, which attains linear convergence. However, SAG, as well as other incremental aggregated gradient methods such as SAGA~\cite{defazio:saga}, needs to store all gradients, so that $O(nd)$ memory is required in general~\cite{babanezhad:vrsg}. Similarly, SDCA~\cite{shalev-Shwartz:sdca} requires storage of all dual variables~\cite{johnson:svrg}, which uses $O(n)$ memory. In contrast, SVRG proposed by Johnson and Zhang~\cite{johnson:svrg}, as well as Prox-SVRG~\cite{xiao:prox-svrg}, has the similar convergence rate to SAG and SDCA, but without the memory requirements of all gradients and dual variables. In particular, the SVRG estimator in~\eqref{equ051} may be the most popular choice for \emph{stochastic gradient estimators}. The update rule of SVRG for the case of Problem \eqref{equ01} when $g(\cdot)\!\equiv\!0$ is
\begin{equation}\label{equ052}
x^{s}_{k+1}=x^{s}_{k}-\eta \widetilde{\nabla}\! f_{i^{s}_{k}}\!(x^{s}_{k}).
\end{equation}
When the smooth regularizer $g(\cdot)\!\neq\!0$, the update rule in (\ref{equ052}) becomes: $x^{s}_{k+1}\!=\!x^{s}_{k}\!-\!\eta[\widetilde{\nabla}\! f_{i^{s}_{k}}\!(x^{s}_{k})+\nabla\!g(x^{s}_{k})]$. Although the original SVRG in~\cite{johnson:svrg} only has convergence guarantees for the special case of Problem \eqref{equ01}, when each $f_{i}(x)$ is $L$-smooth, $f(x)$ is $\mu$-strongly convex, and $g(x)\!\equiv\!0$, we can extend SVRG to the proximal setting by introducing the proximal operator in~(\ref{equ03}), as shown in Line 7 of Algorithm~\ref{alg1}.

Based on the SVRG estimator in~\eqref{equ051}, some accelerated algorithms~\cite{nitanda:svrg,lin:vrsg,zhu:Katyusha} have been proposed. The proximal update rules of Katyusha~\cite{zhu:Katyusha} are formulated as follows:
\addtocounter{equation}{1}
\begin{align}
&\!x^{s}_{k+\!1}\!=w_{1}y^{s}_{k}+w_{2}\widetilde{x}^{s-\!1}+(1-w_{1}-w_{2})z^{s}_{k}\tag{\theequation a},\label{equ061}\\
&\!y^{s}_{k+\!1}\!=\!\mathop{\arg\min}_{y\in\mathbb{R}^{d}}\!\left\{\frac{1}{2\eta}\|y\!-\!y^{s}_{k}\|^2\!+\!y^{T}\widetilde{\nabla}\! f_{i^{s}_{k}}\!(x^{s}_{k+\!1})\!+\!g(y)\right\}\!,\tag{\theequation b}\label{equ062}\\
&\!z^{s}_{k+\!1}\!=\!\mathop{\arg\min}_{z\in\mathbb{R}^{d}}\!\left\{\!\frac{3L}{2}\|z\!-\!x^{s}_{k+\!1}\|^2\!+\!z^{T}\widetilde{\nabla}\! f_{i^{s}_{k}}\!(x^{s}_{k+\!1})\!+\!g(z)\!\right\}\tag{\theequation c}\label{equ063}
\end{align}
where $w_{1},w_{2}\!\in\![0,1]$ are two parameters. To eliminate the need for parameter tuning, $\eta$ is set to $1/(3w_{1}L)$, and $w_{2}$ is fixed to $0.5$ in~\cite{zhu:Katyusha}. In addition, \cite{zhu:cca,shamir:pca,garber:svd} applied efficient stochastic solvers to compute leading eigenvectors of a symmetric matrix or generalized eigenvectors of two symmetric matrices. The first such method is VR-PCA proposed by Shamir~\cite{shamir:pca}, and the convergence properties of VR-PCA for such a non-convex problem are also provided. Garber \emph{et al.}~\cite{garber:svd} analyzed the convergence rate of SVRG when $f(\cdot)$ is a convex function that is a sum of non-convex component functions. Moreover, \cite{zhu:vrnc,reddi:svrnc} and \cite{reddi:saga} proved that SVRG and SAGA with minor modifications can converge asymptotically to a stationary point of non-convex functions. Some sparse approximation, parallel and distributed variants~\cite{zhou:fsvrg,mania:svrg,reddi:sgd,lian:psgd,lee:dsgd} of accelerated SGD methods have also been proposed.

An important class of stochastic methods is the \emph{proximal stochastic gradient} (Prox-SG) method, such as Prox-SVRG~\cite{xiao:prox-svrg}, SAGA~\cite{defazio:saga}, and Katyusha~\cite{zhu:Katyusha}. Different from standard variance reduction SGD methods such as SVRG, the Prox-SG method has a unified update rule for both smooth and non-smooth cases of $g(\cdot)$. For instance, the update rule of Prox-SVRG~\cite{xiao:prox-svrg} is formulated as follows:
\begin{equation}\label{equ14}
x^{s}_{k+1}\!=\!\mathop{\arg\min}_{y\in\mathbb{R}^{d}}\left\{g(y)\!+\!y^{T}\widetilde{\nabla}\!f_{i^{s}_{k}}\!(x^{s}_{k})\!+\!\frac{1}{2\eta}\|y\!-\!x^{s}_{k}\|^2\right\}.
\end{equation}
For the sake of completeness, the details of Prox-SVRG~\cite{xiao:prox-svrg} are shown in Algorithm~\ref{alg1} with Option II. When $g(\cdot)$ is the widely used $\ell_{2}$-norm regularizer, i.e., $g(\cdot)=(\lambda_{1}/2)\|\cdot\|^{2}$, the proximal update formula in \eqref{equ14} becomes
\begin{equation}\label{equ15}
x^{s}_{k+1}=\frac{1}{1+\lambda_{1}\eta}\left[x^{s}_{k}-\eta\widetilde{\nabla}\!f_{i^{s}_{k}}(x^{s}_{k})\right].
\end{equation}

\section{Variance Reduced SGD}
In this section, we propose an efficient variance reduced stochastic gradient descent (VR-SGD) algorithm, as shown in Algorithm~\ref{alg2}. Different from the choices of the snapshot and starting points in SVRG \cite{johnson:svrg} and Prox-SVRG \cite{xiao:prox-svrg}, the two vectors of each epoch in VR-SGD are set to the average and last iterate of the previous epoch, respectively. Moreover, unlike existing proximal stochastic methods, we design two different update rules for smooth and non-smooth objective functions, respectively.

\subsection{Snapshot and Starting Points}
Like SVRG, VR-SGD is also divided into $S$ epochs, and each epoch consists of $m$ stochastic gradient steps, where $m$ is usually chosen to be $\Theta(n)$, as suggested in~\cite{johnson:svrg,xiao:prox-svrg,zhu:Katyusha}. Within each epoch, we need to compute the full gradient $\nabla\! f(\widetilde{x}^{s})$ at the snapshot $\widetilde{x}^{s}$ and use it to define the variance reduced stochastic gradient estimator $\widetilde{\nabla}\! f_{i^{s}_{k}}\!(x^{s}_{k})$ in~\eqref{equ051}. Unlike SVRG, whose snapshot is set to the last iterate of the previous epoch, the \emph{snapshot} $\widetilde{x}^{s}$ of VR-SGD is set to the \emph{average} of the previous epoch, e.g., $\widetilde{x}^{s}\!=\!\frac{1}{m}\!\sum^{m}_{k=1}x^{s}_{k}$ in Option I of Algorithm~\ref{alg2}, which leads to better robustness to gradient noise\footnote{It should be emphasized that the noise introduced by random sampling is inevitable, and generally slows down the convergence speed in this sense. However, SGD and its variants are probably the mostly used optimization algorithms for deep learning~\cite{bengio:deep}. In particular, \cite{ge:sgd} has shown that by adding gradient noise at each step, noisy gradient descent can escape the saddle points efficiently and converge to a local minimum of the non-convex minimization problem, e.g., the application of deep neural networks in~\cite{neelakantan:noise}.}, as also suggested in~\cite{shang:fsvrg,shang:vrsgd,flammarion:sgd}. In fact, the choice of Option II in Algorithm~\ref{alg2}, i.e., $\widetilde{x}^{s}\!=\!\frac{1}{m-\!1}\!\sum^{m-\!1}_{k=1}x^{s}_{k}$, also works well in practice, as shown in Fig.\ 2 in the Supplementary Material. Therefore, we provide the convergence guarantees for our algorithm with either Option I or Option II in the next section. In particular, we find that one of the effects of the choice in Option I or Option II of Algorithm~\ref{alg2} is to allow taking much larger learning rates or step sizes than SVRG in practice, e.g., $1/L$ for VR-SGD vs.\ $1/(10L)$ for SVRG (see Fig.\ \ref{figs01}). Actually, a larger learning rate enjoyed by VR-SGD means that the variance of its stochastic gradient estimator goes asymptotically to zero faster.

Unlike Prox-SVRG~\cite{xiao:prox-svrg} whose starting point is initialized to the average of the previous epoch, the starting point of VR-SGD is set to the last iterate of the previous epoch. That is, in VR-SGD, the last iterate of the previous epoch becomes the new starting point, while the two points of Prox-SVRG are completely different, thereby leading to relatively slow convergence in general. Both the starting and snapshot points of SVRG \cite{johnson:svrg} are set to the last iterate of the previous epoch{\footnote{Note that the theoretical convergence of the original SVRG \cite{johnson:svrg} relies on its Option II, i.e., both $\widetilde{x}^{s}$ and $x^{s+\!1}_{0}$ are set to $x^{s}_{k}$, where $k$ is randomly chosen from $\{1,2,\ldots,m\}$. However, the empirical results in \cite{johnson:svrg} suggest that Option I is a better choice than its Option II, and the convergence guarantee of SVRG with Option I for strongly convex objective functions is provided in \cite{tan:sgd}.}}, while the two points of Prox-SVRG \cite{xiao:prox-svrg} are set to the average of the previous epoch (also suggested in \cite{johnson:svrg}). By setting the starting and snapshot points in VR-SGD to the two different vectors mentioned above, the convergence analysis of VR-SGD becomes significantly more challenging than that of SVRG and Prox-SVRG, as shown in Section \ref{sec4}.

\begin{algorithm}[t]
\caption{VR-SGD for solving smooth problems}
\label{alg2}
\renewcommand{\algorithmicrequire}{\textbf{Input:}}
\renewcommand{\algorithmicensure}{\textbf{Initialize:}}
\renewcommand{\algorithmicoutput}{\textbf{Output:}}
\begin{algorithmic}[1]
\REQUIRE The number of epochs $S$, and the number of iterations $m$ per epoch.\\
\ENSURE $x^{1}_{0}=\widetilde{x}^{0}$, and $\{\eta_{s}\}$.\\
\FOR{$s=1,2,\ldots,S$}
\STATE {$\widetilde{\mu}^{s}=\frac{1}{n}\!\sum^{n}_{i=1}\!\nabla\!f_{i}(\widetilde{x}^{s-1})$; \hfill $/\!/$\:\emph{Compute the full gradient}}
\FOR{$k=0,1,\ldots,m-1$}
\STATE {Pick $i^{s}_{k}$ uniformly at random from $[n]$;}
\STATE {$\widetilde{\nabla}\! f_{i^{s}_{k}}(x^{s}_{k})=\nabla\! f_{i^{s}_{k}}(x^{s}_{k})-\nabla\! f_{i^{s}_{k}}(\widetilde{x}^{s-1})+\widetilde{\mu}^{s}$;}
\STATE {$x^{s}_{k+\!1}\!\!=\!x^{s}_{k}\!-\!\eta_{s}[\widetilde{\nabla}\!f_{i^{s}_{k}}\!(x^{s}_{k})+\!\nabla\! g(x^{s}_{k})]$;}
\ENDFOR
\STATE {Option I:\,$\widetilde{x}^{s}\!=\!\frac{1}{m}\!\sum^{m}_{k=1}\!x^{s}_{k}$; \!\!\hfill $/\!/$\:\emph{Iterate averaging for $\widetilde{x}$}}
\STATE {Option II:\,$\widetilde{x}^{s}\!=\!\frac{1}{m-\!1}\!\sum^{m-\!1}_{k=1}\!x^{s}_{k}$; \!\!\hfill $/\!/$\:\emph{Iterate averaging for $\widetilde{x}$}}
\STATE {$x^{s+1}_{0}\!=x^{s}_{m}$; \hfill $/\!/$\:\emph{Initiate $x^{s+1}_{0}$ for the next epoch}}
\ENDFOR
\OUTPUT {$\widehat{x}^{S}\!=\!\widetilde{x}^{S}$, if $F(\widetilde{x}^{S})\!\leq\! F(\frac{1}{S}\!\sum^{S}_{s=1}\!\widetilde{x}^{s})$, and $\widehat{x}^{S}\!=\!\frac{1}{S}\!\sum^{S}_{s=1}\!\widetilde{x}^{s}$ otherwise.}
\end{algorithmic}
\end{algorithm}

\subsection{The VR-SGD Algorithm}
In this part, we propose an efficient VR-SGD algorithm to solve Problem (\ref{equ01}), as outlined in \textbf{Algorithm} \ref{alg2} for the case of smooth objective functions. It is well known that the original SVRG~\cite{johnson:svrg} only works for the case of smooth minimization problems. However, in many machine learning applications, e.g., elastic net regularized logistic regression, the strongly convex objective function $F(x)$ is non-smooth. To solve this class of problems, the proximal variant of SVRG, Prox-SVRG~\cite{xiao:prox-svrg}, was subsequently proposed. Unlike the original SVRG, VR-SGD can not only solve \emph{smooth} objective functions, but also directly tackle \emph{non-smooth} ones. That is, when the regularizer $g(x)$ is smooth (e.g., the $\ell_{2}$-norm regularizer), the key update rule of VR-SGD is
\begin{equation}\label{equ21}
x^{s}_{k+1}=x^{s}_{k}-\eta_{s}[\widetilde{\nabla}\!f_{i^{s}_{k}}(x^{s}_{k})+\nabla\! g(x^{s}_{k})].
\end{equation}
When $g(x)$ is non-smooth (e.g., the $\ell_{1}$-norm regularizer), the key update rule of VR-SGD in \textbf{Algorithm} \ref{alg2} becomes
\begin{equation}\label{equ22}
x^{s}_{k+1}=\textrm{Prox}^{\,g}_{\,\eta_{s}}\!\left(x^{s}_{k}-\eta_{s}\widetilde{\nabla}\!f_{i^{s}_{k}}(x^{s}_{k})\right).
\end{equation}

\begin{figure}[t]
\centering
\subfigure[Logistic regression: $\lambda=10^{-4}$ (left) \;and\; $\lambda=10^{-5}$ (right)]{
\includegraphics[width=0.489\columnwidth]{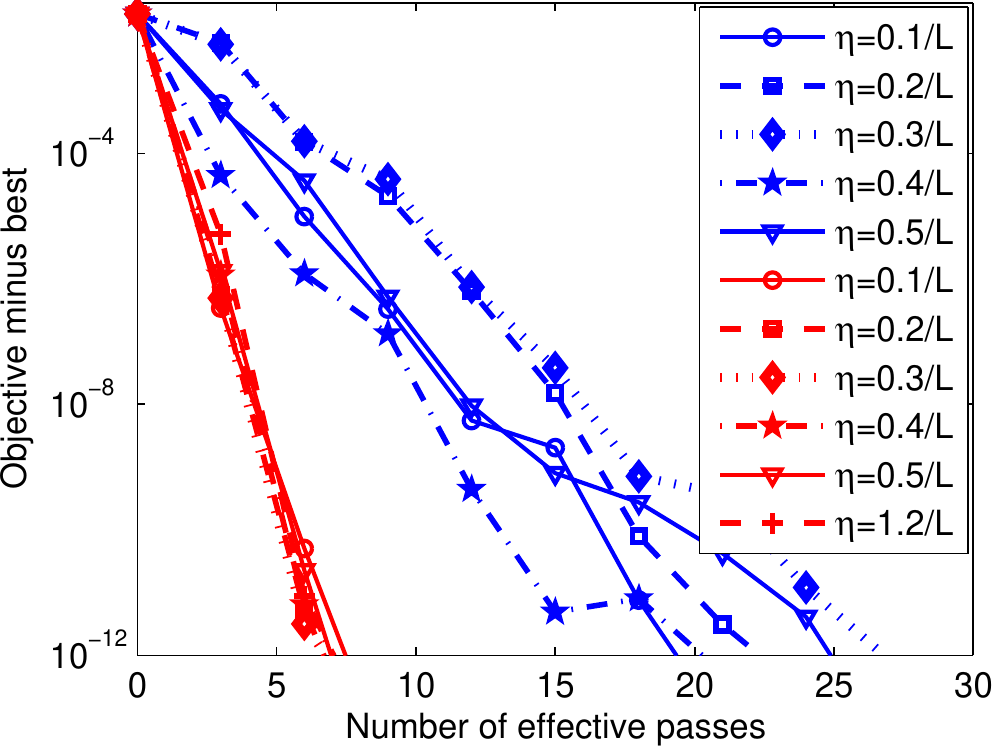}
\includegraphics[width=0.489\columnwidth]{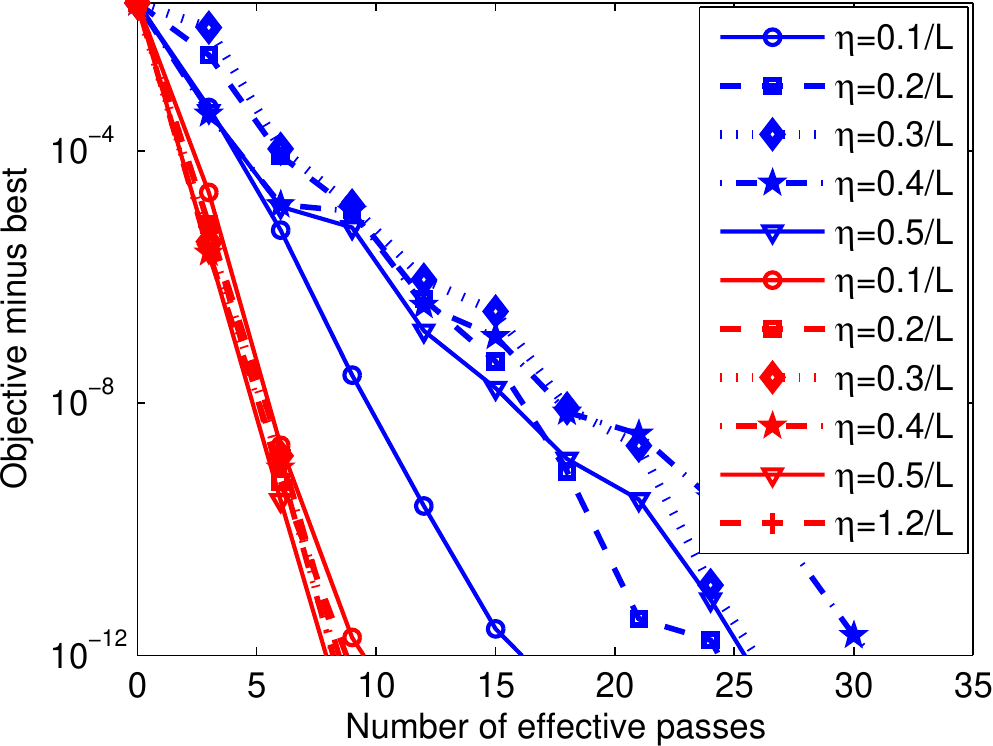}}
\vspace{0.36mm}
\subfigure[Ridge regression: $\lambda=10^{-4}$ (left) \;and\; $\lambda=10^{-5}$ (right)]{
\includegraphics[width=0.489\columnwidth]{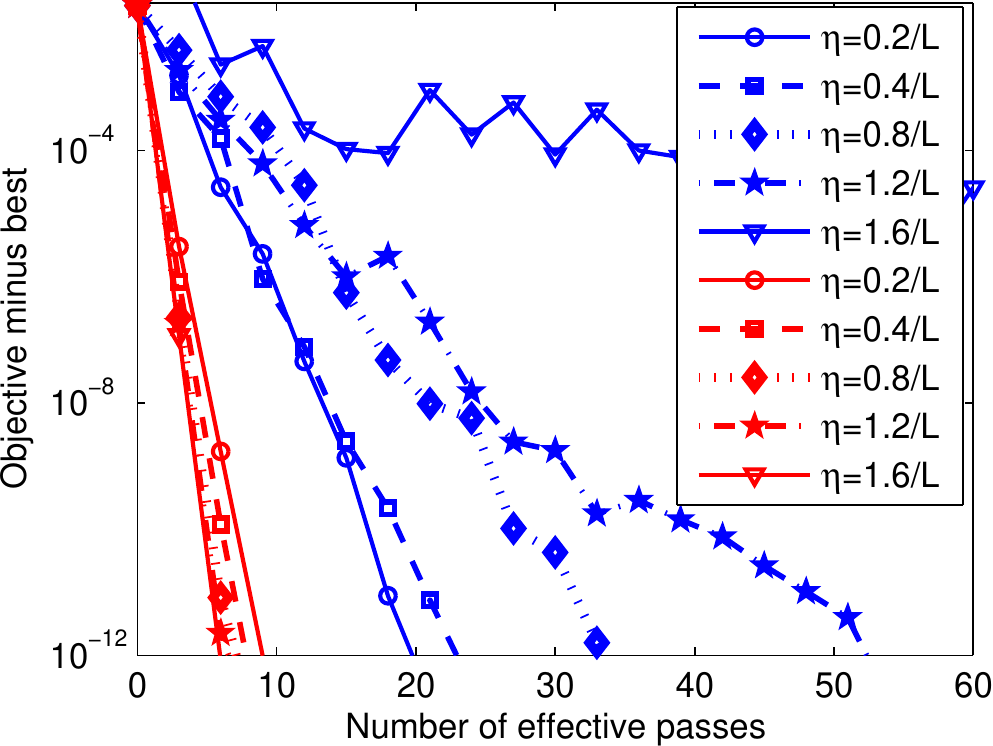}
\includegraphics[width=0.489\columnwidth]{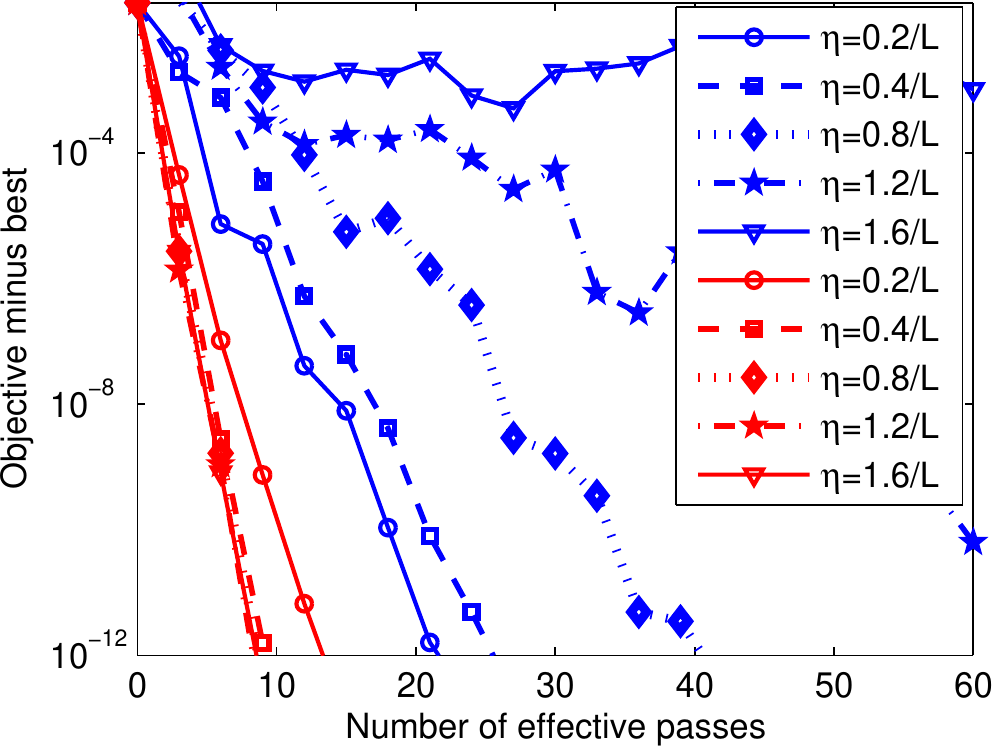}\label{figs01a}}
\caption{Comparison of SVRG~\cite{johnson:svrg} and VR-SGD with different learning rates for solving $\ell_{2}$-norm regularized logistic regression and ridge regression on Covtype. Note that the \emph{blue} lines stand for the results of SVRG, while the \emph{red} lines correspond to the results of VR-SGD (best viewed in colors).}
\label{figs01}
\end{figure}

Unlike the proximal stochastic methods such as Prox-SVRG~\cite{xiao:prox-svrg}, all of which have a unified update rule as in (\ref{equ14}) for both the smooth and non-smooth cases of $g(\cdot)$, VR-SGD has two different update rules for the two cases, as in (\ref{equ21}) and (\ref{equ22}). Fig.\ \ref{figs01} demonstrates that VR-SGD has a significant advantage over SVRG in terms of robustness of learning rate selection. That is, VR-SGD yields good performance within the range of the learning rate from $0.2/L$ to $1.2/L$, whereas the performance of SVRG is very \emph{sensitive} to the selection of learning rates. Thus, VR-SGD is convenient to be applied in various real-world problems of machine learning. In fact, VR-SGD can use much larger learning rates than SVRG for ridge regression problems in practice, e.g., $8/(5L)$ for VR-SGD vs.\ $1/(5L)$ for SVRG, as shown in Fig.\ \ref{figs01a}.

\subsection{VR-SGD for Non-Strongly Convex Objectives}
Although many stochastic variance reduced methods have been proposed, most of them, including SVRG and Prox-SVRG, only have convergence guarantees for the case of Problem (\ref{equ01}), when $F(x)$ is strongly convex. However, $F(x)$ may be non-strongly convex in many machine learning applications, such as Lasso and $\ell_{1}$-norm regularized logistic regression. As suggested in~\cite{zhu:Katyusha,zhu:box}, this class of problems can be transformed into strongly convex ones by adding a proximal term $(\tau/2)\|x\!-\!x^{s}_{0}\|^2$, which can be efficiently solved by Algorithm~\ref{alg2}. However, the reduction technique may degrade the performance of the involved algorithms both in theory and in practice~\cite{zhu:univr}. Thus, we use VR-SGD to directly solve \emph{non-strongly convex} problems.

The learning rate $\eta_{s}$ of Algorithm~\ref{alg2} can be fixed to a constant. Inspired by existing accelerated stochastic algorithms~\cite{shang:fsvrg,zhu:Katyusha}, the learning rate in Algorithm~\ref{alg2} can be gradually increased in early iterations for both strongly convex and non-strongly convex problems, which leads to faster convergence (see Fig.\ 3 in the Supplementary Material). Different from SGD and Katyusha~\cite{zhu:Katyusha}, where the learning rate of the former requires to be gradually decayed and that of the latter needs to be gradually increased, the update rule of $\eta_{s}$ in \textbf{Algorithm} \ref{alg2} is defined as follows: $\eta_{0}$ is an initial learning rate, and for any $s\!\geq\!1$,
\begin{equation}\label{equ23}
\eta_{s}=\eta_{0}/\max\{\alpha,\,2/(s+1)\}
\end{equation}
where $0\!<\!\alpha\!\leq\!1$ is a given constant, e.g., $\alpha\!=\!0.2$.

\subsection{Extensions of VR-SGD}
It has been shown in~\cite{koneeny:mini,nitanda:svrg} that \emph{mini-batching} can effectively decrease the variance of stochastic gradient estimates. Therefore, we first extend the proposed VR-SGD method to the mini-batch setting, as well as its convergence results below. Here, we denote by $b$ the mini-batch size and $I^{s}_{k}$ the selected random index set $I_{k}\!\subset\![n]$ for each outer-iteration $s\!\in\![S]$ and inner-iteration $k\!\in\!\{0,1,\ldots,m\!-\!1\}$.

\begin{definition}\label{def2}
The stochastic variance reduced gradient estimator in the mini-batch setting is defined as
\begin{equation}
\widetilde{\nabla}\! f_{I^{s}_{k}}(x^{s}_{k})=\frac{1}{b}\!\sum_{i\in I^{s}_{k}}\!\left[\nabla\!f_{i}(x^{s}_{k})\!-\!\nabla\! f_{i}(\widetilde{x}^{s-\!1})\right]\!+\!{\nabla}\! f(\widetilde{x}^{s-\!1})
\end{equation}
where $I^{s}_{k}\!\subset\![n]$ is a mini-batch of size $b$.
\end{definition}

If some component functions are non-smooth, we can use the proximal operator oracle \cite{zhu:box} or the Nesterov's smoothing \cite{nesterov:smooth} and homotopy smoothing \cite{xu:hs} techniques to smoothen them, and thereby obtain their smoothed approximations. In addition, we can directly extend our VR-SGD method to the non-smooth setting as in \cite{shang:fsvrg} (e.g., Algorithm 3 in \cite{shang:fsvrg}) without using any smoothing techniques.

Considering that each component function $f_{i}(x)$ may have different degrees of smoothness, picking the random index $i^{s}_{k}$ from a non-uniform distribution is a much better choice than commonly used uniform random sampling \cite{zhao:prox-smd,needell:sgd}, as well as without-replacement sampling vs.\ with-replacement sampling \cite{shamir:sgd}. This can be done using the same techniques in \cite{xiao:prox-svrg,zhu:Katyusha}, i.e., the sampling probabilities for all $f_{i}(x)$ are proportional to their Lipschitz constants, i.e., $p_{i}\!=\!L_{i}/\!\sum^{n}_{j=1}\!L_{j}$. VR-SGD can also be combined with other accelerated techniques used for SVRG. For instance, the epoch length of VR-SGD can be automatically determined by the techniques in \cite{zhu:univr,konecny:s2gd}, instead of a fixed epoch length. We can reduce the number of gradient calculations in early iterations as in \cite{babanezhad:vrsg,zhu:univr}, which leads to faster convergence in general (see Section~\ref{sec55} for details). Moreover, we can introduce the Nesterov's acceleration techniques in \cite{hu:sgd,nitanda:svrg,lin:vrsg,frostig:sgd,lan:rpdg} and momentum acceleration tricks in \cite{shang:fsvrg,zhu:Katyusha,shang:asvrg} to further improve the performance of VR-SGD.

\section{Algorithm Analysis}
\label{sec4}
In this section, we provide the convergence guarantees of VR-SGD for solving both smooth and non-smooth general convex problems, and extend the results to the mini-batch setting. We also study the convergence properties of VR-SGD for solving both smooth and non-smooth strongly convex objective functions. Moreover, we discuss the equivalent relationship between VR-SGD and its momentum accelerated variant, as well as some of its extensions.

\subsection{Convergence Properties: Non-strongly Convex}
In this part, we analyze the convergence properties of VR-SGD for solving more general non-strongly convex problems. Considering that the proposed algorithm (i.e., Algorithm~\ref{alg2}) has two different update rules for smooth and non-smooth cases, we give the convergence guarantees of VR-SGD for the two cases as follows.

\subsubsection{Smooth Objective Functions}
We first provide the convergence guarantee of our algorithm for solving Problem \eqref{equ01} when $F(x)$ is \emph{smooth}. In order to simplify analysis, we denote $F(x)$ by $f(x)$, that is, $f_{i}(x):=f_{i}(x)+g(x)$ for all $i\!=\!1,2,\ldots,n$, and then $g(x)\equiv0$.

\begin{lemma}[Variance bound]
\label{lemm2}
Let $x^{*}$ be the optimal solution of Problem \eqref{equ01}. Suppose Assumption~\ref{assum1} holds. Then the following inequality holds
\begin{equation*}
\mathbb{E}\!\!\left[\|\widetilde{\nabla}\! f_{i^{s}_{k}}\!(x^{s}_{k})\!-\!\nabla\! f(x^{s}_{k})\|^{2}\right]\!\!\leq\!4L[f(x^{s}_{k})\!-\!f(\!x^{*}\!)\!+\!f(\widetilde{x}^{s-\!1})\!-\!f(\!x^{*}\!)].
\end{equation*}
\end{lemma}

The proofs of this lemma, the lemmas and theorems below are all included in the Supplementary Material. Lemma~\ref{lemm2} provides the upper bound on the expected variance of the variance reduced gradient estimator in \eqref{equ051}, i.e., the SVRG estimator. For Algorithm \ref{alg2} with Option II and a fixed learning rate $\eta$, we have the following result.

\begin{theorem}[Smooth objectives]
\label{the1}
Suppose Assumption~\ref{assum1} holds. Then the following inequality holds
\begin{equation*}
\begin{split}
\!\!\mathbb{E}\!\left[f(\widehat{x}^{S})\right]-f(x^{*})
\leq&\:\frac{2(m+1)}{[\gamma-4m+2]S}[f(\widetilde{x}^{0})-f(x^{*})]\\
&\:+\frac{\beta(\beta-1)L}{2[\gamma-4m+2]S}\|\widetilde{x}^{0}\!-x^{*}\|^{2}
\end{split}
\end{equation*}
where $\gamma=(\beta\!-\!1)(m\!-\!1)$, and $\beta\!=\!1/(L\eta)$.
\end{theorem}

From Theorem~\ref{the1} and its proof, one can see that our convergence analysis is very different from that of existing stochastic methods, such as SVRG~\cite{johnson:svrg}, Prox-SVRG~\cite{xiao:prox-svrg}, and SVRG++~\cite{zhu:univr}. Similarly, the convergence of Algorithm \ref{alg2} with Option I and a fixed learning rate can be guaranteed, as stated in Theorem 6 in the Supplementary Material. All the results show that VR-SGD attains a convergence rate of $\mathcal{O}(1/T)$ for non-strongly convex functions.

\subsubsection{Non-Smooth Objective Functions}
We also provide the convergence guarantee of Algorithm \ref{alg2} with Option I and (\ref{equ22}) for solving Problem \eqref{equ01} when $F(x)$ is \emph{non-smooth} and non-strongly convex, as shown below.

\begin{theorem}[Non-smooth objectives]
\label{the3}
Suppose Assumption~\ref{assum1} holds. Then the following inequality holds
\begin{equation*}
\begin{split}
&\mathbb{E}\!\left[F(\widehat{x}^{S})\right]-F(x^{*})\\
\leq&\,\frac{2(m\!+\!1)}{(\beta\!-\!5)mS}[F(\widetilde{x}^{0})\!-F(x^{*})]+\frac{\beta(\beta\!-\!1)L}{2(\beta\!-\!5)mS}\|\widetilde{x}^{0}\!-x^{*}\|^{2}.
\end{split}
\end{equation*}
\end{theorem}

Similarly, the convergence of Algorithm \ref{alg2} with Option II and a fixed learning rate can be guaranteed, as stated in Corollary 4 in the Supplementary Material.

\subsubsection{Mini-Batch Settings}
The upper bound on the variance of $\widetilde{\nabla}\!f_{i^{s}_{k}}(x^{s}_{k})$ is extended to the \emph{mini-batch setting} as follows.

\begin{corollary}[Variance bound of mini-batch]
\label{lemm8}
If each $f_{i}(\cdot)$ is convex and $L$-smooth, then the following inequality holds
\begin{equation*}
\begin{split}
&\,\mathbb{E}\!\left[\|\widetilde{\nabla}\! f_{I^{s}_{k}}(x^{s}_{k})-\nabla\! f(x^{s}_{k})\|^{2}\right]\\
\leq&\,4L\delta(b)[F(x^{s}_{k})-F(x^{*})+F(\widetilde{x}^{s-1})-F(x^{*})]
\end{split}
\end{equation*}
where $\delta(b)=(n\!-\!b)/[(n\!-\!1)b]$.
\end{corollary}

This corollary is essentially identical to Theorem 4 in~\cite{koneeny:mini}, and hence its proof is omitted. It is not hard to verify that $0\leq\delta(b)\leq1$. Based on the variance upper bound, we further analyze the convergence properties of VR-SGD in the mini-batch setting, as shown below.

\begin{theorem}[Mini-batch]
\label{the5}
If each $f_{i}(\cdot)$ is convex and $L$-smooth, then the following inequality holds
\begin{equation*}\label{equ55}
\begin{split}
&\mathbb{E}\!\left[F(\widehat{x}^{S})\right]-F(x^{*})\\
\leq&\frac{2\delta(b)(m\!+\!1)}{\zeta mS}\mathbb{E}\!\left[F(\widetilde{x}^{0})\!-\!F(x^{*})\right]\!+\!\frac{\beta(\beta\!-\!1)L}{2\zeta mS}\mathbb{E}\!\left[\|x^{*}\!-\!\widetilde{x}^{0}\|^2\right]
\end{split}
\end{equation*}
where $\zeta=\beta\!-\!1\!-\!4\delta(b)$.
\end{theorem}

From Theorem~\ref{the5}, one can see that when $b\!=\!n$ (i.e., the batch setting), $\delta(n)\!=\!0$, and the first term on the right-hand side of the above inequality diminishes. That is, VR-SGD degenerates to a batch method. When $b\!=\!1$, we have $\delta(1)\!=\!1$, and thus Theorem~\ref{the5} degenerates to Theorem~\ref{the3}.

\subsection{Convergence Properties: Strongly Convex}
We also analyze the convergence properties of VR-SGD for solving \emph{strongly convex} problems. We first give the following convergence result for Algorithm~\ref{alg2} with Option II.

\begin{theorem}[Strongly convex]
\label{the6}
Suppose Assumptions~\ref{assum1}, \ref{assum2} and 3 in the Supplementary Material hold, and $m$ is sufficiently large so that
\begin{equation*}
\rho:=\,\frac{2L\eta(m\!+\!c)}{(m\!-\!1)(1\!-\!3L\eta)}+\frac{c(1\!-\!L\eta)}{\mu\eta(m\!-\!1)(1\!-\!3L\eta)}< 1
\end{equation*}
where $c$ is a constant. Then Algorithm~\ref{alg2} with Option II has the following geometric convergence in expectation:
\begin{equation*}
\mathbb{E}\left[F(\widehat{x}^{S})-F(x^{*})\right]\leq \rho^{S}\!\left[F(\widetilde{x}^{0})-F(x^{*})\right].
\end{equation*}
\end{theorem}

We also provide the linear convergence guarantees for Algorithm~\ref{alg2} with Option I for solving non-smooth and strongly convex functions, as stated in Theorem 7 in the Supplementary Material. Similarly, the linear convergence of Algorithm~\ref{alg2} can be guaranteed for the smooth strongly-convex case. All the theoretical results show that VR-SGD attains a \emph{linear} convergence rate and at most the oracle complexity of $\mathcal{O}\!\left((n\!+\!{L}/{\mu})\log({1}/{\epsilon})\right)$ for both \emph{smooth} and \emph{non-smooth} strongly convex functions. In contrast, the convergence of SVRG~\cite{johnson:svrg} is only guaranteed for smooth and strongly convex problems.

Although the learning rate in Theorem~\ref{the6} needs to be less than $1/(3L)$, we can use much larger learning rates in practice, e.g., $\eta\!=\!1/L$. However, it can be easily verified that the learning rate of SVRG should be less than $1/(4L)$ in theory, and adopting a larger learning rate for SVRG is not always helpful in practice, which means that VR-SGD can use much larger learning rates than SVRG both in theory and in practice. In other words, although they have the same theoretical convergence rate, VR-SGD converges significantly faster than SVRG in practice, as shown by our experiments. Note that similar to the convergence analysis in \cite{zhu:vrnc,reddi:svrnc,reddi:saga}, the convergence of VR-SGD for some non-convex problems can also be guaranteed.

\subsection{Equivalent to Its Momentum Accelerated Variant}
Inspired by the success of the momentum technique in our previous work~\cite{liu:avrrg,zhou:fsvrg,shang:asvrg}, we present a momentum accelerated variant of Algorithm~\ref{alg2}, as shown in Algorithm~\ref{alg3}. Unlike existing momentum techniques, e.g., \cite{nesterov:fast,beck:fista,hu:sgd,nitanda:svrg,lin:vrsg,zhu:Katyusha}, we use the convex combination of the snapshot $\widetilde{x}^{s-\!1}$ and latest iterate $v^{s}_{k}$ for acceleration, i.e., $\widetilde{x}^{s-\!1}\!+\!w_{s}(v^{s}_{k+\!1}\!-\!\widetilde{x}^{s-\!1})\!=\!w_{s}v^{s}_{k}\!+\!(1\!-\!w_{s})\widetilde{x}^{s-\!1}$. It is not hard to verify that Algorithm~\ref{alg2} with Option I is equivalent to its variant (i.e., Algorithm~\ref{alg3} with Option I), when $w_{s}\!=\!\max\{\alpha, 2/(s\!+\!1)\}$ and $\alpha$ is sufficiently small (see the Supplementary Material for their equivalent analysis). We emphasize that the only difference between Options I and II in Algorithm~\ref{alg3} is the initialization of $x^{s}_{0}$ and $v^{s}_{0}$.

\begin{theorem}
\label{the7}
Suppose Assumption~\ref{assum1} holds. Then the following inequality holds:
\begin{equation*}
\begin{split}
&\mathbb{E}[F(\widehat{x}^{S})-F(x^{*})]\\
\leq&\,\frac{4(1\!-\!w_{1})}{w^{2}_{1}(S\!+\!1)^{2}}[F(\widetilde{x}^{0})-F(x^{*})]+\!\frac{2}{m\eta_{0}(S\!+\!1)^{2}}\|x^{*}\!-\widetilde{x}^{0}\|^2.
\end{split}
\end{equation*}
Choosing $m\!=\!\Theta(n)$, Algorithm~\ref{alg3} with Option II achieves an $\epsilon$-suboptimal solution (i.e., $\mathbb{E}[F(\widehat{x}^{S})]\!-\!F(x^{*})\!\!\leq\!\! \varepsilon$) using at most $\mathcal{O}(n\sqrt{[F(\widetilde{x}^{0})\!-\!F(x^{*})]/\varepsilon}\!+\!\sqrt{nL/\varepsilon}\|\widetilde{x}^{0}\!-\!x^{*}\|)$ iterations.
\end{theorem}

\begin{figure}[t]
\centering
\subfigure[Small dataset: MNIST]{\includegraphics[width=0.456\columnwidth]{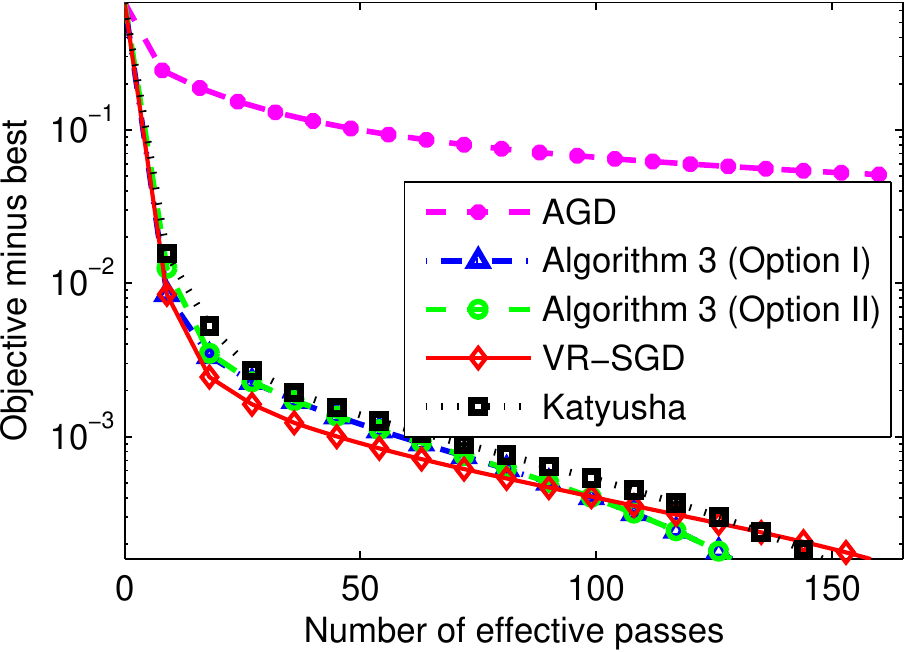}}\;\,
\subfigure[Large dataset: Epsilon]{\includegraphics[width=0.456\columnwidth]{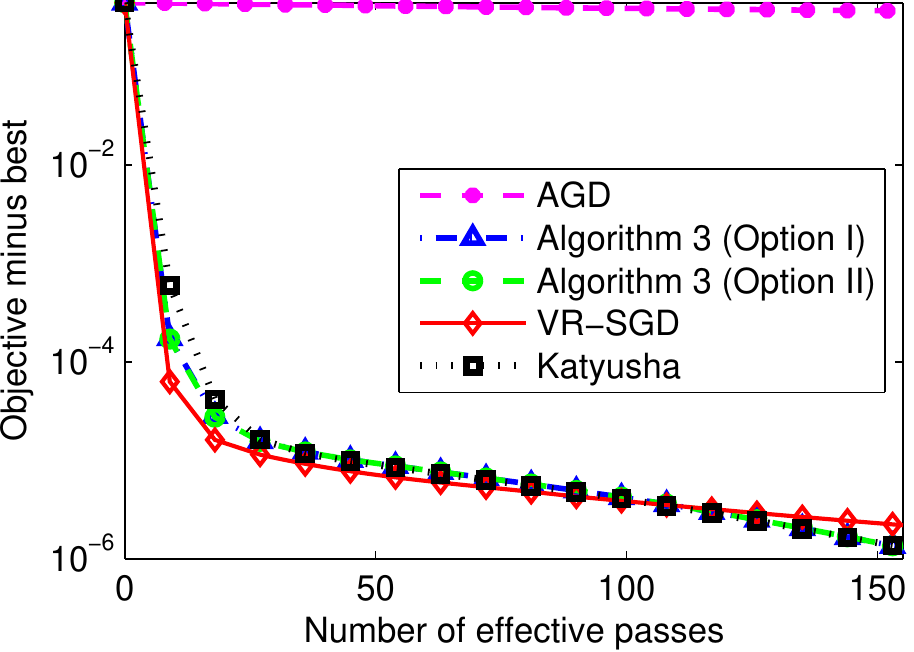}}
\caption{Comparison of AGD~\cite{nesterov:co}, Katyusha~\cite{zhu:Katyusha}, Algorithm~\ref{alg3} with Option I and II, and VR-SGD for solving logistic regression with $\lambda=0$.}
\label{figs02}
\end{figure}

This theorem shows that the oracle complexity of Algorithm \ref{alg3} with Option II is consistent with that of Katyusha~\cite{zhu:Katyusha}, and is better than that of accelerated deterministic methods (e.g., AGD~\cite{nesterov:co}), (i.e., $\mathcal{O}(n\sqrt{L/\varepsilon})$), which are also verified by the experimental results in Fig.\ \ref{figs02}. Our algorithm also achieves the optimal convergence rate $\mathcal{O}(1/T^2)$ for non-strongly convex functions as in~\cite{zhu:Katyusha,hien:asmd}. Fig.\ \ref{figs02} shows that Katyusha and Algorithm \ref{alg3} with Option II have similar performance as Algorithms \ref{alg2} and \ref{alg3} with Option I ($\eta_{0}\!\!=\!\!3/(5L)$) in terms of number of effective passes. Clearly, Algorithm \ref{alg3} and Katyusha have higher complexity per iteration than Algorithm \ref{alg2}. Thus, we only report the results of VR-SGD (i.e., Algorithm \ref{alg2}) in Section \ref{sec5}.

\begin{algorithm}[t]
\caption{The momentum accelerated algorithm}
\label{alg3}
\renewcommand{\algorithmicrequire}{\textbf{Input:}}
\renewcommand{\algorithmicensure}{\textbf{Initialize:}}
\renewcommand{\algorithmicoutput}{\textbf{Output:}}
\begin{algorithmic}[1]
\REQUIRE $S$ and $m$.\\
\ENSURE $x^{1}_{0}=v^{1}_{0}=\widetilde{x}^{0}$, $\{w_{s}\}$, $\alpha\!>\!0$, and $\eta_{0}$.\\
\FOR{$s=1,2,\ldots,S$}
\STATE {$\widetilde{\mu}^{s}=\frac{1}{n}\!\sum^{n}_{i=1}\!\nabla\!f_{i}(\widetilde{x}^{s-\!1})$, $\eta_{s}=\eta_{0}/\max\{\alpha, 2/(s\!+\!1)\}$;}
\STATE {Option I: $v^{s}_{0}\!=\!x^{s}_{0}$, or Option II: $x^{s}_{0}\!=\!w_{s}v^{s}_{0}\!+\!(1\!-\!w_{s})\widetilde{x}^{s-\!1}$;}
\FOR{$k=0,1,\ldots,m-1$}
\STATE {Pick $i^{s}_{k}$ uniformly at random from $[n]$;}
\STATE {$\widetilde{\nabla}\! f_{i^{s}_{k}}(x^{s}_{k})=\nabla\! f_{i^{s}_{k}}(x^{s}_{k})-\nabla\! f_{i^{s}_{k}}(\widetilde{x}^{s-1})+\widetilde{\mu}^{s}$; }
\STATE {$v^{s}_{k+1}=v^{s}_{k}-\eta_{s}[\widetilde{\nabla}\!f_{i^{s}_{k}}\!(x^{s}_{k})+\nabla\! g(x^{s}_{k})]$;}
\STATE {$x^{s}_{k+1}=\widetilde{x}^{s-1}+w_{s}(v^{s}_{k+1}-\widetilde{x}^{s-1})$;}
\ENDFOR
\STATE {$\widetilde{x}^{s}=\frac{1}{m}\!\sum^{m}_{k=1}\!x^{s}_{k}$;}
\STATE {Option I: $x^{s+1}_{0}\!=x^{s}_{m}$, or Option II: $v^{s+1}_{0}\!=v^{s}_{m}$;}
\ENDFOR
\OUTPUT {$\widehat{x}^{S}=\widetilde{x}^{S}$, if $F(\widetilde{x}^{S})\!\leq \!F(\frac{1}{S}\!\sum^{S}_{s=1}\!\widetilde{x}^{s})$, and $\widehat{x}^{S}\!=\!\frac{1}{S}\!\sum^{S}_{s=1}\!\widetilde{x}^{s}$ otherwise.}
\end{algorithmic}
\end{algorithm}

\subsection{Complexity Analysis}
From Algorithm~\ref{alg2}, we can see that the per-iteration cost of VR-SGD is dominated by the computation of $\nabla\! f_{i^{s}_{k}}\!(x^{s}_{k})$, $\nabla\! f_{i^{s}_{k}}\!(\widetilde{x}^{s-1})$, and $\nabla\!g(x^{s}_{k})$ or the proximal update in~\eqref{equ22}. Thus, the complexity is $O(d)$, which is as low as that of SVRG~\cite{johnson:svrg} and Prox-SVRG~\cite{xiao:prox-svrg}. In fact, for some ERM problems, we can save the intermediate gradients $\nabla\! f_{i}(\widetilde{x}^{s-1})$ in the computation of $\widetilde{\mu}^{s}$, which generally requires $O(n)$ additional storage. As a result, each epoch only requires $(n\!+\!m)$ component gradient evaluations. In addition, for extremely sparse data, we can introduce the lazy update tricks in~\cite{koneeny:mini,carpenter:lazy,langford:online} to our algorithm, and perform the update steps in (\ref{equ21}) and (\ref{equ22}) only for the non-zero dimensions of each sample, rather than all dimensions. In other words, the per-iteration complexity of VR-SGD can be improved from $O(d)$ to $O(d')$, where $d'\!\leq\!d$ is the sparsity of feature vectors. Moreover, VR-SGD has a much lower per-iteration complexity than existing accelerated stochastic variance reduction methods such as Katyusha~\cite{zhu:Katyusha}, which have more updating steps for additional variables, as shown in~\eqref{equ061}-\eqref{equ063}.

\section{Experimental Results}
\label{sec5}
In this section, we evaluate the performance of VR-SGD for solving a number of convex and non-convex ERM problems (such as logistic regression, Lasso and ridge regression), and compare its performance with several state-of-the-art stochastic variance reduced methods (including SVRG~\cite{johnson:svrg}, Prox-SVRG~\cite{xiao:prox-svrg}, SAGA~\cite{defazio:saga}) and accelerated methods, such as Catalyst~\cite{lin:vrsg} and Katyusha~\cite{zhu:Katyusha}. Moreover, we apply VR-SGD to solve other machine learning problems, such as ERM with non-convex loss and leading eigenvalue computation.

\begin{figure}[t]
\centering
\subfigure[$\ell_{2}$-norm regularized logistic regression: $\lambda\!=\!10^{-5}$]{\includegraphics[width=0.486\columnwidth]{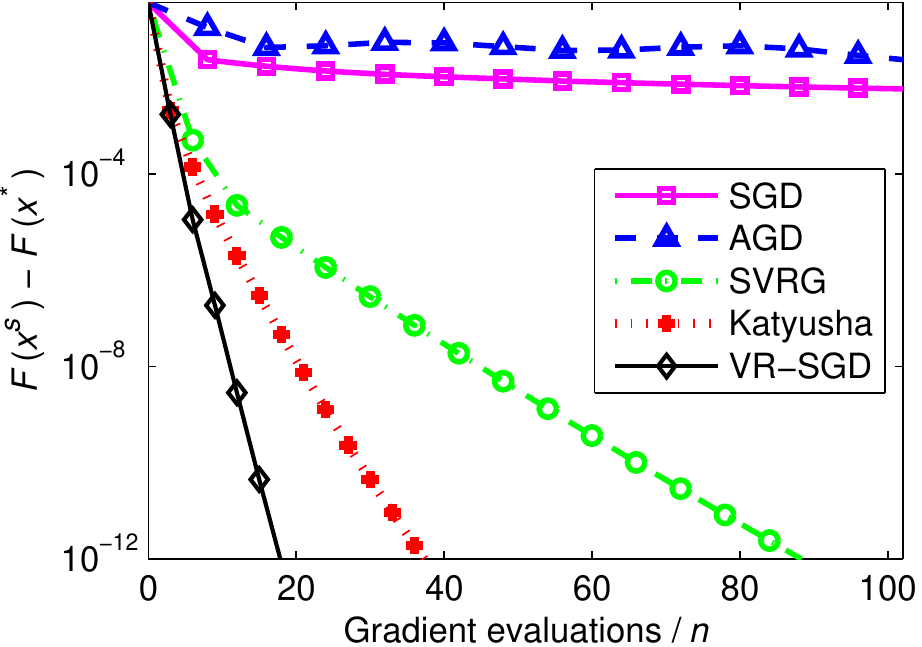}\includegraphics[width=0.486\columnwidth]{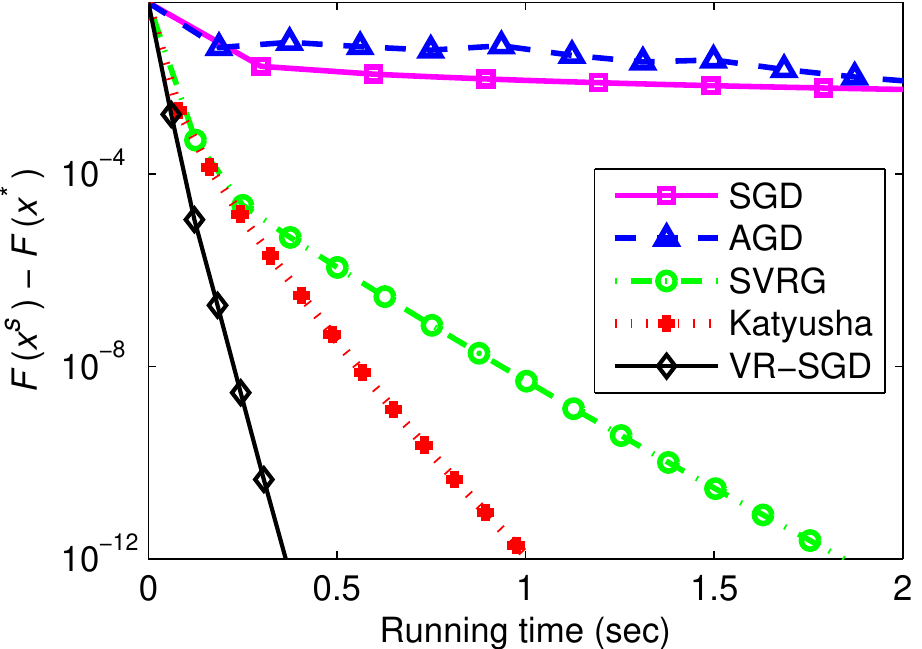}}
\subfigure[$\ell_{1}$-norm regularized logistic regression: $\lambda\!=\!10^{-5}$]{\includegraphics[width=0.486\columnwidth]{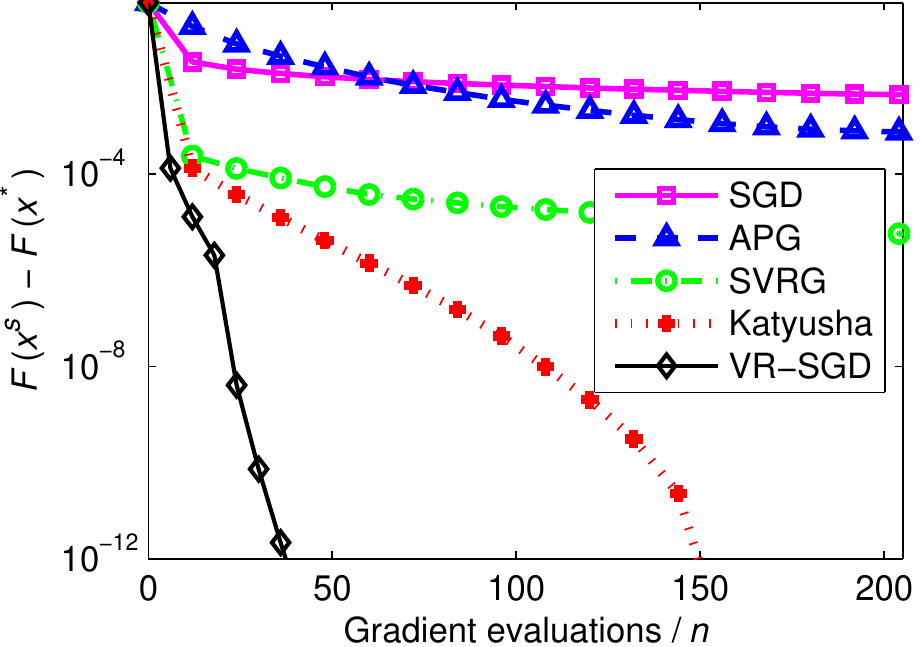}\includegraphics[width=0.486\columnwidth]{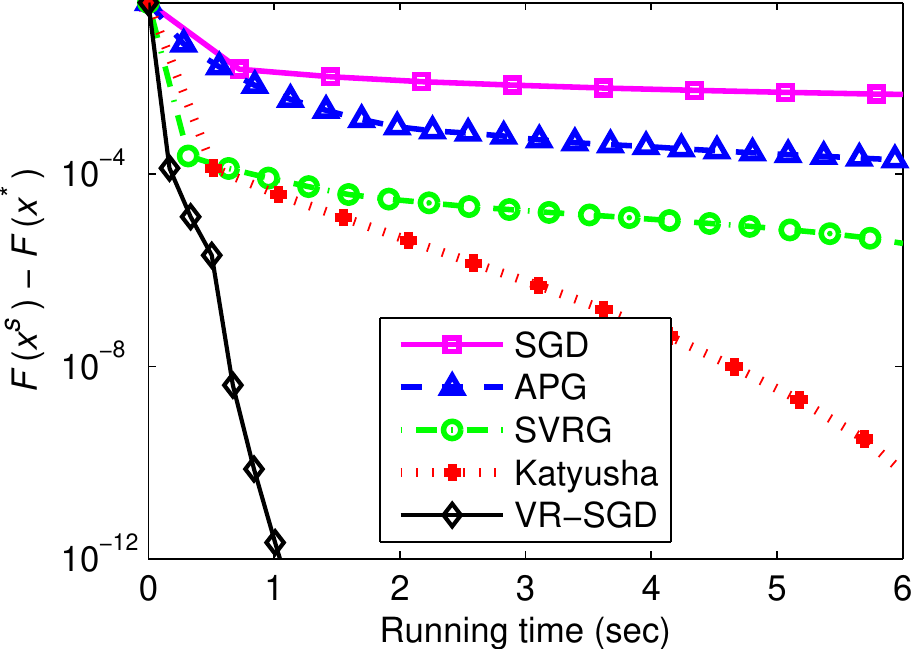}}
\caption{Comparison of deterministic and stochastic methods on Adult.}
\label{figs03}
\end{figure}

\begin{table*}[t]
\centering
\caption{The three choices of snapshot and starting points for stochastic variance reduction optimization.}
\label{tab2}
\setlength{\tabcolsep}{10.6pt}
\renewcommand\arraystretch{1.56}
\begin{tabular}{c|c|c}
\hline
Option I    & Option II  & Option III\\
\hline
\ $\widetilde{x}^{s}\!=x^{s}_{m}$ \;and\; $x^{s+1}_{0}\!=x^{s}_{m}$        & $\widetilde{x}^{s}\!=\frac{1}{m}\!\sum^{m}_{k=1}\!x^{s}_{k}$ \;and\; $x^{s+1}_{0}\!=\frac{1}{m}\!\sum^{m}_{k=1}\!x^{s}_{k}$            & $\widetilde{x}^{s}\!=\frac{1}{m}\!\sum^{m}_{k=1}\!x^{s}_{k}$ \;and\; $x^{s+1}_{0}\!=x^{s}_{m}$\\
\hline
\end{tabular}
\end{table*}

\begin{figure*}[t]
\centering
\subfigure[Ridge regression: $\lambda\!=\!10^{-5}$ (left) \;and\; $\lambda\!=\!10^{-6}$ (right)]{\includegraphics[width=0.492\columnwidth]{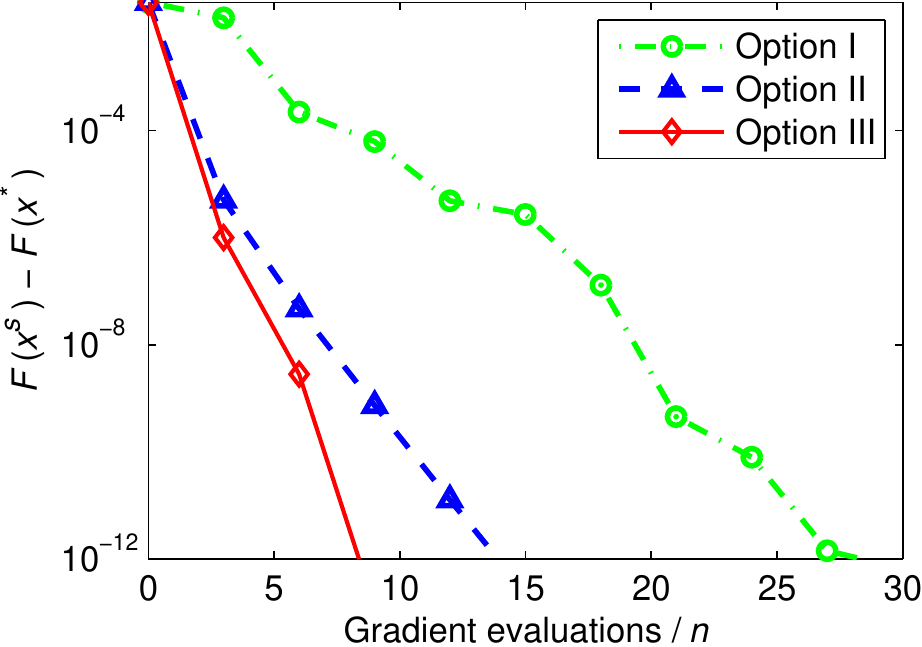}\:\includegraphics[width=0.492\columnwidth]{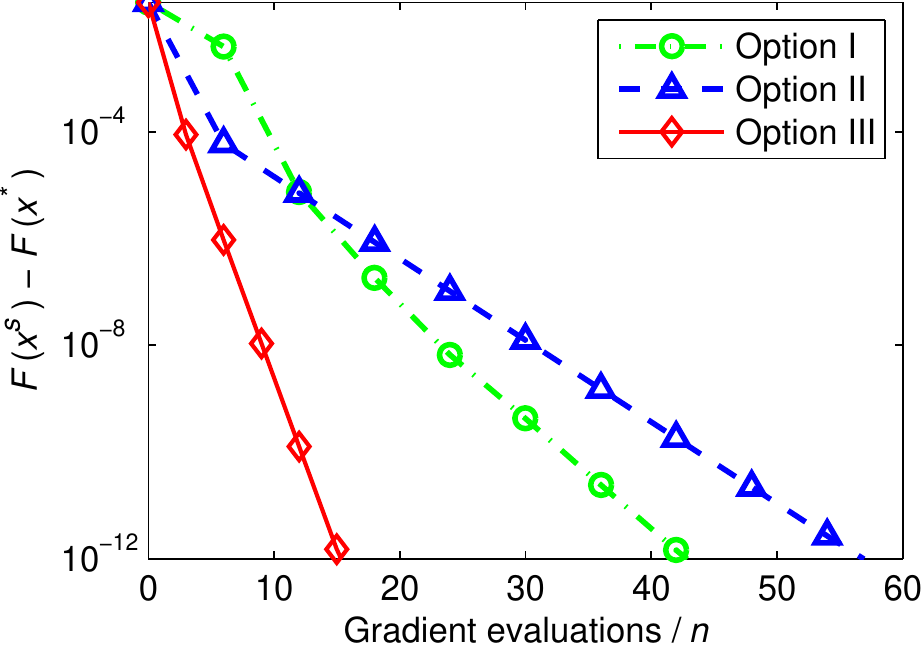}}\;\;
\subfigure[Lasso: $\lambda\!=\!10^{-4}$ (left) \;and\; $\lambda\!=\!10^{-5}$ (right)]{\includegraphics[width=0.492\columnwidth]{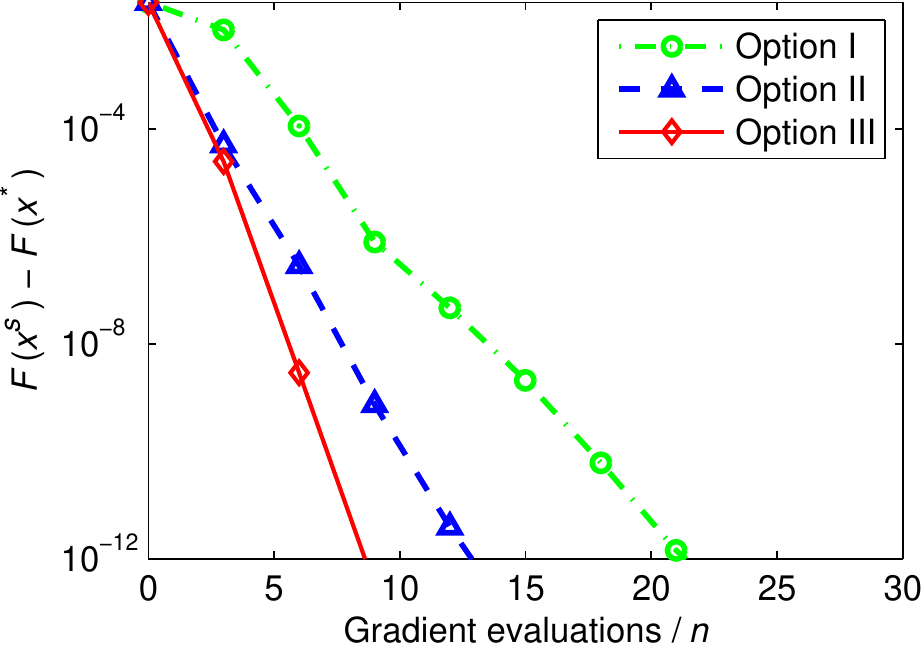}\:\includegraphics[width=0.492\columnwidth]{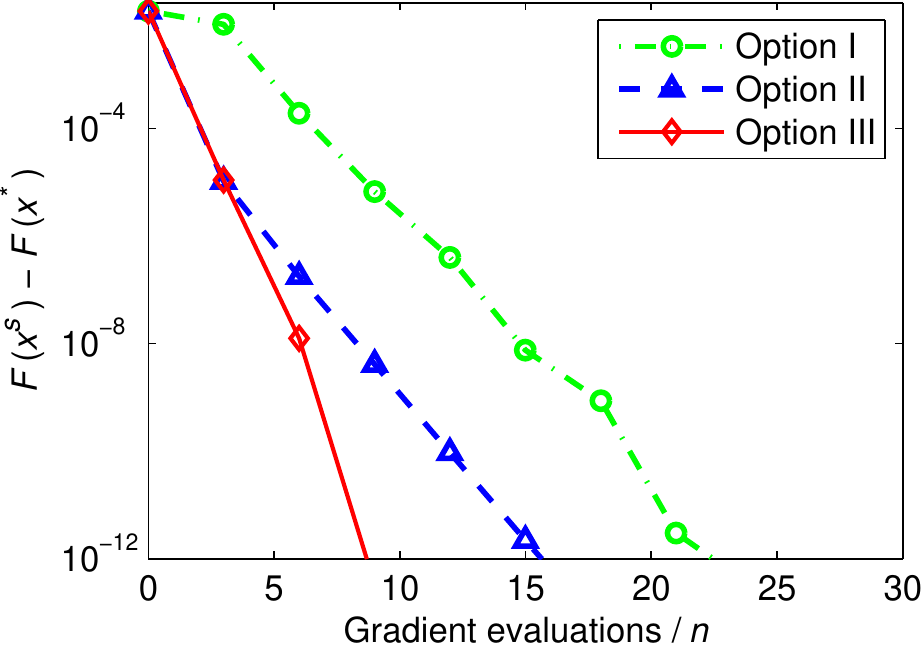}}
\caption{Comparison of the algorithms with Options I, II, and III for solving ridge regression and Lasso on Covtype. In each plot, the vertical axis shows the objective value minus the minimum, and the horizontal axis denotes the number of effective passes.}
\label{figs04}
\end{figure*}

\subsection{Experimental Setup}
We used several publicly available data sets in the experiments: Adult (also called a9a), Covtype, Epsilon, MNIST, and RCV1, all of which can be downloaded from the LIBSVM Data website\footnote{\url{https://www.csie.ntu.edu.tw/~cjlin/libsvm/}}. \emph{It should be noted that each sample of these date sets was normalized so that they have unit length as in~\cite{xiao:prox-svrg,shang:fsvrg}, which leads to the same upper bound on the Lipschitz constants $L_{i}$, i.e., $L\!=\!L_{i}$ for all $i\!=\!1,\ldots,n$}. As suggested in~\cite{johnson:svrg,xiao:prox-svrg,zhu:Katyusha}, the epoch length is set to $m\!=\!2n$ for the stochastic variance reduced methods, SVRG~\cite{johnson:svrg}, Prox-SVRG~\cite{xiao:prox-svrg}, Catalyst~\cite{lin:vrsg}, and Katyusha~\cite{zhu:Katyusha}, as well as VR-SGD. Then the only parameter we have to tune by hand is the learning rate, $\eta$. More specifically, we select learning rates from $\{10^{j},2.5\times10^{j},5\times10^{j},7.5\times10^{j},10^{j+1}\}$, where $j\!\in\!\{-2,-1,0\}$. Since Katyusha has a much higher per-iteration complexity than SVRG and VR-SGD, we compare their performance in terms of both the number of effective passes and running time (seconds), where computing a single full gradient or evaluating $n$ component gradients is considered as one effective pass over the data. For fair comparison, we implemented SVRG, Prox-SVRG, SAGA, Catalyst, Katyusha, and VR-SGD in C++ with a Matlab interface, as well as their sparse versions with lazy update tricks, and performed all the experiments on a PC with an Intel i5-4570 CPU and 16GB RAM. The source code of all the methods is available at \url{https://github.com/jnhujnhu/VR-SGD}.

\begin{figure*}[t]
\centering
\subfigure[Adult: $\lambda=10^{-3}$ (left) and $\lambda=10^{-4}$ (right)]{\includegraphics[width=0.492\columnwidth]{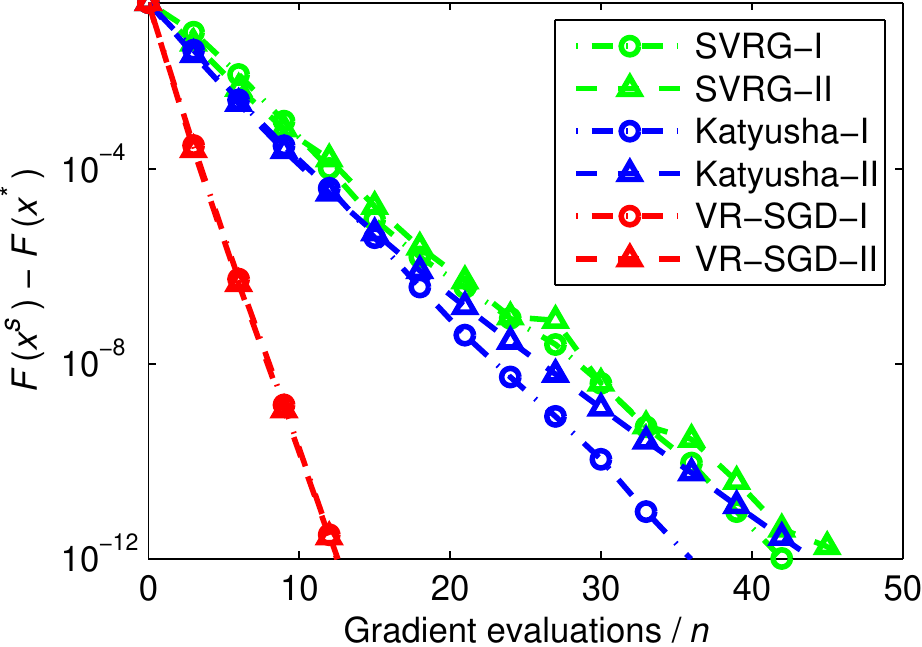}\,\includegraphics[width=0.492\columnwidth]{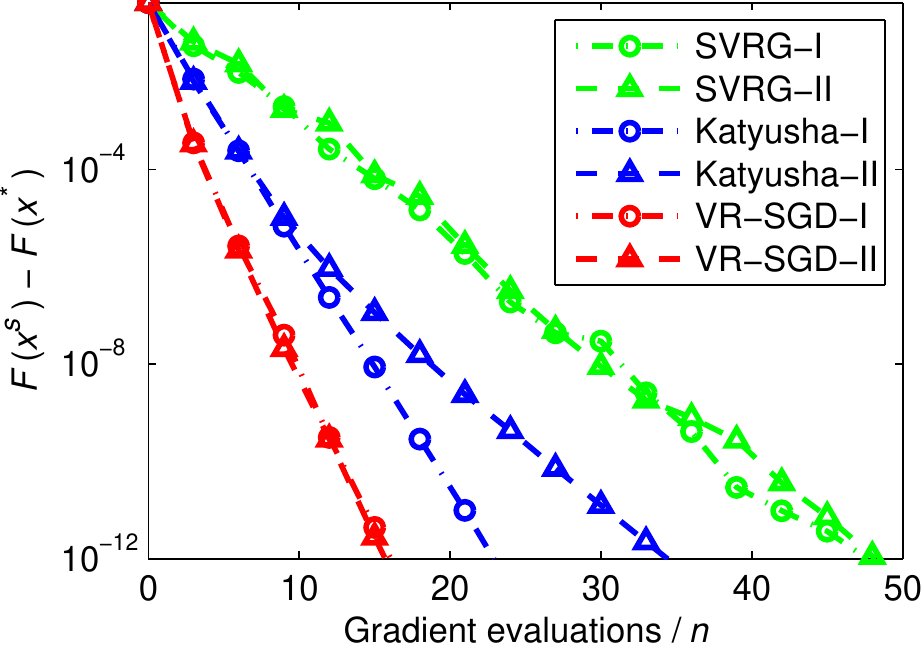}}\;\;
\subfigure[Covtype: $\lambda=10^{-5}$ (left) and $\lambda=10^{-6}$ (right)]{\includegraphics[width=0.492\columnwidth]{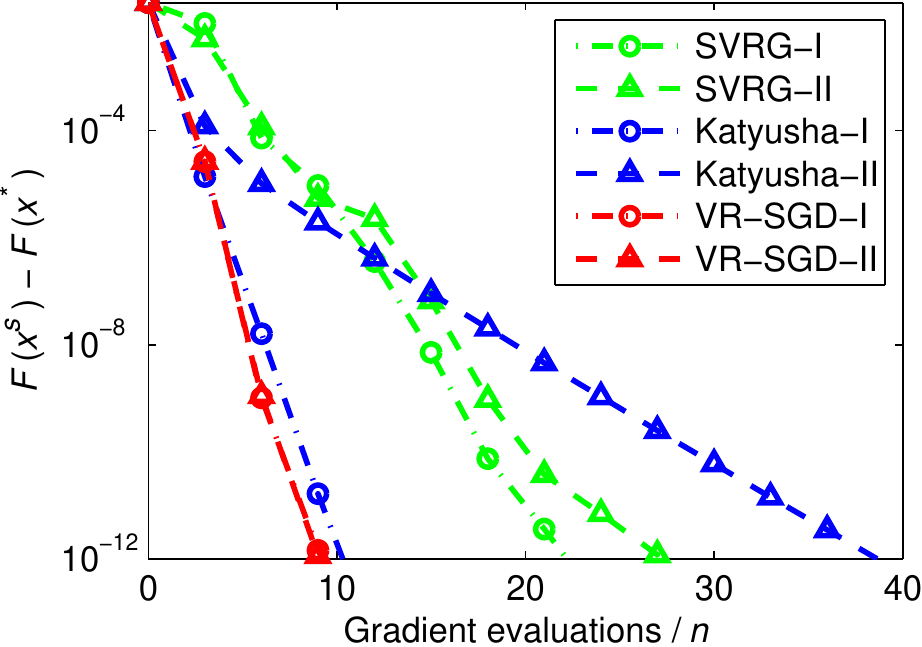}\,\includegraphics[width=0.492\columnwidth]{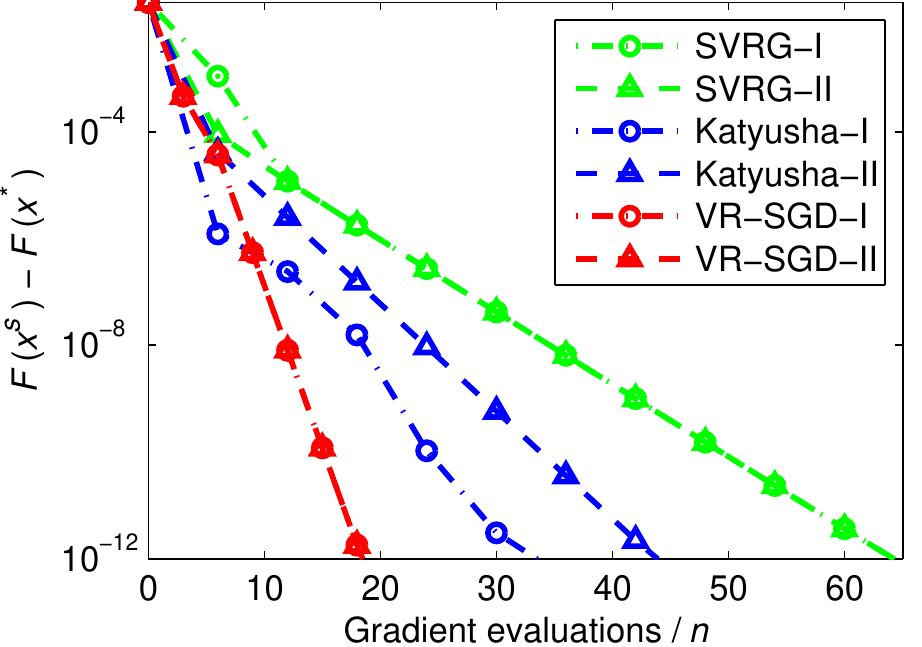}}
\caption{Comparison of SVRG~\cite{johnson:svrg}, Katyusha~\cite{zhu:Katyusha}, VR-SGD and their proximal versions for solving ridge regression problems. In each plot, the vertical axis shows the objective value minus the minimum, and the horizontal axis is the number of effective passes over data.}
\label{figs05}
\end{figure*}

\subsection{Deterministic Methods vs.\ Stochastic Methods}
In this subsection, we compare the performance of stochastic methods (including SGD, SVRG, Katyusha, and VR-SGD) with that of deterministic methods such as AGD~\cite{nesterov:fast,nesterov:co} and APG~\cite{beck:fista} for solving strongly and non-strongly convex problems. Note that the important momentum parameter $w$ of AGD is $w\!=\!(\sqrt{L}\!-\!\sqrt{\mu})/(\sqrt{L}\!+\!\sqrt{\mu})$ as in~\cite{su:nag}, while that of APG is defined as follows: $w_{k}=(\alpha_{k}\!-\!1)/\alpha_{k+\!1}$ for all $k\!\geq\!1$ \cite{beck:fista}, where $\alpha_{k+\!1}\!=\!(1\!+\!\sqrt{1\!+\!4\alpha^{2}_{k}})/2$, and $\alpha_{1}\!=\!1$.

Fig.\ \ref{figs03} shows the the objective gap (i.e., $F(x^{s})\!-\!F(x^{*})$) of those deterministic and stochastic methods for solving $\ell_{2}$-norm and $\ell_{1}$-norm regularized logistic regression problems (see the Supplementary Material for more results). It can be seen that the accelerated deterministic methods and SGD have similar convergence speed, and APG usually performs slightly better than SGD for non-strongly convex problems. The variance reduction methods (e.g., SVRG, Katyusha and VR-SGD) significantly outperform the accelerated deterministic methods and SGD for both strongly and non-strongly convex cases, suggesting the importance of variance reduction techniques. Although accelerated deterministic methods have a faster theoretical speed than SVRG for general convex problems, as discussed in Section~\ref{sec12}, APG converges much slower in practice. VR-SGD consistently outperforms the other methods (including Katyusha) in all the settings, which verifies the effectiveness of VR-SGD.

\subsection{Different Choices for Snapshot and Starting Points}
\label{sec52}
In the practical implementation of SVRG~\cite{johnson:svrg}, both the snapshot $\widetilde{x}^{s}$ and starting point $x^{s+\!1}_{0}$ in each epoch are set to the last iterate $x^{s}_{m}$ of the previous epoch (i.e., Option I in Algorithm~\ref{alg1}), while the two vectors in~\cite{xiao:prox-svrg} are set to the average point of the previous epoch, $\frac{1}{m}\!\sum^{m}_{k=1}\!x^{s}_{k}$ (i.e., Option II in Algorithm~\ref{alg1}). In contrast, $\widetilde{x}^{s}$ and $x^{s+\!1}_{0}$ in our algorithm are set to $\frac{1}{m}\!\sum^{m}_{k=1}\!x^{s}_{k}$ and $x^{s}_{m}$ (denoted by Option III, i.e., Option I{\footnote{As Options I and II in Algorithm~\ref{alg2} achieve very similar performance, we only report the results of our algorithm with Option I.}} in Algorithm~\ref{alg2}), respectively.

We compare the performance of the algorithms with the three settings (i.e., the Options I, II and III listed in Table~\ref{tab2}) for solving ridge regression and Lasso problems, as shown in Fig.\ \ref{figs04} (see the Supplementary Material for more results). Except for the three different settings for snapshot and starting points, we use the update rules in (\ref{equ21}) and (\ref{equ22}) for ridge regression and Lasso problems, respectively. We can see that the algorithm with Option III (i.e., Algorithm~\ref{alg2} with Option I) consistently converges much faster than the algorithms with Options I and II for both strongly convex and non-strongly convex cases. This indicates that the setting of Option III suggested in this paper is a better choice than Options I and II for stochastic optimization.

\subsection{Common SG Updates vs.\ Prox-SG Updates}
\label{sec54}
In this subsection, we compare the original Katyusha algorithm in~\cite{zhu:Katyusha} with the slightly modified Katyusha algorithm (denoted by Katyusha-I). In Katyusha-I, only the following two update rules are used to replace the original proximal stochastic gradient update rules in~\eqref{equ062} and \eqref{equ063}.
\begin{equation}\label{equ53}
\begin{split}
&y^{s}_{k+1}=y^{s}_{k}-\eta [\widetilde{\nabla}\! f_{i_{k}}\!(x^{s}_{k+1})+\nabla\!g(x^{s}_{k+1})],\\
&z^{s}_{k+1}=x^{s}_{k+1}-[\widetilde{\nabla}\! f_{i_{k}}\!(x^{s}_{k+1})+\nabla\!g(x^{s}_{k+1})]/(3L).
\end{split}
\end{equation}
Similarly, we also implement the proximal versions{\footnote{Here, the proximal variant of SVRG is different from Prox-SVRG~\cite{xiao:prox-svrg}, and their main difference is the choices of both the snapshot point and starting point. That is, the two vectors of the former are set to the last iterate $x^{s}_{m}$, while those of Prox-SVRG are set to the average point of the previous epoch, i.e., $\frac{1}{m}\!\sum^{m}_{k=1}\!x^{s}_{k}$.} for the original SVRG (called SVRG-I) and VR-SGD (denoted by VR-SGD-I) methods, and denote their proximal variants by SVRG-II and VR-SGD-II, respectively. In addition, the original Katyusha is denoted by Katyusha-II.

\begin{figure}[t]
\centering
\subfigure[$\lambda=10^{-5}$]{\includegraphics[width=0.486\columnwidth]{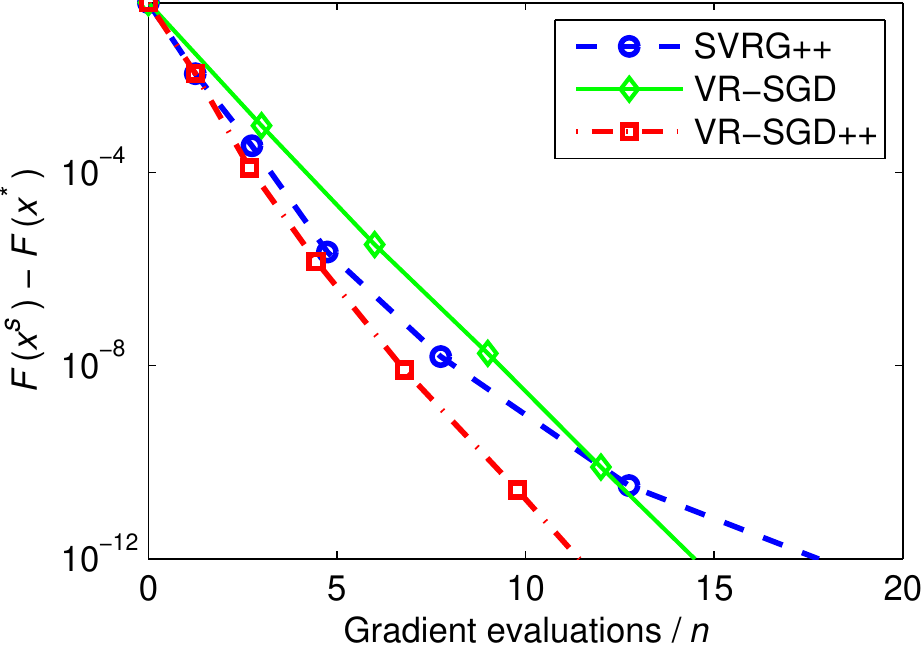}}
\subfigure[$\lambda=10^{-6}$]{\includegraphics[width=0.486\columnwidth]{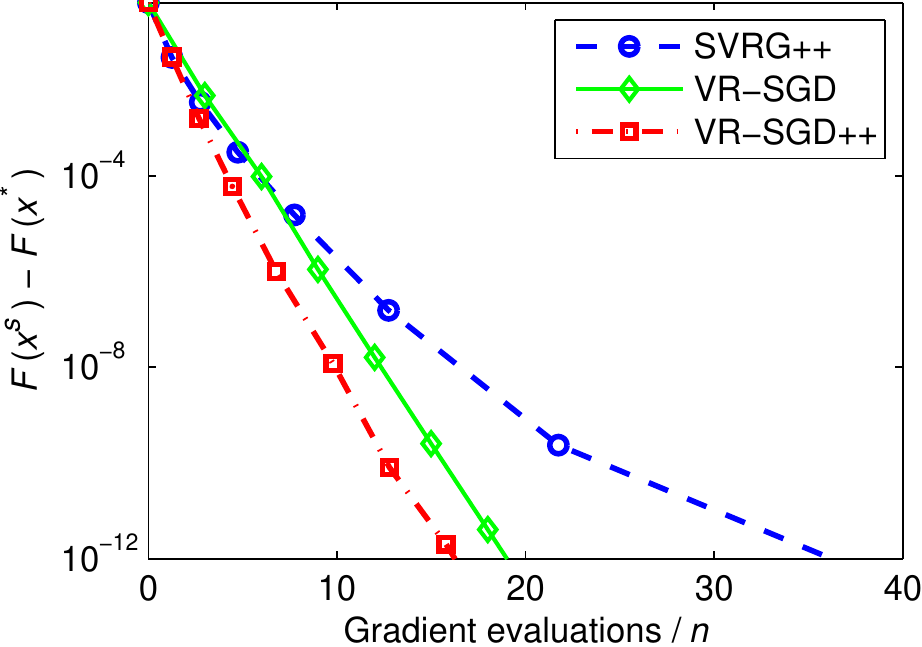}}
\caption{Comparison of SVRG++~\cite{zhu:univr}, VR-SGD and VR-SGD++ for solving logistic regression problems on Epsilon.}
\label{figs06}
\end{figure}

\begin{figure*}[t]
\centering
\subfigure[Epsilon: $\lambda_{1}=10^{-6}$ and $\lambda_{2}=0$]{\includegraphics[width=0.492\columnwidth]{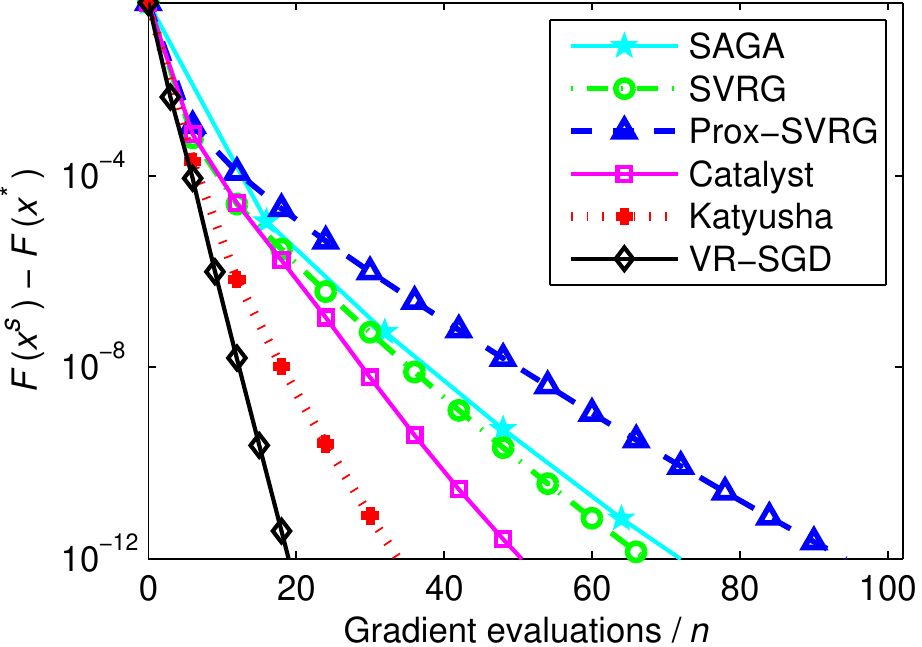}\:\includegraphics[width=0.492\columnwidth]{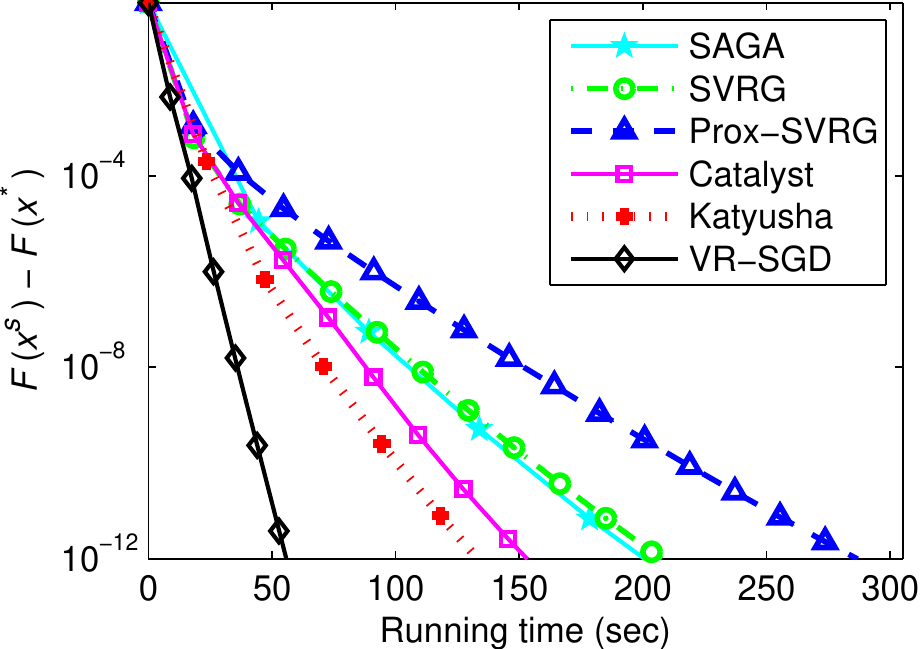}}\;\;
\subfigure[RCV1: $\lambda_{1}=10^{-5}$ and $\lambda_{2}=0$]{\includegraphics[width=0.492\columnwidth]{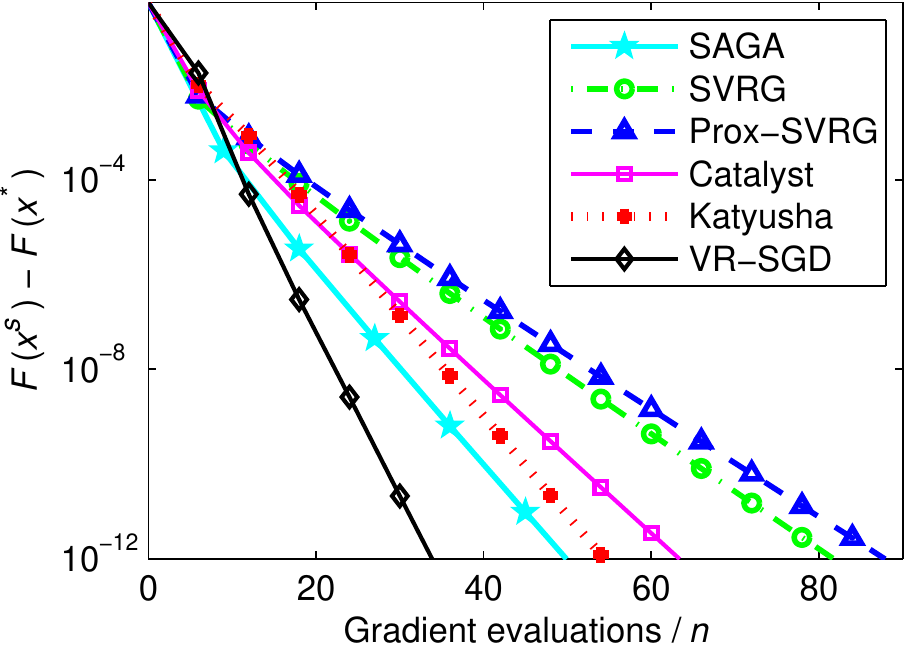}\:\includegraphics[width=0.492\columnwidth]{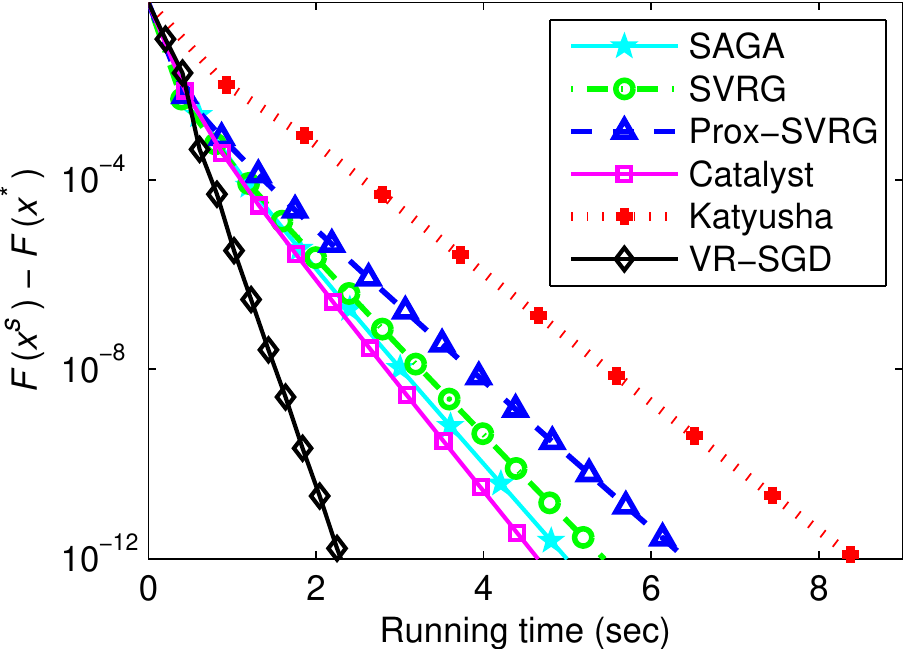}\label{figs07b}}

\subfigure[Covtype: $\lambda_{1}=0$ and $\lambda_{2}=10^{-5}$]{\includegraphics[width=0.492\columnwidth]{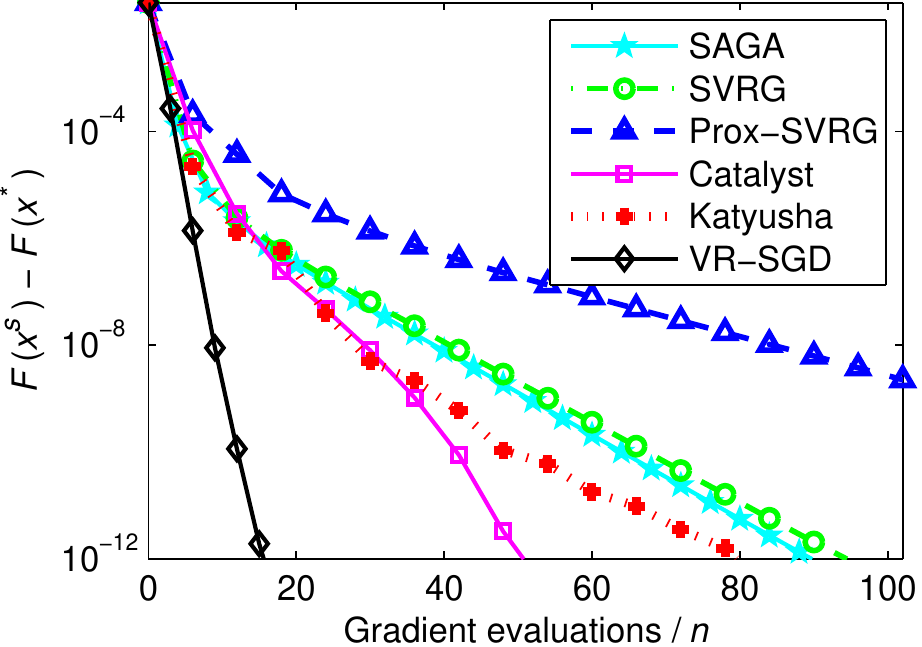}\:\includegraphics[width=0.492\columnwidth]{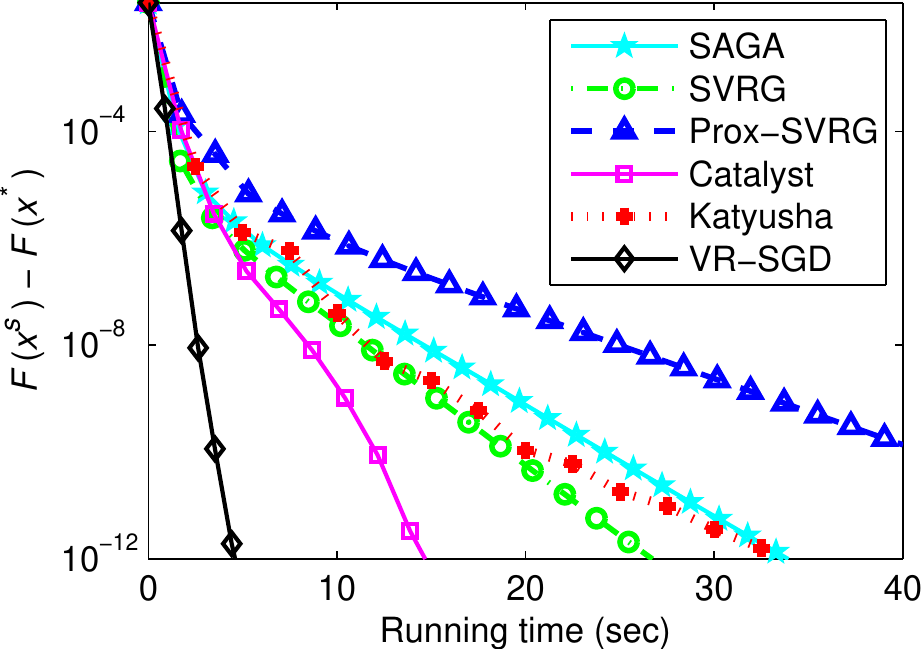}}\;\;\subfigure[Adult: $\lambda_{1}=10^{-6}$ \;and\; $\lambda_{2}=10^{-5}$]{\includegraphics[width=0.492\columnwidth]{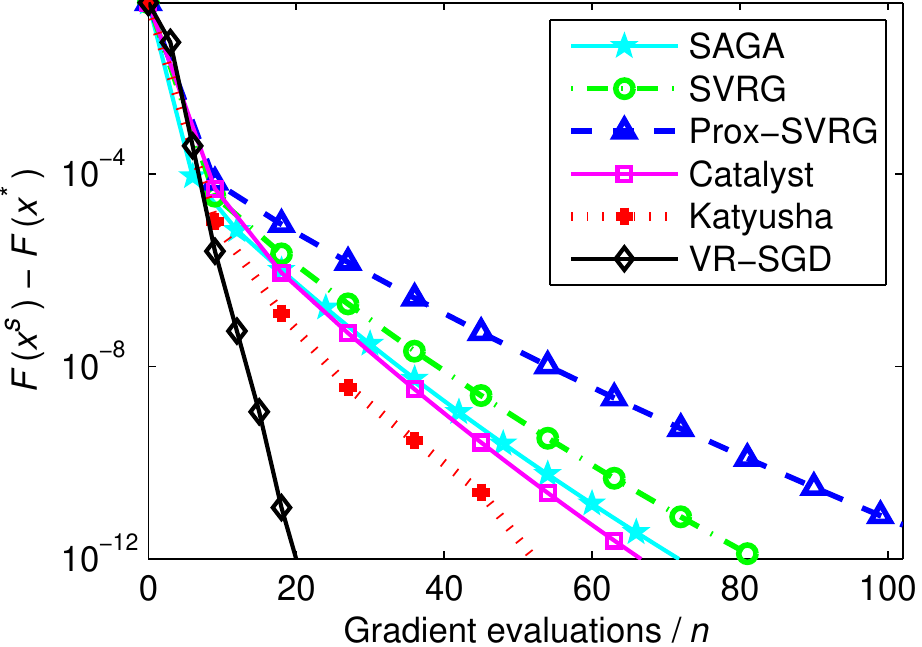}\:\includegraphics[width=0.492\columnwidth]{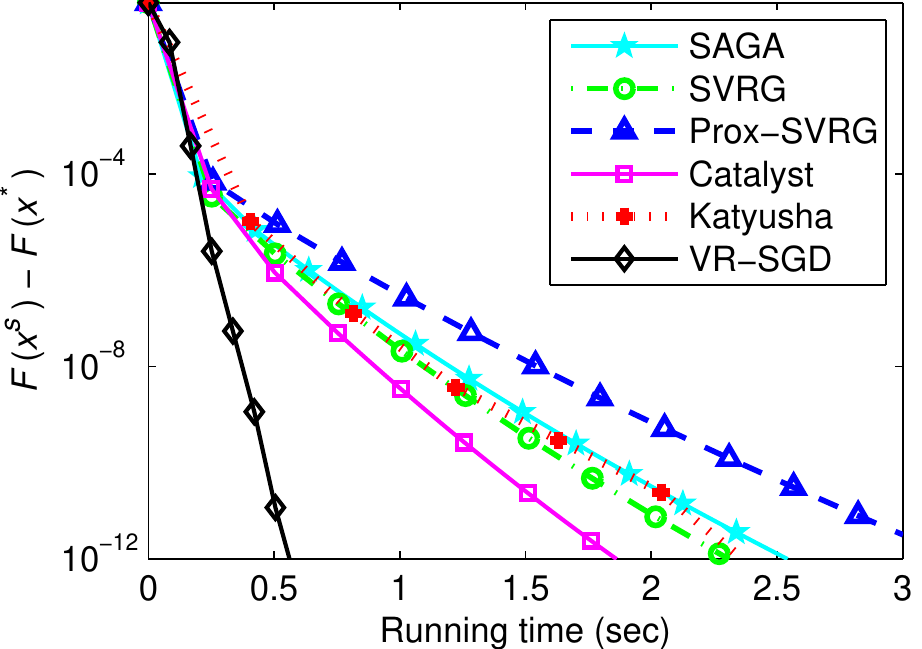}}
\caption{Comparison of SAGA~\cite{defazio:saga}, SVRG~\cite{johnson:svrg}, Prox-SVRG~\cite{xiao:prox-svrg}, Catalyst~\cite{lin:vrsg}, Katyusha~\cite{zhu:Katyusha}, and VR-SGD for solving $\ell_{2}$-norm (the first row), $\ell_{1}$-norm (c), and elastic net (d) regularized logistic regression problems. In each plot, the vertical axis shows the objective value minus the minimum, and the horizontal axis is the number of effective passes (left) or running time (right).}
\label{figs07}
\end{figure*}

Fig.\ \ref{figs05} shows the performance of Katyusha-I and Katyusha-II for solving ridge regression on the two popular data sets: Adult and Covtype. We also report the results of SVRG, VR-SGD, and their proximal variants. It is clear that Katyusha-I usually performs better than Katyusha-II (i.e., the original Katyusha~\cite{zhu:Katyusha}), and converges significantly faster in the case when the regularization parameter is $10^{-4}$ or $10^{-6}$. This seems to be the main reason why Katyusha has inferior performance when the regularization parameter is relatively large, as shown in Section~\ref{sec53}. In contrast, VR-SGD and its proximal variant have similar performance, and the former slightly outperforms the latter in most cases (similar results are also observed for SVRG vs. its proximal variant). This suggests that stochastic gradient update rules as in (\ref{equ21}) and (\ref{equ53}) are better choices than proximal update rules as in (\ref{equ14}), \eqref{equ062} and \eqref{equ063} for smooth objective functions. We also believe that our new insight can help us to design accelerated stochastic optimization methods.

Both Katyusha-I and Katyusha-II usually outperform SVRG and its proximal variant, especially when the regularization parameter is relatively small, e.g., $\lambda\!=\!10^{-6}$. Moreover, it can be seen that both VR-SGD and its proximal variant achieve much better performance than the other methods in most cases, and are also comparable to Katyusha-I and Katyusha-II in the remaining cases. This further verifies that VR-SGD is suitable for various large-scale machine learning.

\subsection{Growing Epoch Size Strategy in Early Iterations}
\label{sec55}
In this subsection, we present a general growing epoch size strategy in early iterations (i.e., If $m_{s}\!<\!2n$, $m_{s+1}\!=\!\lfloor\rho m_{s}\rfloor$ with the factor $\rho\!>\!1$. Otherwise, $m_{s+1}\!=\!m_{s}$). Different from the doubling-epoch technique used in SVRG++~\cite{zhu:univr} (i.e., $m_{s+1}\!=\!2m_{s}$), we gradually increase the epoch size in only the early iterations. Similar to the convergence analysis in Section~\ref{sec4}, VR-SGD with the growing epoch size strategy (called VR-SGD++) can be guaranteed to converge. As suggested in~\cite{zhu:univr}, we set $m_{1}\!=\!\lfloor n/4\rfloor$ for both SVRG++ and VR-SGD++, and $\rho\!=\!1.75$ for VR-SGD++. Note that they use the same initial learning rate. We compare their performance for solving $\ell_{2}$-norm regularized logistic regression, as shown in Fig.\ \ref{figs06} (see the Supplementary Material for more results). All the results show that VR-SGD++ converges faster than VR-SGD, which means that reducing the number of gradient calculations in early iterations can lead to faster convergence as discussed in~\cite{babanezhad:vrsg}. Moreover, both VR-SGD++ and VR-SGD significantly outperform SVRG++, especially when the regularization parameter is relatively small, e.g., $\lambda\!=\!10^{-6}$.

\begin{figure*}[!th]
\centering
\includegraphics[width=0.492\columnwidth]{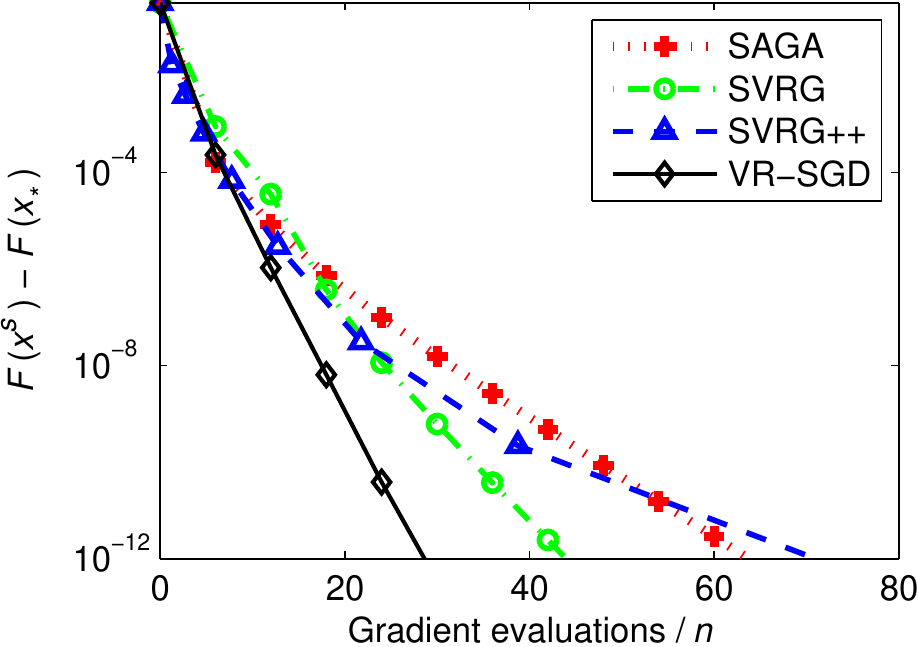}
\includegraphics[width=0.492\columnwidth]{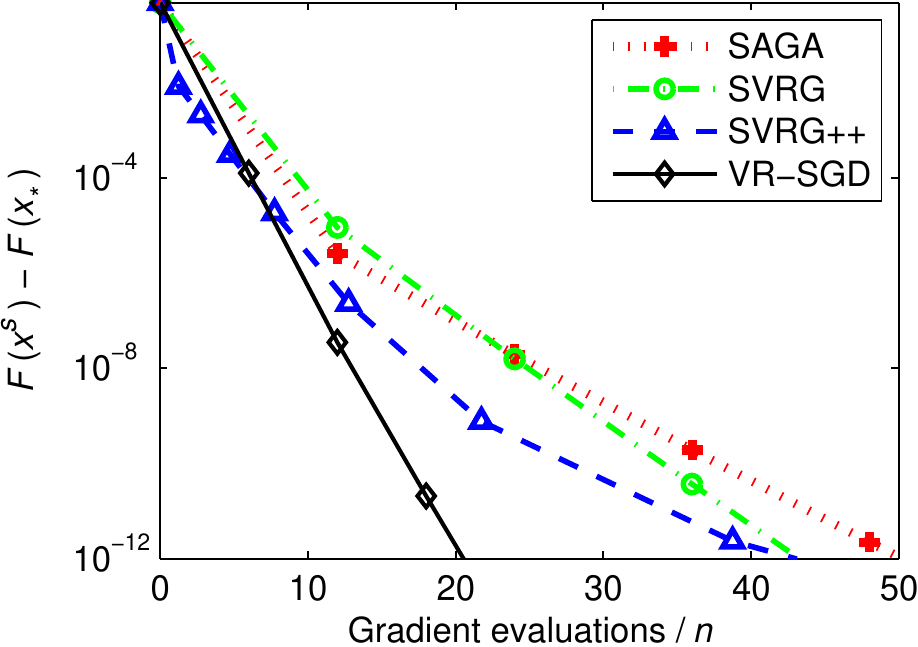}
\includegraphics[width=0.492\columnwidth]{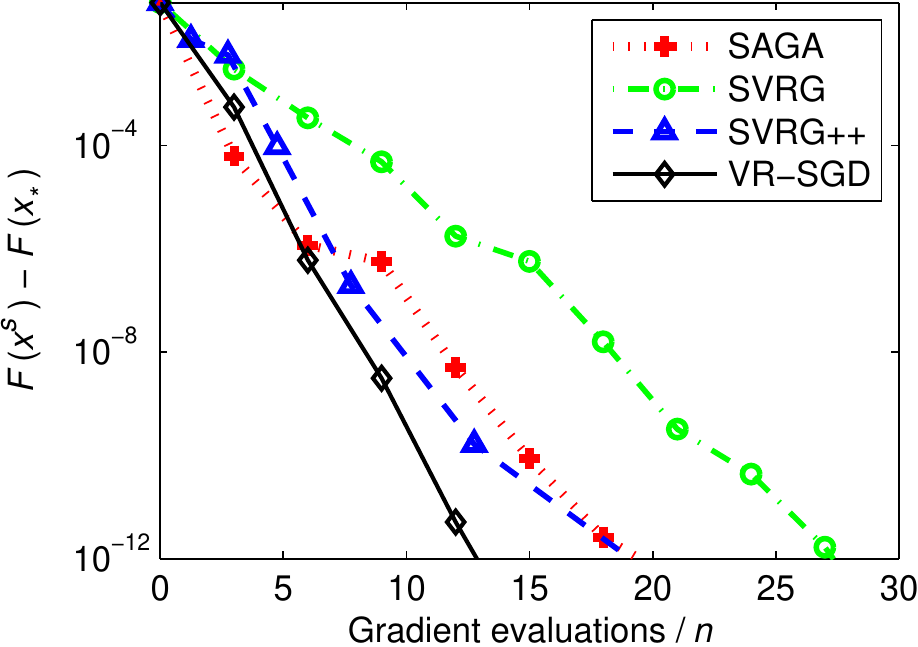}
\includegraphics[width=0.492\columnwidth]{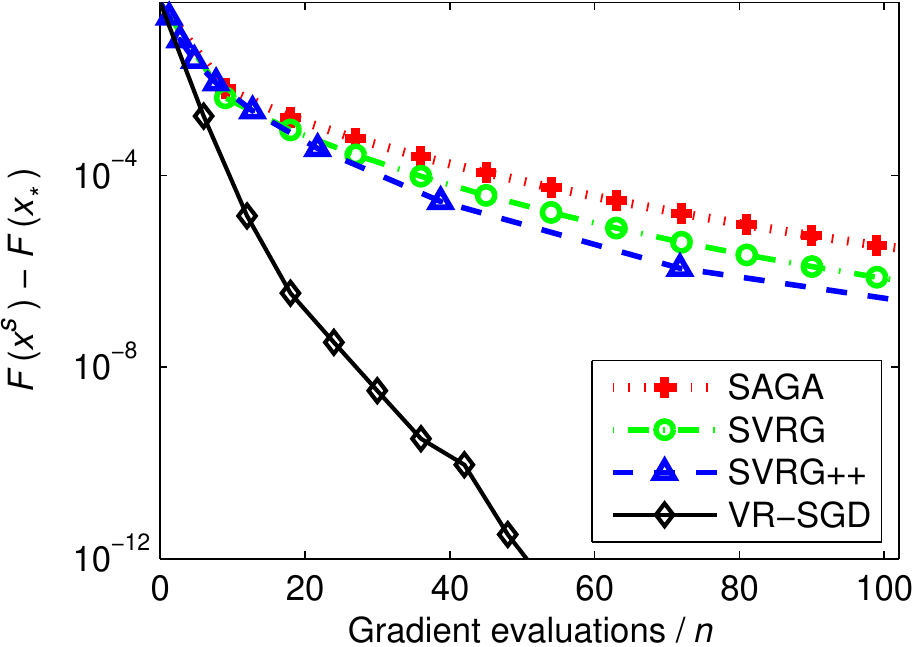}

\subfigure[Adult]{\includegraphics[width=0.492\columnwidth]{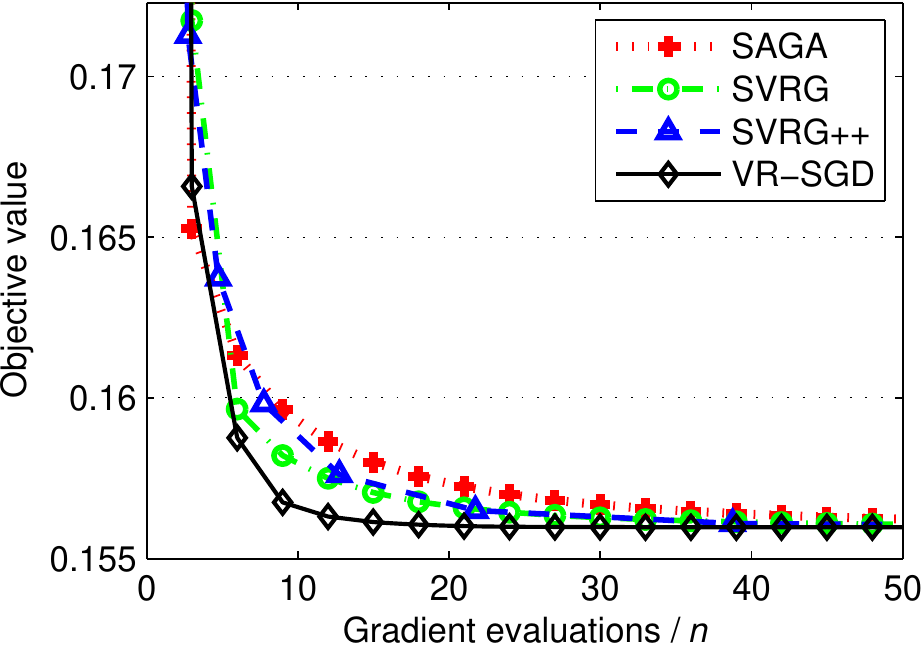}}
\subfigure[MNIST]{\includegraphics[width=0.492\columnwidth]{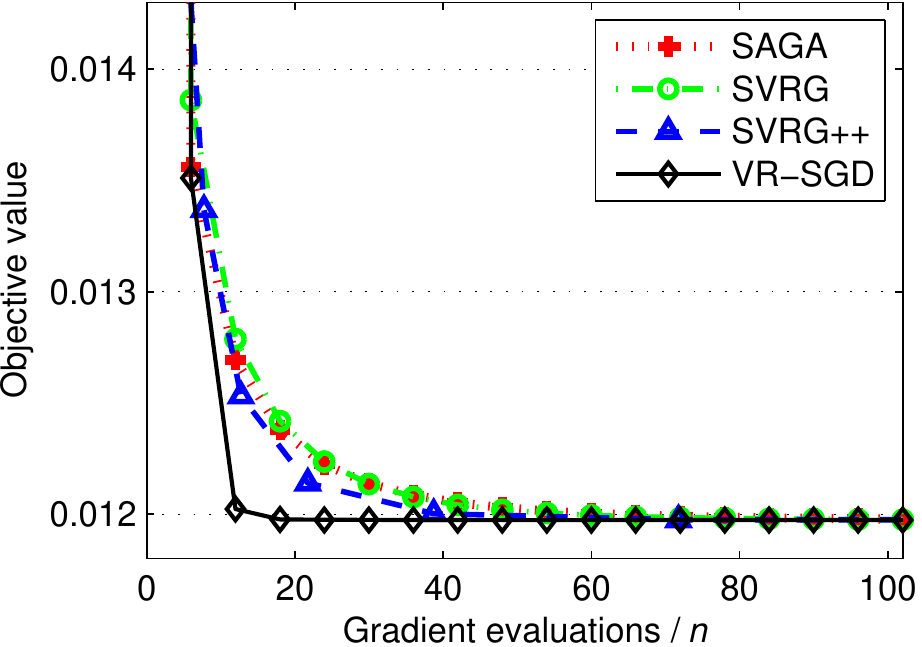}}
\subfigure[Covtype]{\includegraphics[width=0.492\columnwidth]{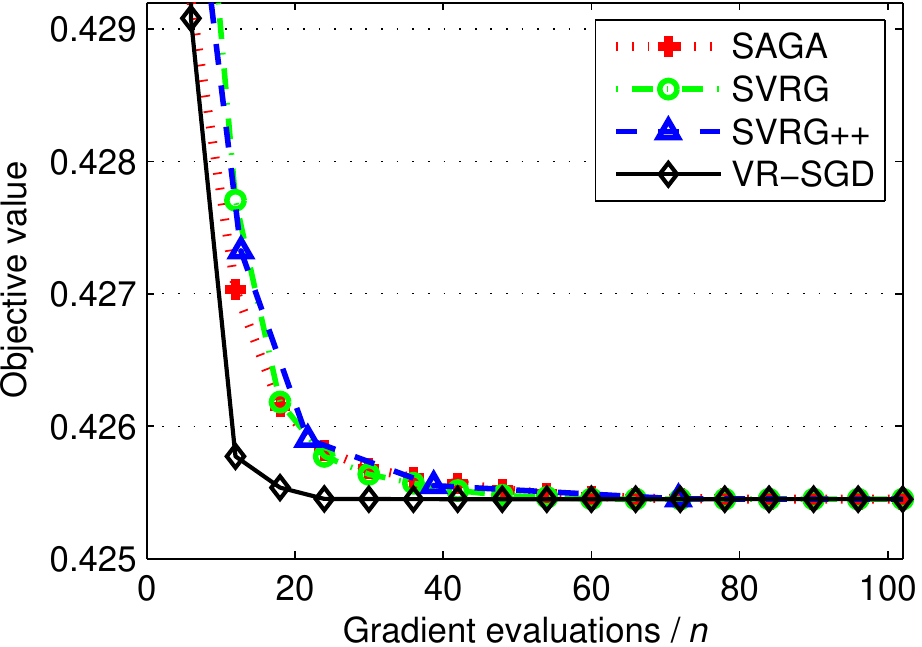}}
\subfigure[RCV1]{\includegraphics[width=0.492\columnwidth]{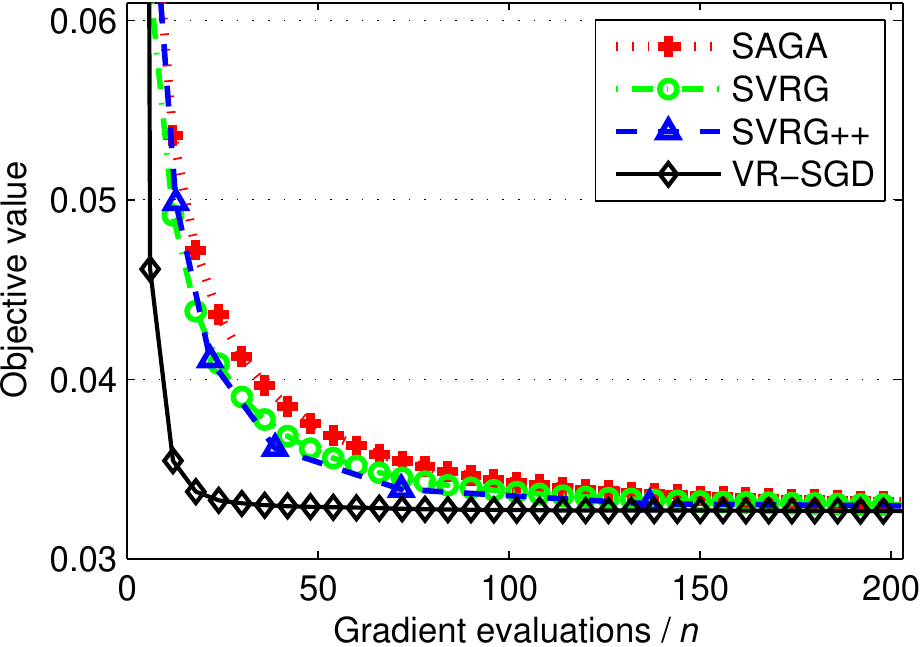}}
\caption{Comparison of SAGA~\cite{reddi:saga}, SVRG~\cite{zhu:vrnc}, SVRG++~\cite{zhu:univr}, and VR-SGD for solving non-convex ERM problems with sigmoid loss: $\lambda=10^{-5}$ (top) and $\lambda=10^{-6}$ (bottom). Note that $x_{*}$ denotes the best solution obtained by running all those methods for a large number of iterations and multiple random initializations.}
\label{figs08}
\end{figure*}

\subsection{Real-world Applications}
In this subsection, we apply VR-SGD to solve a number of machine learning problems, e.g., logistic regression, non-convex ERM and eigenvalue computation.

\subsubsection{Convex Logistic Regression}
\label{sec53}
In this part, we focus on the following generalized logistic regression problem for binary classification,
\begin{equation}\label{equ54}
\min_{x\in\mathbb{R}^{d}}\frac{1}{n}\sum^{n}_{i=1}\log(1+\exp(-b_{i}a^{T}_{i}x))+\frac{\lambda_{1}}{2}\|x\|^{2}+\lambda_{2}\|x\|_{1}
\end{equation}
where $\{(a_{i},b_{i})\}$ is a set of training examples, and $\lambda_{1},\lambda_{2}\!\geq\!0$ are the regularization parameters. Note that when $\lambda_{2}\!>\!0$, $f_{i}(x)\!=\!\log(1\!+\!\exp(-b_{i}a^{T}_{i}x))\!+\!(\lambda_{1}/{2})\|x\|^{2}$. The formulation (\ref{equ54}) includes the $\ell_{2}$-norm (i.e., $\lambda_{2}\!=\!0$),  $\ell_{1}$-norm (i.e., $\lambda_{1}\!=\!0$), and elastic net (i.e., $\lambda_{1}\!\neq\!0$ and $\lambda_{2}\!\neq\!0$) regularized logistic regression problems. Fig.\ \ref{figs07} shows how the objective gap decreases for the $\ell_{2}$-norm, $\ell_{1}$-norm, and elastic-net regularized logistic regression problems, respectively (see the Supplementary Material for more results). From all the results, we make the following observations.
\begin{itemize}
\item When the problems are well-conditioned (e.g., $\lambda_{1}\!=\!10^{-4}$ or $\lambda_{2}\!\!=\!\!10^{-4}$), Prox-SVRG usually converges faster than SVRG for both strongly convex (e.g., $\ell_{2}$-norm regularized logistic regression) and non-strongly convex (e.g., $\ell_{1}$-norm regularized logistic regression) cases. On the contrary, SVRG often outperforms Prox-SVRG, when the problems are ill-conditioned, e.g., $\lambda_{1}\!\!=\!\!10^{-6}$ or $\lambda_{2}\!\!=\!\!10^{-6}$ (see Figs.\ 11 and 12 in the Supplementary Material). The main reason is that they have different initialization settings, i.e., $\widetilde{x}^{s}\!\!=\!\!x^{s}_{m}$ and $x^{s+\!1}_{0}\!\!=\!\!x^{s}_{m}$ for SVRG vs.\ $\widetilde{x}^{s}=\!\;\frac{1}{m}\!\sum^{m}_{k=1}\!x^{s}_{k}$ and $x^{s+\!1}_{0}\!\!=\!\!\frac{1}{m}\!\sum^{m}_{k=1}\!x^{s}_{k}$ for Prox-SVRG.
\item Katyusha converges much faster than SAGA, SVRG, Prox-SVRG, and Catalyst in the cases when the problems are ill-conditioned, e.g., $\lambda_{1}\!=\!10^{-6}$, whereas it often achieves similar or inferior performance when the problems are well-conditioned, e.g., $\lambda_{1}\!=\!10^{-4}$ (see Figs.\ 11, 12 and 13 in the Supplementary Material). Note that we implemented the original algorithm with Option I in~\cite{zhu:Katyusha} for Katyusha. Obviously, the above observation matches the convergence properties of Katyusha provided in~\cite{zhu:Katyusha}, that is, only if $m\mu/L\!\leq\!3/4$, Katyusha attains the oracle complexity of $\mathcal{O}((n\!+\!\!\sqrt{n{L}/{\mu}})\log({1}/{\epsilon}))$ for strongly convex problems.
\item VR-SGD converges significantly faster than SAGA, SVRG, Prox-SVRG and Catalyst, especially when the problems are ill-conditioned, e.g., $\lambda_{1}\!=\!10^{-6}$ or $\lambda_{2}\!=\!10^{-6}$ (see Figs.\ 11 and 12 in the Supplementary Material). The main reason is that VR-SGD can use much larger learning rates than them (e.g., $1/L$ for VR-SGD vs.\ $1/(10L)$ for SVRG), which leads to faster convergence. This further verifies that the settings of both snapshot and starting points in our algorithm (i.e., Algorithm~\ref{alg2}) are better choices than Options I and II in Algorithm~\ref{alg1}.
\item In particular, VR-SGD consistently outperforms the best-known stochastic method, Katyusha, in terms of the number of passes through the data, especially when the problems are well-conditioned, e.g., $10^{-4}$ and $10^{-5}$ (see Figs.\ 11 and 12 in the Supplementary Material). Since VR-SGD has a much lower per-iteration complexity than Katyusha, VR-SGD has more obvious advantage over Katyusha in terms of running time, especially in the case of sparse data (e.g., RCV1), as shown in Fig.\ \ref{figs07b}. From the algorithms of Katyusha proposed in~\cite{zhu:Katyusha}, we can see that the learning rate of Katyusha is at least set to $1/(3L)$. Similarly, the learning rate used in VR-SGD is comparable to that of Katyusha, which may be the main reason why the performance of VR-SGD is much better than that of Katyusha. This also implies that the algorithms (including VR-SGD) that enjoy larger learning rates can converge faster in general.
\end{itemize}

\subsubsection{ERM with Non-Convex Loss}
In this part, we apply VR-SGD to solve the following regularized ERM problem with non-convex sigmoid loss:
\begin{equation}
\min_{x\in\mathbb{R}^{d}}\;\frac{1}{n}\sum^{n}_{i=1}f_{i}(x)+\frac{\lambda}{2}\|x\|^{2}
\end{equation}
where $f_{i}(x)=1/[1+\exp(b_{i}a^{T}_{i}x)]$. Some work~\cite{zhu:vrnc,shalev:loss} has shown that the sigmoid function usually generalizes better than some other loss functions (such as squared loss, logistic loss and hinge loss) in terms of test accuracy especially when there are outliers. Here, we consider binary classification on the four data sets: Adult, MNIST, Covtype and RCV1. Note that we only consider classifying the first class in MNIST.

We compare the performance (including training objective value and function suboptimality, i.e., $F(x^{s})\!-\!F(x_{*})$) of VR-SGD with that of SAGA~\cite{reddi:saga}, SVRG~\cite{zhu:vrnc}, and SVRG++~\cite{zhu:univr}, as shown in Fig.\ \ref{figs08} (more results are provided in the Supplementary Material), where $x_{*}$ denotes the best solution obtained by running all those methods for a large number of iterations and multiple random initializations. Note that both SAGA and SVRG are two variants of the original SAGA~\cite{defazio:saga} and SVRG~\cite{johnson:svrg}. The results show that our VR-SGD method has faster convergence than the other methods, and its objective value is much lower. This implies that VR-SGD can yield much better solutions than the other methods including SVRG++. Furthermore, we can see that VR-SGD has much greater advantage over the other methods in the cases when the smaller $\lambda$ is, which means that the objective function becomes more ``non-convex''.

Moreover, we report the classification testing accuracies of all those methods on the test sets of Adult and MNIST in Fig.\ \ref{figs09}, as the number of effective passes over datasets increases. Note that the regularization parameter is set to $\lambda\!=\!10^{-4}$. It can be seen that our VR-SGD method obtains higher test accuracies than the other methods with much shorter running time, suggesting faster convergence.

\subsubsection{Eigenvalue Computation}
Finally, we apply VR-SGD to solve the following non-convex leading eigenvalue computation problem:
\begin{equation}
\min_{x\in\mathbb{R}^{d}:x^{T}\!x=1}-x^{T}\left(\frac{1}{n}\sum^{n}_{i=1}a_{i}a^{T}_{i}\right)x.
\end{equation}
We plot the performance of the classical Power iteration method, VR-PCA~\cite{shamir:pca}, and VR-SGD on Epsilon and RCV1 in Fig.\ \ref{figs10}, where the relative error is defined as in~\cite{shamir:pca}, i.e., $\log_{10}(1-\|A^{T}x\|^{2}/(\max_{u:u^{T}u=1}\|A^{T}u\|^{2}))$, and $A\in\mathbb{R}^{d\times n}$ is the data matrix. Note that the epoch length is set to $m\!=\!n$ for VR-PCA and VR-SGD, as suggested in~\cite{shamir:pca}, and both of them use a constant learning rate. The results show that the stochastic variance reduced methods, VR-PCA and VR-SGD, significantly outperform the traditional method, Power. Moreover, our VR-SGD method often converges much faster than VR-PCA.

\begin{figure}[t]
\centering
\includegraphics[width=0.486\columnwidth]{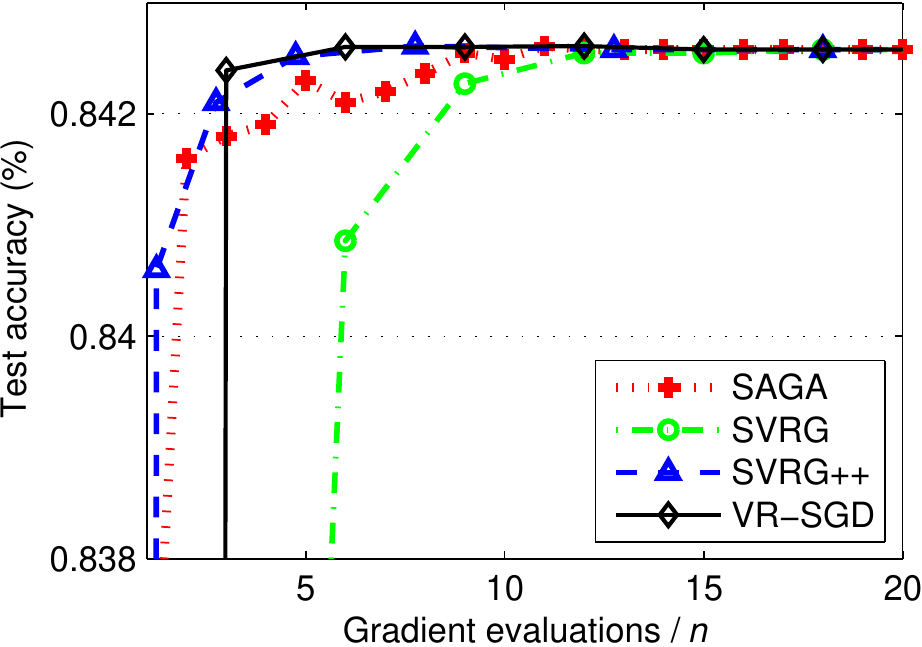}
\includegraphics[width=0.486\columnwidth]{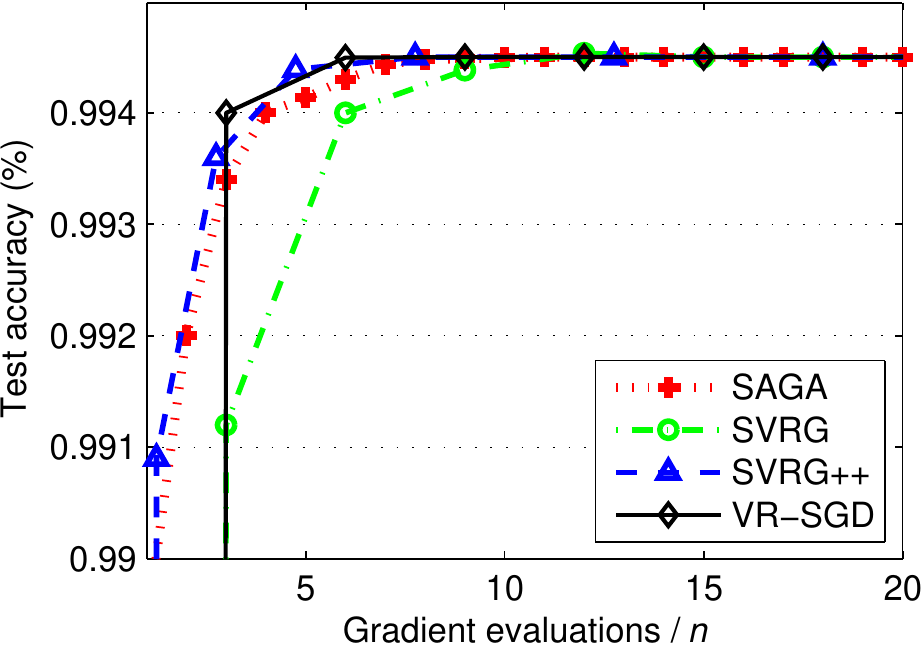}
\caption{Testing accuracy comparison on Adult (left) and MNIST (right).}
\label{figs09}
\end{figure}

\begin{figure}[t]
\centering
\includegraphics[width=0.486\columnwidth]{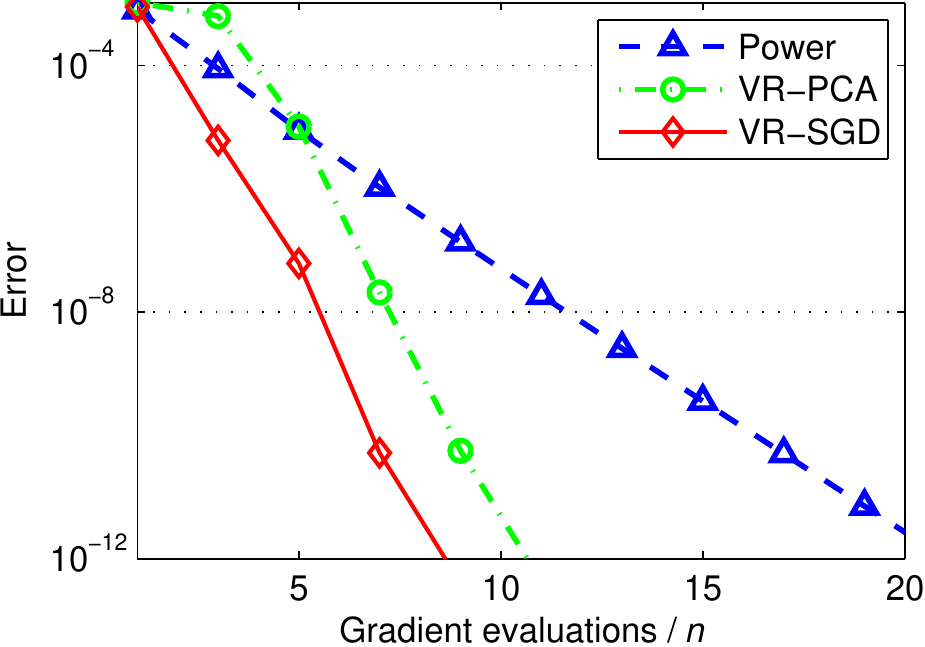}
\includegraphics[width=0.486\columnwidth]{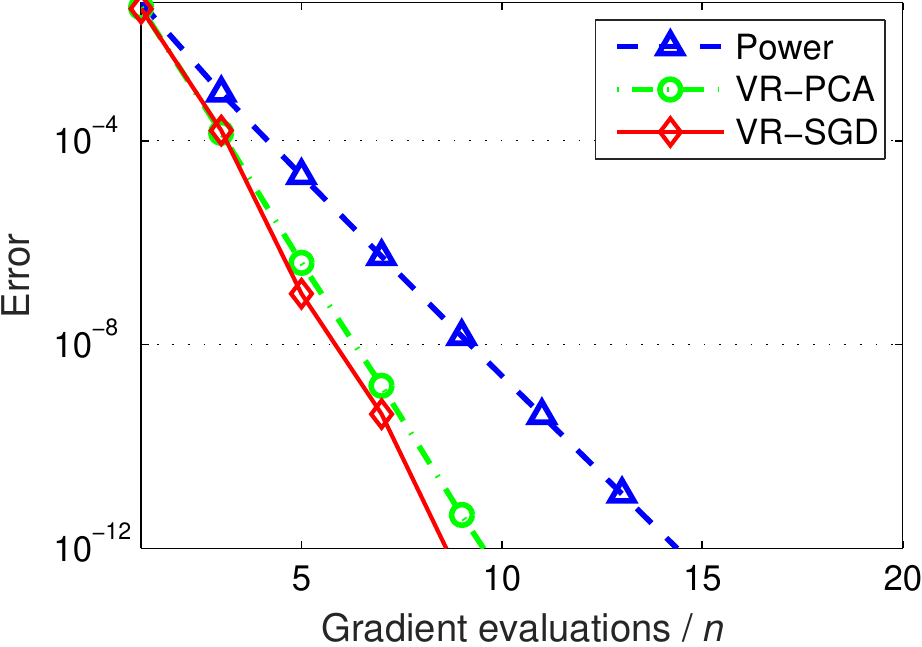}
\caption{Relative error comparison of leading eigenvalue computation on RCV1 (left) and Epsilon (right).}
\label{figs10}
\end{figure}

\section{Conclusions}
We proposed a simple variant of the original SVRG~\cite{johnson:svrg}, called variance reduced stochastic gradient descent (VR-SGD). Unlike the choices of snapshot and starting points in SVRG and Prox-SVRG~\cite{xiao:prox-svrg}, the two points of each epoch in VR-SGD are set to the average and last iterate of the previous epoch, respectively. This setting allows us to use much larger learning rates than SVRG, e.g., $1/L$ for VR-SGD vs.\ $1/(10L)$ for SVRG, and also makes VR-SGD much more robust to learning rate selection. Different from existing proximal stochastic methods such as Prox-SVRG and Katyusha~\cite{zhu:Katyusha}, we designed two different update rules for smooth and non-smooth problems, respectively, which makes VR-SGD suitable for non-smooth and/or non-strongly convex problems without using any reduction techniques as in~\cite{zhu:box}. Our empirical results also showed that for smooth problems stochastic gradient update rules as in (\ref{equ21}) are better choices than proximal update formulas as in \eqref{equ14}.

On the practical side, the choices of the snapshot and starting points make VR-SGD significantly faster than its counterparts, SVRG and Prox-SVRG. On the theoretical side, the setting also makes our convergence analysis more challenging. We analyzed the convergence properties of VR-SGD for strongly convex objective functions, which show that VR-SGD attains a linear convergence rate. Moreover, we provided the convergence guarantees of VR-SGD for non-strongly convex functions, and our experimental results showed that VR-SGD achieves similar performance to its momentum accelerated variant that has the optimal convergence rate $\mathcal{O}(1/T^2)$. In contrast, SVRG and Prox-SVRG cannot directly solve non-strongly convex functions~\cite{zhu:univr}. Various experimental results show that VR-SGD significantly outperforms state-of-the-art variance reduction methods such as SAGA~\cite{reddi:saga}, SVRG~\cite{johnson:svrg} and Prox-SVRG~\cite{xiao:prox-svrg}, and also achieves better or at least comparable performance with recently-proposed acceleration methods, e.g., Catalyst~\cite{lin:vrsg} and Katyusha~\cite{zhu:Katyusha}. Since VR-SGD has a much lower per-iteration complexity than accelerated methods (e.g., Katyusha), it has more obvious advantage over them in terms of running time, especially for high-dimensional sparse data. This further verifies that VR-SGD is suitable for various large-scale machine learning. Furthermore, as the update rules of VR-SGD are much simpler than existing accelerated stochastic variance reduction methods such as Katyusha, it is more friendly to asynchronous parallel and distributed implementation similar to~\cite{mania:svrg,reddi:sgd,lee:dsgd}.

\section*{Acknowledgments}
Fanhua Shang, Hongying Liu and Licheng Jiao were supported in part by Project supported the Foundation for Innovative Research Groups of the National Natural Science Foundation of China (No.\ 61621005), the Major Research Plan of the National Natural Science Foundation of China (Nos.\ 91438201 and 91438103), the National Natural Science Foundation of China (Nos.\ 61876220, 61836009, U1701267, 61871310, 61573267, 61502369, 61876221 and 61473215), the Program for Cheung Kong Scholars and Innovative Research Team in University (No.\ IRT\_15R53), the Fund for Foreign Scholars in University Research and Teaching Programs (the 111 Project) (No.\ B07048), and the Science Foundation of Xidian University (No.\ 10251180018). James Cheng was supported in part by Grants (CUHK 14206715 \& 14222816) from the Hong Kong RGC. Ivor Tsang was supported by ARC FT130100746, DP180100106 and LP150100671. Lijun Zhang was supported by JiangsuSF (BK20160658). Dacheng Tao was supported by Australian Research Council Projects FL-170100117, DP-180103424, and IH180100002.

\bibliographystyle{IEEEtran}
\bibliography{IEEEabrv,ieee17}

\begin{IEEEbiography}[{\includegraphics[width=1in,height=1.25in,clip,keepaspectratio]{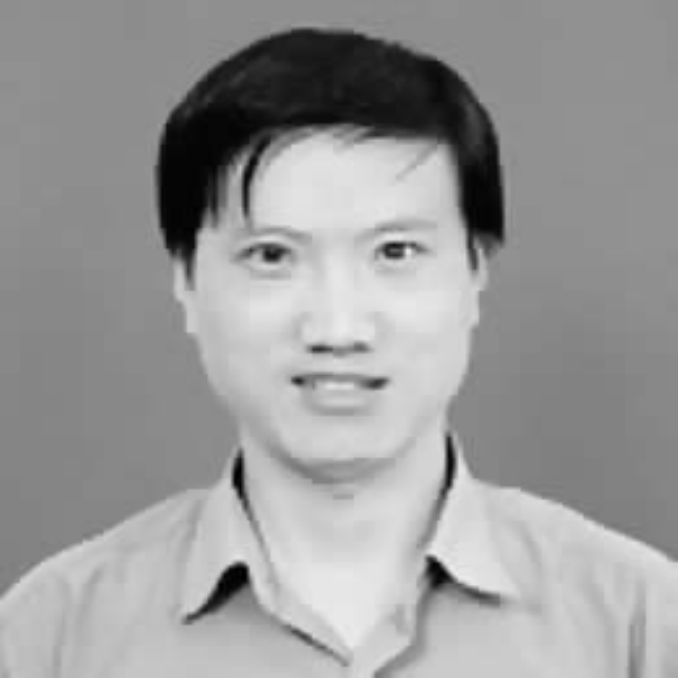}}]{Fanhua Shang} (M'14) received the Ph.D.\ degree in Circuits and Systems from Xidian University, Xi'an, China, in 2012.

He is currently a professor with the School of Artificial Intelligence, Xidian University, China. Prior to joining Xidian University, he was a Research Associate with the Department of Computer Science and Engineering, The Chinese University of Hong Kong. From 2013 to 2015, he was a Post-Doctoral Research Fellow with the Department of Computer Science and Engineering, The Chinese University of Hong Kong. From 2012 to 2013, he was a Post-Doctoral Research Associate with the Department of Electrical and Computer Engineering, Duke University, Durham, NC, USA. His current research interests include machine learning, data mining, pattern recognition, and computer vision.
\end{IEEEbiography}

\begin{IEEEbiography}[{\includegraphics[width=1in,height=1.25in,clip,keepaspectratio]{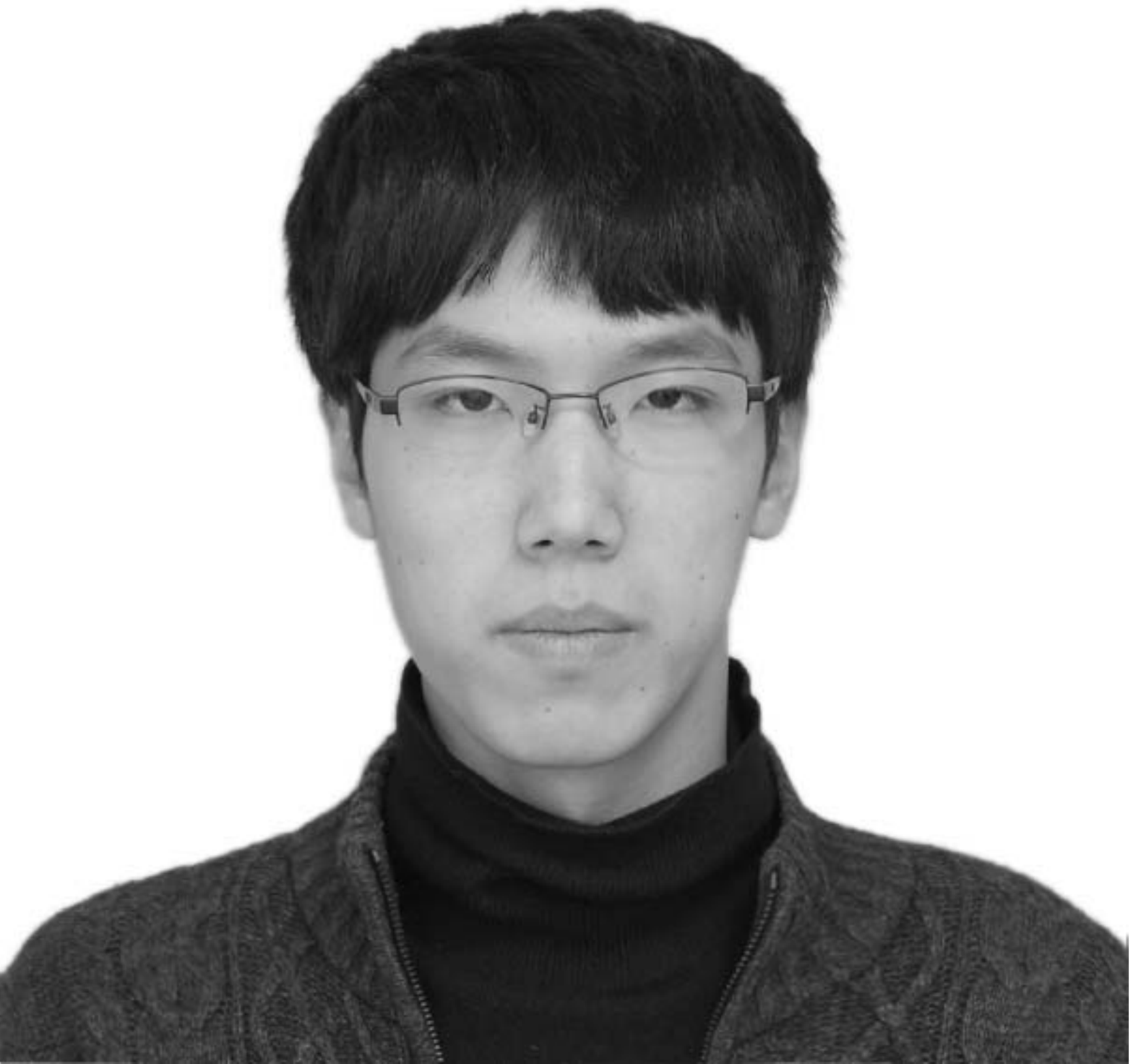}}]{Kaiwen Zhou} received the BS degree in Computer Science and Technology from Fudan University in 2017. He is currently working toward his Master degree in the Department of Computer Science and Engineering, The Chinese University of Hong Kong, Hong Kong. His current research interests include data mining, stochastic optimization for machine learning, large scale machine learning, etc.
\end{IEEEbiography}

\begin{IEEEbiography}[{\includegraphics[width=1in,height=1.25in,clip,keepaspectratio]{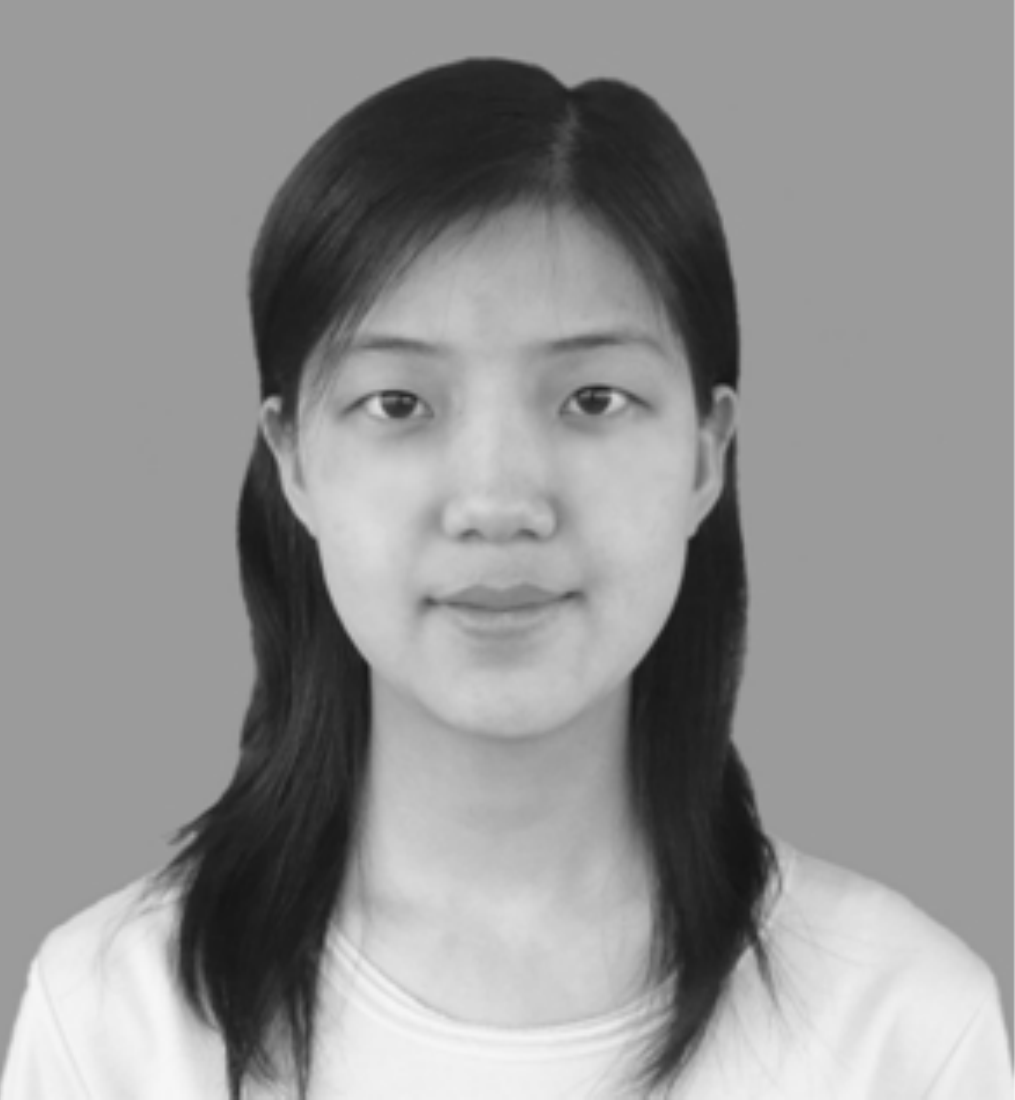}}]{Hongying Liu} (M'10) received her B.E. and M.S. degrees in Computer Science and Technology from Xi¡¯An University of Technology, China, in 2006 and 2009, respectively, and Ph.D. in Engineering from Waseda University, Japan in 2012. Currently, she is a faculty member at the School of Artificial Intelligence, and also with the Key Laboratory of Intelligent Perception and Image Understanding of Ministry of Education, Xidian University, China. In addition, she is a member of IEEE. Her major research interests include image processing, intelligent signal processing, machine learning, etc.
\end{IEEEbiography}

\begin{IEEEbiography}[{\includegraphics[width=1in,height=1.25in,clip,keepaspectratio]{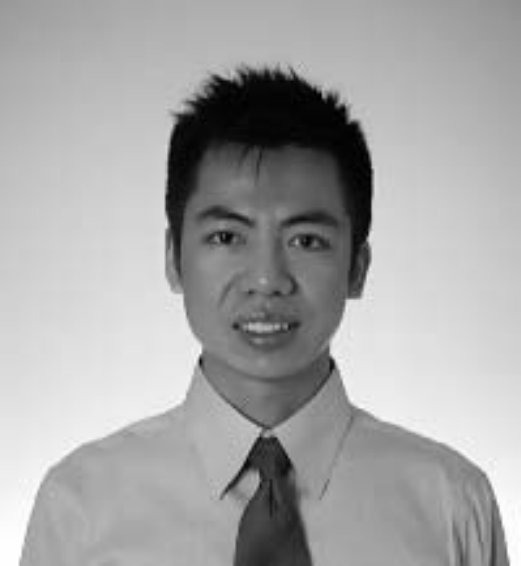}}]{James Cheng} is an Assistant Professor with the Department of Computer Science and Engineering, The Chinese University of Hong Kong, Hong Kong. His current research interests include distributed computing systems, large-scale network analysis, temporal networks, and big data.
\end{IEEEbiography}

\begin{IEEEbiography}[{\includegraphics[width=1in,height=1.25in,clip,keepaspectratio]{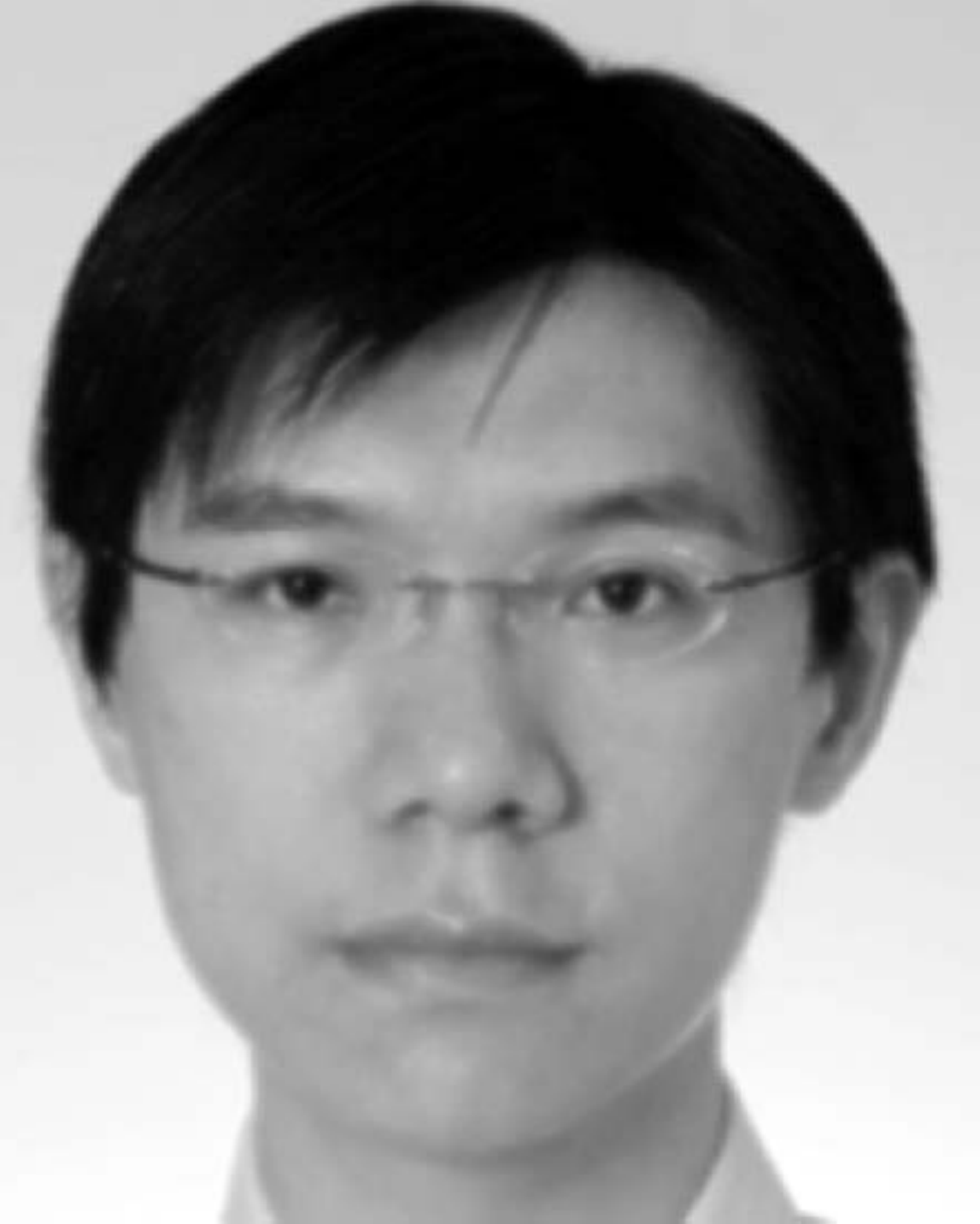}}]{Ivor~W.~Tsang} is an ARC future fellow and a professor of artificial intelligence with the University of Technology Sydney. He is also the research director of the UTS Priority Research Centre for Artificial Intelligence. His research focuses on transfer learning, feature selection, big data analytics for data with trillions of dimensions, and their applications to computer vision and pattern recognition. He has more than 140 research papers published in top-tier journal and conference papers, including the Journal of Machine Learning Research, the Madras Law Journal, the IEEE Transactions on Pattern Analysis and Machine Intelligence, the IEEE Transactions on Neural Networks and Learning Systems, NIPS, ICML, etc. In 2009, he was conferred the 2008 Natural Science Award (Class II) by Ministry of Education, China, which recognized his contributions to kernel methods. In 2013, he received his prestigious Australian Research Council Future Fellowship for his research regarding Machine Learning on Big Data. In addition, he had received the prestigious IEEE Transactions on Neural Networks Outstanding 2004 Paper Award in 2007, the 2014 IEEE Transactions on Multimedia Prize Paper Award, and a number of best paper awards and honors from reputable international conferences, including the Best Student Paper Award at CVPR 2010, and the Best Paper Award at ICTAI 2011. He was also awarded the ECCV 2012 Outstanding Reviewer Award.
\end{IEEEbiography}

\begin{IEEEbiography}[{\includegraphics[width=1in,height=1.25in,clip,keepaspectratio]{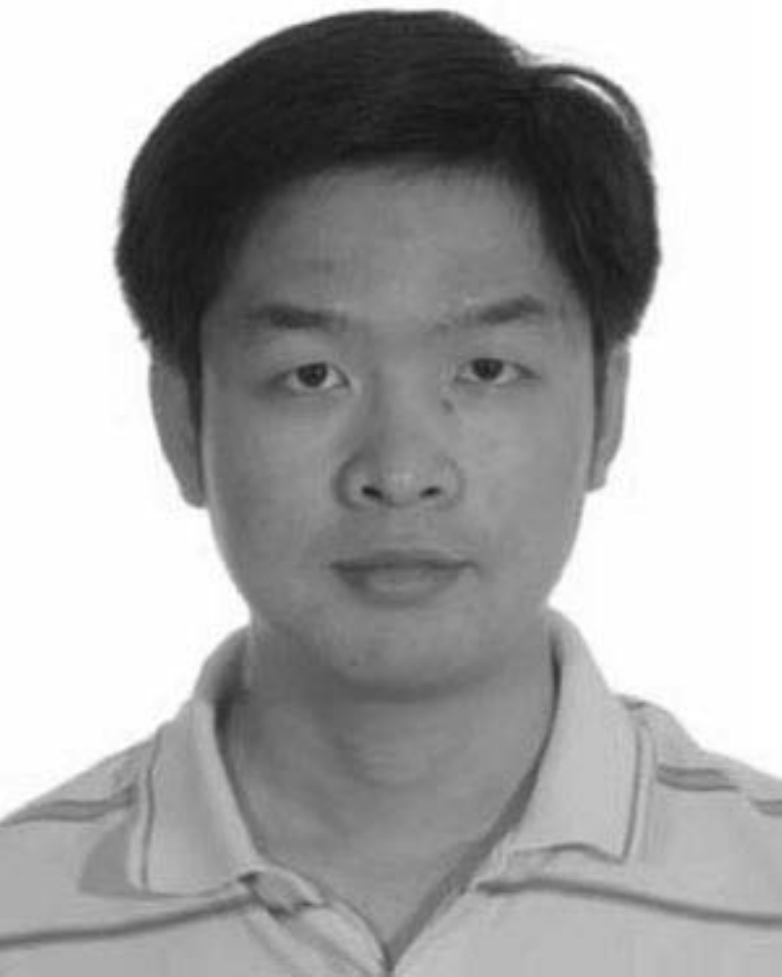}}]{Lijun Zhang} received the BS and PhD degrees in software engineering and computer science from Zhejiang University, China, in 2007 and 2012, respectively. He is currently an associate professor of the Department of Computer Science and Technology, Nanjing University, China. Prior to joining Nanjing University, he was a postdoctoral researcher at the Department of Computer Science and Engineering, Michigan State University. His research interests include machine learning, optimization, information retrieval, and data mining. He is a member of the IEEE.
\end{IEEEbiography}

\begin{IEEEbiography}[{\includegraphics[width=1in,height=1.25in,clip,keepaspectratio]{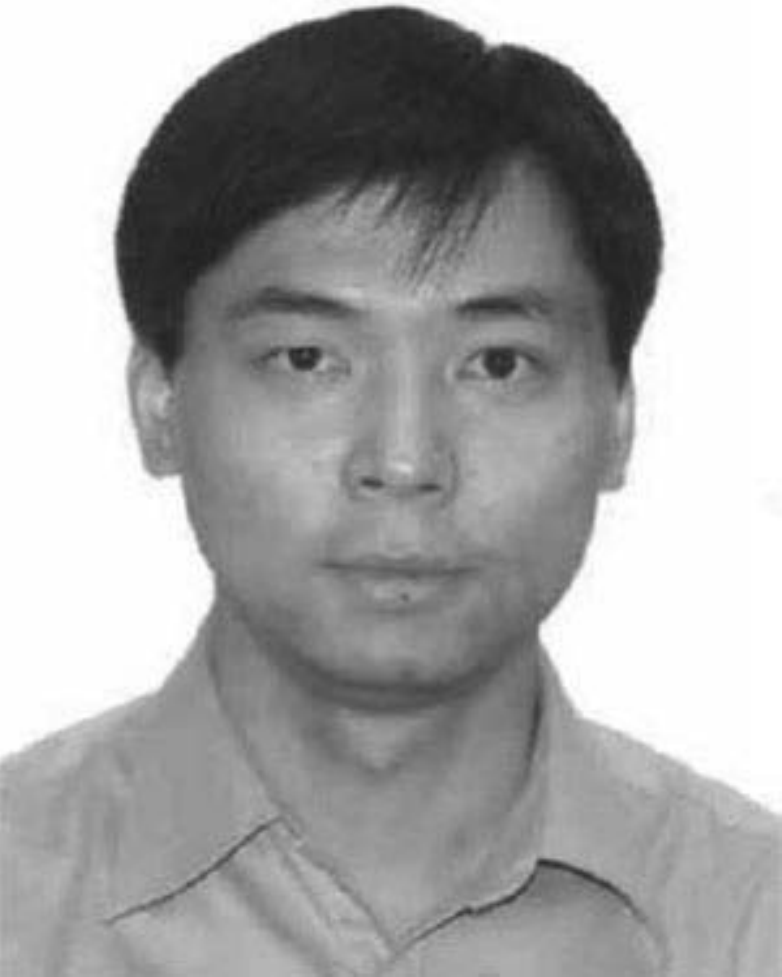}}]{Dacheng Tao} (F'15) is Professor of Computer Science and ARC Laureate Fellow in the School of Information Technologies and the Faculty of Engineering and Information Technologies, and the Inaugural Director of the UBTECH Sydney Artificial Intelligence Centre, at the University of Sydney. He mainly applies statistics and mathematics to Artificial Intelligence and Data Science. His research results have expounded in one monograph and 200+ publications at prestigious journals and prominent conferences, such as IEEE T-PAMI, T-IP, T-NNLS, IJCV, JMLR, NIPS, ICML, CVPR, ICCV, ECCV, ICDM, and ACM SIGKDD, with several best paper awards, such as the best theory/algorithm paper runner up award in IEEE ICDM'07, the best student paper award in IEEE ICDM'13, the distinguished paper award in the 2018 IJCAI, the 2014 ICDM 10-year highest-impact paper award, and the 2017 IEEE Signal Processing Society Best Paper Award. He is a Fellow of the Australian Academy of Science, AAAS, IEEE, IAPR, OSA and SPIE.
\end{IEEEbiography}

\begin{IEEEbiography}[{\includegraphics[width=1in,height=1.25in,clip,keepaspectratio]{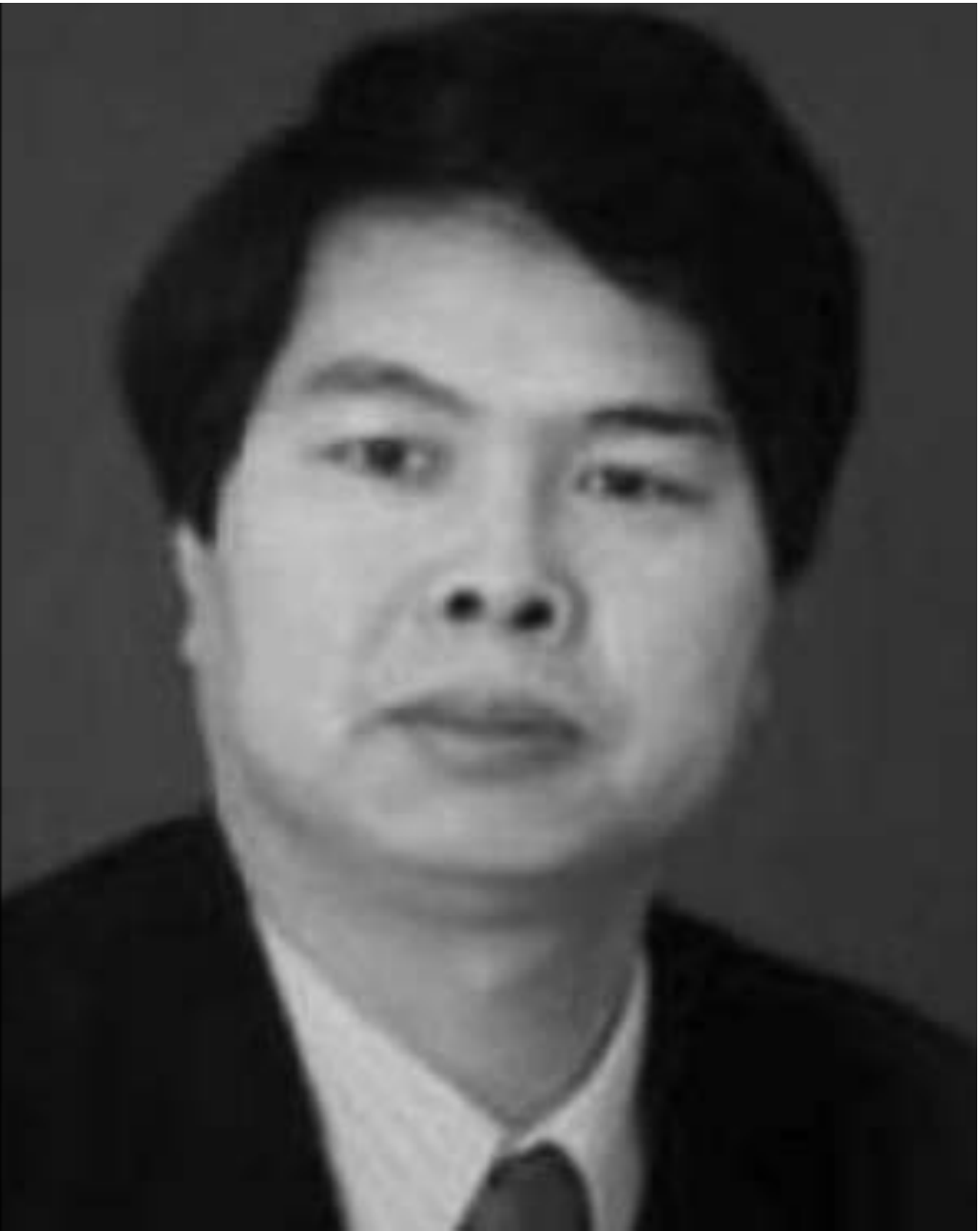}}]{Licheng Jiao} (F'18) received the B.S. degree from Shanghai Jiaotong University, Shanghai, China, in 1982, and the M.S. and Ph.D. degrees from Xi¡¯an Jiaotong University, Xi¡¯an, China, in 1984 and 1990, respectively.

He was a Post-Doctoral Fellow with the National Key Laboratory for Radar Signal Processing, Xidian University, Xi¡¯an, from 1990 to 1991, where he has been a Professor with the School of Electronic Engineering, since 1992, and currently the Director of the Key Laboratory of Intelligent Perception and Image Understanding, Ministry of Education of China. He has charged of about 40 important scientific research projects, and published over 20 monographs and a hundred papers in international journals and conferences. His current research interests include image processing, natural computation, machine learning, and intelligent information processing.

Dr. Jiao is the Chairman of Awards and Recognition Committee, the Vice Board Chairperson of the Chinese Association of Artificial Intelligence, a Councilor of the Chinese Institute of Electronics, a Committee Member of the Chinese Committee of Neural Networks, and an expert of the Academic Degrees Committee of the State Council. He is a fellow of the IEEE.
\end{IEEEbiography}

\newpage
\onecolumn
\linespread{1.469}

\vspace{16mm}
\centerline{\textbf{\large{Supplementary Materials for }}}
\vspace{2mm}
\centerline{\textbf{\large{``VR-SGD: A Simple Stochastic Variance Reduction Method for Machine Learning"}}}
\vspace{6mm}

In this supplementary material, we give the detailed proofs of some lemmas and theorems, and provide convergence analysis for Algorithm 2 with Option I or Option II, Algorithm 3, and Algorithm 4 (i.e., VR-SGD++). In addition, we also report more experimental results for solving various machine learning problems on real-world data sets.
\vspace{1mm}

\setlength{\floatsep}{1pt}
\setlength{\abovecaptionskip}{1pt}
\setlength{\belowcaptionskip}{1pt}

\section*{Appendix A: Proofs of Lemma 1 and Theorem 1}
We first introduce the following lemma, which is useful in our analysis.

\begin{lemma}[3-point property, \cite{lan:sgd}]
\label{lemm1}
Let $\hat{z}$ be the optimal solution of the following minimization problem,
\begin{displaymath}
\min_{z\in\mathbb{R}^{d}}\frac{\tau}{2}\|z-z_{0}\|^{2}+r(z)
\end{displaymath}
where $\tau\!\geq\!0$, and $r(z)$ is a convex function (but possibly non-differentiable). Then for any $z\!\in\!\mathbb{R}^{d}$, the following inequality holds
\begin{displaymath}
r(\hat{z})+\frac{\tau}{2}\|\hat{z}-z_{0}\|^{2}\leq r(z)+\frac{\tau}{2}\|z-z_{0}\|^{2}-\frac{\tau}{2}\|z-\hat{z}\|^{2}.
\end{displaymath}
\end{lemma}
\vspace{1mm}

The proof of Lemma 1 is similar to that of Theorem 1 in~\cite{johnson:svrg}. For completeness, we give the detailed proof of Lemma 1 below. Before proving Lemma 1, we first give and prove the following lemma.

\begin{lemma}\label{lemm2}
Suppose each convex component function $f_{i}(\cdot)$ is $L$-smooth, and let $x^{*}$ be the optimal solution of Problem (1) when $g(x)\equiv0${\footnote{In order to simplify our analysis, we denote $F(x)$ by $f(x)$, that is, $f_{i}(x):=f_{i}(x)+g(x)$ for all $i\!=\!1,2,\ldots,n$, and then $g(x)\equiv0$.}}, then we have
\begin{displaymath}
\mathbb{E}\!\left[\left\|\nabla\! f_{i^{s}_{k}}(x)-\nabla\! f_{i^{s}_{k}}(x^{*})\right\|^{2}\right]\leq 2L\left[f(x)-f(x^{*})\right].
\end{displaymath}
\end{lemma}
\vspace{1mm}

\begin{proof}
Following Theorem 2.1.5 in~\cite{nesterov:co} and Lemma 3.4~\cite{xiao:prox-svrg}, we have
\begin{displaymath}
\begin{split}
\left\|\nabla\! f_{i^{s}_{k}}(x)-\nabla\! f_{i^{s}_{k}}(x^{*})\right\|^{2}
\leq2L\left[f_{i^{s}_{k}}(x)-f_{i^{s}_{k}}(x^{*})-\left\langle \nabla\! f_{i^{s}_{k}}(x^{*}),\;x-x^{*}\right\rangle\right].
\end{split}
\end{displaymath}
Summing the above inequality over $i=1,\ldots,n$, we obtain
\begin{displaymath}
\begin{split}
\mathbb{E}\!\left[\left\|\nabla\! f_{i^{s}_{k}}(x)-\nabla\! f_{i^{s}_{k}}(x^{*})\right\|^{2}\right]=&\;\frac{1}{n}\!\sum^{n}_{i=1}\!\left\|\nabla\! f_{i}(x)-\nabla\! f_{i}(x^{*})\right\|^{2}\\
\leq&\; 2L\left[f(x)-f(x^{*})-\left\langle \nabla\! f(x^{*}),\;x-x^{*}\right\rangle\right].
\end{split}
\end{displaymath}
By the optimality of $x^{*}$, i.e., $x^{*}=\mathop{\arg\min}_{x} f(x)$, we have $\nabla\! f(x^{*})=0$. Then
\begin{displaymath}
\begin{split}
\mathbb{E}\!\left[\left\|\nabla\! f_{i^{s}_{k}}(x)-\nabla\! f_{i^{s}_{k}}(x^{*})\right\|^{2}\right]
\leq&\; 2L\left[f(x)-f(x^{*})-\left\langle \nabla\! f(x^{*}),x-x^{*}\right\rangle\right]\\
=&\;2L\left[f(x)-f(x^{*})\right].
\end{split}
\end{displaymath}
This completes the proof.
\end{proof}
\vspace{3mm}

\textbf{Proof of Lemma 1:}
\begin{proof}
\begin{equation*}
\begin{split}
\mathbb{E}\!\left[\|\widetilde{\nabla}\! f_{i^{s}_{k}}(x^{s}_{k})-\nabla\! f(x^{s}_{k})\|^{2}\right]=&\:\mathbb{E}\!\left[\left\|\nabla\! f_{i^{s}_{k}}(x^{s}_{k})-\nabla\! f_{i^{s}_{k}}(\widetilde{x}^{s-\!1})+\nabla\! f(\widetilde{x}^{s-1})-\nabla\! f(x^{s}_{k})\right\|^{2}\right]\\
=&\:\mathbb{E}\!\left[\left\|\nabla\! f_{i^{s}_{k}}(x^{s}_{k})-\nabla\! f_{i^{s}_{k}}(\widetilde{x}^{s-1})\right\|^{2}\right]-\left\|\nabla\! f(x^{s}_{k})-\nabla\! f(\widetilde{x}^{s-1})\right\|^{2}\\
\leq&\:\mathbb{E}\!\left[\left\|\nabla\! f_{i^{s}_{k}}(x^{s}_{k})-\nabla\! f_{i^{s}_{k}}(\widetilde{x}^{s-1})\right\|^{2}\right]\\
\leq&\:2\mathbb{E}\!\left[\left\|\nabla\! f_{i^{s}_{k}}(x^{s}_{k})-\nabla\! f_{i^{s}_{k}}(x^{*})\right\|^{2}\right]+2\mathbb{E}\!\left[\left\|\nabla\! f_{i^{s}_{k}}(\widetilde{x}^{s-\!1})-\nabla\! f_{i^{s}_{k}}(x^{*})\right\|^{2}\right]\\
\leq&\: 4L\!\left[f(x^{s}_{k})-f(x^{*})+f(\widetilde{x}^{s-1})-f(x^{*})\right]
\end{split}
\end{equation*}
where the second equality holds due to the fact that $\mathbb{E}[\|x\!-\!\mathbb{E}[x]\|^{2}]\!=\!\mathbb{E}[\|x\|^{2}]\!-\!\|\mathbb{E}[x]\|^{2}$; the second inequality holds due to the fact that $\|a-b\|^{2}\leq2(\|a\|^{2}+\|b\|^{2})$; and the last inequality follows from Lemma~\ref{lemm2}.
\end{proof}
\vspace{6mm}

Before proving Theorem 1, we first give and prove the following lemma.

\begin{lemma}[Smooth objectives and Option II]
\label{lemm4}
Let $\beta\!=\!1/(L\eta)$. If each $f_{i}(\cdot)$ is convex and $L$-smooth, then the following inequality holds for all $s=1,2,\ldots,S$,
\begin{equation*}
\begin{split}
&\;\frac{\beta\!-\!3}{\beta\!-\!1}\mathbb{E}\!\left[f(\widetilde{x}^{s})-f(x^{*})\right]+\frac{1}{m\!-\!1}\mathbb{E}\!\left[f(x^{s}_{m})-f(x^{*})\right]\\
\leq&\;\frac{2m}{\gamma}\mathbb{E}\!\left[f(\widetilde{x}^{s-1})-f(x^{*})\right]+\frac{2}{\gamma}\mathbb{E}\!\left[f(x^{s}_{0})-f(x^{*})\right]+\frac{L\beta }{2(m\!-\!1)}\mathbb{E}\!\left[\|x^{*}-x^{s}_{0}\|^2-\|x^{*}-x^{s}_{m}\|^2\right]
\end{split}
\end{equation*}
where $\gamma=(\beta\!-\!1)(m\!-\!1)$.
\end{lemma}
\vspace{3mm}

\textbf{Proof of Lemma~\ref{lemm4}:}
\begin{proof}
In order to simplify the notations, the stochastic gradient estimator is defined as: $\widetilde{\nabla}_{\!i^{s}_{k}}\!:=\!\nabla\! f_{i^{s}_{k}}(x^{s}_{k})-\nabla\! f_{i^{s}_{k}}(\widetilde{x}^{s-\!1})+\nabla\! f(\widetilde{x}^{s-\!1})$. Since each function $f_{i}(x)$ is $L$-smooth, which implies that the gradient of the average function $f(x)$ is Lipschitz continuous with parameter $L$, i.e., for all $x,y\!\in\! \mathbb{R}^{d}$, $\|\nabla f(x)-\nabla f(y)\|\leq L\|x-y\|$, whose equivalent form is
\begin{displaymath}
f(y)\leq f(x)+\langle\nabla f(x),\;y-x\rangle+\frac{L}{2}\|y-x\|^{2}.
\end{displaymath}
Using the above smoothness inequality, we have
\begin{equation}\label{equ31}
\begin{split}
f(x^{s}_{k+1})\leq\,& f(x^{s}_{k})+\left\langle\nabla f(x^{s}_{k}),\,x^{s}_{k+1}-x^{s}_{k}\right\rangle+\frac{L}{2}\!\left\|x^{s}_{k+1}-x^{s}_{k}\right\|^{2}\\
=\,& f(x^{s}_{k})+\left\langle\nabla f(x^{s}_{k}),\,x^{s}_{k+1}-x^{s}_{k}\right\rangle+\frac{L\beta }{2}\!\left\|x^{s}_{k+1}-x^{s}_{k}\right\|^{2}-\frac{L(\beta \!-\!1)}{2}\!\left\|x^{s}_{k+1}-x^{s}_{k}\right\|^{2}\\
=\,& f(x^{s}_{k})+\left\langle \widetilde{\nabla}_{\!i^{s}_{k}},\,x^{s}_{k+1}-x^{s}_{k}\right\rangle+\frac{L\beta }{2}\|x^{s}_{k+1}-x^{s}_{k}\|^2\\
&+\left\langle\nabla f(x^{s}_{k})-\widetilde{\nabla}_{\!i^{s}_{k}},\,x^{s}_{k+1}-x^{s}_{k}\right\rangle-\frac{L(\beta \!-\!1)}{2}\|x^{s}_{k+1}-x^{s}_{k}\|^{2}
\end{split}
\end{equation}
where $\beta=1/(L\eta)>3$ is a constant. By Lemma 1, then we get
\begin{equation}\label{equ32}
\begin{split}
&\mathbb{E}\!\left[\left\langle\nabla\! f(x^{s}_{k})-\widetilde{\nabla}_{\!i^{s}_{k}},\,x^{s}_{k+1}-x^{s}_{k}\right\rangle-\frac{L(\beta \!-\!1)}{2}\|x^{s}_{k+1}-x^{s}_{k}\|^{2}\right]\\
\leq\,& \mathbb{E}\!\left[\frac{1}{2L(\beta \!-\!1)}\|\nabla\!f(x^{s}_{k})-\widetilde{\nabla}_{\!i^{s}_{k}}\|^{2}+\frac{L(\beta \!-\!1)}{2}\|x^{s}_{k+1}\!-\!x^{s}_{k}\|^{2}-\frac{L(\beta \!-\!1)}{2}\|x^{s}_{k+1}\!-\!x^{s}_{k}\|^{2}\right]\\
\leq\,& \frac{2}{\beta \!-\!1}\!\left[f(x^{s}_{k})-f(x^{*})+f(\widetilde{x}^{s-1})-f(x^{*})\right]
\end{split}
\end{equation}
where the first inequality holds due to the Young's inequality (i.e., $a^{T}b\!\leq\!{\|a\|^2}/{(2\gamma)}\!+\!{\gamma\|b\|^2}/{2}$ for all $\gamma\!>\!0$ and $a,b\!\in\! \mathbb{R}^{d}$), and the second inequality follows from Lemma 1.

Substituting the inequality in \eqref{equ32} into the inequality in \eqref{equ31}, and taking the expectation with respect to the random choice $i^{s}_{k}$, we have
\begin{equation*}
\begin{split}
&\quad\;\, \mathbb{E}[f(x^{s}_{k+1})]\\
&\leq f(x^{s}_{k})+\mathbb{E}\!\left[\left\langle \widetilde{\nabla}_{\!i^{s}_{k}}, \,x^{s}_{k+\!1}\!-\!x^{s}_{k}\right\rangle+\frac{L\beta }{2}\|x^{s}_{k+\!1}\!-\!x^{s}_{k}\|^2\right]+\frac{2}{\beta \!-\!1}\!\left[f(x^{s}_{k})\!-\!f(x^{*})\!+\!f(\widetilde{x}^{s-\!1})\!-\!f(x^{*})\right]\\
&\leq f(x^{s}_{k})+\mathbb{E}\!\left[\left\langle \widetilde{\nabla}_{\!i^{s}_{k}}, \,x^{*}\!-\!x^{s}_{k}\right\rangle+\frac{L\beta }{2}(\|x^{*}\!-\!x^{s}_{k}\|^2-\|x^{*}\!-\!x^{s}_{k+\!1}\|^2)\right]+\frac{2}{\beta \!-\!1}\!\left[f(x^{s}_{k})\!-\!f(x^{*})\!+\!f(\widetilde{x}^{s-\!1})\!-\!f(x^{*})\right]\\
&\leq f(x^{s}_{k})+\left\langle \nabla\! f(x^{s}_{k}), \,x^{*}\!-\!x^{s}_{k}\right\rangle+\mathbb{E}\!\left[\frac{L\beta }{2}(\|x^{*}\!-\!x^{s}_{k}\|^2-\|x^{*}\!-\!x^{s}_{k+\!1}\|^2)\right]+\frac{2}{\beta \!-\!1}\!\left[f(x^{s}_{k})\!-\!f(x^{*})\!+\!f(\widetilde{x}^{s-\!1})\!-\!f(x^{*})\right]\\
&\leq f(x^{*})+\frac{L\beta }{2}\mathbb{E}\!\left[\left(\|x^{*}\!-x^{s}_{k}\|^2-\|x^{*}\!-x^{s}_{k+1}\|^2\right)\right]+\frac{2}{\beta \!-\!1}\!\left[f(x^{s}_{k})-f(x^{*})+f(\widetilde{x}^{s-1})-f(x^{*})\right].
\end{split}
\end{equation*}
Here, the first inequality holds due to the inequality in \eqref{equ31} and the inequality in \eqref{equ32}; the second inequality follows from Lemma 2 with $\hat{z}=x^{s}_{k+1}$, $z=x^{*}$, $z_{0}=x^{s}_{k}$, $\tau=L\beta=1/\eta$, and $r(z):=\langle \widetilde{\nabla}_{\!i^{s}_{k}}, \,z-x^{s}_{k}\rangle$; the third inequality holds due to the fact that $\mathbb{E}[\widetilde{\nabla}_{\!i^{s}_{k}}]=\nabla\! f(x^{s}_{k})$; and the last inequality follows from the convexity of the smooth function $f(\cdot)$, i.e., $f(x^{s}_{k})+\langle\nabla\! f(x^{s}_{k}), \,x^{*}-x^{s}_{k}\rangle\leq f(x^{*})$. The above inequality can be rewritten as follows:
\begin{equation*}
\begin{split}
\mathbb{E}[f(x^{s}_{k+1})]-f(x^{*})
\leq \frac{2}{\beta \!-\!1}\!\left[f(x^{s}_{k})-f(x^{*})+f(\widetilde{x}^{s-1})-f(x^{*})\right]+ \frac{L\beta }{2}\mathbb{E}\!\left[\|x^{*}-x^{s}_{k}\|^2-\|x^{*}-x^{s}_{k+1}\|^2\right].
\end{split}
\end{equation*}
Summing the above inequality over $k=0,1,\ldots,m\!-\!1$, then
\begin{equation}\label{equ33}
\begin{split}
\sum^{m}_{k=1}\mathbb{E}\!\left[f(x^{s}_{k})-f(x^{*})\right]
\leq \sum^{m}_{k=1}\!\left\{\frac{2}{\beta \!-\!1}\!\left[f(x^{s}_{k-1})-f(x^{*})+f(\widetilde{x}^{s-1})-f(x^{*})\right]+\frac{L\beta }{2}\mathbb{E}\!\left[\|x^{*}\!-x^{s}_{k-1}\|^2-\|x^{*}\!-x^{s}_{k}\|^2\right]\right\}.
\end{split}
\end{equation}

Due to the setting of $\widetilde{x}^{s}\!=\!\frac{1}{m-1}\sum^{m-1}_{k=1}x^{s}_{k}$ in Option II, and the convexity of $f(\cdot)$, then we have
\begin{equation}\label{equ34}
f(\widetilde{x}^{s})\leq \frac{1}{m-1}\sum^{m-1}_{k=1}f(x^{s}_{k}).
\end{equation}
The left and right hand sides of the inequality in \eqref{equ33} can be rewritten as follows:
\begin{equation*}
\begin{split}
&\qquad\quad\qquad\sum^{m}_{k=1}\mathbb{E}\!\left[f(x^{s}_{k})-f(x^{*})\right]=\sum^{m-1}_{k=1}\mathbb{E}\!\left[f(x^{s}_{k})-f(x^{*})\right]+\mathbb{E}\!\left[f(x^{s}_{m})-f(x^{*})\right],\\
&\sum^{m}_{k=1}\left\{\frac{2}{\beta \!-\!1}\!\left[f(x^{s}_{k-1})-f(x^{*})+f(\widetilde{x}^{s-1})-f(x^{*})\right]+\frac{L\beta }{2}\mathbb{E}\!\left[\|x^{*}-x^{s}_{k-1}\|^2-\|x^{*}-x^{s}_{k}\|^2\right]\right\}\\
=&\; \frac{2}{\beta\!-\!1}\sum^{m-1}_{k=1}\!\left[f(x^{s}_{k})\!-\!f(x^{*})\right]+\frac{2}{\beta\!-\!1}\!\left\{f(x^{s}_{0})\!-\!f(x^{*})+m[f(\widetilde{x}^{s-1})\!-\!f(x^{*})]\right\}+\frac{L\beta }{2}\mathbb{E}\!\left[\|x^{*}\!-\!x^{s}_{0}\|^2\!-\!\|x^{*}\!-\!x^{s}_{m}\|^2\right]\!.
\end{split}
\end{equation*}
Subtracting $\frac{2}{\beta-1}\sum^{m-1}_{k=1}\!\left[f(x^{s}_{k})\!-\!f(x^{*})\right]$ from both sides of the inequality in \eqref{equ33}, then we obtain
\begin{equation*}
\begin{split}
&\left(1-\frac{2}{\beta\!-\!1}\right)\sum^{m-1}_{k=1}\mathbb{E}\!\left[f(x^{s}_{k})-f(x^{*})\right]+\mathbb{E}\!\left[f(x^{s}_{m})-f(x^{*})\right]\\
\leq& \frac{2}{\beta\!-\!1}\mathbb{E}\!\left\{f(x^{s}_{0})\!-\!f(x^{*})+m[f(\widetilde{x}^{s-1})\!-\!f(x^{*})]\right\}+\frac{L\beta }{2}\mathbb{E}\!\left[\|x^{*}\!-\!x^{s}_{0}\|^2-\|x^{*}\!-\!x^{s}_{m}\|^2\right].
\end{split}
\end{equation*}
Using the inequality in \eqref{equ34}, we have
\begin{equation*}
\begin{split}
&\left(1-\frac{2}{\beta\!-\!1}\right)(m\!-\!1)\mathbb{E}\!\left[f(\widetilde{x}^{s})-f(x^{*})\right]+\mathbb{E}\!\left[f(x^{s}_{m})-f(x^{*})\right]\\
\leq&\left(1-\frac{2}{\beta\!-\!1}\right)\sum^{m-1}_{k=1}\mathbb{E}\!\left[f(x^{s}_{k})-f(x^{*})\right]+\mathbb{E}\!\left[f(x^{s}_{m})-f(x^{*})\right]\\
\leq& \frac{2}{\beta\!-\!1}\mathbb{E}\!\left\{f(x^{s}_{0})\!-\!f(x^{*})+m[f(\widetilde{x}^{s-1})\!-\!f(x^{*})]\right\}+\frac{L\beta }{2}\mathbb{E}\!\left[\|x^{*}\!-\!x^{s}_{0}\|^2-\|x^{*}\!-\!x^{s}_{m}\|^2\right].
\end{split}
\end{equation*}
Dividing both sides of the above inequality by $(m\!-\!1)$, we arrive at
\begin{equation*}
\begin{split}
&\:\left(1-\frac{2}{\beta\!-\!1}\right)\mathbb{E}\!\left[f(\widetilde{x}^{s})-f(x^{*})\right]+\frac{1}{m\!-\!1}\mathbb{E}\!\left[f(x^{s}_{m})-f(x^{*})\right]\\
\leq&\: \frac{2}{(\beta\!-\!1)(m\!-\!1)}\mathbb{E}\!\left[f(x^{s}_{0})\!-\!f(x^{*})\right]\!+\!\frac{2m}{(\beta\!-\!1)(m\!-\!1)}\mathbb{E}\!\left[f(\widetilde{x}^{s-\!1})\!-\!f(x^{*})\right]\!+\!\frac{L\beta }{2(m\!-\!1)}\mathbb{E}\!\left[\|x^{*}\!-\!x^{s}_{0}\|^2\!-\!\|x^{*}\!-\!x^{s}_{m}\|^2\right].
\end{split}
\end{equation*}
This completes the proof.
\end{proof}
\vspace{3mm}

\textbf{Proof of Theorem 1:}
\begin{proof}
Since $2/(\beta\!-\!1)<1$, it is easy to verify that
\begin{equation}
\frac{2}{(\beta\!-\!1)(m\!-\!1)}\left\{\mathbb{E}[f(x^{s}_{m})]-f(x^{*})\right\}\leq\frac{1}{m\!-\!1}\left\{\mathbb{E}[f(x^{s}_{m})]-f(x^{*})\right\}.
\end{equation}
Using the above inequality and Lemma~\ref{lemm4}, we have
\begin{equation*}
\begin{split}
&\left(1-\frac{2}{\beta\!-\!1}\right)\mathbb{E}\!\left[f(\widetilde{x}^{s})-f(x^{*})\right]+\frac{2}{(\beta\!-\!1)(m\!-\!1)}\mathbb{E}[f(x^{s}_{m})-f(x^{*})]\\
\leq&\left(1-\frac{2}{\beta\!-\!1}\right)\mathbb{E}\!\left[f(\widetilde{x}^{s})-f(x^{*})\right]+\frac{1}{m\!-\!1}\mathbb{E}[f(x^{s}_{m})-f(x^{*})]\\
\leq&\frac{2}{(\beta\!-\!1)(m\!-\!1)}\mathbb{E}\!\left[f(x^{s}_{0})-f(x^{*})\right]+\frac{2m}{(\beta\!-\!1)(m\!-\!1)}\mathbb{E}\!\left[f(\widetilde{x}^{s-1})-f(x^{*})\right]+\frac{L\beta}{2(m\!-\!1)}\mathbb{E}\!\left[\|x^{*}\!-x^{s}_{0}\|^{2}-\|x^{*}\!-x^{s}_{m}\|^{2}\right].
\end{split}
\end{equation*}
Summing the above inequality over $s\!=\!1,2,\ldots,S$, taking expectation with respect to the history of random variables $i^{s}_{k}$, and using the setting of $x^{s+1}_{0}\!=\!x^{s}_{m}$, we obtain
\begin{equation*}
\begin{split}
&\sum^{S}_{s=1}\left(1-\frac{2}{\beta\!-\!1}\right)\mathbb{E}\!\left[f(\widetilde{x}^{s})-f(x^{*})\right]\\
\leq&\:\sum^{S}_{s=1}\left\{\frac{2}{(\beta\!-\!1)(m\!-\!1)}\mathbb{E}\!\left[f(x^{s}_{0})-f(x^{*})-(f(x^{s}_{m})-f(x^{*}))\right]+\frac{2m}{(\beta\!-\!1)(m\!-\!1)}\mathbb{E}[f(\widetilde{x}^{s-1})-f(x^{*})]\right\}\\
&+\frac{L\beta}{2(m\!-\!1)}\sum^{S}_{s=1}\mathbb{E}\!\left[\|x^{*}-x^{s}_{0}\|^{2}-\|x^{*}-x^{s}_{m}\|^{2}\right].
\end{split}
\end{equation*}

Subtracting $\frac{2m}{(\beta-1)(m-1)}\!\sum^{S-1}_{s=1}\left[f(\widetilde{x}^{s})\!-\!f(x^{*})\right]$ from both sides of the above inequality, we have
\begin{equation*}
\begin{split}
&\:\frac{2m}{(\beta\!-\!1)(m\!-\!1)}{\mathbb{E}[f(\widetilde{x}^{S})-f(x^{*})}]+\sum^{S}_{s=1}\left(1-\frac{4}{\beta\!-\!1}-\frac{2}{(\beta\!-\!1)(m\!-\!1)}\right)\mathbb{E}\!\left[f(\widetilde{x}^{s})-f(x^{*})\right]\\
\leq&\:\frac{2}{(\beta\!-\!1)(m\!-\!1)}\mathbb{E}\!\left[f(x^{1}_{0})-f(x^{*})-(f(x^{S}_{m})-f(x^{*}))\right]+\frac{2m}{(\beta\!-\!1)(m\!-\!1)}\mathbb{E}[f(\widetilde{x}^{0})-f(x^{*})]\\
&+\frac{L\beta}{2(m-1)}\mathbb{E}\!\left[\|x^{*}-x^{1}_{0}\|^{2}-\|x^{*}-x^{S}_{m}\|^{2}\right].
\end{split}
\end{equation*}
Dividing both sides of the above inequality by $S$, and using the setting of $\widetilde{x}^{0}=x^{1}_{0}$, we arrive at
\begin{equation*}
\begin{split}
&\:\frac{1}{S}\left(1-\frac{4}{\beta\!-\!1}-\frac{2}{(\beta\!-\!1)(m\!-\!1)}\right)\sum^{S}_{s=1}\mathbb{E}\!\left[f(\widetilde{x}^{s})-f(x^{*})\right]\\
\leq&\,\frac{2m}{(\beta\!-\!1)(m\!-\!1)S}{\mathbb{E}[f(\widetilde{x}^{S})-f(x^{*})}]+\frac{1}{S}\!\left(1-\frac{4}{\beta\!-\!1}-\frac{2}{(\beta\!-\!1)(m\!-\!1)}\right)\sum^{S}_{s=1}\mathbb{E}\!\left[f(\widetilde{x}^{s})-f(x^{*})\right]\\
\leq&\,\frac{2}{(\beta\!-\!1)(m\!-\!1)S}[f(x^{1}_{0})-f(x^{*})]+\frac{2m}{(\beta\!-\!1)(m\!-\!1)S}[f(\widetilde{x}^{0})-f(x^{*})]+\frac{L\beta}{2(m\!-\!1)S}\|x^{*}-x^{1}_{0}\|^{2}\\
=&\,\frac{2(m+1)}{(\beta\!-\!1)(m\!-\!1)S}[f(\widetilde{x}^{0})-f(x^{*})]+\frac{L\beta}{2(m\!-\!1)S}\|\widetilde{x}^{0}-x^{*}\|^{2}
\end{split}
\end{equation*}
where the first inequality holds due to the fact that $f(\widetilde{x}^{S})\!-\!f(x^{*})\geq0$; the second inequality holds due to the facts that $f(x^{S}_{m})\!-\!f(x^{*})\geq0$ and $\|x^{*}\!-\!x^{S}_{m}\|^{2}\geq0$; and the last equality follows from the setting of $\widetilde{x}^{0}=x^{1}_{0}$.

Due to the definition of $\overline{x}^{S}$ (i.e., $\overline{x}^{S}=\frac{1}{S}\!\sum^{S}_{s=1}\!\widetilde{x}^{s}$) and the convexity of $f(\cdot)$, we have $f(\overline{x}^{S})\!\leq\! \frac{1}{S}\sum^{S}_{s=1}f(\widetilde{x}^{s})$, and therefore the above inequality becomes:
\begin{equation}
\begin{split}
&\:\left(1-\frac{4}{\beta\!-\!1}-\frac{2}{(\beta\!-\!1)(m\!-\!1)}\right)\mathbb{E}\!\left[f(\overline{x}^{S})-f(x^{*})\right]\\
\leq&\,\frac{2(m+1)}{(\beta\!-\!1)(m\!-\!1)S}[f(\widetilde{x}^{0})-f(x^{*})]+\frac{L\beta}{2(m\!-\!1)S}\|\widetilde{x}^{0}-x^{*}\|^{2}.
\end{split}
\end{equation}
Dividing both sides of the above inequality by $c_1\!=\!1\!-\!\frac{4}{\beta-1}\!-\!\frac{2}{(\beta-1)(m-1)}>0$, we have
\begin{equation*}
\mathbb{E}\!\left[f(\overline{x}^{S})\right]-f(x^{*})\leq\frac{2(m+1)}{c_1(\beta\!-\!1)(m\!-\!1)S}[f(\widetilde{x}^{0})-f(x^{*})]+\frac{L\beta}{2c_1(m\!-\!1)S}\|\widetilde{x}^{0}-x^{*}\|^{2}.
\end{equation*}
Due to the setting for the output of Algorithm 2, $\widehat{x}^{S}=\widetilde{x}^{S}$ if $f(\widetilde{x}^{S})\leq f(\overline{x}^{S})$. Then
\begin{equation*}
\begin{split}
\mathbb{E}\!\left[f(\widehat{x}^{S})\right]-f(x^{*})\leq\mathbb{E}\!\left[f(\overline{x}^{S})\right]-f(x^{*})\leq\frac{2(m+1)}{c_1(\beta\!-\!1)(m\!-\!1)S}[f(\widetilde{x}^{0})-f(x^{*})]+\frac{L\beta}{2c_1(m\!-\!1)S}\|\widetilde{x}^{0}-x^{*}\|^{2}.
\end{split}
\end{equation*}
Alternatively, when $f(\widetilde{x}^{S})\geq f(\overline{x}^{S})$, let $\widehat{x}^{S}=\overline{x}^{S}$, and the above inequality still holds.

This completes the proof.
\end{proof}
\vspace{3mm}

\section*{Appendix B: Convergence Analysis of Algorithm 2 with Option I}
Similar to Algorithm 2 with Option II, we also analyze the convergence properties of Algorithm 2 with Option I for smooth and non-strongly functions. We first give and prove the following lemma.

\begin{lemma}[Smooth objectives and Option I]
\label{lemm5}
If each $f_{i}(\cdot)$ is convex and $L$-smooth, then the following inequality holds for all $s=1,2,\ldots,S$,
\begin{equation*}
\begin{split}
&\;\frac{\beta\!-\!3}{\beta\!-\!1}\mathbb{E}\!\left[f(\widetilde{x}^{s})\!-\!f(x^{*})\right]+\frac{2}{(\beta\!-\!1)m}\mathbb{E}[f(x^{s}_{m})\!-\!f(x^{*})]\\
\leq&\;\frac{2}{(\beta\!-\!1)}\mathbb{E}\!\left[f(\widetilde{x}^{s-\!1})\!-\!f(x^{*})\right]\!+\!\frac{2}{(\beta\!-\!1)m}\mathbb{E}\!\left[f(x^{s}_{0})\!-\!f(x^{*})\right]+\frac{L\beta}{2m}\mathbb{E}\!\left[\|x^{*}-x^{s}_{0}\|^2-\|x^{*}-x^{s}_{m}\|^2\right].
\end{split}
\end{equation*}
\end{lemma}

This lemma is a slight generalization of Lemma~\ref{lemm4}, and we give its detailed proof for completeness.

\begin{proof}
For convenience, the stochastic gradient estimator is defined as: $\widetilde{\nabla}_{\!i^{s}_{k}}\!:=\!\nabla\! f_{i^{s}_{k}}(x^{s}_{k})\!-\!\nabla\! f_{i^{s}_{k}}(\widetilde{x}^{s-1})\!+\!\nabla\! f(\widetilde{x}^{s-1})$. Since each function $f_{i}(x)$ is $L$-smooth, which implies that the gradient of the average function $f(x)$ is Lipschitz continuous with parameter $L$, i.e., for all $x,y\!\in\! \mathbb{R}^{d}$,
\begin{displaymath}
\|\nabla f(x)-\nabla f(y)\|\leq L\|x-y\|,
\end{displaymath}
whose equivalent form is
\begin{displaymath}
f(y)\leq f(x)+\langle\nabla f(x),\;y-x\rangle+\frac{L}{2}\|y-x\|^{2}.
\end{displaymath}
Using the above smoothness inequality, we have
\begin{equation}\label{equ35}
\begin{split}
f(x^{s}_{k+1})\leq\,& f(x^{s}_{k})+\left\langle\nabla f(x^{s}_{k}),\,x^{s}_{k+1}-x^{s}_{k}\right\rangle+\frac{L}{2}\!\left\|x^{s}_{k+1}-x^{s}_{k}\right\|^{2}\\
=\,& f(x^{s}_{k})+\left\langle\nabla f(x^{s}_{k}),\,x^{s}_{k+1}-x^{s}_{k}\right\rangle+\frac{L\beta }{2}\!\left\|x^{s}_{k+1}-x^{s}_{k}\right\|^{2}-\frac{L(\beta \!-\!1)}{2}\!\left\|x^{s}_{k+1}-x^{s}_{k}\right\|^{2}\\
=\,& f(x^{s}_{k})+\left\langle \widetilde{\nabla}_{\!i^{s}_{k}},\,x^{s}_{k+1}-x^{s}_{k}\right\rangle+\frac{L\beta }{2}\|x^{s}_{k+1}-x^{s}_{k}\|^2\\
&+\left\langle\nabla f(x^{s}_{k})-\widetilde{\nabla}_{\!i^{s}_{k}},\,x^{s}_{k+1}-x^{s}_{k}\right\rangle-\frac{L(\beta \!-\!1)}{2}\|x^{s}_{k+1}-x^{s}_{k}\|^{2}.
\end{split}
\end{equation}
Using Lemma 1, then we get
\begin{equation}\label{equ36}
\begin{split}
&\mathbb{E}\!\left[\left\langle\nabla\! f(x^{s}_{k})-\widetilde{\nabla}_{\!i^{s}_{k}},\,x^{s}_{k+1}-x^{s}_{k}\right\rangle-\frac{L(\beta \!-\!1)}{2}\|x^{s}_{k+1}-x^{s}_{k}\|^{2}\right]\\
\leq\,& \mathbb{E}\!\left[\frac{1}{2L(\beta \!-\!1)}\|\nabla\!f(x^{s}_{k})-\widetilde{\nabla}_{\!i^{s}_{k}}\|^{2}+\frac{L(\beta \!-\!1)}{2}\|x^{s}_{k+1}\!-\!x^{s}_{k}\|^{2}-\frac{L(\beta \!-\!1)}{2}\|x^{s}_{k+1}\!-\!x^{s}_{k}\|^{2}\right]\\
\leq\,& \frac{2}{\beta \!-\!1}\!\left[f(x^{s}_{k})-f(x^{*})+f(\widetilde{x}^{s-1})-f(x^{*})\right]
\end{split}
\end{equation}
where the first inequality holds due to the Young's inequality (i.e., $y^{T}z\!\leq\!{\|y\|^2}/{(2\theta)}\!+\!{\theta\|z\|^2}/{2}$ for all $\theta\!>\!0$ and $y,z\!\in\! \mathbb{R}^{d}$), and the second inequality follows from Lemma 1.

Taking the expectation over the random choice of $i^{s}_{k}$ and substituting the inequality in \eqref{equ36} into the inequality in \eqref{equ35}, we have
\begin{equation*}
\begin{split}
&\quad\;\, \mathbb{E}[f(x^{s}_{k+1})]\\
&\leq f(x^{s}_{k})+\mathbb{E}\!\left[\left\langle \widetilde{\nabla}_{\!i^{s}_{k}}, \,x^{s}_{k+1}\!-x^{s}_{k}\right\rangle+\frac{L\beta }{2}\|x^{s}_{k+1}\!-x^{s}_{k}\|^2\right]+\frac{2}{\beta \!-\!1}\!\left[f(x^{s}_{k})-f(x^{*})+f(\widetilde{x}^{s-1})-f(x^{*})\right]\\
&\leq f(x^{s}_{k})+\mathbb{E}\!\left[\left\langle \widetilde{\nabla}_{\!i^{s}_{k}}, \,x^{*}\!-x^{s}_{k}\right\rangle+\frac{L\beta }{2}(\|x^{*}-x^{s}_{k}\|^2-\|x^{*}-x^{s}_{k+1}\|^2)\right]+\frac{2}{\beta \!-\!1}\!\left[f(x^{s}_{k})-f(x^{*})+f(\widetilde{x}^{s-1})-f(x^{*})\right]\\
&\leq f(x^{s}_{k})+\left\langle \nabla\! f(x^{s}_{k}), \,x^{*}\!-x^{s}_{k}\right\rangle+\mathbb{E}\!\left[\frac{L\beta }{2}(\|x^{*}\!-x^{s}_{k}\|^2-\|x^{*}\!-x^{s}_{k+1}\|^2)\right]+\frac{2}{\beta \!-\!1}\!\left[f(x^{s}_{k})-f(x^{*})+f(\widetilde{x}^{s-1})-f(x^{*})\right]\\
&\leq f(x^{*})+\mathbb{E}\!\left[\frac{L\beta }{2}(\|x^{*}-x^{s}_{k}\|^2-\|x^{*}-x^{s}_{k+1}\|^2)\right]+\frac{2}{\beta \!-\!1}\!\left[f(x^{s}_{k})-f(x^{*})+f(\widetilde{x}^{s-1})-f(x^{*})\right]\\
&= f(x^{*})+\frac{L\beta }{2}\mathbb{E}\!\left[\left(\|x^{*}-x^{s}_{k}\|^2-\|x^{*}-x^{s}_{k+1}\|^2\right)\right]+\frac{2}{\beta \!-\!1}\left[f(x^{s}_{k})-f(x^{*})+f(\widetilde{x}^{s-1})-f(x^{*})\right]
\end{split}
\end{equation*}
where the first inequality holds due to the inequality in \eqref{equ35} and the inequality in \eqref{equ36}; the second inequality follows from Lemma 2 with $\hat{z}=x^{s}_{k+1}$, $z=x^{*}$, $z_{0}=x^{s}_{k}$, $\tau=L\beta=1/\eta$, and $r(z):=\langle \widetilde{\nabla}_{\!i^{s}_{k}}, \,z-x^{s}_{k}\rangle$; the third inequality holds due to the fact that $\mathbb{E}[\widetilde{\nabla}_{\!i^{s}_{k}}]=\nabla\! f(x^{s}_{k})$; and the last inequality follows from the convexity of $f(\cdot)$, i.e., $f(x^{s}_{k})+\langle\nabla\! f(x^{s}_{k}), \,x^{*}-x^{s}_{k}\rangle\leq f(x^{*})$. The above inequality can be rewritten as follows:
\begin{equation*}
\begin{split}
\mathbb{E}[f(x^{s}_{k+1})]-f(x^{*})
\leq\frac{2}{\beta \!-\!1}\!\left[f(x^{s}_{k})-f(x^{*})+f(\widetilde{x}^{s-1})-f(x^{*})\right]+ \frac{L\beta}{2}\mathbb{E}\!\left[\|x^{*}\!-x^{s}_{k}\|^2-\|x^{*}\!-x^{s}_{k+1}\|^2\right].
\end{split}
\end{equation*}
Summing the above inequality over $k=0,1,\ldots,m\!-\!1$, we obtain
\begin{equation*}
\begin{split}
&\sum^{m-1}_{k=0}\left\{\mathbb{E}[f(x^{s}_{k+1})]-f(x^{*})\right\}\\
\leq& \sum^{m-1}_{k=0}\left\{\frac{2}{\beta \!-\!1}\!\left[f(x^{s}_{k})-f(x^{*})+f(\widetilde{x}^{s-1})-f(x^{*})\right]+ \frac{L\beta }{2}\mathbb{E}\!\left[\|x^{*}-x^{s}_{k}\|^2-\|x^{*}-x^{s}_{k+1}\|^2\right]\right\}.
\end{split}
\end{equation*}
Then
\begin{equation}\label{equ37}
\begin{split}
&\left(1-\frac{2}{\beta\!-\!1}\right)\sum^{m}_{k=1}\left\{\mathbb{E}[f(x^{s}_{k})]-f(x^{*})\right\}+\frac{2}{\beta\!-\!1}\sum^{m}_{k=1}\left\{\mathbb{E}[f(x^{s}_{k})]-f(x^{*})\right\}\\
\leq& \sum^{m}_{k=1}\left\{\frac{2}{\beta \!-\!1}\!\left[f(x^{s}_{k-1})-f(x^{*})+f(\widetilde{x}^{s-1})-f(x^{*})\right]\right\}+ \frac{L\beta }{2}\mathbb{E}\!\left[\|x^{*}-x^{s}_{0}\|^2-\|x^{*}-x^{s}_{m}\|^2\right].
\end{split}
\end{equation}

Due to the setting of $\widetilde{x}^{s}\!=\!\frac{1}{m}\sum^{m}_{k=1}x^{s}_{k}$ in Option I, and the convexity of $f(\cdot)$, then
\begin{equation*}
f(\widetilde{x}^{s})\leq \frac{1}{m}\sum^{m}_{k=1}f(x^{s}_{k}).
\end{equation*}
Using the above inequality, the inequality in \eqref{equ37} becomes
\begin{equation*}
\begin{split}
&m\left(1-\frac{2}{\beta\!-\!1}\right)\mathbb{E}\!\left[f(\widetilde{x}^{s})-f(x^{*})\right]+\frac{2}{\beta\!-\!1}\sum^{m}_{k=1}\left\{\mathbb{E}[f(x^{s}_{k})]-f(x^{*})\right\}\\
\leq& \sum^{m}_{k=1}\left\{\frac{2}{\beta \!-\!1}\!\left[f(x^{s}_{k-1})-f(x^{*})+f(\widetilde{x}^{s-1})-f(x^{*})\right]\right\}+ \frac{L\beta }{2}\mathbb{E}\!\left[\|x^{*}-x^{s}_{0}\|^2-\|x^{*}-x^{s}_{m}\|^2\right].
\end{split}
\end{equation*}
Dividing both sides of the above inequality by $m$ and subtracting $\frac{2}{(\beta-1)m}\sum^{m-1}_{k=1}\!\left[f(x^{s}_{k})\!-\!f(x^{*})\right]$ from both sides, we arrive at
\begin{equation*}
\begin{split}
&\left(1-\frac{2}{\beta\!-\!1}\right)\mathbb{E}\!\left[f(\widetilde{x}^{s})-f(x^{*})\right]+\frac{2}{(\beta\!-\!1)m}\mathbb{E}[f(x^{s}_{m})-f(x^{*})]\\
\leq &\, \frac{2}{(\beta\!-\!1)}\mathbb{E}\!\left[f(\widetilde{x}^{s-1})-f(x^{*})\right]+\frac{2}{(\beta\!-\!1)m}\mathbb{E}\!\left[f(x^{s}_{0})-f(x^{*})\right]+\frac{L\beta}{2m}\mathbb{E}\!\left[\|x^{*}-x^{s}_{0}\|^2-\|x^{*}-x^{s}_{m}\|^2\right].
\end{split}
\end{equation*}

This completes the proof.
\end{proof}
\vspace{3mm}

Similar to Theorem 1, we also provide the convergence guarantees of Algorithm 2 with Option I for solving smooth and non-strongly convex functions as follows.

\begin{theorem}[Smooth objectives and Option I]
\label{the2}
If each $f_{i}(\cdot)$ is convex and $L$-smooth, then the following inequality holds
\begin{equation*}
\begin{split}
\mathbb{E}\!\left[f(\widehat{x}^{S})\right]-f(x^{*})
\leq\frac{2(m\!+\!1)}{mS(\beta\!-\!5)}[f(\widetilde{x}^{0})\!-\!f(x^{*})]+\frac{L\beta(\beta\!-\!1)}{2mS(\beta\!-\!5)}\|\widetilde{x}^{0}\!-x^{*}\|^{2}.
\end{split}
\end{equation*}
\end{theorem}
\vspace{3mm}

\begin{proof}
Using Lemma~\ref{lemm5}, we have
\begin{equation*}
\begin{split}
&(1-\frac{2}{\beta\!-\!1})\mathbb{E}\!\left[f(\widetilde{x}^{s})-f(x^{*})\right]+\frac{2}{(\beta\!-\!1)m}\mathbb{E}[f(x^{s}_{m})-f(x^{*})]\\
\leq&\,\frac{2}{(\beta\!-\!1)m}\mathbb{E}\!\left[f(x^{s}_{0})-f(x^{*})\right]+\frac{2}{\beta\!-\!1}\mathbb{E}\!\left[f(\widetilde{x}^{s-1})-f(x^{*})\right]+\frac{L\beta}{2m}\mathbb{E}[\|x^{*}-x^{s}_{0}\|^{2}-\|x^{*}-x^{s}_{m}\|^{2}].
\end{split}
\end{equation*}
Summing the above inequality over $s\!=\!1,2,\ldots,S$, taking expectation with respect to the history of random variables $i^{s}_{k}$, and using the setting of $x^{s+1}_{0}=x^{s}_{m}$ in Algorithm 2, we arrive at
\begin{equation*}
\begin{split}
&\sum^{S}_{s=1}(1\!-\!\frac{2}{\beta\!-\!1})\mathbb{E}\!\left[f(\widetilde{x}^{s})-f(x^{*})\right]\\
\leq&\,\sum^{S}_{s=1}\left\{\frac{2}{(\beta\!-\!1)m}\mathbb{E}\!\left\{f(x^{s}_{0})-f(x^{*})-\left[f(x^{s}_{m})-f(x^{*})\right]\right\}+\frac{2}{\beta\!-\!1}\mathbb{E}\!\left[f(\widetilde{x}^{s-1})-f(x^{*})\right]\right\}\\
&\,+\frac{L\beta}{2m}\sum^{S}_{s=1}\mathbb{E}\!\left[\|x^{*}-x^{s}_{0}\|^{2}-\|x^{*}-x^{s}_{m}\|^{2}\right].
\end{split}
\end{equation*}
Subtracting $\frac{2}{\beta-1}\sum^{S}_{s=1}\!\left[f(\widetilde{x}^{s})-f(x^{*})\right]$ from both sides of the above inequality, we obtain
\begin{equation*}
\begin{split}
&\sum^{S}_{s=1}\left(1-\frac{4}{\beta\!-\!1}\right)\mathbb{E}\!\left[f(\widetilde{x}^{s})-f(x^{*})\right]\\
\leq&\,\frac{2}{(\beta\!-\!1)m}\mathbb{E}\!\left\{f(x^{1}_{0})-f(x^{*})-[f(x^{S}_{m})-f(x^{*})]\right\}+\frac{2}{\beta\!-\!1}\mathbb{E}[f(\widetilde{x}^{0})-f(\widetilde{x}^{S})]\\
&\,+\frac{L\beta}{2m}\mathbb{E}\!\left[\|x^{*}-x^{1}_{0}\|^{2}-\|x^{*}-x^{S}_{m}\|^{2}\right].
\end{split}
\end{equation*}

It is not hard to verify that $\mathbb{E}[f(\widetilde{x}^{0})\!-\!f(\widetilde{x}^{S})]\!\leq\!f(\widetilde{x}^{0})\!-\!f(x^{*})$. Dividing both sides of the above inequality by $S$, and using the choice of $\widetilde{x}^{0}=x^{1}_{0}$, we have
\begin{equation}\label{equ38}
\begin{split}
&\frac{1}{S}\left(1-\frac{4}{\beta\!-\!1}\right)\sum^{S}_{s=1}\mathbb{E}\!\left[f(\widetilde{x}^{s})-f(x^{*})\right]\\
\leq&\,\frac{2}{(\beta\!-\!1)mS}\mathbb{E}\!\left\{f(x^{1}_{0})-f(x^{*})-[f(x^{S}_{m})-f(x^{*})]\right\}+\frac{2}{(\beta\!-\!1)S}[f(\widetilde{x}^{0})-f(x^{*})]+\frac{L\beta}{2mS}\|x^{*}-x^{1}_{0}\|^{2}\\
\leq&\,\frac{2}{(\beta\!-\!1)mS}\!\left[f(\widetilde{x}^{0})-f(x^{*})\right]+\frac{2}{(\beta\!-\!1)S}[f(\widetilde{x}^{0})-f(x^{*})]+\frac{L\beta}{2mS}\|x^{*}-\widetilde{x}^{0}\|^{2}\\
=&\,\frac{2(m\!+\!1)}{(\beta\!-\!1)mS}[f(\widetilde{x}^{0})-f(x^{*})]+\frac{L\beta}{2mS}\|\widetilde{x}^{0}-x^{*}\|^{2}
\end{split}
\end{equation}
where the first inequality holds due to the facts that $\mathbb{E}[f(\widetilde{x}^{0})\!-\!f(\widetilde{x}^{S})]\leq f(\widetilde{x}^{0})\!-\!f(x^{*})$ and $\mathbb{E}\!\left[\|x^{*}\!-\!x^{S}_{m}\|^{2}\right]\geq0$, and the last inequality uses the facts that $\mathbb{E}\!\left[f(x^{S}_{m})\!-\!f(x^{*})\right]\geq0$ and $\widetilde{x}^{0}=x^{1}_{0}$.

Since $\overline{x}^{S}=\frac{1}{S}\!\sum^{S}_{s=1}\widetilde{x}^{s}$, and using the convexity of $f(\cdot)$, we have $f(\overline{x}^{S})\leq \frac{1}{S}\!\sum^{S}_{s=1}\!f(\widetilde{x}^{s})$, and therefore the inequality in \eqref{equ38} becomes
\begin{equation*}
\begin{split}
\left(1-\frac{4}{\beta\!-\!1}\right)\mathbb{E}\!\left[f(\overline{x}^{S})-f(x^{*})\right]\leq&\: \frac{1}{S}\left(1-\frac{4}{\beta\!-\!1}\right)\sum^{S}_{s=1}\mathbb{E}\!\left[f(\widetilde{x}^{s})-f(x^{*})\right]\\
\leq&\,\frac{2(m\!+\!1)}{(\beta\!-\!1)mS}[f(\widetilde{x}^{0})-f(x^{*})]+\frac{L\beta}{2mS}\|\widetilde{x}^{0}-x^{*}\|^{2}.
\end{split}
\end{equation*}
Dividing both sides of the above inequality by $(1\!-\!\frac{4}{\beta-1})>0$ (i.e., $\eta<1/(5L)$), we arrive at
\begin{equation*}
\mathbb{E}\!\left[f(\overline{x}^{S})-f(x^{*})\right]\leq\frac{2(m\!+\!1)}{mS(\beta\!-\!5)}[f(\widetilde{x}^{0})-f(x^{*})]+\frac{L\beta(\beta\!-\!1)}{2mS(\beta\!-\!5)}\|\widetilde{x}^{0}-x^{*}\|^{2}.
\end{equation*}

If $f(\widetilde{x}^{S})\leq f(\overline{x}^{S})$, then $\widehat{x}^{S}=\widetilde{x}^{S}$, and
\begin{equation*}
\begin{split}
\mathbb{E}\!\left[f(\widehat{x}^{S})-f(x^{*})\right]\leq&\,\mathbb{E}\!\left[f(\overline{x}^{S})-f(x^{*})\right]\\
\leq&\,\frac{2(m\!+\!1)}{mS(\beta\!-\!5)}[f(\widetilde{x}^{0})-f(x^{*})]+\frac{L\beta(\beta\!-\!1)}{2mS(\beta\!-\!5)}\|\widetilde{x}^{0}-x^{*}\|^{2}.
\end{split}
\end{equation*}
Alternatively, if $f(\widetilde{x}^{S})\!\geq\!f(\overline{x}^{S})$, then $\widehat{x}^{S}\!=\!\overline{x}^{S}$, and the above inequality still holds.

This completes the proof.
\end{proof}
\vspace{3mm}

\section*{Appendix C: Proof of Theorem 2}
We first extend Lemma 1 to the non-smooth setting as follows~\cite{xiao:prox-svrg,shang:vrsgd}.

\begin{corollary}[Variance bound of non-smooth objectives]
\label{lemm51}
If each $f_{i}(\cdot)$ is $L$-smooth, then the following inequality holds
\begin{equation*}
\begin{split}
\mathbb{E}\!\left[\|\widetilde{\nabla}\! f_{i^{s}_{k}}(x^{s}_{k})-\nabla\! f(x^{s}_{k})\|^{2}\right]\leq 4L[F(x^{s}_{k})-F(x^{*})+F(\widetilde{x}^{s-1})-F(x^{*})].
\end{split}
\end{equation*}
\end{corollary}
\vspace{3mm}

Before proving Theorem 2, we first give and prove the following lemma.

\begin{lemma}[Non-smooth objectives]
\label{lemm6}
If each $f_{i}(\cdot)$ is $L$-smooth, then the following inequality holds for all $s=1,2,\ldots,S$,
\begin{equation*}\label{equ39}
\begin{split}
&\frac{\beta\!-\!3}{\beta\!-\!1}\mathbb{E}\!\left[F(\widetilde{x}^{s})\!-\!F(x^{*})\right]+\frac{2}{(\beta\!-\!1)m}\mathbb{E}\!\left[F(x^{s}_{m})\!-\!F(x^{*})\right]\\
\!\!\!\leq&\frac{2}{(\beta\!-\!1)m}\mathbb{E}\!\left[F(x^{s}_{0})\!\!-\!F(x^{*})\right]\!+\!\frac{2}{\beta\!-\!1}\mathbb{E}\!\left[F(\widetilde{x}^{s-\!1})\!-\!F(x^{*})\right]+\frac{L\beta}{2m}\mathbb{E}\!\left[\|x^{*}\!-\!x^{s}_{0}\|^2-\|x^{*}\!-\!x^{s}_{m}\|^2\right].
\end{split}
\end{equation*}
\end{lemma}
\vspace{3mm}

\textbf{Proof of Lemma~\ref{lemm6}:}
\begin{proof}
Since the average function $f(x)$ is $L$-smooth, then for all $x,y\!\in\! \mathbb{R}^{d}$,
\begin{displaymath}
f(y)\leq f(x)+\langle\nabla f(x),\;y-x\rangle+\frac{L}{2}\|y-x\|^{2},
\end{displaymath}
which then implies
\begin{displaymath}
f(x^{s}_{k+1})\leq f(x^{s}_{k})+\left\langle\nabla f(x^{s}_{k}),\,x^{s}_{k+1}-x^{s}_{k}\right\rangle+\frac{L}{2}\!\left\|x^{s}_{k+1}\!-x^{s}_{k}\right\|^{2}.
\end{displaymath}
Using the above inequality, we have
\begin{equation}\label{equ41}
\begin{split}
F(x^{s}_{k+1})=\,&f(x^{s}_{k+1})+g(x^{s}_{k+1})\\
\leq\,& f(x^{s}_{k})+g(x^{s}_{k+1})+\left\langle\nabla f(x^{s}_{k}),\,x^{s}_{k+1}-x^{s}_{k}\right\rangle+\frac{L\beta}{2}\!\left\|x^{s}_{k+1}-x^{s}_{k}\right\|^{2}-\frac{L(\beta\!-\!1)}{2}\!\left\|x^{s}_{k+1}-x^{s}_{k}\right\|^{2}\\
=\,& f(x^{s}_{k})+g(x^{s}_{k+1})+\left\langle \widetilde{\nabla}_{\!i^{s}_{k}},\,x^{s}_{k+1}-x^{s}_{k}\right\rangle+\frac{L\beta}{2}\|x^{s}_{k+1}-x^{s}_{k}\|^2\\
&+\left\langle\nabla f(x^{s}_{k})-\widetilde{\nabla}_{\!i^{s}_{k}},\,x^{s}_{k+1}-x^{s}_{k}\right\rangle-\frac{L(\beta\!-\!1)}{2}\|x^{s}_{k+1}-x^{s}_{k}\|^{2}.
\end{split}
\end{equation}
According to Corollary 2, then we obtain
\begin{equation}\label{equ42}
\begin{split}
&\mathbb{E}\!\left[\left\langle\nabla\! f(x^{s}_{k})-\widetilde{\nabla}_{\!i^{s}_{k}},\,x^{s}_{k+1}-x^{s}_{k}\right\rangle-\frac{L(\beta\!-\!1)}{2}\|x^{s}_{k+1}-x^{s}_{k}\|^{2}\right]\\
\leq\,& \mathbb{E}\!\left[\frac{1}{2L(\beta\!-\!1)}\|\nabla\!f(x^{s}_{k})-\widetilde{\nabla}_{\!i^{s}_{k}}\|^{2}+\frac{L(\beta\!-\!1)}{2}\|x^{s}_{k+1}\!-\!x^{s}_{k}\|^{2}-\frac{L(\beta\!-\!1)}{2}\|x^{s}_{k+1}\!-\!x^{s}_{k}\|^{2}\right]\\
\leq\,& \frac{2}{\beta\!-\!1}\!\left[F(x^{s}_{k})-F(x^{*})+F(\widetilde{x}^{s-1})-F(x^{*})\right],
\end{split}
\end{equation}
where the first inequality holds due to the Young's inequality, and the second inequality follows from Corollary 2. Substituting the inequality \eqref{equ42} into the inequality \eqref{equ41}, and taking the expectation over the random choice $i^{s}_{k}$, we arrive at
\begin{equation*}
\begin{split}
\mathbb{E}\!\left[F(x^{s}_{k+1})\right]
\leq&\; \mathbb{E}\!\left[f(x^{s}_{k})\right]+\mathbb{E}\!\left[g(x^{s}_{k+1})\right]+\mathbb{E}\!\left[\left\langle \widetilde{\nabla}_{\!i^{s}_{k}}, \;x^{s}_{k+1}-x^{s}_{k}\right\rangle+\frac{L\beta}{2}\|x^{s}_{k+1}-x^{s}_{k}\|^2\right]\\
&\;+\frac{2}{\beta\!-\!1}\!\left[F(x^{s}_{k})-F(x^{*})+F(\widetilde{x}^{s-1})-F(x^{*})\right]\\
\leq&\; \mathbb{E}\!\left[f(x^{s}_{k})\right]+g(x^{*})+\mathbb{E}\!\left[\left\langle \widetilde{\nabla}_{\!i^{s}_{k}}, \;x^{*}-x^{s}_{k}\right\rangle+\frac{L\beta}{2}(\|x^{*}-x^{s}_{k}\|^2-\|x^{*}-x^{s}_{k+1}\|^2)\right]\\
&\;+\frac{2}{\beta\!-\!1}\!\left[F(x^{s}_{k})-F(x^{*})+F(\widetilde{x}^{s-1})-F(x^{*})\right]\\
\leq&\; f(x^{*})+g(x^{*})+\mathbb{E}\!\left[\frac{L\beta}{2}(\|x^{*}-x^{s}_{k}\|^2-\|x^{*}-x^{s}_{k+1}\|^2)\right]\\
&\;+\frac{2}{\beta\!-\!1}\!\left[F(x^{s}_{k})-F(x^{*})+F(\widetilde{x}^{s-1})-F(x^{*})\right]\\
=&\;F(x^{*})+\frac{L\beta}{2}\mathbb{E}\!\left[\|x^{*}\!-x^{s}_{k}\|^2-\|x^{*}\!-x^{s}_{k+1}\|^2\right]+\frac{2}{\beta\!-\!1}\left[F(x^{s}_{k})-F(x^{*})+F(\widetilde{x}^{s-1})-F(x^{*})\right],\\
\end{split}
\end{equation*}
where the first inequality holds due to the inequality \eqref{equ41} and the inequality \eqref{equ42}; the second inequality follows from Lemma 2 with $\hat{z}=x^{s}_{k+1}$, $z=x^{*}$, $z_{0}=x^{s}_{k}$, $\tau=L\beta =1/\eta$, and $r(z):=\langle \widetilde{\nabla}_{\!i^{s}_{k}}, \,z-x^{s}_{k}\rangle+g(z)$; and the third inequality holds due to the fact that $\mathbb{E}[\widetilde{\nabla}_{\!i^{s}_{k}}]=\nabla\! f(x^{s}_{k})$ and the convexity of $f(\cdot)$, i.e., $f(x^{s}_{k})+\langle\nabla\! f(x^{s}_{k}), \,x^{*}-x^{s}_{k}\rangle\leq f(x^{*})$. Then the above inequality is rewritten as follows:
\begin{equation}
\begin{split}
&\:\mathbb{E}[F(x^{s}_{k+1})]-F(x^{*})\\
\leq&\:\frac{2}{\beta\!-\!1}\!\left[F(x^{s}_{k})-F(x^{*})+F(\widetilde{x}^{s-1})-F(x^{*})\right]+\frac{L\beta}{2}\mathbb{E}\!\left[\|x^{*}\!-x^{s}_{k}\|^2-\|x^{*}\!-x^{s}_{k+1}\|^2\right].
\end{split}
\end{equation}

Summing the above inequality over $k=0,1,\ldots,(m\!-\!1)$ and taking expectation over whole history, we have
\begin{equation*}
\begin{split}
&\sum^{m-1}_{k=0}\left\{\mathbb{E}[F(x^{s}_{k+1})]-F(x^{*})\right\}\\
\leq& \sum^{m-1}_{k=0}\left\{\frac{2}{\beta\!-\!1}\!\left[F(x^{s}_{k})-F(x^{*})+F(\widetilde{x}^{s-1})-F(x^{*})\right]+ \frac{L\beta}{2}\mathbb{E}\!\left[\|x^{*}-x^{s}_{k}\|^2-\|x^{*}-x^{s}_{k+1}\|^2\right]\right\}.
\end{split}
\end{equation*}
Subtracting $\frac{2}{\beta-1}\!\sum^{m-2}_{k=0}\mathbb{E}\!\left[F(x^{s}_{k+1})\!-\!F(x^{*})\right]$ from both sides of the above inequality, we obtain
\begin{equation*}
\begin{split}
&\sum^{m-1}_{k=0}\mathbb{E}\!\left[F(x^{s}_{k+1})-F(x^{*})\right]-\frac{2}{\beta\!-\!1}\!\sum^{m-1}_{k=0}\mathbb{E}\!\left[F(x^{s}_{k+1})\!-\!F(x^{*})\right]+\frac{2}{\beta\!-\!1}\mathbb{E}\!\left[F(x^{s}_{m})\!-\!F(x^{*})\right]\\
\leq& \sum^{m-1}_{k=0}\!\left\{\frac{2}{\beta\!-\!1}\!\left[F(x^{s}_{k})\!-\!F(x^{*})\!+\!F(\widetilde{x}^{s-1})\!-\!F(x^{*})\right]\right\}\!-\!\frac{2}{\beta\!-\!1}\!\sum^{m-2}_{k=0}\!\mathbb{E}\!\left[F(x^{s}_{k+1})\!-\!F(x^{*})\right]\!+\! \frac{L\beta}{2}\mathbb{E}\!\left[\|x^{*}\!-\!x^{s}_{0}\|^2\!-\!\|x^{*}\!-\!x^{s}_{m}\|^2\right].
\end{split}
\end{equation*}
Then
\begin{equation*}
\begin{split}
&\left(1-\frac{2}{\beta\!-\!1}\right)\sum^{m}_{k=1}\mathbb{E}\!\left[F(x^{s}_{k})-F(x^{*})\right]+\frac{2}{\beta\!-\!1}\mathbb{E}\!\left[F(x^{s}_{m})\!-\!F(x^{*})\right]\\
\leq& \frac{2}{\beta\!-\!1}\mathbb{E}\!\left[F(x^{s}_{0})-F(x^{*})\right]+\frac{2m}{\beta\!-\!1}\mathbb{E}\!\left[F(\widetilde{x}^{s-1})\!-\!F(x^{*})\right]+ \frac{L\beta}{2}\mathbb{E}\!\left[\|x^{*}\!-\!x^{s}_{0}\|^2\!-\!\|x^{*}\!-\!x^{s}_{m}\|^2\right].
\end{split}
\end{equation*}
Due to the settings of $\widetilde{x}^{s}\!=\!\frac{1}{m}\sum^{m}_{k=1}x^{s}_{k}$ and $x^{s+1}_{0}\!=\!x^{s}_{m}$, and the convexity of the objective function $F(\cdot)$, we have $F(\widetilde{x}^{s})\leq \frac{1}{m}\sum^{m}_{k=1}F(x^{s}_{k})$, and
\begin{equation*}
\begin{split}
&m\left(1-\frac{2}{\beta\!-\!1}\right)\mathbb{E}\!\left[F(\widetilde{x}^{s})-F(x^{*})\right]+\frac{2}{\beta\!-\!1}\mathbb{E}\!\left[F(x^{s}_{m})\!-\!F(x^{*})\right]\\
\leq&\frac{2}{\beta\!-\!1}\mathbb{E}\!\left[F(x^{s}_{0})-F(x^{*})\right]+\frac{2m}{\beta\!-\!1}\mathbb{E}\!\left[F(\widetilde{x}^{s-1})\!-\!F(x^{*})\right]+ \frac{L\beta}{2}\mathbb{E}\!\left[\|x^{*}\!-\!x^{s}_{0}\|^2\!-\!\|x^{*}\!-\!x^{s}_{m}\|^2\right].
\end{split}
\end{equation*}
Dividing both sides of the above inequality by $m$, we arrive at
\begin{equation}\label{equ69}
\begin{split}
&\left(1-\frac{2}{\beta\!-\!1}\right)\mathbb{E}\!\left[F(\widetilde{x}^{s})-F(x^{*})\right]+\frac{2}{(\beta\!-\!1)m}\mathbb{E}\!\left[F(x^{s}_{m})\!-\!F(x^{*})\right]\\
\leq&\frac{2}{(\beta\!-\!1)m}\mathbb{E}\!\left[F(x^{s}_{0})-F(x^{*})\right]+\frac{2}{\beta\!-\!1}\mathbb{E}\!\left[F(\widetilde{x}^{s-1})\!-\!F(x^{*})\right]+ \frac{L\beta}{2m}\mathbb{E}\!\left[\|x^{*}\!-\!x^{s}_{0}\|^2\!-\!\|x^{*}\!-\!x^{s}_{m}\|^2\right].
\end{split}
\end{equation}

This completes the proof.
\end{proof}
\vspace{3mm}

\textbf{Proof of Theorem 2:}
\begin{proof}
Summing the inequality in (\ref{equ69}) over $s\!=\!1,2,\ldots,S$, and taking expectation with respect to the history of $i^{s}_{k}$, we have
\begin{equation*}
\begin{split}
&\sum^{S}_{s=1}\left(1\!-\!\frac{2}{\beta\!-\!1}\right)\!\mathbb{E}\!\left[F(\widetilde{x}^{s})-F(x^{*})\right]+\sum^{S}_{s=1}\frac{2}{(\beta\!-\!1)m}\mathbb{E}\!\left[F(x^{s}_{m})-F(x^{*})\right]\\
\leq&\,\sum^{S}_{s=1}\left\{\frac{2}{(\beta\!-\!1)m}\mathbb{E}\!\left[F(x^{s}_{0})-F(x^{*})\right]+\frac{2}{\beta\!-\!1}\mathbb{E}\!\left[F(\widetilde{x}^{s-1})-F(x^{*})\right]\right\}\\
&\,+\frac{L\beta}{2m}\sum^{S}_{s=1}\mathbb{E}\!\left[\|x^{*}-x^{s}_{0}\|^{2}-\|x^{*}-x^{s}_{m}\|^{2}\right].
\end{split}
\end{equation*}
Subtracting $\sum^{S}_{s=1}\!\frac{2}{(\beta-1)m}\mathbb{E}\!\left[F(x^{s}_{m})\!-\!F(x^{*})\right]+\frac{2}{\beta-1}\!\sum^{S}_{s=1}\!\left[F(\widetilde{x}^{s})\!-\!F(x^{*})\right]$ from both sides of the above inequality, and using the setting of $x^{s+1}_{0}=x^{s}_{m}$, we arrive at
\begin{equation*}
\begin{split}
&\sum^{S}_{s=1}\left(1-\frac{4}{\beta\!-\!1}\right)\mathbb{E}\!\left[F(\widetilde{x}^{s})-F(x^{*})\right]\\
\leq&\,\frac{2}{(\beta\!-\!1)m}\mathbb{E}\!\left[F(x^{1}_{0})\!-\!F(x^{S}_{m})\right]+\frac{2}{\beta\!-\!1}\mathbb{E}\!\left[F(\widetilde{x}^{0})\!-\!F(\widetilde{x}^{S})\right]+\frac{L\beta}{2m}\mathbb{E}\!\left[\|x^{*}\!-\!x^{1}_{0}\|^{2}-\|x^{*}\!-\!x^{S}_{m}\|^{2}\right].
\end{split}
\end{equation*}
It is easy to verify that $\mathbb{E}[F(\widetilde{x}^{0})\!-\!F(\widetilde{x}^{S})]\leq F(\widetilde{x}^{0})\!-\!F(x^{*})$. Dividing both sides of the above inequality by $S$, and using the choice $\widetilde{x}^{0}=x^{1}_{0}$, we obtain
\begin{equation}\label{equ43}
\begin{split}
&\:\left(1-\frac{4}{\beta\!-\!1}\right)\frac{1}{S}\sum^{S}_{s=1}\mathbb{E}\!\left[F(\widetilde{x}^{s})-F(x^{*})\right]\\
\leq&\,\frac{2}{(\beta\!-\!1)mS}\mathbb{E}\!\left[F(x^{1}_{0})\!-\!F(x^{S}_{m})\right]+\frac{2}{(\beta\!-\!1)S}[F(\widetilde{x}^{0})-F(\widetilde{x}^{S})]+\frac{L\beta}{2mS}\|x^{*}-x^{1}_{0}\|^{2}\\
\leq&\,\frac{2}{(\beta\!-\!1)mS}\!\left[F(\widetilde{x}^{0})-F(x^{*})\right]+\frac{2}{(\beta\!-\!1)S}[F(\widetilde{x}^{0})-F(x^{*})]+\frac{L\beta}{2mS}\|x^{*}-\widetilde{x}^{0}\|^{2}\\
=&\,\frac{2(m\!+\!1)}{(\beta\!-\!1)mS}[F(\widetilde{x}^{0})-F(x^{*})]+\frac{L\beta}{2mS}\|\widetilde{x}^{0}-x^{*}\|^{2}
\end{split}
\end{equation}
where the first inequality uses the fact that $\|x^{*}-x^{S}_{m}\|^{2}\geq0$; and the last inequality holds due to the facts that $\mathbb{E}\!\left[F(x^{1}_{0})\!-\!F(x^{S}_{m})\right]$ $\leq\! F(x^{1}_{0})\!-\!F(x^{*})$, $\mathbb{E}[F(\widetilde{x}^{0})\!-\!F(\widetilde{x}^{S})]\leq F(\widetilde{x}^{0})\!-\!F(x^{*})$, and $\widetilde{x}^{0}=x^{1}_{0}$.

Using the definition of $\overline{x}^{S}\!=\!\frac{1}{S}\!\sum^{S}_{s=1}\widetilde{x}^{s}$ and the convexity of the objective function $F(\cdot)$, we have $F(\overline{x}^{S})\!\leq\! \frac{1}{S}\!\sum^{S}_{s=1}\!F(\widetilde{x}^{s})$, and therefore we can rewrite the above inequality in \eqref{equ43} as follows:
\begin{equation*}
\begin{split}
\left(1-\frac{4}{\beta\!-\!1}\right)\mathbb{E}\!\left[F(\overline{x}^{S})-F(x^{*})\right]\leq&\: \left(1-\frac{4}{\beta\!-\!1}\right)\frac{1}{S}\sum^{S}_{s=1}\mathbb{E}\!\left[F(\widetilde{x}^{s})-F(x^{*})\right]\\
\leq&\,\frac{2(m\!+\!1)}{(\beta\!-\!1)mS}[F(\widetilde{x}^{0})-F(x^{*})]+\frac{L\beta}{2mS}\|\widetilde{x}^{0}-x^{*}\|^{2}.
\end{split}
\end{equation*}
Dividing both sides of the above inequality by $(1\!-\!\frac{4}{\beta-1})\!>\!0$, we have
\begin{equation*}
\mathbb{E}\!\left[F(\overline{x}^{S})\right]-F(x^{*})\leq\frac{2(m\!+\!1)}{(\beta\!-\!5)mS}[F(\widetilde{x}^{0})-F(x^{*})]+\frac{\beta(\beta\!-\!1)L}{2(\beta\!-\!5)mS}\|\widetilde{x}^{0}-x^{*}\|^{2}.
\end{equation*}

When $F(\widetilde{x}^{S})\!\leq\!F(\overline{x}^{S})$, then $\widehat{x}^{S}\!=\!\widetilde{x}^{S}$, and
\begin{equation*}
\begin{split}
\mathbb{E}\!\left[F(\widehat{x}^{S})\right]-F(x^{*})\leq\,\frac{2(m\!+\!1)}{(\beta\!-\!5)mS}[F(\widetilde{x}^{0})-F(x^{*})]+\frac{\beta(\beta\!-\!1)L}{2(\beta\!-\!5)mS}\|\widetilde{x}^{0}-x^{*}\|^{2}.
\end{split}
\end{equation*}
Alternatively, if $F(\widetilde{x}^{S})\!\geq\!F(\overline{x}^{S})$, then $\widehat{x}^{S}\!=\!\overline{x}^{S}$, and the above inequality still holds.

This completes the proof.
\end{proof}
\vspace{3mm}

\section*{Appendix D: Convergence Analysis of Algorithm 2 with Option II}

\begin{corollary}[Option II and non-smooth objectives]
\label{lemm7}
If each $f_{i}(\cdot)$ is convex and $L$-smooth, then the following inequality holds for all $s=1,2,\ldots,S$,
\begin{equation*}
\begin{split}
&\,\frac{\beta\!-\!3}{\beta\!-\!1}\mathbb{E}\!\left[F(\widetilde{x}^{s})-F(x^{*})\right]+\frac{1}{m\!-\!1}\mathbb{E}\!\left[F(x^{s}_{m})-F(x^{*})\right]\\
\leq&\;\frac{2m}{\gamma}\mathbb{E}\!\left[F(\widetilde{x}^{s-\!1})-F(x^{*})\right]+\frac{2}{\gamma}\mathbb{E}\!\left[F(x^{s}_{0})-F(x^{*})\right]+\frac{L\beta }{2(m\!-\!1)}\mathbb{E}\!\left[\|x^{*}-x^{s}_{0}\|^2-\|x^{*}-x^{s}_{m}\|^2\right].
\end{split}
\end{equation*}
\end{corollary}

\begin{corollary}[Option II and non-smooth objectives]
\label{the4}
Suppose Assumption 1 holds. Then the following inequality holds
\begin{equation*}
\begin{split}
\mathbb{E}\!\left[F(\widehat{x}^{S})\right]-F(x^{*})\leq\frac{2(m+1)}{[\gamma\!-\!4m\!+\!2]S}[F(\widetilde{x}^{0})-F(x^{*})]+\frac{\beta(\beta-1)L}{2[\gamma\!-\!4m\!+\!2]S}\|\widetilde{x}^{0}-x^{*}\|^{2}.
\end{split}
\end{equation*}
\end{corollary}

Corollaries~\ref{lemm7} and~\ref{the4} can be viewed as the generalizations of Lemma~\ref{lemm6} and Theorem 2, respectively, and hence their proofs are omitted. From Theorem 2 and Corollary~\ref{the4}, one can see that Algorithm 2 also achieves a convergence rate $\mathcal{O}(1/T)$ for non-strongly convex and non-smooth functions.

\section*{Appendix E: Proof of Theorem 3}
Before proving Theorem 3, we first give and prove the following lemma.

\begin{lemma}[Mini-batch]
\label{lemm9}
Using the same notation as in Corollary 2, we have
\begin{equation*}
\begin{split}
&\frac{2\delta(b)}{(\beta\!-\!\!1)m}\!\!\left\{\mathbb{E}[F(x^{s}_{m})]\!-\!F(x^{*})\!\right\}\!+\!(1\!-\!\frac{2\delta(b)}{\beta\!-\!\!1})\!\left\{\mathbb{E}[F(\widetilde{x}^{s})]\!-\!F(x^{*})\!\right\}\\
\leq&\, \frac{2\delta(b)}{(\beta\!-\!1)}\!\left[F(\widetilde{x}^{s-1})-F(x^{*})\right]+\frac{2\delta(b)}{(\beta\!-\!1)m}\left[F(x^{s}_{0})-F(x^{*})\right]+\frac{L\beta}{2m}\mathbb{E}\!\left[\|x^{*}-x^{s}_{0}\|^2-\|x^{*}\!-x^{s}_{m}\|^2\right].
\end{split}
\end{equation*}
\end{lemma}
\vspace{3mm}

\textbf{Proof of Lemma~\ref{lemm9}:}
\begin{proof}
In order to simplify notation, the stochastic gradient estimator of mini-batch is defined as:
\begin{equation*}
\widetilde{\nabla}_{\!I^{s}_{k}}:=\frac{1}{b}\!\sum_{i\in{I}^{s}_{k}}\!\left[\nabla\! f_{i}(x^{s}_{k})-\nabla\! f_{i}(\widetilde{x}^{s-1})\right]+\nabla\! f(\widetilde{x}^{s-1}).
\end{equation*}
\begin{equation}\label{equ61}
\begin{split}
F(x^{s}_{k+1})\leq\,&\, g(x^{s}_{k+1})+f(x^{s}_{k})+\left\langle\nabla f(x^{s}_{k}),\,x^{s}_{k+1}-x^{s}_{k}\right\rangle+\frac{L\beta}{2}\!\left\|x^{s}_{k+1}-x^{s}_{k}\right\|^{2}-\frac{L(\beta\!-\!1)}{2}\!\left\|x^{s}_{k+1}-x^{s}_{k}\right\|^{2}\\
=&\, g(x^{s}_{k+1})+f(x^{s}_{k})+\left\langle \widetilde{\nabla}_{\!I^{s}_{k}},\,x^{s}_{k+1}-x^{s}_{k}\right\rangle+\frac{L\beta}{2}\|x^{s}_{k+1}-x^{s}_{k}\|^2\\
&\,+\left\langle\nabla f(x^{s}_{k})-\widetilde{\nabla}_{\!I^{s}_{k}},\,x^{s}_{k+1}-x^{s}_{k}\right\rangle-\frac{L(\beta\!-\!1)}{2}\|x^{s}_{k+1}-x^{s}_{k}\|^{2}.
\end{split}
\end{equation}
Using Corollary 1, then we obtain
\begin{equation}\label{equ62}
\begin{split}
&\mathbb{E}\!\left[\left\langle\nabla\! f(x^{s}_{k})-\widetilde{\nabla}_{\!I^{s}_{k}},\,x^{s}_{k+1}-x^{s}_{k}\right\rangle-\frac{L(\beta\!-\!1)}{2}\|x^{s}_{k+1}-x^{s}_{k}\|^{2}\right]\\
\leq&\, \mathbb{E}\!\left[\frac{1}{2L(\beta\!-\!1)}\|\nabla\!f(x^{s}_{k})-\widetilde{\nabla}_{\!I^{s}_{k}}\|^{2}+\frac{L(\beta\!-\!1)}{2}\|x^{s}_{k+1}\!-\!x^{s}_{k}\|^{2}-\frac{L(\beta\!-\!1)}{2}\|x^{s}_{k+1}\!-\!x^{s}_{k}\|^{2}\right]\\
\leq&\, \frac{2\delta(b)}{\beta\!-\!1}\!\left[F(x^{s}_{k})-F(x^{*})+F(\widetilde{x}^{s-1})-F(x^{*})\right]
\end{split}
\end{equation}
where the first inequality holds due to the Young's inequality, and the second inequality follows from Corollary 1. Substituting the inequality \eqref{equ62} into the inequality \eqref{equ61}, and taking the expectation over the random mini-batch set $I^{s}_{k}$, we have
\begin{equation*}
\begin{split}
&\,\mathbb{E}[F(x^{s}_{k+1})]\\
\leq&\,\mathbb{E}[g(x^{s}_{k+\!1})]+f(x^{s}_{k})+\mathbb{E}\!\left[\left\langle \widetilde{\nabla}_{\!I^{s}_{k}}, \,x^{s}_{k+\!1}\!-\!x^{s}_{k}\right\rangle+\!\frac{L\beta}{2}\|x^{s}_{k+\!1}\!-\!x^{s}_{k}\|^2\right]+\frac{2\delta(b)}{\beta\!-\!1}\!\left[F(x^{s}_{k})\!-\!F(x^{*})\!+\!F(\widetilde{x}^{s-\!1})\!-\!F(x^{*})\right]\\
\leq&\, g(x^{*})+f(x^{s}_{k})+\mathbb{E}\!\left[\left\langle \widetilde{\nabla}_{\!I^{s}_{k}}, \,x^{*}\!-\!x^{s}_{k}\right\rangle+\frac{L\beta}{2}(\|x^{*}\!-\!x^{s}_{k}\|^2\!-\!\|x^{*}\!-\!x^{s}_{k+1}\|^2)\right]+\frac{2\delta(b)}{\beta\!-\!1}\!\left[F(x^{s}_{k})\!-\!F(x^{*})\!+\!F(\widetilde{x}^{s-\!1})\!-\!F(x^{*})\right]\\
\leq&\, g(x^{*})+ f(x^{*})+\mathbb{E}\!\left[\frac{L\beta}{2}(\|x^{*}-x^{s}_{k}\|^2-\|x^{*}-x^{s}_{k+1}\|^2)\right]+\frac{2\delta(b)}{\beta\!-\!1}\!\left[F(x^{s}_{k})\!-\!F(x^{*})\!+\!F(\widetilde{x}^{s-\!1})\!-\!F(x^{*})\right]\\
=&\, F(x^{*})+\frac{L\beta}{2}\mathbb{E}\!\left[\left(\|x^{*}-x^{s}_{k}\|^2-\|x^{*}-x^{s}_{k+1}\|^2\right)\right]+\frac{2\delta(b)}{\beta-1}\left[F(x^{s}_{k})\!-\!F(x^{*})\!+\!F(\widetilde{x}^{s-\!1})\!-\!F(x^{*})\right]
\end{split}
\end{equation*}
where the second inequality holds from Lemma 2. Then the above inequality is rewritten as follows:
\begin{equation}\label{equ63}
\begin{split}
\mathbb{E}\!\left[F(x^{s}_{k+1})\right]-F(x^{*})\leq \frac{2\delta(b)}{\beta\!-\!1}\!\left[F(x^{s}_{k})-F(x^{*})+F(\widetilde{x}^{s-1})-F(x^{*})\right]+ \frac{L\beta}{2}\mathbb{E}\!\left[\|x^{*}-x^{s}_{k}\|^2-\|x^{*}-x^{s}_{k+1}\|^2\right].
\end{split}
\end{equation}
Summing the above inequality over $k=0,1,\cdots,(m-1)$, then
\begin{equation*}
\begin{split}
&\,\sum^{m-1}_{k=0}\left\{\mathbb{E}[F(x^{s}_{k+1})]-F(x^{*})\right\}\\
\leq&\, \sum^{m-1}_{k=0}\left\{\frac{2\delta(b)}{\beta\!-\!1}\!\left[F(x^{s}_{k})-F(x^{*})+F(\widetilde{x}^{s-1})-F(x^{*})\right]+ \frac{L\beta}{2}\mathbb{E}\!\left[\|x^{*}-x^{s}_{k}\|^2-\|x^{*}-x^{s}_{k+1}\|^2\right]\right\}.
\end{split}
\end{equation*}

Since $\widetilde{x}^{s}=\frac{1}{m}\sum^{m}_{k=1}x^{s}_{k}$, we have $F(\widetilde{x}^{s})\leq \frac{1}{m}\sum^{m}_{k=1}F(x^{s}_{k})$, and
\begin{equation*}
\begin{split}
&\:\frac{2\delta(b)}{(\beta\!-\!1)m}\left\{\mathbb{E}[F(x^{s}_{m})]-F(x^{*})\right\}+\left(1-\frac{2\delta(b)}{\beta\!-\!1}\right)\left\{\mathbb{E}[F(\widetilde{x}^{s})]-F(x^{*})\right\}\\
\leq&\: \frac{2\delta(b)}{(\beta\!-\!1)}\left[F(\widetilde{x}^{s-1})-F(x^{*})\right]+\frac{2\delta(b)}{(\beta\!-\!1)m}\left[F(x^{s}_{0})-F(x^{*})\right]+\frac{L\beta}{2m}\mathbb{E}\!\left[\|x^{*}-x^{s}_{0}\|^2-\|x^{*}-x^{s}_{m}\|^2\right].
\end{split}
\end{equation*}

This completes the proof.
\end{proof}
\vspace{3mm}

\textbf{Proof of Theorem 3:}
\begin{proof}
Using Lemma~\ref{lemm9}, we have
\begin{equation*}
\begin{split}
&\,\left(1-\frac{2\delta(b)}{\beta\!-\!1}\right)\mathbb{E}\!\left[F(\widetilde{x}^{s})-F(x^{*})\right]\\
\leq&\,\frac{2\delta(b)}{(\beta\!-\!1)}\left[F(\widetilde{x}^{s-1})-F(x^{*})\right]+\frac{2\delta(b)}{(\beta\!-\!1)m}\left[F(x^{s}_{0})-F(x^{*})\right]+\frac{L\beta}{2m}\mathbb{E}\!\left[\|x^{*}\!-x^{s}_{0}\|^2-\|x^{*}\!-x^{s}_{m}\|^2\right].
\end{split}
\end{equation*}
Summing the above inequality over $s=1,2,\ldots,S$, taking expectation over whole history of $I^{s}_{k}$, and using $x^{s+1}_{0}=x^{s}_{m}$, we obtain
\begin{equation*}
\begin{split}
&\,\sum^{S}_{s=1}\left(1-\frac{2\delta(b)}{\beta\!-\!1}\right)\mathbb{E}\left[F(\widetilde{x}^{s})-F(x^{*})\right]\\
\leq&\, \sum^{S}_{s=1}\frac{2\delta(b)}{(\beta\!-\!1)}\!\left[F(\widetilde{x}^{s-1})-F(x^{*})\right]+\sum^{S}_{s=1}\frac{2\delta(b)}{(\beta\!-\!1)m}\mathbb{E}\!\left[F(x^{s}_{0})-F(x^{s}_{m})\right]+\frac{L\beta}{2m}\sum^{S}_{s=1}\mathbb{E}\!\left[\|x^{*}\!-x^{s}_{0}\|^2-\|x^{*}\!-x^{s}_{m}\|^2\right].
\end{split}
\end{equation*}

Subtracting $\frac{2\delta(b)}{\beta-1}\!\sum^{S-1}_{s=1}\mathbb{E}\!\left[F(\widetilde{x}^{s})\!-\!F(x^{*})\right]$ from both sides of the above inequality, we have
\begin{equation*}
\begin{split}
&\:\frac{2\delta(b)}{\beta\!-\!1}\mathbb{E}\!\left[F(\widetilde{x}^{S})-F(x^{*})\right]+\sum^{S}_{s=1}\left(1-\frac{4\delta(b)}{\beta\!-\!1}\right)\mathbb{E}\!\left[F(\widetilde{x}^{s})-F(x^{*})\right]\\
\leq&\:\frac{2\delta(b)}{(\beta\!-\!1)}\left[F(\widetilde{x}^{0})-F(x^{*})\right]+\frac{2\delta(b)}{(\beta\!-\!1)m}\mathbb{E}\!\left[F(x^{1}_{0})-F(x^{S}_{m})\right]+\frac{L\beta}{2m}\mathbb{E}\!\left[\|x^{*}\!-x^{1}_{0}\|^2-\|x^{*}\!-x^{S}_{m}\|^2\right]\\
\end{split}
\end{equation*}
Dividing both sides of the above inequality by $S$ and using $\mathbb{E}[F(\overline{x})]\leq \frac{1}{S}\sum^{S}_{s=1}F(\widetilde{x}^{s})$, we arrive at
\begin{equation*}
\begin{split}
&\,\frac{2\delta(b)}{(\beta\!-\!1)S}\mathbb{E}[F(\widetilde{x}^{S})-F(x^{*})]+\left(1-\frac{4\delta(b)}{\beta\!-\!1}\right)\mathbb{E}\!\left[F(\overline{x})-F(x^{*})\right]\\
\leq&\, \frac{2\delta(b)}{(\beta\!-\!1)S}\!\left[F(\widetilde{x}^{0})-F(x^{*})\right]+\frac{2\delta(b)}{(\beta\!-\!1)mS}\mathbb{E}\!\left[F(x^{1}_{0})-F(x^{S}_{m})\right]+\frac{L\beta}{2mS}\mathbb{E}\!\left[\|x^{*}\!-x^{1}_{0}\|^2-\|x^{*}\!-x^{S}_{m}\|^2\right].
\end{split}
\end{equation*}
Subtracting $\frac{2\delta(b)}{(\beta\!-\!1)S}\mathbb{E}[F(\widetilde{x}^{S})\!-\!F(x^{*})]$ from both sides of the above inequality, we have
\begin{equation*}
\begin{split}
&\,\left(1-\frac{4\delta(b)}{\beta\!-\!1}\right)\mathbb{E}\!\left[F(\overline{x})-F(x^{*})\right]\\
\leq&\, \frac{2\delta(b)}{(\beta\!-\!1)S}\mathbb{E}\!\left[F(\widetilde{x}^{0})-F(\widetilde{x}^{S})\right]+\frac{2\delta(b)}{(\beta\!-\!1)mS}\mathbb{E}\!\left[F(x^{1}_{0})-F(x^{S}_{m})\right]+\frac{L\beta}{2mS}\mathbb{E}\!\left[\|x^{*}-x^{1}_{0}\|^2-\|x^{*}-x^{S}_{m}\|^2\right].
\end{split}
\end{equation*}

Dividing both sides of the above inequality by $(1\!-\!\frac{4\delta(b)}{\beta-1})>0$, we arrive at
\begin{equation*}
\begin{split}
&\,\mathbb{E}\!\left[F(\overline{x})\right]-F(x^{*})\\
\leq&\, \frac{2\delta(b)}{(\beta\!-\!1\!-\!4\delta(b))S}\mathbb{E}\!\left[F(\widetilde{x}^{0})-F(\widetilde{x}^{S})\right]+\!\frac{2\delta(b)}{(\beta\!-\!1\!-\!4\delta(b))mS}\mathbb{E}\!\left[F(\widetilde{x}^{0})-F(x^{S}_{m})\right]+\!\frac{L\beta(\beta\!-\!1)}{2(\beta\!-\!1\!-\!4\delta(b))mS}\mathbb{E}\!\left[\|x^{*}-\widetilde{x}^{0}\|^2\right]\\
\leq&\, \frac{2\delta(b)}{(\beta\!-\!1\!-\!4\delta(b))S}\mathbb{E}\!\left[F(\widetilde{x}^{0})-F(x^{*})\right]+\!\frac{2\delta(b)}{(\beta\!-\!1\!-\!4\delta(b))mS}\mathbb{E}\!\left[F(\widetilde{x}^{0})-F(x^{*})\right]+\!\frac{L\beta(\beta\!-\!1)}{2(\beta\!-\!1\!-\!4\delta(b))mS}\mathbb{E}\!\left[\|x^{*}-\widetilde{x}^{0}\|^2\right]\\
=&\, \frac{2\delta(b)(m+1)}{(\beta\!-\!1\!-\!4\delta(b))mS}\mathbb{E}\!\left[F(\widetilde{x}^{0})-F(x^{*})\right]+\!\frac{L\beta(\beta\!-\!1)}{2(\beta\!-\!1\!-\!4\delta(b))mS}\mathbb{E}\!\left[\|x^{*}-\widetilde{x}^{0}\|^2\right].
\end{split}
\end{equation*}

When $F(\widetilde{x}^{S})\leq F(\overline{x}^{S})$, then $\widehat{x}^{S}=\widetilde{x}^{S}$, and
\begin{equation*}
\begin{split}
\mathbb{E}\!\left[F(\widehat{x}^{S})\right]-F(x^{*})\leq\,\frac{2\delta(b)(m+1)}{(\beta\!-\!1\!-\!4\delta(b))mS}\mathbb{E}\!\left[F(\widetilde{x}^{0})-F(x^{*})\right]+\!\frac{L\beta(\beta\!-\!1)}{2(\beta\!-\!1\!-\!4\delta(b))mS}\mathbb{E}\!\left[\|x^{*}-\widetilde{x}^{0}\|^2\right].
\end{split}
\end{equation*}
Alternatively, if $F(\widetilde{x}^{S})\!\geq\!F(\overline{x}^{S})$, then $\widehat{x}^{S}\!=\!\overline{x}^{S}$, and the above inequality still holds.

This completes the proof.
\end{proof}

Theorem 3 shows that the mini-batch version of Algorithm 2 with Option I attains a convergence rate $\mathcal{O}(1/T)$ for non-strongly convex and non-smooth functions. In addition, we can also provide the convergence properties of Algorithm 2 for non-strongly convex and smooth functions.
\vspace{3mm}

\begin{figure}[t]
\centering
\subfigure[Adult: $\lambda=10^{-4}$ (left) \;and\; $\lambda=10^{-5}$ (right)]{\includegraphics[width=0.439\columnwidth]{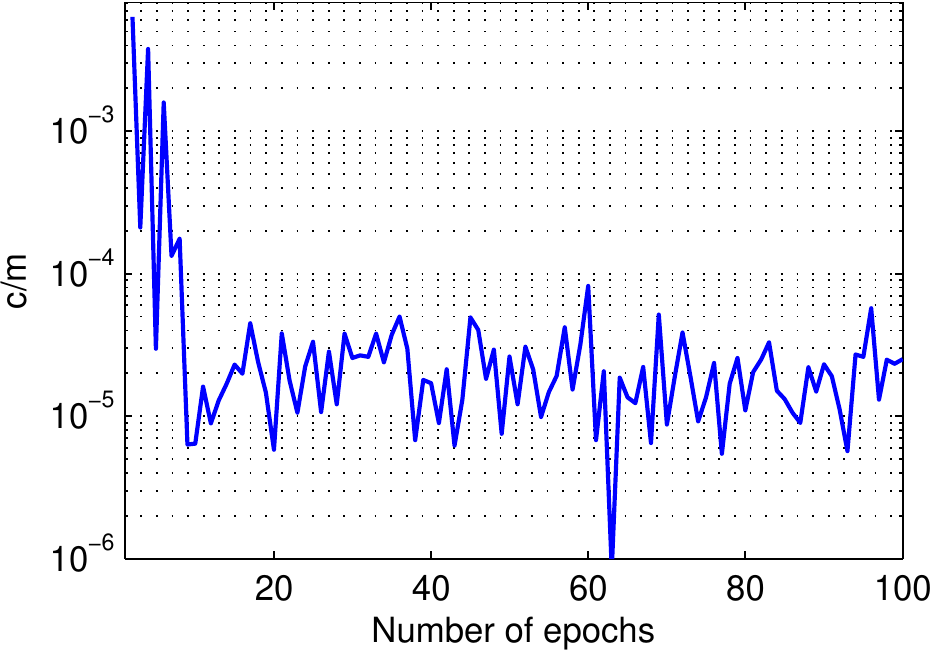}\;\;\;\includegraphics[width=0.439\columnwidth]{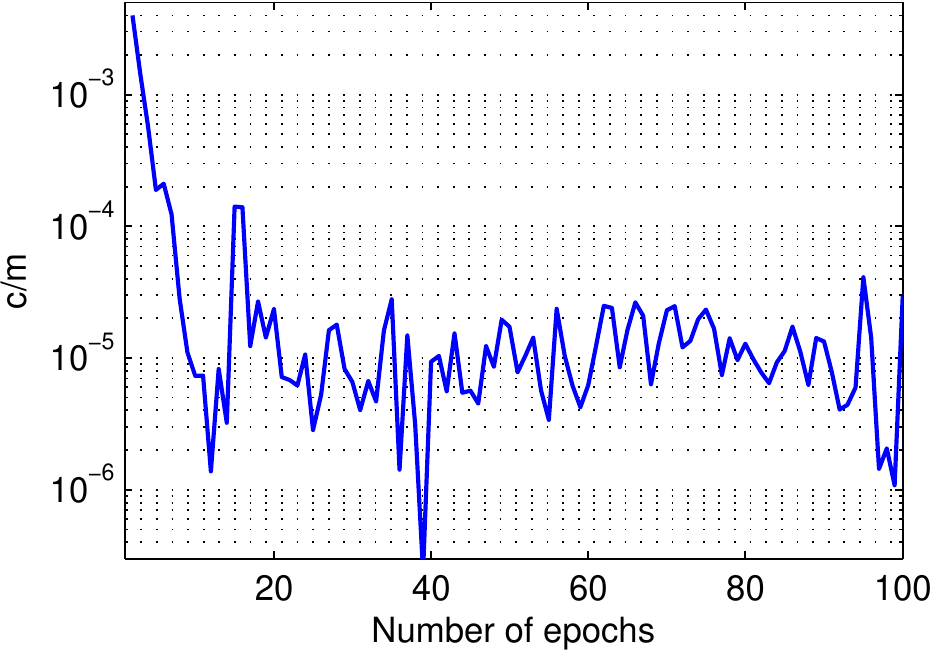}}
\subfigure[Covtype: $\lambda=10^{-4}$ (left) \;and\; $\lambda=10^{-5}$ (right)]{\includegraphics[width=0.439\columnwidth]{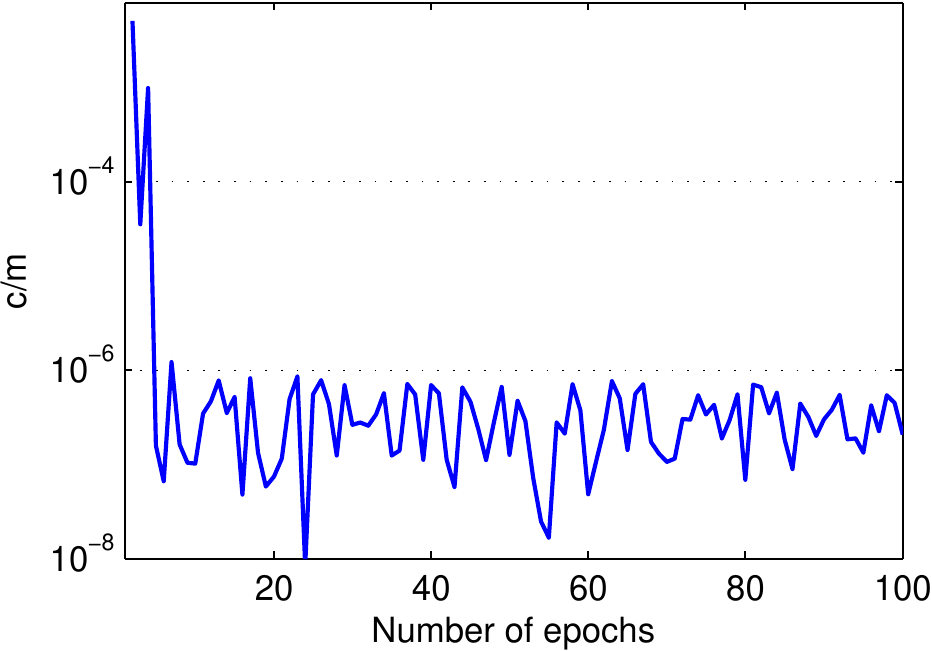}\;\;\;\includegraphics[width=0.439\columnwidth]{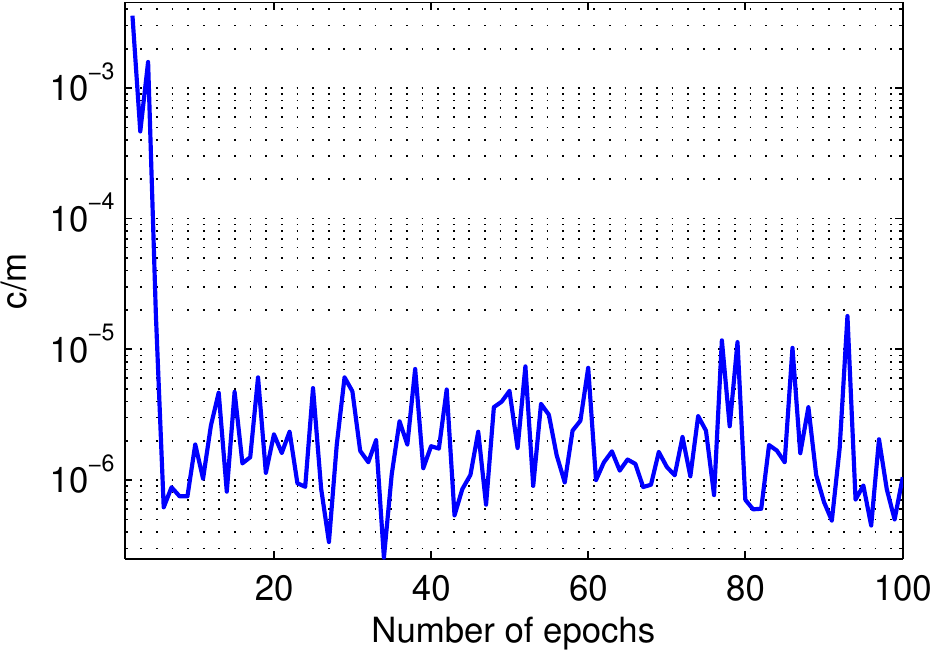}}
\caption{The value of $c/m$ vs.\ the number of epochs for VR-SGD, where $m=2n$.}
\label{figs01}
\end{figure}

\section*{Appendix F: Proofs of Theorems 4 and 7}
We also provide the convergence guarantees for VR-SGD under the strongly convex condition. We first give the following assumption.

\begin{assumption}\label{assum3}
For all $s\!=\!1,2,\ldots,S$, the following inequality holds
\begin{equation*}
\mathbb{E}\!\left[F(x^{s}_{0})-F(x^{*})\right]\leq c\;\!\mathbb{E}\!\left[F(\widetilde{x}^{s-\!1})-F(x^{*})\right]
\end{equation*}
where $0<c\ll m$ is a constant{\footnote{This assumption shows the relationship of the gaps between the function values at the starting and snapshot points of each epoch and the optimal value of the objective function. In fact, as our algorithm progresses, $\widetilde{x}^{s-\!1}$ and $x^{s-\!1}_{m}$ (i.e., $x^{s}_{0}$) both converge toward the same optimal point $x^{*}$, and thus $c$ is far less than $m$, i.e., $c\ll m$, as shown in Fig.\ \ref{figs01}. In fact, we can choose a simple restart technique as in~\cite{zhu:Katyusha} instead of this assumption and achieve the same convergence guarantee.}}.
\end{assumption}
\vspace{3mm}

\textbf{Proof of Theorem 4:}
\begin{proof}
Since each $f_{i}(\cdot)$ is convex and $L$-smooth, then Corollary~\ref{lemm7} holds, which then implies that
\begin{equation}\label{equ56}
\begin{split}
&\left(1-\frac{2}{\beta\!-\!1}\right)\mathbb{E}\!\left[F(\widetilde{x}^{s})-F(x^{*})\right]+\frac{1}{m\!-\!1}\mathbb{E}\!\left[F(x^{s+1}_{0})-F(x^{*})\right]\!+\!\frac{L\beta }{2(m\!-\!1)}\mathbb{E}\!\left[\|x^{s+1}_{0}\!-x^{*}\|^2\right]\\
\leq& \frac{2m}{(m\!-\!1)(\beta\!-\!1)}\mathbb{E}\!\left[F(\widetilde{x}^{s-\!1})-F(x^{*})\right]\!+\!\frac{2}{(m\!-\!1)(\beta\!-\!1)}\mathbb{E}\!\left[F(x^{s}_{0})-F(x^{*})\right]\!+\!\frac{L\beta }{2(m\!-\!1)}\mathbb{E}\!\left[\|x^{s}_{0}-x^{*}\|^2\right].
\end{split}
\end{equation}
Due to the strong convexity of $F(\cdot)$, we have $\|x^{s}_{0}-x^{*}\|^2\leq({2}/{\mu})[F(x^{s}_{0})-F(x^{*})]$. Then the inequality in (\ref{equ56}) can be rewritten as follows:
\begin{equation*}
\begin{split}
&\:\left(1-\frac{2}{\beta\!-\!1}\right)\mathbb{E}\!\left[F(\widetilde{x}^{s})-F(x^{*})\right]\\
\leq&\: \frac{2m}{(m\!-\!1)(\beta\!-\!1)}\mathbb{E}\!\left[F(\widetilde{x}^{s-\!1})-F(x^{*})\right]\!+\!\left(\frac{2}{(m\!-\!1)(\beta\!-\!1)}+\frac{L\beta }{\mu(m\!-\!1)}\right)\mathbb{E}\!\left[F(x^{s}_{0})-F(x^{*})\right]\\
\leq&\: \frac{2m}{(m\!-\!1)(\beta\!-\!1)}\mathbb{E}\!\left[F(\widetilde{x}^{s-\!1})-F(x^{*})\right]\!+\!\left(\frac{2c}{(m\!-\!1)(\beta\!-\!1)}+\frac{cL\beta }{\mu(m\!-\!1)}\right)\mathbb{E}\!\left[F(\widetilde{x}^{s-\!1})-F(x^{*})\right]\\
=&\:\left(\frac{2(m\!+\!c)}{(m\!-\!1)(\beta\!-\!1)}+\frac{cL\beta }{\mu(m\!-\!1)}\right)\mathbb{E}\!\left[F(\widetilde{x}^{s-\!1})-F(x^{*})\right]
\end{split}
\end{equation*}
where the first inequality holds due to the fact that $\|x^{s}_{0}-x^{*}\|^2\leq({2}/{\mu})[F(x^{s}_{0})-F(x^{*})]$, and the second inequality follows from Assumption 3. Dividing both sides of the above inequality by $[1\!-\!{2}/({\beta\!-\!1})]\!>\!0$ (that is, $\beta$ is required to be larger than 3) and using the definition of $\beta\!=\!1/(L\eta)$, we arrive at
\begin{equation*}
\begin{split}
\mathbb{E}\!\left[F(\widehat{x}^{s})-F(x^{*})\right]\leq&\:\mathbb{E}\!\left[F(\widetilde{x}^{s})-F(x^{*})\right]\\
\leq&\:\left(\frac{2(m\!+\!c)}{(m\!-\!1)(\beta\!-\!3)}+\frac{cL\beta(\beta\!-\!1)}{\mu(m\!-\!1)(\beta\!-\!3)}\right)\mathbb{E}\!\left[F(\widetilde{x}^{s-\!1})-F(x^{*})\right]\\
=&\:\left(\frac{2(m\!+\!c)L\eta}{(m\!-\!1)(1\!-\!3L\eta)}+\frac{c(1\!-\!L\eta)}{\mu\eta(m\!-\!1)(1\!-\!3L\eta)}\right)\mathbb{E}\!\left[F(\widetilde{x}^{s-\!1})-F(x^{*})\right].
\end{split}
\end{equation*}
This completes the proof.
\end{proof}
\vspace{3mm}

Similar to Theorem 4, we give the following convergence result for Algorithm 2 with Option I for non-smooth and strongly-convex functions.

\begin{theorem}[Option I]
\label{the7}
Suppose Assumptions 1, 2, and 3 hold, and $m$ is sufficiently large so that
\begin{equation*}
\rho_{I}:=\,\frac{2L\eta(m\!+\!c)}{m(1\!-\!3L\eta)}+\frac{c(1\!-\!L\eta)}{m\mu\eta(1\!-\!3L\eta)}< 1.
\end{equation*}
Then Algorithm 2 with Option I has the following geometric convergence in expectation:
\begin{equation*}
\mathbb{E}\left[F(\widehat{x}^{S})-F(x^{*})\right]\leq\,\rho^{S}_{I}\left[F(\widetilde{x}^{0})-F(x^{*})\right].
\end{equation*}
\end{theorem}
\vspace{3mm}

\textbf{Proof of Theorem~\ref{the7}:}
\begin{proof}
Since each $f_{i}(\cdot)$ is convex and $L$-smooth, then we have
\begin{equation*}
\begin{split}
&\left(1-\frac{2}{\beta\!-\!1}\right)\mathbb{E}\!\left[F(\widetilde{x}^{s})-F(x^{*})\right]+\frac{2}{m(\beta\!-\!1)}\mathbb{E}\!\left[F(x^{s+1}_{0})-F(x^{*})\right]+\frac{L\beta}{2m}\mathbb{E}\!\left[\|x^{s+1}_{0}-x^{*}\|^2\right]\\
\leq&\,\frac{2}{\beta\!-\!1}\mathbb{E}\!\left[F(\widetilde{x}^{s-\!1})-F(x^{*})\right]+\frac{2}{m(\beta\!-\!1)}\mathbb{E}\!\left[F(x^{s}_{0})-F(x^{*})\right]+ \frac{L\beta}{2m}\mathbb{E}\!\left[\|x^{s}_{0}-x^{*}\|^2\right]\\
\leq&\,\frac{2}{\beta\!-\!1}\mathbb{E}\!\left[F(\widetilde{x}^{s-\!1})-F(x^{*})\right]+\left(\frac{2}{m(\beta\!-\!1)}+\frac{L\beta}{m\mu}\right)\mathbb{E}\!\left[F(x^{s}_{0})-F(x^{*})\right]\\
\leq&\,\left(\frac{2(m\!+\!c)}{m(\beta\!-\!1)}+\frac{cL\beta}{m\mu}\right)\mathbb{E}\!\left[F(\widetilde{x}^{s-1})-F(x^{*})\right]
\end{split}
\end{equation*}
where the first inequality follows from Lemma~\ref{lemm6}; the second inequality holds due to the fact that $\|x^{s}_{0}-x^{*}\|^2\leq({2}/{\mu})[F(x^{s}_{0})-F(x^{*})]$; and the last inequality follows from Assumption 3.

Due to the definition of $\beta=1/(L\eta)$, the above inequality is rewritten as follows:
\begin{equation*}
\begin{split}
\frac{1\!-\!3L\eta}{1\!-\!L\eta}\mathbb{E}\!\left[F(\widetilde{x}^{s})-F(x^{*})\right]\leq\left(\frac{2L\eta(m\!+\!c)}{m(1\!-\!L\eta)}+\frac{c}{m\mu\eta}\right)\mathbb{E}\!\left[F(\widetilde{x}^{s-1})-F(x^{*})\right].
\end{split}
\end{equation*}
Dividing both sides of the above inequality by $(1\!-\!3L\eta)(1\!-\!L\eta)>0$, we arrive at
\begin{equation*}
\begin{split}
\mathbb{E}\!\left[F(\widehat{x}^{s})-F(x^{*})\right]\leq&\:\mathbb{E}\!\left[F(\widetilde{x}^{s})-F(x^{*})\right]\\
\leq&\:\left(\frac{2L\eta(m\!+\!c)}{m(1\!-\!3L\eta)}+\frac{c(1\!-\!L\eta)}{m\mu\eta(1\!-\!3L\eta)}\right)\mathbb{E}\!\left[F(\widetilde{x}^{s-1})-F(x^{*})\right].
\end{split}
\end{equation*}
This completes the proof.
\end{proof}
It is clear that Theorem~\ref{the7} shows that VR-SGD attains a \emph{linear} convergence rate and at most the oracle complexity of $\mathcal{O}\!\left((n\!+\!{L}/{\mu})\log({1}/{\epsilon})\right)$ for \emph{non-smooth} and strongly convex functions.
\vspace{3mm}

\section*{Appendix G: Proof of Theorem 5 and Equivalent Relationship}
Before proving Theorem 5, we first give and prove the following lemmas.

\begin{lemma}
\label{lemm10}
Suppose each convex function $f_{i}(\cdot)$ is $L$-smooth. Then the following inequality holds:
\begin{equation*}
\begin{split}
\mathbb{E}\!\left[\left\|\widetilde{\nabla}\! f_{i^{s}_{k}}(x^{s}_{k})-\nabla\! f(x^{s}_{k})\right\|^{2}\right]
\leq2L\left[f(\widetilde{x}^{s-\!1})-f(x^{s}_{k})-\left\langle\nabla\!f(x^{s}_{k}),\;\widetilde{x}^{s-\!1}-x^{s}_{k}\right\rangle\right].
\end{split}
\end{equation*}
\end{lemma}

Lemma~\ref{lemm10} is essentially identical to Lemma 3.4 in~\cite{zhu:Katyusha}. This lemma provides a tighter upper bound on the expected variance of the variance-reduced gradient estimator $\widetilde{\nabla}\! f_{i^{s}_{k}}\!(x^{s}_{k})$ than Lemma 1 in this paper and that of~\cite{xiao:prox-svrg,zhu:univr2}, e.g., Lemma A.2 in~\cite{zhu:univr2}.
\vspace{1mm}

\begin{lemma}
\label{lemm11}
Suppose Assumption 1 holds. Let $x^{*}$ be an optimal solution of Problem (1), and $\{(v^{s}_{k},x^{s}_{k})\}$ be the sequence generated by Algorithm 3 with Option II. Then for all $s\!=\!1,\ldots,S$,
\begin{equation}\label{equ70}
\begin{split}
\mathbb{E}\!\left[F(\widetilde{x}^{s})-F(x^{*})\right]\,\leq&\;(1-w_{s})\!\left[F(\widetilde{x}^{s-\!1})-F(x^{*})\right]\\
&\;+\frac{w^{2}_{s}}{2m\eta_{0}}\mathbb{E}\!\left[\|x^{*}-v^{s}_{0}\|^2-\|x^{*}-v^{s}_{m}\|^2\right].
\end{split}
\end{equation}
\end{lemma}
\vspace{3mm}

\begin{proof}
Because Assumption 1 holds, the smoothness inequality in Assumption 1 has the following equivalent form,
\begin{displaymath}
f_{i}(y)\leq f_{i}(x)+\langle\nabla\! f_{i}(x),\;y-x\rangle+\frac{\widetilde{L}}{2}\|y-x\|^{2},\;\;\forall x,y\in\mathbb{R}^{d},\;i\in[n].
\end{displaymath}
According to the definition of $f(x)$ (i.e., $f(x)\!:=\!\frac{1}{n}\!\sum^{n}_{i=1}\!f_{i}(x)$), we have
\begin{displaymath}
f(y)\leq f(x)+\langle\nabla\! f(x),\;y-x\rangle+\frac{\widetilde{L}}{2}\|y-x\|^{2},\;\;\forall x,y\in\mathbb{R}^{d},
\end{displaymath}
\begin{displaymath}
g(y)\leq g(x)+\langle\nabla\! g(x),\;y-x\rangle+\frac{L'}{2}\|y-x\|^{2},\;\;\forall x,y\in\mathbb{R}^{d}
\end{displaymath}
where the smooth convex regularizer $g(x)$ is $L'$-smooth. Then
\begin{displaymath}
F(y)\leq F(x)+\langle\nabla\! F(x),\;y-x\rangle+\frac{L}{2}\|y-x\|^{2},\;\;\forall x,y\in\mathbb{R}^{d}
\end{displaymath}
where $L=\widetilde{L}+L'\geq \widetilde{L}$. In other words, $F(x)$ is $L$-smooth. Let $\widetilde{\nabla}_{i^{s}_{k}}:=\nabla\! f_{i^{s}_{k}}(x^{s}_{k})-\nabla\! f_{i^{s}_{k}}(\widetilde{x}^{s-1})+\nabla\! f(\widetilde{x}^{s-1})+\nabla\!g(x^{s}_{k})$, $\alpha_{1}>2$ be an appropriate constant, and $\alpha_{2}=\alpha_{1}-1>1$. Using the above inequality, we have
\begin{equation}\label{equ71}
\begin{split}
F(x^{s}_{k+1})\leq&\; F(x^{s}_{k})+\langle\nabla\!F(x^{s}_{k}),\;x^{s}_{k+1}-x^{s}_{k}\rangle+\frac{{L}}{2}\|x^{s}_{k+1}-x^{s}_{k}\|^{2}\\
=&\; F(x^{s}_{k})+\left\langle\nabla\! F(x^{s}_{k}),\;x^{s}_{k+1}-x^{s}_{k}\right\rangle+\frac{\alpha_{1}{L}}{2}\!\left\|x^{s}_{k+1}-x^{s}_{k}\right\|^{2}-\frac{\alpha_{2}{L}}{2}\left\|x^{s}_{k+1}-x^{s}_{k}\right\|^{2}\\
=&\, F(x^{s}_{k})+\left\langle \widetilde{\nabla}_{i^{s}_{k}}, x^{s}_{k+1}-x^{s}_{k}\right\rangle+\frac{\alpha_{1}{L}}{2}\|x^{s}_{k+1}-x^{s}_{k}\|^2
+\left\langle\nabla\! F(x^{s}_{k})-\widetilde{\nabla}_{i^{s}_{k}},\;x^{s}_{k+1}-x^{s}_{k}\right\rangle-\frac{\alpha_{2}{L}}{2}\|x^{s}_{k+1}-x^{s}_{k}\|^{2}.
\end{split}
\end{equation}

\begin{equation}\label{equ72}
\begin{split}
&\mathbb{E}\!\left[\left\langle\nabla\! F(x^{s}_{k})-\widetilde{\nabla}_{i^{s}_{k}},\;x^{s}_{k+1}-x^{s}_{k}\right\rangle\right]\\
\leq\,& \mathbb{E}\!\left[\frac{1}{2\alpha_{2}{L}}\|\nabla\!F(x^{s}_{k})-\widetilde{\nabla}_{i^{s}_{k}}\|^{2}+\frac{\alpha_{2}{L}}{2}\|x^{s}_{k+1}-x^{s}_{k}\|^{2}\right]\\
\leq\,& \frac{1}{\alpha_{2}}\!\left[F(\widetilde{x}^{s-\!1})-F(x^{s}_{k})-\left\langle\nabla\! F(x^{s}_{k}),\;\widetilde{x}^{s-\!1}-x^{s}_{k}\right\rangle\right]+\frac{\alpha_{2}{L}}{2}\;\!\mathbb{E}\!\left[\|x^{s}_{k+1}-x^{s}_{k}\|^{2}\right]
\end{split}
\end{equation}
where the first inequality follows from the Young's inequality, i.e.,
\begin{displaymath}
x^{T}y\leq \frac{\|x\|^2}{2\alpha}+\frac{\alpha\|y\|^2}{2},\;\;\textup{for all}\;\;\alpha>0,
\end{displaymath}
and the second inequality holds due to $\widetilde{\nabla}_{i^{s}_{k}}\!=\!\nabla\! F_{i^{s}_{k}}(x^{s}_{k})-\nabla\! F_{i^{s}_{k}}(\widetilde{x}^{s-1})+\nabla\! F(\widetilde{x}^{s-1})$ and Lemma~\ref{lemm10}, where $F_{i}(x)=f_{i}(x)+g(x)$, and $F_{i}(x)$ is $L$-smooth. Substituting the inequality (\ref{equ72}) into the inequality (\ref{equ71}), and taking expectation over $i^{s}_{k}$, we have
\begin{equation}\label{equ73}
\begin{split}
&\quad\:\mathbb{E}\!\left[F(x^{s}_{k+1})\right]-F(x^{s}_{k})\\
&\leq\mathbb{E}\!\left[\left\langle\widetilde{\nabla}_{i^{s}_{k}},\:x^{s}_{k+1}\!-x^{s}_{k}\right\rangle+ \frac{\alpha_{1}L}{2}\|x^{s}_{k+1}\!-x^{s}_{k}\|^2\right]+\frac{1}{\alpha_{2}}\!\left[F(\widetilde{x}^{s-\!1})-F(x^{s}_{k})-\left\langle\nabla\! F(x^{s}_{k}),\,\widetilde{x}^{s-\!1}-x^{s}_{k}\right\rangle\right]\\
&=\mathbb{E}\!\left[\left\langle w_{s}\widetilde{\nabla}_{i^{s}_{k}},\: v^{s}_{k+1}\!-v^{s}_{k}\right\rangle+\frac{\alpha_{1}Lw_{s}^{2}}{2}\|v^{s}_{k+1}\!-v^{s}_{k}\|^2\right]+\frac{1}{\alpha_{2}}\!\left[F(\widetilde{x}^{s-\!1})-F(x^{s}_{k})-\left\langle\nabla\! F(x^{s}_{k}),\:\widetilde{x}^{s-\!1}-x^{s}_{k}\right\rangle\right]\\
&\leq\mathbb{E}\!\left[\left\langle w_{s}\widetilde{\nabla}_{\!i^{s}_{k}},\;x^{*}-v^{s}_{k}\right\rangle+\frac{\alpha_{1}L w_{s}^{2}}{2}\left(\|v^{s}_{k}-x^{*}\|^2-\|v^{s}_{k+1}-x^{*}\|^2\right)\right]\\
&\qquad+\frac{1}{\alpha_{2}}\!\left[F(\widetilde{x}^{s-\!1})-F(x^{s}_{k})-\left\langle\nabla\! F(x^{s}_{k}),\;\widetilde{x}^{s-\!1}-x^{s}_{k}\right\rangle\right]\\
&=\mathbb{E}\!\left[\frac{\alpha_{1}Lw_{s}^{2}}{2}\!\left(\|v^{s}_{k}-x^{*}\|^2-\|v^{s}_{k+1}\!-x^{*}\|^2\right)\right]+\left\langle\nabla\! F(x^{s}_{k}),\:w_{s} x^{*}+(1\!-\!w_{\!s})\widetilde{x}^{s-\!1}-x^{s}_{k}-\frac{1}{\alpha_{2}}(\widetilde{x}^{s-\!1}-x^{s}_{k})\right\rangle\\
&\quad\;+\mathbb{E}\!\left[\left\langle\nabla\! f(\widetilde{x}^{s-\!1})-\nabla\!f_{i^{s}_{k}}(\widetilde{x}^{s-\!1}),\:w_{s} x^{*}+(1\!-\!w_{s})\widetilde{x}^{s-\!1}-x^{s}_{k}\right\rangle\right]+\frac{1}{\alpha_{2}}\!\left[F(\widetilde{x}^{s-\!1})-F(x^{s}_{k})\right]\\
&=\mathbb{E}\!\left[\frac{\alpha_{1}L w_{s}^{2}}{2}\!\left(\|v^{s}_{k}-x^{*}\|^2-\|v^{s}_{k+1}\!-x^{*}\|^2\right)\right]+\left\langle\nabla\! F(x^{s}_{k}),\:w_{s} x^{*}+(1\!-\!w_{s})\widetilde{x}^{s-\!1}-x^{s}_{k}-\frac{1}{\alpha_{2}}(\widetilde{x}^{s-\!1}-x^{s}_{k})\right\rangle\\
&\quad\;+\frac{1}{\alpha_{2}}\!\left[F(\widetilde{x}^{s-\!1})-F(x^{s}_{k})\right]
\end{split}
\end{equation}
where the first equality holds due to the fact that $x^{s}_{k+\!1}-x^{s}_{k}=w_{s}(v^{s}_{k+\!1}-v^{s}_{k})$ ($x^{s}_{k+\!1}=\widetilde{x}^{s-\!1}+w_{s}(v^{s}_{k+\!1}-\widetilde{x}^{s-\!1})=w_{s}v^{s}_{k+\!1}+(1\!-\!w_{s})\widetilde{x}^{s-\!1}$); the second inequality holds due to Lemma 2 with $z^{*}\!=\!v^{s}_{k+1}$, $z\!=\!x^{*}$, $z_{0}\!=\!v^{s}_{k}$, $\tau\!=\!\alpha_{1}Lw_{\!s}$, and $r(z)\!:=\!\langle \widetilde{\nabla}_{\!i^{s}_{k}},\,z\!-\!v^{s}_{k}\rangle$. The second equality follows from the facts that
\begin{displaymath}
\begin{split}
\left\langle w_{s}\widetilde{\nabla}_{i^{s}_{k}},\; x^{*}-v^{s}_{k}\right\rangle=&\,\left\langle \widetilde{\nabla}_{i^{s}_{k}},\; w_{s}(x^{*}-v^{s}_{k})\right\rangle=\left\langle \widetilde{\nabla}_{i^{s}_{k}},\; w_{s} x^{*}\!+\!(1\!-\!w_{s})\widetilde{x}^{s-\!1}\!-x^{s}_{k}\right\rangle\\
=&\;\left\langle\nabla\! F_{i^{s}_{k}}\!(x^{s}_{k}),\, w_{s} x^{*}\!+\!(1\!-\!w_{s})\widetilde{x}^{s-\!1}\!-\!x^{s}_{k}\right\rangle+\left\langle\nabla\! f(\widetilde{x}^{s-\!1})\!-\!\nabla\! f_{i^{s}_{k}}\!(\widetilde{x}^{s-\!1}),\, w_{s} x^{*}\!+\!(1\!-\!w_{s})\widetilde{x}^{s-\!1}\!-\!x^{s}_{k}\right\rangle,
\end{split}
\end{displaymath}
$\widetilde{\nabla}_{i^{s}_{k}}\!=\!\nabla\! F_{i^{s}_{k}}(x^{s}_{k})\!-\!\nabla\! F_{i^{s}_{k}}(\widetilde{x}^{s-1})\!+\!\nabla\! F(\widetilde{x}^{s-1})\!=\!\nabla\! F_{i^{s}_{k}}(x^{s}_{k})\!-\!\nabla\! f_{i^{s}_{k}}(\widetilde{x}^{s-1})\!+\!\nabla\! f(\widetilde{x}^{s-1})$, and $\mathbb{E}[\nabla\! F_{i^{s}_{k}}\!(x^{s}_{k})]\!=\!\nabla\! F(x^{s}_{k})$. The last equality holds due to the facts that $\mathbb{E}\!\left[\nabla\! f(\widetilde{x}^{s-\!1})\!-\!\nabla\! f_{i^{s}_{k}}\!(\widetilde{x}^{s-\!1})\right]=0$ and
$\mathbb{E}\!\left[\langle-\nabla\! f_{i^{s}_{k}}\!(\widetilde{x}^{s-\!1})\!+\!\nabla\! f(\widetilde{x}^{s-\!1}),\,w_{s} x^{*}\!+\!(1\!-\!w_{s})\widetilde{x}^{s-\!1}\!\!-\!x^{s}_{k}\rangle\right]=0$.

\begin{equation}\label{equ75}
\begin{split}
&\;\left\langle\nabla\! F(x^{s}_{k}),\;w_{s} x^{*}+(1\!-\!w_{s})\widetilde{x}^{s-\!1}-x^{s}_{k}+\frac{1}{\alpha_{2}}\left(x^{s}_{k}-\widetilde{x}^{s-\!1}\right)\right\rangle\\
=&\;\left\langle\nabla\! F(x^{s}_{k}),\;w_{s} x^{*}+(1\!-\!w_{s}-\frac{1}{\alpha_{2}})\widetilde{x}^{s-\!1}+\frac{1}{\alpha_{2}}x^{s}_{k}-x^{s}_{k}\right\rangle\\
\leq &\;F\!\left(w_{s} x^{*}+(1\!-\!w_{s}-\frac{1}{\alpha_{2}})\widetilde{x}^{s-\!1}+\frac{1}{\alpha_{2}}x^{s}_{k}\right)-F(x^{s}_{k})\\
\leq &\;w_{s} F(x^{*})+\left(1\!-\!w_{s}-\frac{1}{\alpha_{2}}\right)F(\widetilde{x}^{s-\!1})+\frac{1}{\alpha_{2}}F(x^{s}_{k})-F(x^{s}_{k})
\end{split}
\end{equation}
where the first inequality holds due to the fact that $\langle \nabla\! F(x_{1}),$ $x_{2}-x_{1}\rangle\!\leq\! F(x_{2})\!-\!F(x_{1})$, and the last inequality follows from the convexity of $F(\cdot)$ and $1\!-\!w_{s}\!-\!1/\alpha_{2}\!\geq\!0$. By combining the above two inequalities in (\ref{equ73}) and (\ref{equ75}), we have
\begin{equation*}\label{equ76}
\begin{split}
\mathbb{E}\!\left[F(x^{s}_{k+1})\right]\leq w_{s}F(x^{*})+(1\!-\!w_{s})F(\widetilde{x}^{s-\!1})+\frac{\alpha_{1}L w_{s}^{2}}{2}\mathbb{E}\!\left[\|v^{s}_{k}-x^{*}\|^2-\|v^{s}_{k+1}-x^{*}\|^2\right].
\end{split}
\end{equation*}
Subtracting $F(x^{*})$ from both sides of the above inequality, then
\begin{equation*}
\begin{split}
\mathbb{E}\!\left[F(x^{s}_{k+1})-F(x^{*})\right]\leq(1\!-\!w_{s})[F(\widetilde{x}^{s-\!1})-F(x^{*})]+\frac{\alpha_{1}L w_{s}^{2}}{2}\mathbb{E}\!\left[\|v^{s}_{k}-x^{*}\|^2-\|v^{s}_{k+1}-x^{*}\|^2\right].
\end{split}
\end{equation*}
Due to the convexity of $F(\cdot)$ and $\widetilde{x}^{s}=\frac{1}{m}\sum^{m}_{k=1}\!x^{s}_{k}$, then
\begin{equation*}
F(\widetilde{x}^{s})=F\!\left(\frac{1}{m}\sum^{m}_{k=1}\!x^{s}_{k}\right)\leq\frac{1}{m}\!\sum^{m}_{k=1}F(x^{s}_{k}).
\end{equation*}
Taking expectation over the history of $i^{s}_{1},\ldots,i^{s}_{m}$ on the above inequality, and summing it up over $k\!=\!1,\ldots,m$ at the $s$-th epoch, we have
\begin{equation*}
\begin{split}
\mathbb{E}\!\left[F(\widetilde{x}^{s})-F(x^{*})\right]
\leq (1\!-\!w_{s})\!\left[F(\widetilde{x}^{s-\!1})-F(x^{*})\right]+\frac{w_{s}^{2}}{2\eta_{0}m}\mathbb{E}\!\left[\|v^{s}_{0}-x^{*}\|^2-\|v^{s}_{m}-x^{*}\|^2\right]
\end{split}
\end{equation*}
where $\eta_{0}=1/(\alpha_{1}L)$, e.g., $\eta_{0}=1/(3L)$. This completes the proof.
\end{proof}
\vspace{3mm}

\textbf{Proof of Theorem 5:}
\begin{proof}
Because $w_{s}\!=\!\max\{\alpha, 2/(s\!+\!1)\}$, and $\alpha$ is sufficiently small, it is easy to verify that
\begin{equation}\label{equ104}
({1-w_{s+1}})/{w^{2}_{s+1}}\leq{1}/{w^{2}_{s}}.
\end{equation}
Dividing both sides of the inequality in (\ref{equ70}) by $w^{2}_{s}$, we have
\begin{equation*}
\begin{split}
\mathbb{E}[F(\widetilde{x}^{s})-F(x^{*})]/w^{2}_{s}
\leq\frac{1\!-\!w_{s}}{w^{2}_{s}}[F(\widetilde{x}^{s-\!1})-F(x^{*})]+\frac{1}{2m\eta_{0}}\mathbb{E}\!\left[\|x^{*}-v^{s}_{0}\|^2-\|x^{*}-v^{s}_{m}\|^2\right]\!,
\end{split}
\end{equation*}
for all $s=1,\ldots,S$. Using the above inequality, we obtain
\begin{equation*}
\begin{split}
\mathbb{E}[F(\widetilde{x}^{S})-F(x^{*})]/w^{2}_{S}
\leq\frac{1\!-\!w_{S}}{w^{2}_{S}}[F(\widetilde{x}^{S-\!1})-F(x^{*})]+\frac{1}{2m\eta_{0}}\mathbb{E}\!\left[\|x^{*}-v^{S}_{0}\|^2-\|x^{*}-v^{S}_{m}\|^2\right]\!,
\end{split}
\end{equation*}
\begin{equation*}
\begin{split}
\mathbb{E}[F(\widetilde{x}^{S-1})-F(x^{*})]/w^{2}_{S-1}
\leq\frac{1\!-\!w_{S-1}}{w^{2}_{S-1}}[F(\widetilde{x}^{S-\!2})-F(x^{*})]+\frac{1}{2m\eta_{0}}\mathbb{E}\!\left[\|x^{*}-v^{S-1}_{0}\|^2-\|x^{*}-v^{S-1}_{m}\|^2\right]\!.
\end{split}
\end{equation*}

Using the inequality in (\ref{equ104}), e.g., $({1-w_{S}})/{w^{2}_{S}}\!\leq\!{1}/{w^{2}_{S-1}}$, and $v^{s+\!1}_{0}=v^{s}_{m}$, e.g., $v^{S}_{0}=v^{S-1}_{m}$, we have
\begin{equation*}
\begin{split}
\mathbb{E}\!\left[F(\widetilde{x}^{S})-F(x^{*})\right]/w^{2}_{S}
\leq\frac{1\!-\!w_{S-1}}{w^{2}_{S-1}}[F(\widetilde{x}^{S-\!2})-F(x^{*})]+\frac{1}{2m\eta_{0}}\mathbb{E}\!\left[\|x^{*}-v^{S-1}_{0}\|^2-\|x^{*}-v^{S}_{m}\|^2\right]\!,
\end{split}
\end{equation*}
\begin{equation*}
\begin{split}
\mathbb{E}\!\left[F(\widetilde{x}^{S})-F(x^{*})\right]/w^{2}_{S}
\leq\frac{1\!-\!w_{1}}{w^{2}_{1}}[F(\widetilde{x}^{0})-F(x^{*})]+\frac{1}{2m\eta_{0}}\mathbb{E}\!\left[\|x^{*}-v^{1}_{0}\|^2-\|x^{*}-v^{S}_{m}\|^2\right]\!.
\end{split}
\end{equation*}

Due to the definition of $w^{2}_{S}$ and $v^{1}_{0}=\widetilde{x}^{0}$, then
\begin{equation*}
\begin{split}
\mathbb{E}[F(\widehat{x}^{S})-F(x^{*})]\leq&\;\mathbb{E}\!\left[F(\widetilde{x}^{S})-F(x^{*})\right]\\
\leq&\;\frac{4(1\!-\!w_{1})}{w^{2}_{1}(S\!+\!1)^{2}}[F(\widetilde{x}^{0})-F(x^{*})]+\frac{2}{m\eta_{0}(S\!+\!1)^{2}}\mathbb{E}\!\left[\|x^{*}-v^{1}_{0}\|^2-\|x^{*}-v^{S}_{m}\|^2\right]\\
\leq&\;\frac{4(1\!-\!w_{1})}{w^{2}_{1}(S\!+\!1)^{2}}[F(\widetilde{x}^{0})-F(x^{*})]+\frac{2}{m\eta_{0}(S\!+\!1)^{2}}\left[\|x^{*}-\widetilde{x}^{0}\|^2\right]\!.
\end{split}
\end{equation*}
This completes the proof.
\end{proof}

Theorem 5 shows that Algorithm 3 with Option II attains the optimal convergence rate $\mathcal{O}(1/T^2)$ for smooth and non-strongly convex functions. Similarly, the convergence of Algorithm 3 with Option II can be guaranteed for non-smooth and non-strongly convex functions. In addition, this theorem also requires that the learning rate $\eta$ must be less than $1/(3L)$, while that of SVRG should be less than $1/(4L)$. This means that our VR-SGD method can use much larger learning rates than SVRG both in theory and in practice.

\vspace{3mm}

\textbf{Equivalent Relationship}\\
Here, we discuss the equivalent relationship between Algorithm 3 with Option I (i.e., $v^{s}_{0}=x^{s}_{0}$ and $x^{s+1}_{0}\!=x^{s}_{m}$) and Algorithm 2 with Option I and a fixed learning rate $\eta_{0}$. Suppose $w_{s}=\max\{\alpha, 2/(s\!+\!1)\}$ and $\alpha$ is sufficiently small. Then Algorithm 3 with Option I has the following update rule:
\begin{displaymath}
\begin{split}
x^{s}_{k+1}-x^{s}_{k}&=w_{s}\left(v^{s}_{k+1}-\widetilde{x}^{s-1}\right)-w_{s}\left(v^{s}_{k}-\widetilde{x}^{s-1}\right)\\
&=w_{s}\left(v^{s}_{k+1}-v^{s}_{k}\right)=-w_{s}\eta_{s}\!\left[\widetilde{\nabla}\!f_{i^{s}_{k}}\!(x^{s}_{k})+\nabla g(x^{s}_{k})\right]\\
&=-\eta_{0}\!\left[\widetilde{\nabla}\!f_{i^{s}_{k}}\!(x^{s}_{k})+\nabla g(x^{s}_{k})\right].
\end{split}
\end{displaymath}
In other words, $x^{s}_{k+1}\!=x^{s}_{k}-\eta_{0}\!\left[\widetilde{\nabla}\!f_{i^{s}_{k}}\!(x^{s}_{k})+\nabla g(x^{s}_{k})\right]$. In addition, if Algorithms 2 and 3 have the same settings, e.g., the same initial point $\widetilde{x}^{0}$ and $x^{s+1}_{0}\!=x^{s}_{m}$, together with $v^{s}_{0}=x^{s}_{0}$ for Algorithm 3. Therefore, Algorithm 3 with Option I is theoretically equivalent to Algorithm 2 with Option I.
\vspace{3mm}

\section*{Appendix H: Convergence Analysis of VR-SGD++}
As shown in Section 5.5 of the main paper, VR-SGD++ is the variant of VR-SGD, and uses a general growing epoch size strategy in early iterations (i.e., If $m_{s}\!<\!m$, $m_{s+1}\!=\!\lfloor\rho m_{s}\rfloor$ with the factor $\rho\!>\!1$. Otherwise, $m_{s+1}\!=\!m_{s}$), as shown in Algorithm~\ref{alg4}. Note that the factor $\rho$ in our VR-SGD++ method can be any constant satisfying $\rho\!>\!1$, while the factor in SVRG++~\cite{zhu:univr} only is equal to $2$. In particular, unlike the doubling-epoch technique used in SVRG++~\cite{zhu:univr} (i.e., $m_{s+1}\!=\!2m_{s}$), we gradually increase the epoch size in only the early iterations of the epochs when $m_{s}\leq m$. And all the early iterations can viewed as the initialization steps of Algorithm~\ref{alg4}. Similar to Algorithm 2, the convergence of Algorithm~\ref{alg4} (i.e., VR-SGD++) can also be guaranteed.

\begin{algorithm}[t]
\caption{VR-SGD++}
\label{alg4}
\renewcommand{\algorithmicrequire}{\textbf{Input:}}
\renewcommand{\algorithmicensure}{\textbf{Initialize:}}
\renewcommand{\algorithmicoutput}{\textbf{Output:}}
\begin{algorithmic}[1]
\REQUIRE The number of epochs $S$.\\
\ENSURE $x^{1}_{0}=\widetilde{x}^{0}$, $\rho\!>\!1$, $m$, $m_{1}=\lfloor n/4\rfloor$, and $\{\eta_{s}\}$.\\
\FOR{$s=1,2,\ldots,S$}
\STATE {$\widetilde{\mu}^{s}=\frac{1}{n}\!\sum^{n}_{i=1}\!\nabla\!f_{i}(\widetilde{x}^{s-1})$;}
\FOR{$k=0,1,\ldots,m_{s}-1$}
\STATE {Pick $i^{s}_{k}$ uniformly at random from $[n]$;}
\STATE {$\widetilde{\nabla}\! f_{i^{s}_{k}}(x^{s}_{k})=\nabla\! f_{i^{s}_{k}}(x^{s}_{k})-\nabla\! f_{i^{s}_{k}}(\widetilde{x}^{s-1})+\widetilde{\mu}^{s}$;}
\STATE {$x^{s}_{k+\!1}=x^{s}_{k}-\eta_{s}[\widetilde{\nabla}\!f_{i^{s}_{k}}\!(x^{s}_{k})+\nabla\! g(x^{s}_{k})]$ \;or\; $x^{s}_{k+1}=\textrm{Prox}^{g}_{\eta}\!\left(x^{s}_{k}-\eta_{s}\widetilde{\nabla}\!f_{i^{s}_{k}}(x^{s}_{k})\right)$;}
\ENDFOR
\STATE {$\widetilde{x}^{s}\!=\!\frac{1}{m_{s}}\!\sum^{m_{s}}_{k=1}\!x^{s}_{k}$, \;$x^{s+1}_{0}\!=x^{s}_{m_{s}}$;}
\IF {$m_{s}<m$}
\STATE $m_{s+1}=\lfloor\rho m_{s}\rfloor$;
\ELSE
\STATE {$m_{s+1}=m_{s}$;}
\ENDIF
\ENDFOR
\OUTPUT {$\widehat{x}^{S}=\widetilde{x}^{S}$, if $F(\widetilde{x}^{S})\leq F(\frac{1}{S}\!\sum^{S}_{s=1}\!\widetilde{x}^{s})$, and $\widehat{x}^{S}=\frac{1}{S}\!\sum^{S}_{s=1}\!\widetilde{x}^{s}$ otherwise.}
\end{algorithmic}
\end{algorithm}

\section*{Appendix I: More Experimental Results}

\subsection*{Experimental Setup}
In this paper, we used five publicly available data sets in the experiments: Adult (also called a9a), Covtype, Epsilon, RCV1, and MNIST, as listed in Table~\ref{tab1}. For fair comparison, we implemented the state-of-the-art stochastic methods such as SAGA~\cite{defazio:saga}, SVRG~\cite{johnson:svrg}, Prox-SVRG~\cite{xiao:prox-svrg}, Catalyst~\cite{lin:vrsg}, and Katyusha~\cite{zhu:Katyusha}, and our VR-SGD method in C++ with a Matlab interface, and conducted all the experiments on a PC with an Intel i5-4570 CPU and 16GB RAM. All the codes of VR-SGD and related methods can be downloaded from \url{https://github.com/jnhujnhu/VR-SGD}. Furthermore, we also implemented the accelerated deterministic methods such as AGD~\cite{nesterov:co} and APG~\cite{beck:fista}, and other stochastic methods such as SGD and SVRG++~\cite{zhu:univr}.

\begin{table}[th]
\centering
\caption{Summary of data sets used for our experiments.}
\label{tab1}
\setlength{\tabcolsep}{12.9pt}
\renewcommand\arraystretch{1.19}
\begin{tabular}{lcccccc}
\hline
\ Data sets   & Adult & Covtype & Epsilon & RCV1 & MNIST\\
\hline
\ Number of training samples, $n$ & 32,562  & 581,012 & 400,000 & 20,242 & 60,000 \\
\ Number of dimensions, $d$       & 123     & 54 & 2,000 & 47,236 & 784 \\
\ Sparsity                        & 11.28\% & 22.12\% & 100\% & 0.16\% & 19.12\% \\
\ Size                            & 733K & 50M & 11G & 13M & 12M\\
\hline
\end{tabular}
\end{table}

\begin{figure}[t]
\centering
\subfigure[Logistic regression]{\includegraphics[width=0.439\columnwidth]{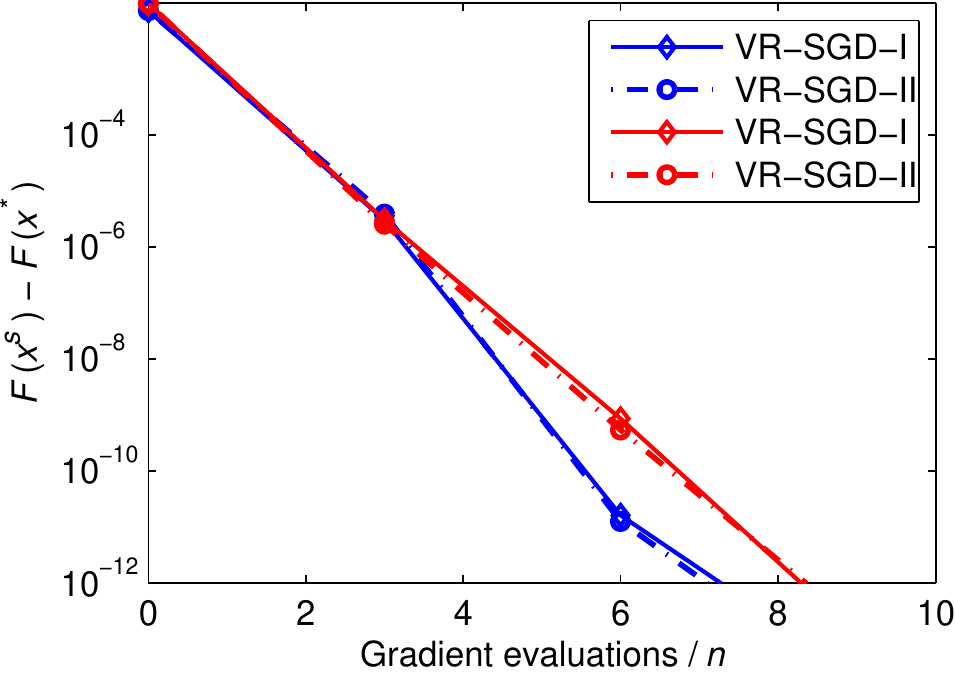}\label{figs11a}}\;\;\;
\subfigure[Ridge regression]{\includegraphics[width=0.439\columnwidth]{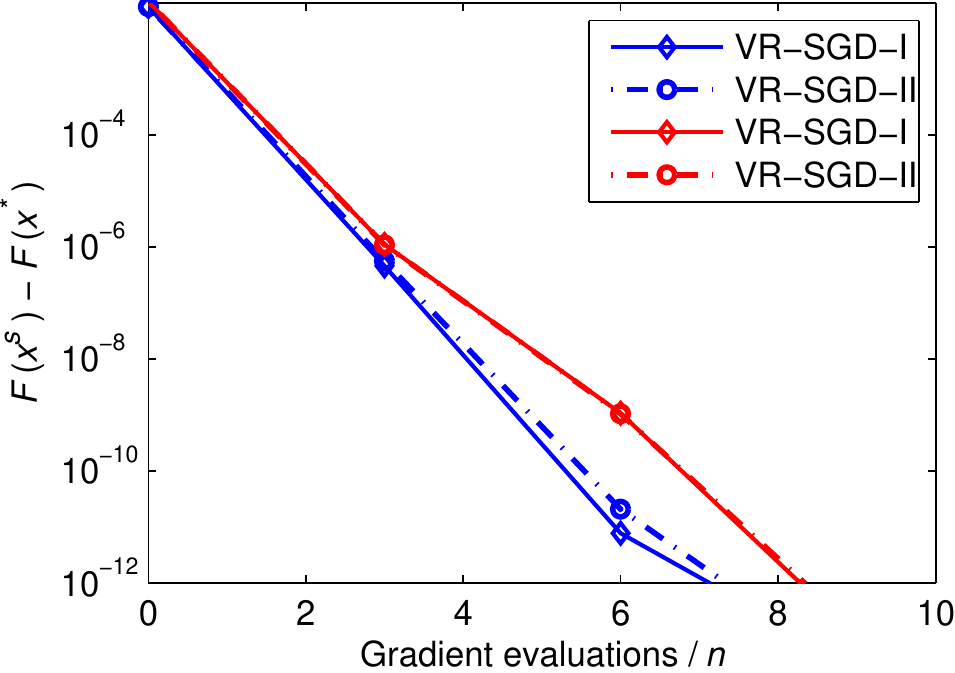}}
\caption{Comparison of our VR-SGD method with Option I (called VR-SGD-I) and Option II (called VR-SGD-II) for solving $\ell_{2}$-norm (i.e., $(\lambda/2)\|\!\cdot\!\|^{2}$) regularized logistic regression and ridge regression problems on the Covtype data set. In each plot, the vertical axis shows the objective value minus the minimum, and the horizontal axis is the number of effective passes. Note that the blue lines stand for the results for $\lambda\!=\!10^{-4}$, while the red lines correspond to the results for $\lambda\!=\!10^{-5}$ (best viewed in colors).}
\label{figs00}
\end{figure}

\subsection*{Comparison of VR-SGD with Options I and II}
We first compared the performance of VR-SGD with Option I and VR-SGD with Option II, as shown in Fig.\ \ref{figs00}. The results show that the performance of VR-SGD with Option I is almost identical to that of VR-SGD with Option II.

\begin{figure}[th]
\centering
\subfigure[Results of $\ell_{2}$-norm regularized logistic regression with $\lambda=10^{-3},10^{-4},10^{-5},10^{-6}$ (from left to right) on Adult.]{
\includegraphics[width=0.245\columnwidth]{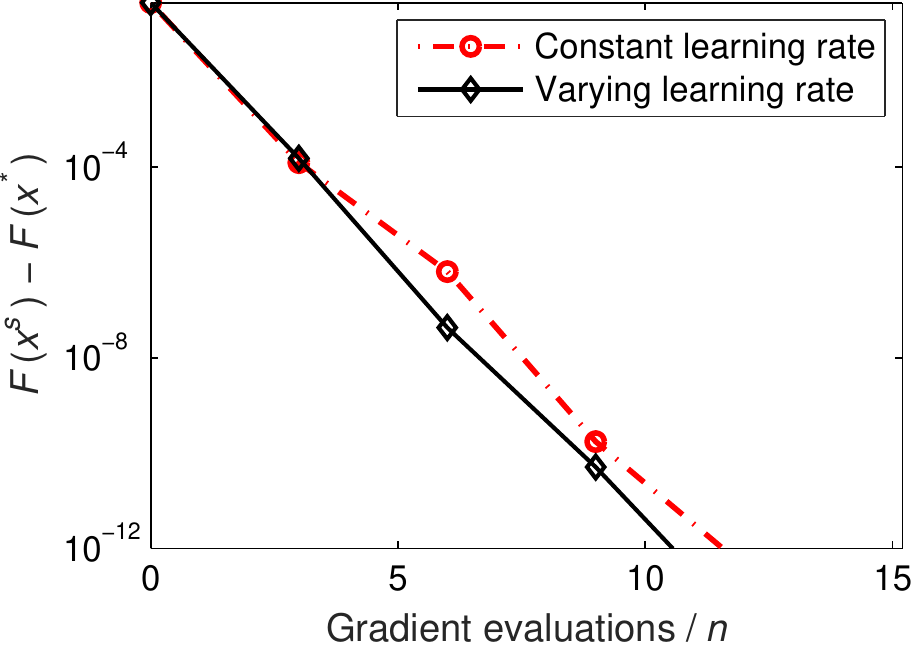}\:\includegraphics[width=0.245\columnwidth]{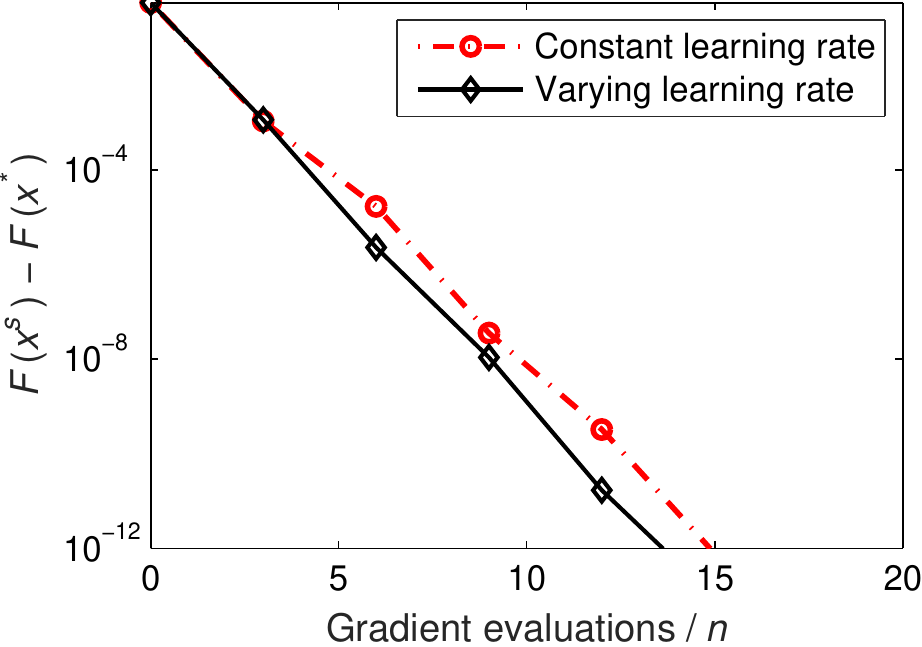}\:\includegraphics[width=0.245\columnwidth]{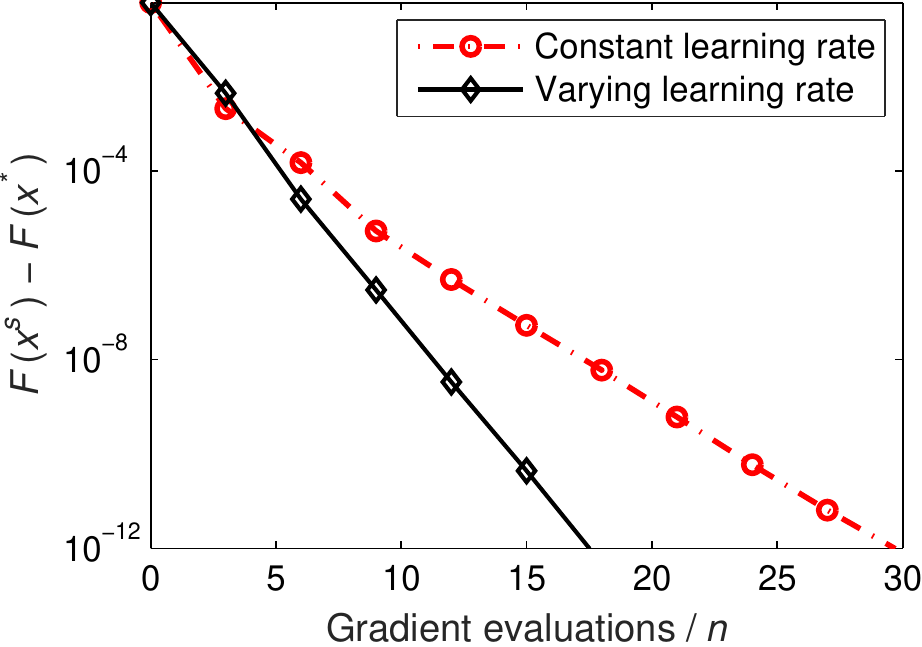}\:\includegraphics[width=0.245\columnwidth]{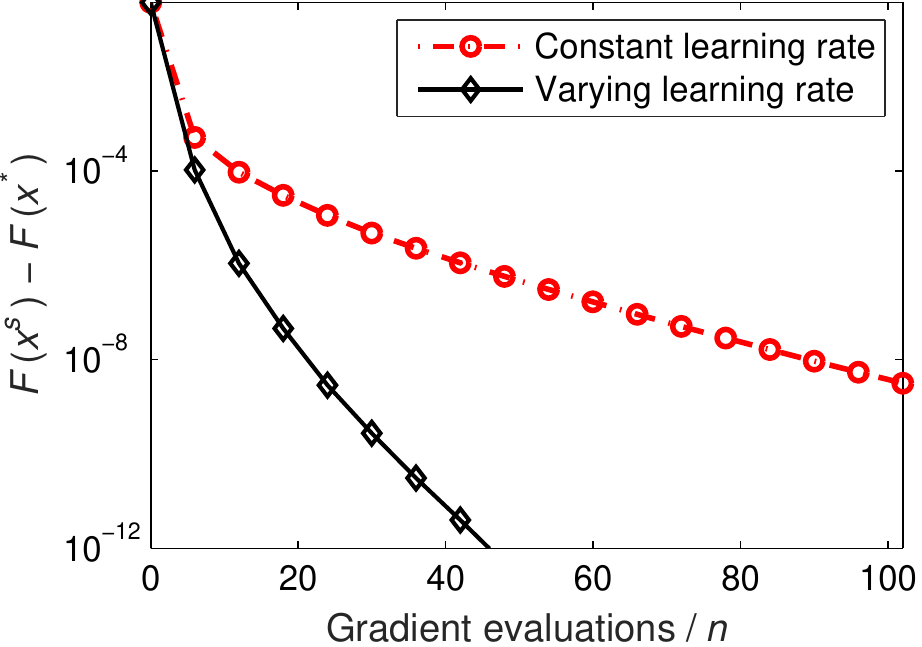}}
\subfigure[Results of $\ell_{2}$-norm regularized logistic regression with $\lambda=10^{-4},10^{-5},10^{-6},10^{-7}$ (from left to right) on Covtype.]{
\includegraphics[width=0.245\columnwidth]{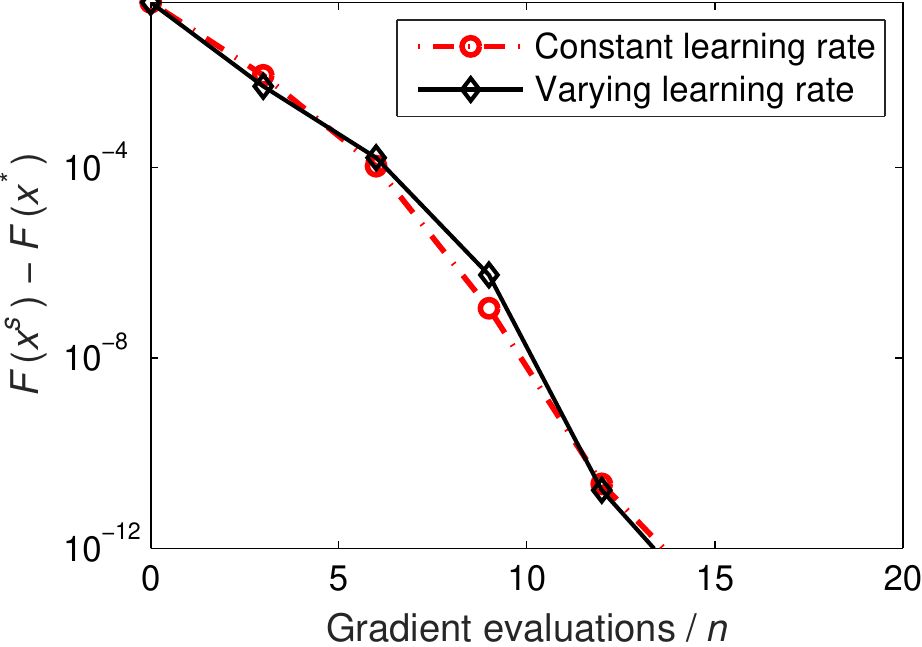}\:\includegraphics[width=0.245\columnwidth]{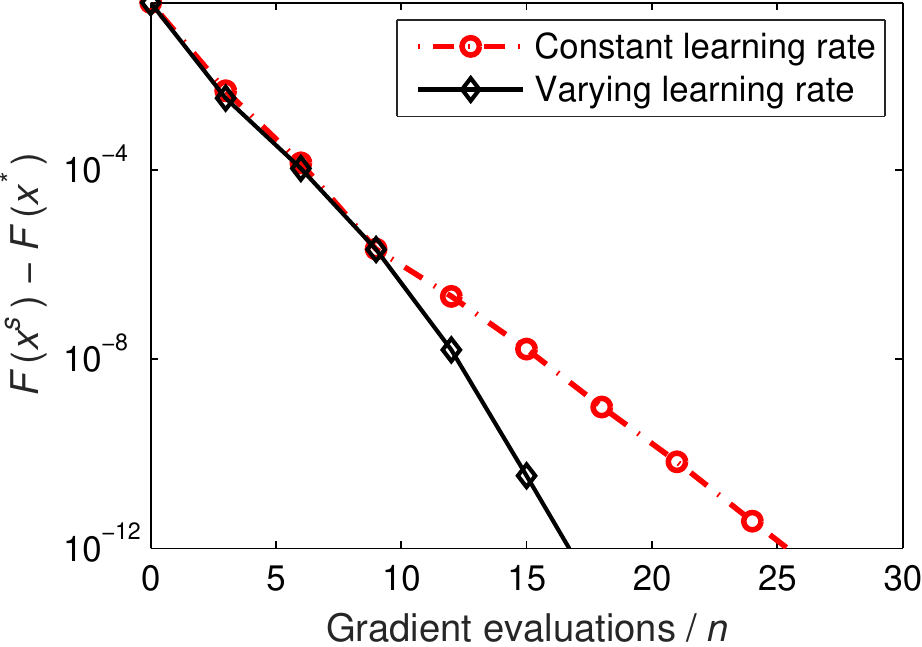}\:\includegraphics[width=0.245\columnwidth]{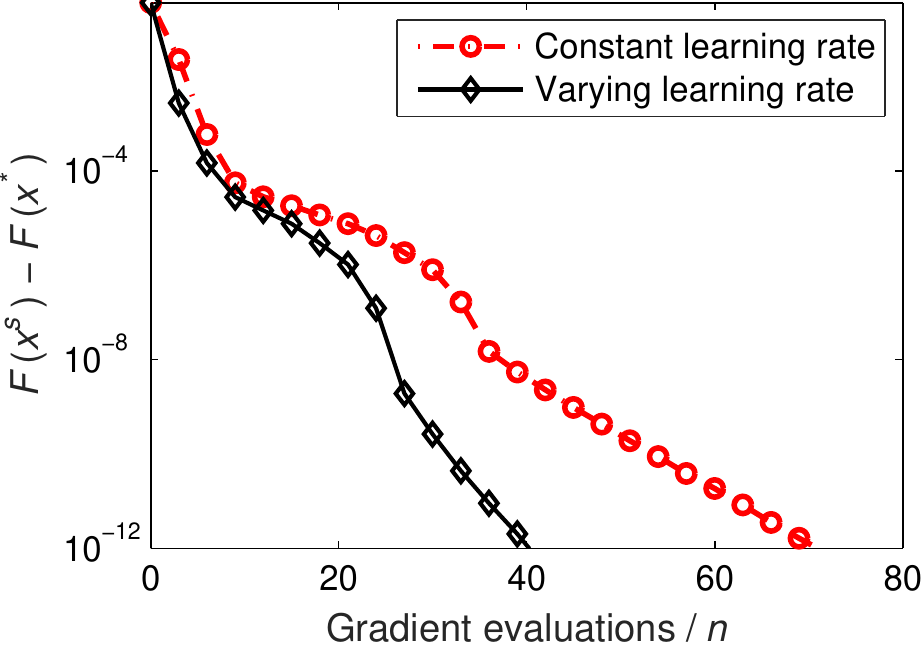}\:\includegraphics[width=0.245\columnwidth]{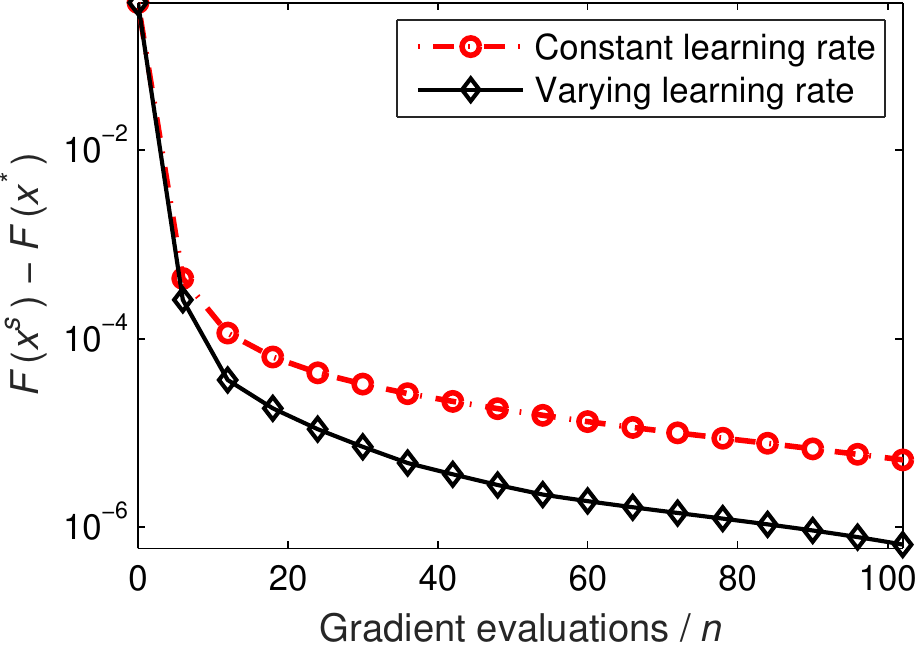}}
\vspace{0.6mm}

\subfigure[Results of $\ell_{1}$-norm regularized logistic regression with $\lambda=10^{-3},10^{-4},10^{-5},10^{-6}$ (from left to right) on Adult.]{
\includegraphics[width=0.245\columnwidth]{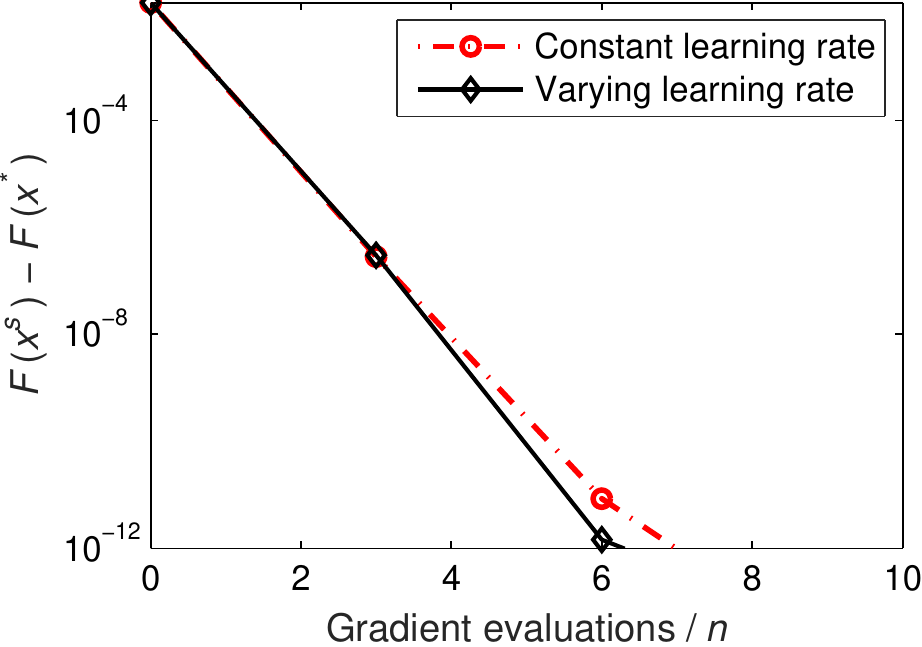}\:\includegraphics[width=0.245\columnwidth]{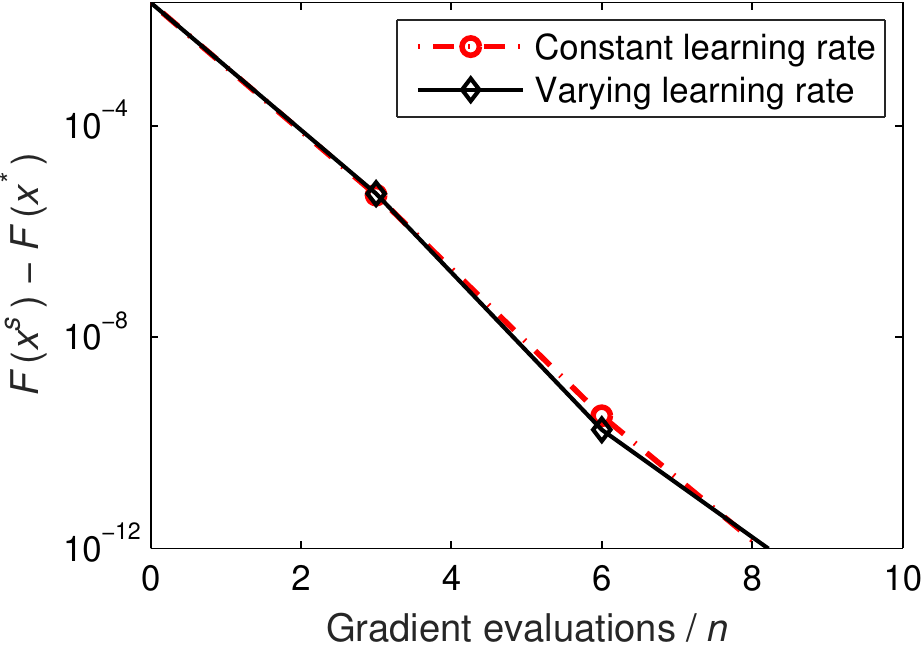}\:\includegraphics[width=0.245\columnwidth]{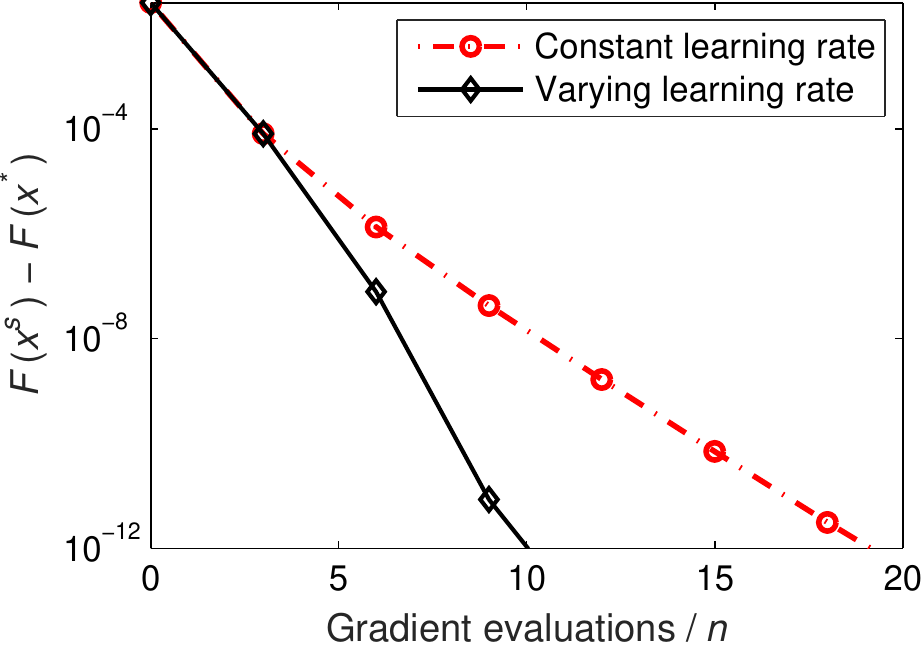}\:\includegraphics[width=0.245\columnwidth]{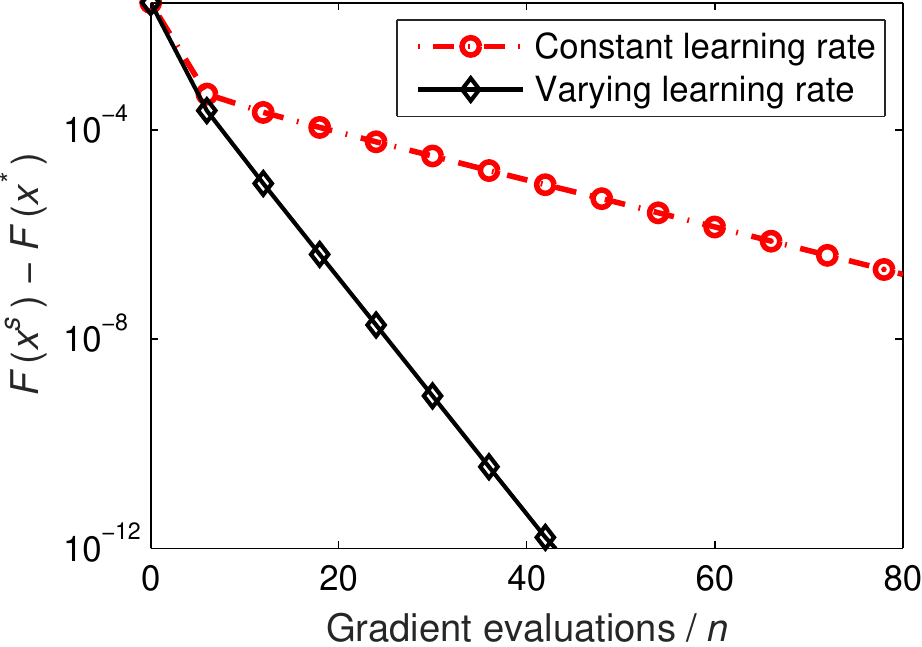}}
\subfigure[Results of $\ell_{1}$-norm regularized logistic regression with $\lambda=10^{-3},10^{-4},10^{-5},10^{-6}$ (from left to right) on Covtype.]{
\includegraphics[width=0.245\columnwidth]{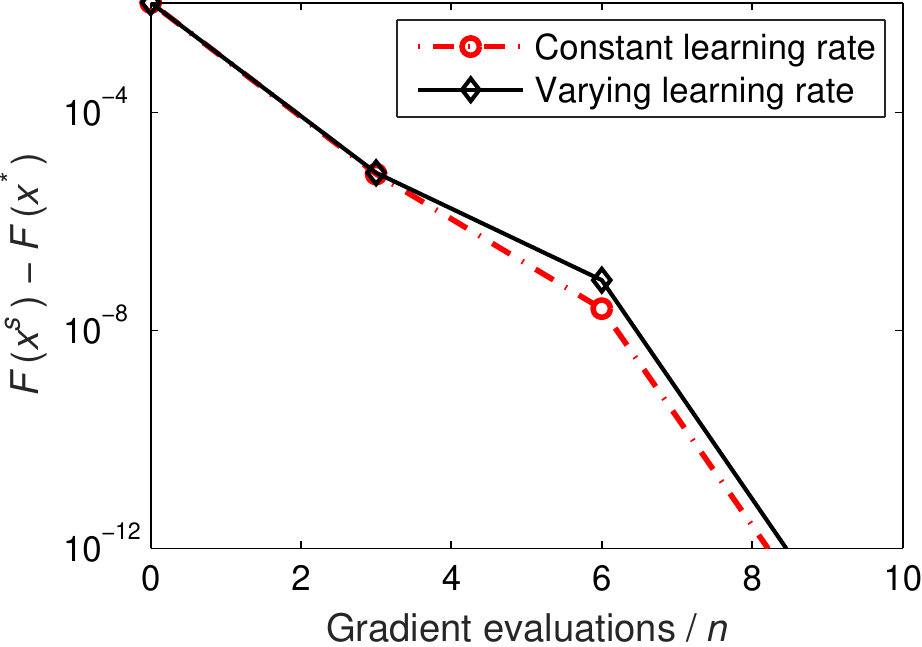}\:\includegraphics[width=0.245\columnwidth]{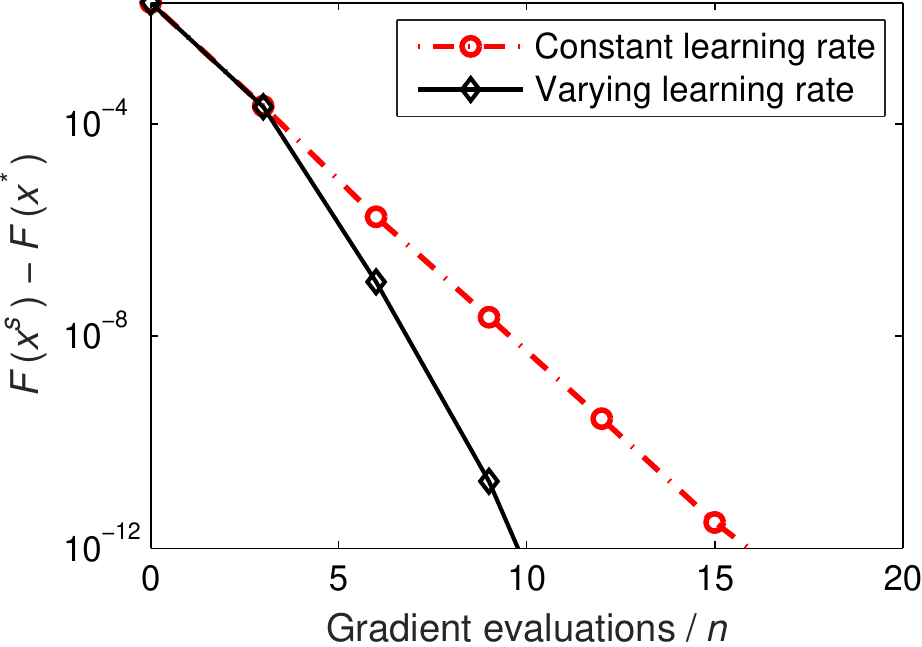}\:\includegraphics[width=0.245\columnwidth]{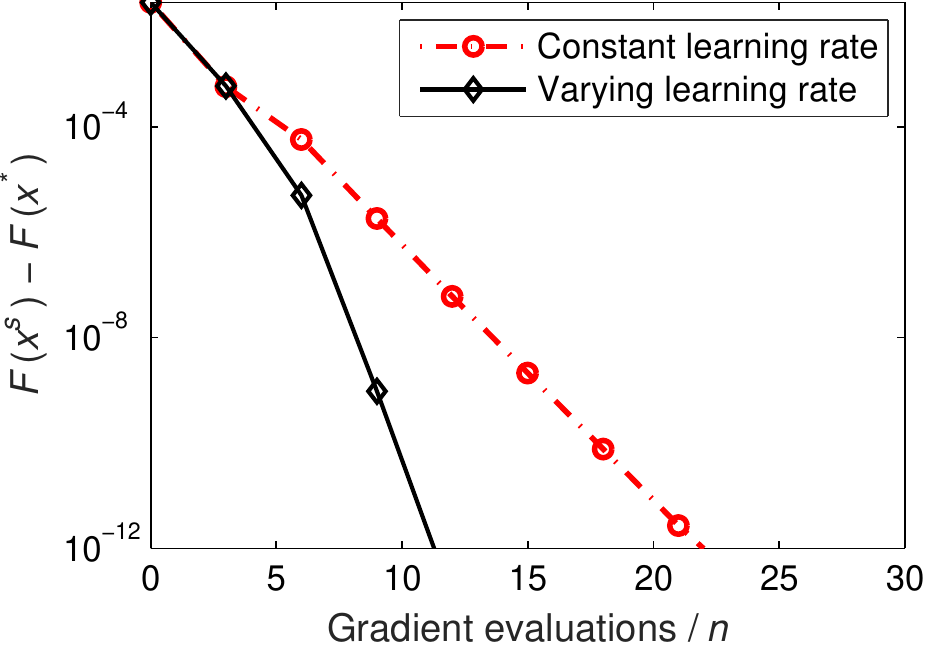}\:\includegraphics[width=0.245\columnwidth]{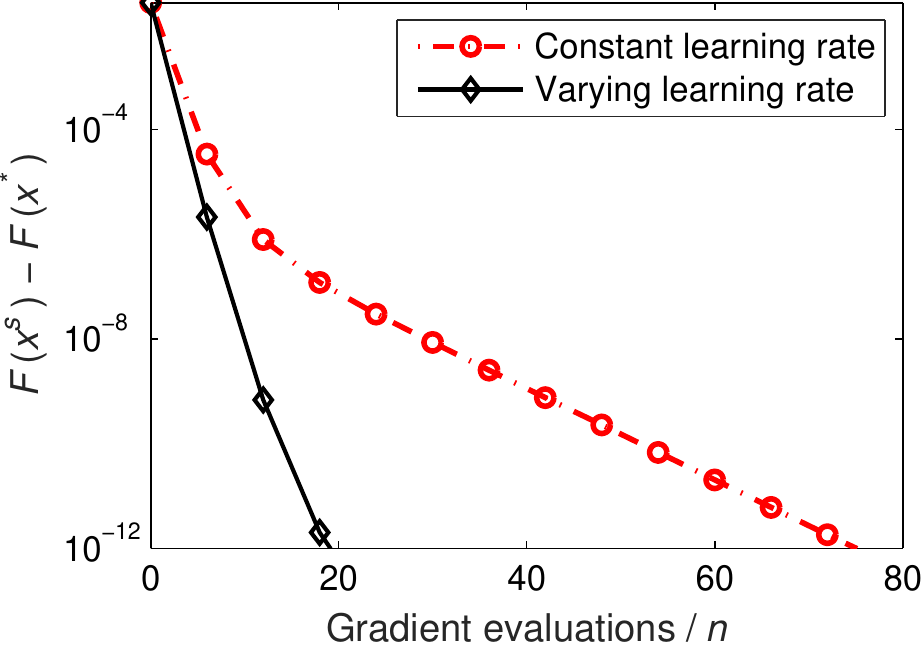}}
\caption{Comparison of our VR-SGD method (i.e., Algorithm 2) with fixed and varying learning rates for solving $\ell_{2}$-norm (i.e., $(\lambda/2)\|\cdot\|^{2}$) and $\ell_{1}$-norm (i.e., $\lambda\|\cdot\|_{1}$) regularized logistic regression problems.}
\label{figs22}
\end{figure}

\subsection*{Impact of Increasing Learning Rates}
\label{sec55}
We compared the performance of Algorithm 2 with constant and varying learning rates for solving $\ell_{2}$-norm and $\ell_{1}$-norm regularized logistic regression problems, as shown in Fig.\ \ref{figs22}. Note that the learning rate in Algorithm 2 is varied according to the update formula in (13) (i.e., $\eta_{s}\!=\!\eta_{0}/\max\{\alpha,\,2/(s+1)\}$), and the initial learning rate $\eta_{0}$ is set to the same value for the two cases. We can observe that Algorithm 2 with varying learning rates converges much faster than Algorithm 2 with fixed learning rates in most cases, especially when the regularization parameter is relatively small, e.g., $\lambda=10^{-6}$ or $10^{-7}$. This empirically verifies the importance of increasing the learning rate.

\subsection*{More Results of Stochastic and Accelerated Deterministic Methods}
In this part, we compared the well-known accelerated deterministic methods including AGD~\cite{nesterov:co} and APG~\cite{beck:fista} with some stochastic optimization methods, including SGD, one of stochastic variance reduced methods, SVRG~\cite{johnson:svrg}, one of accelerated variance reduction methods, Katyusha~\cite{zhu:Katyusha}, and VR-SGD. Note that AGD and APG are used to solve smooth and non-smooth objective functions, respectively. Fig.\ \ref{figs11} shows the experimental results of SGD, AGD, SVRG, Katyusha, and VR-SGD for solving $\ell_{2}$-norm regularized logistic regression problems with different regularization parameters. Furthermore, we also reported their experimental results for solving $\ell_{1}$-norm regularized logistic regression problems in Fig.\ \ref{figs12}. All the results show that the stochastic variance reduction methods, including SVRG, Katyusha, and VR-SGD, significantly outperform the accelerated deterministic methods, as well as SGD, for both strongly-convex and non-strongly convex cases, which empirically verifies the importance of variance reduction techniques.

\begin{figure}[th]
\centering
\includegraphics[width=0.245\columnwidth]{Fig701}
\includegraphics[width=0.245\columnwidth]{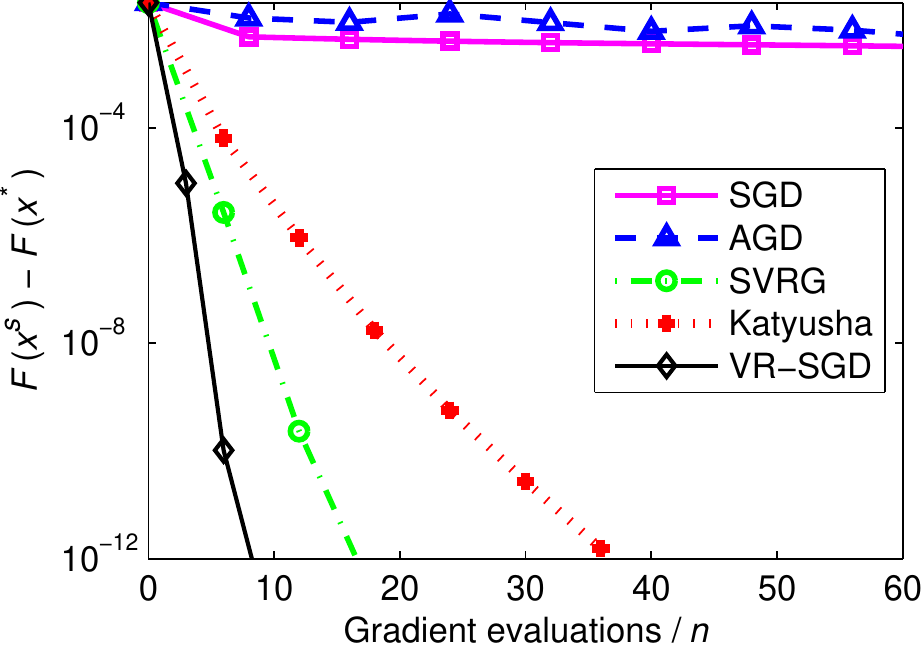}
\includegraphics[width=0.245\columnwidth]{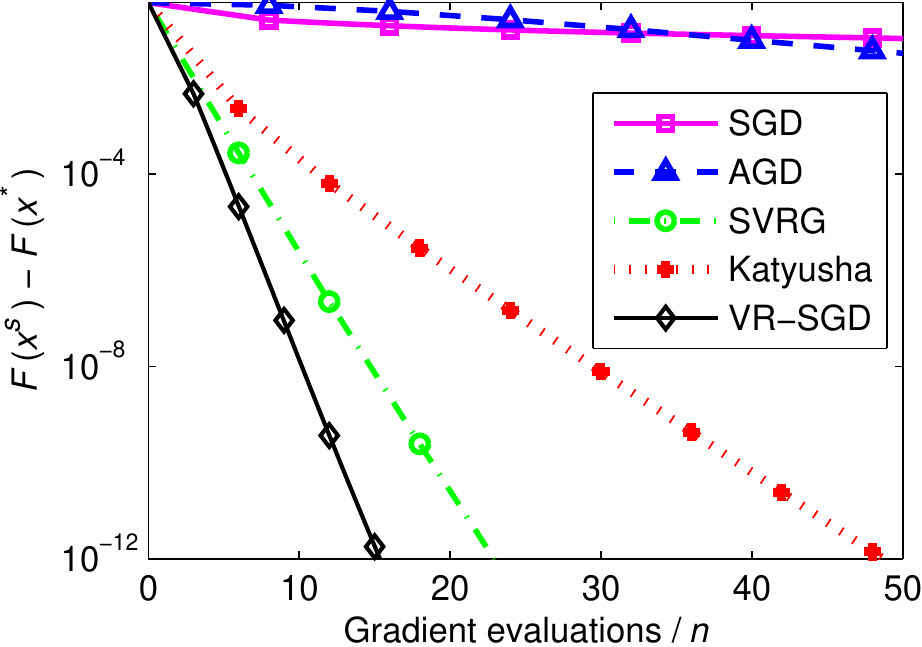}
\includegraphics[width=0.245\columnwidth]{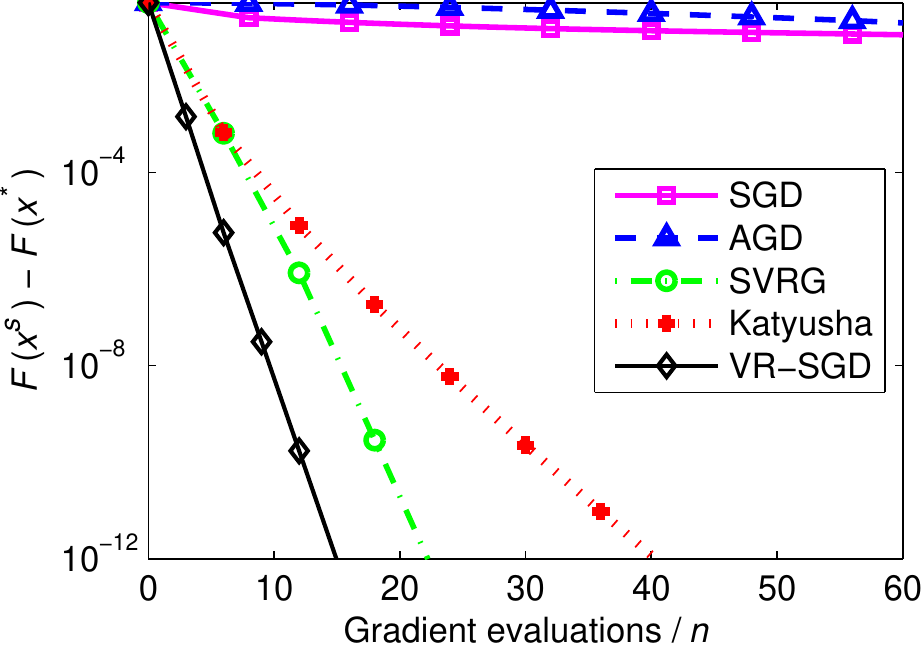}

\subfigure[Adult: $\lambda=10^{-5}$]{\includegraphics[width=0.245\columnwidth]{Fig702}}
\subfigure[Covtype: $\lambda=10^{-5}$]{\includegraphics[width=0.245\columnwidth]{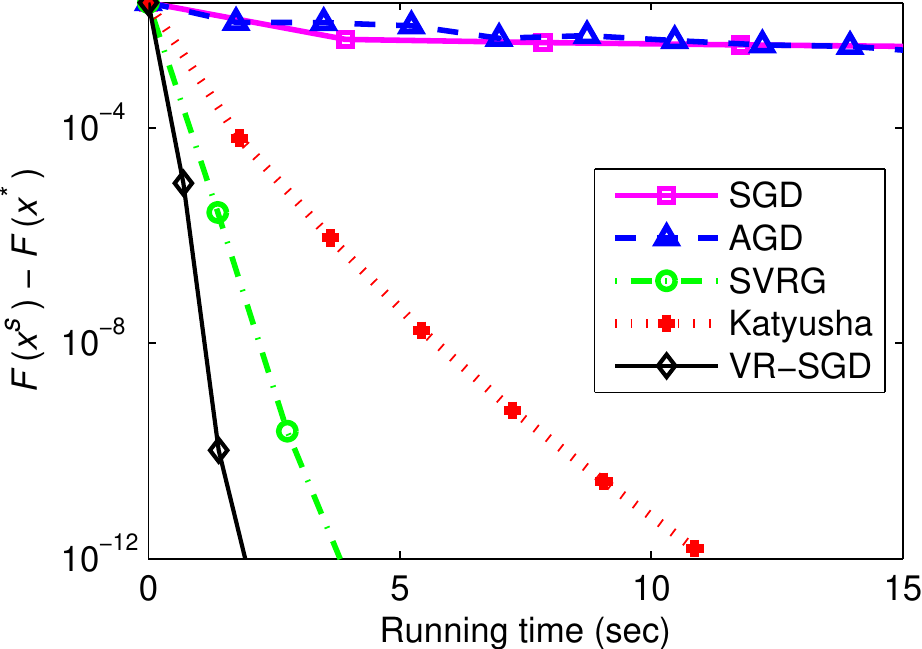}}
\subfigure[RCV1: $\lambda=10^{-4}$]{\includegraphics[width=0.245\columnwidth]{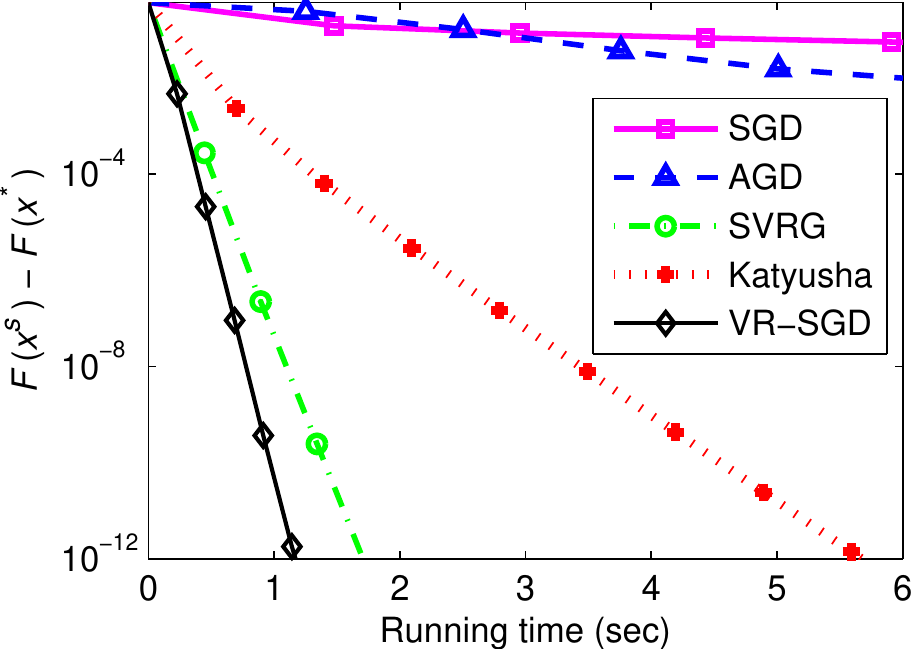}}
\subfigure[Epsilon: $\lambda=10^{-5}$]{\includegraphics[width=0.245\columnwidth]{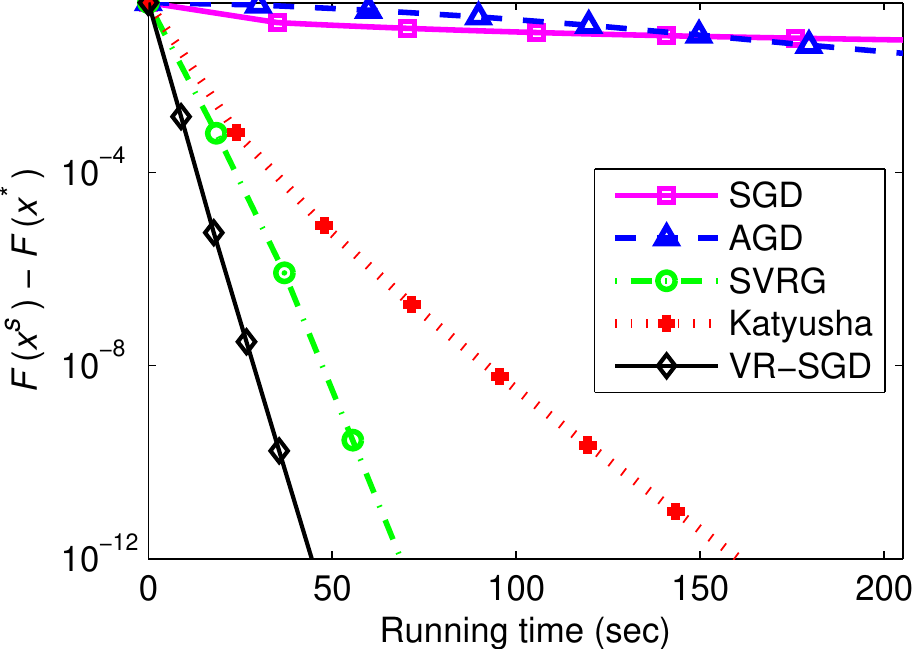}}
\vspace{1.6mm}

\includegraphics[width=0.245\columnwidth]{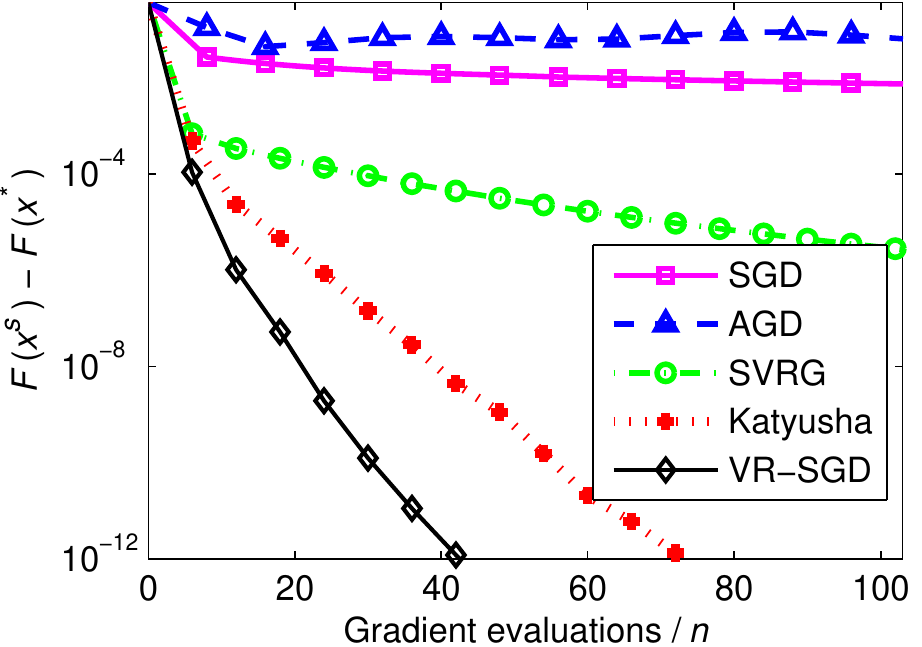}
\includegraphics[width=0.245\columnwidth]{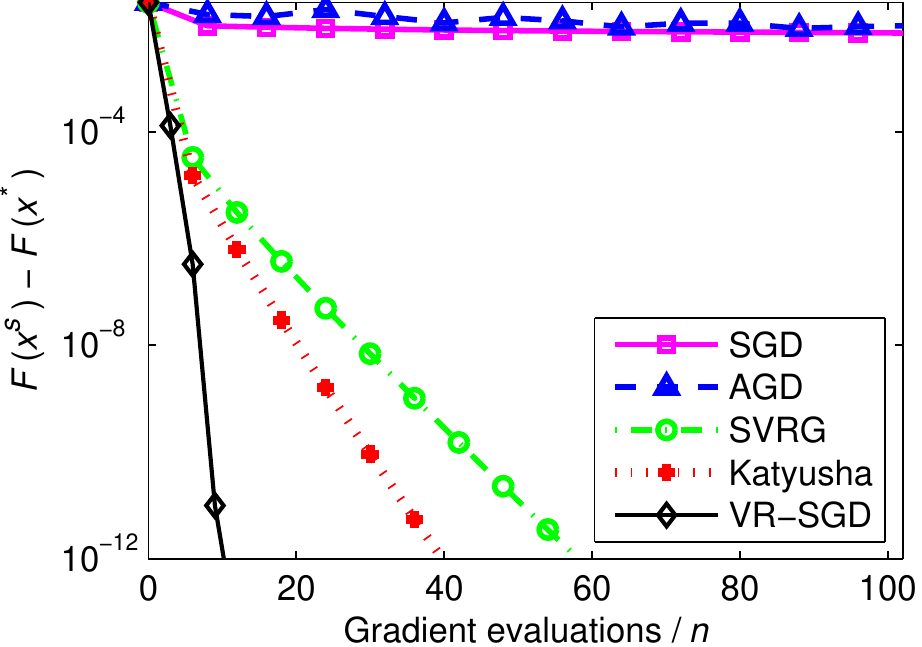}
\includegraphics[width=0.245\columnwidth]{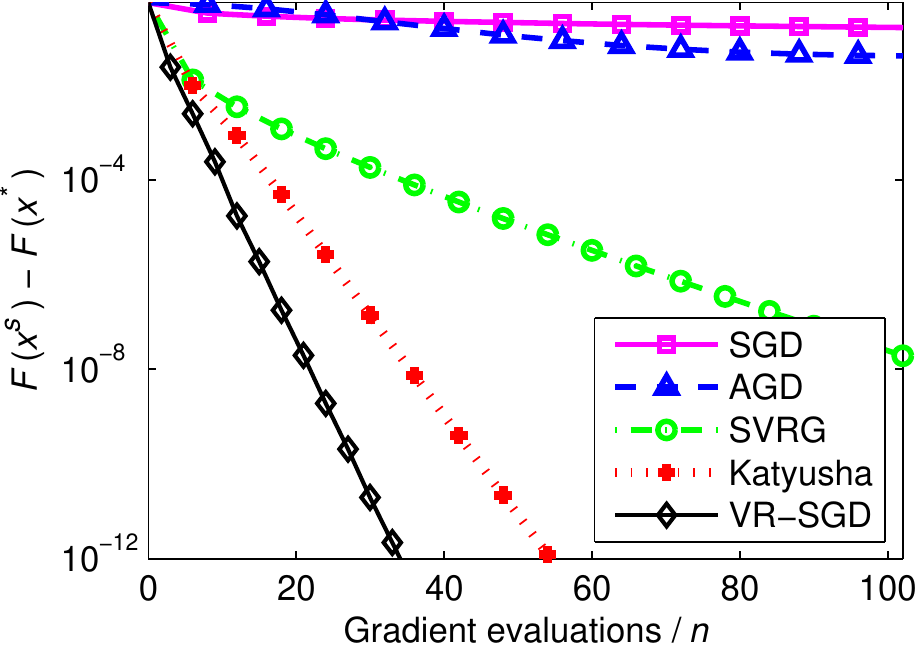}
\includegraphics[width=0.245\columnwidth]{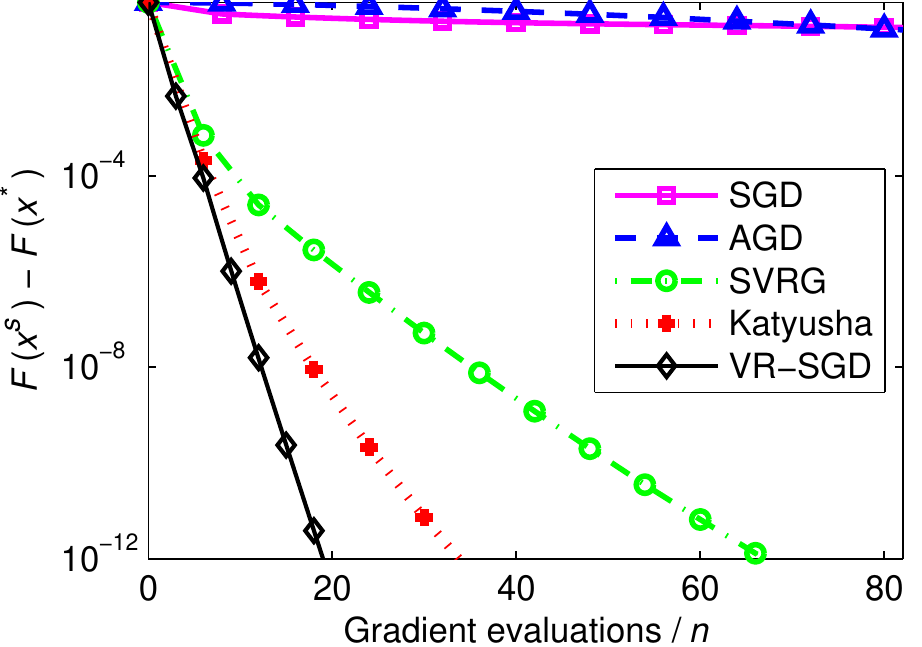}

\subfigure[Adult: $\lambda=10^{-6}$]{\includegraphics[width=0.245\columnwidth]{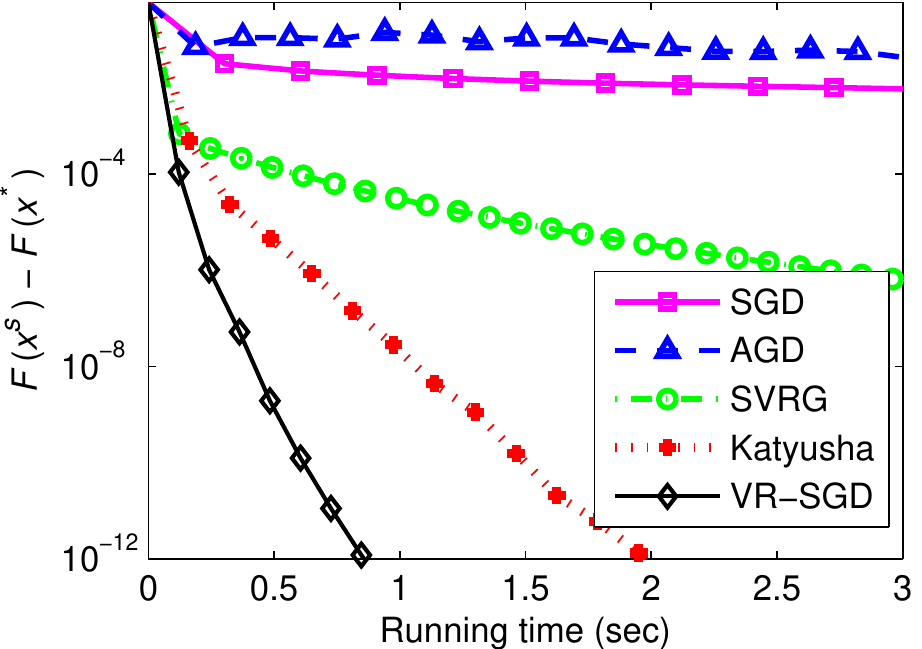}}
\subfigure[Covtype: $\lambda=10^{-6}$]{\includegraphics[width=0.245\columnwidth]{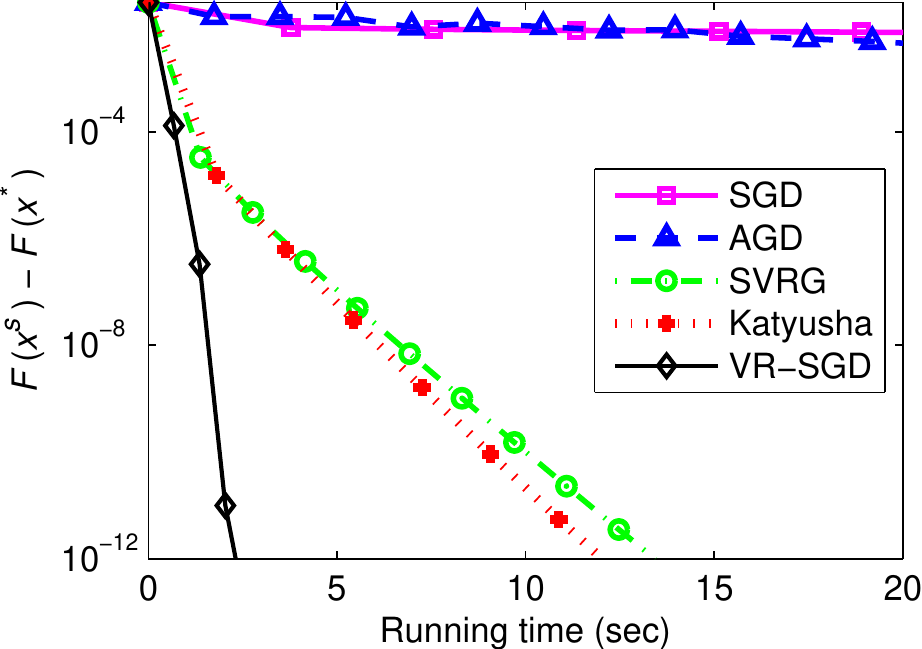}}
\subfigure[RCV1: $\lambda=10^{-5}$]{\includegraphics[width=0.245\columnwidth]{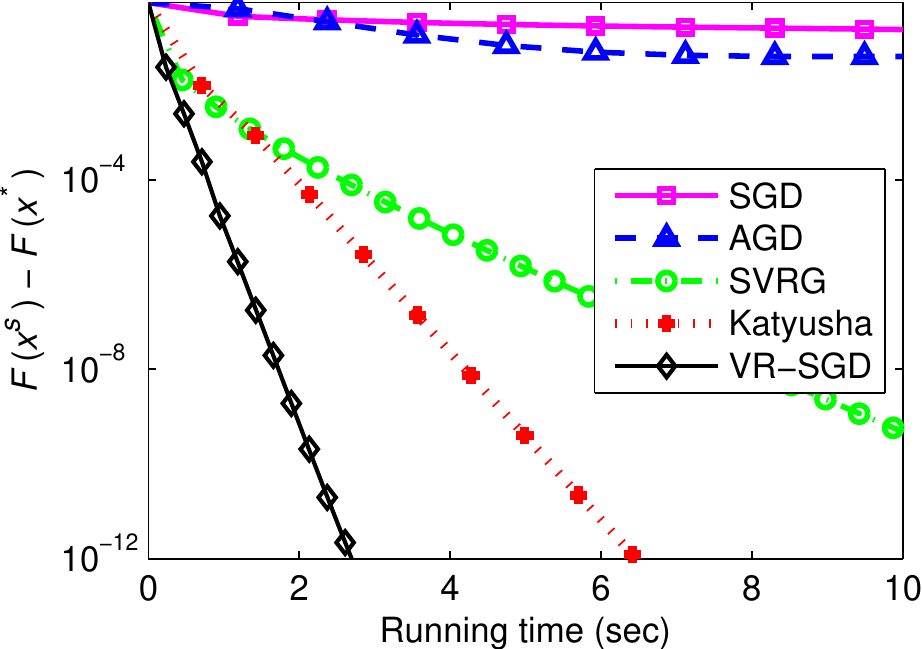}}
\subfigure[Epsilon: $\lambda=10^{-6}$]{\includegraphics[width=0.245\columnwidth]{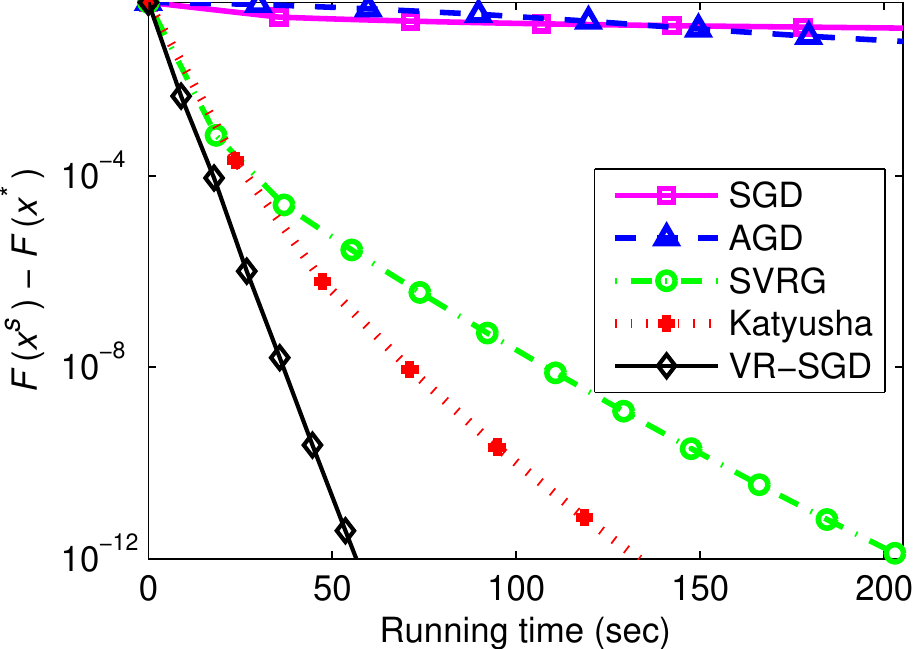}}
\caption{Comparison of SGD, AGD~\cite{nesterov:co}, SVRG~\cite{johnson:svrg}, Katyusha~\cite{zhu:Katyusha}, and our VR-SGD method for solving $\ell_{2}$-norm regularized logistic regression problems. In each plot, the vertical axis shows the objective value minus the minimum, and the horizontal axis is the number of effective passes (top) or running time (bottom).}
\label{figs11}
\end{figure}

\begin{figure}[!th]
\centering
\includegraphics[width=0.245\columnwidth]{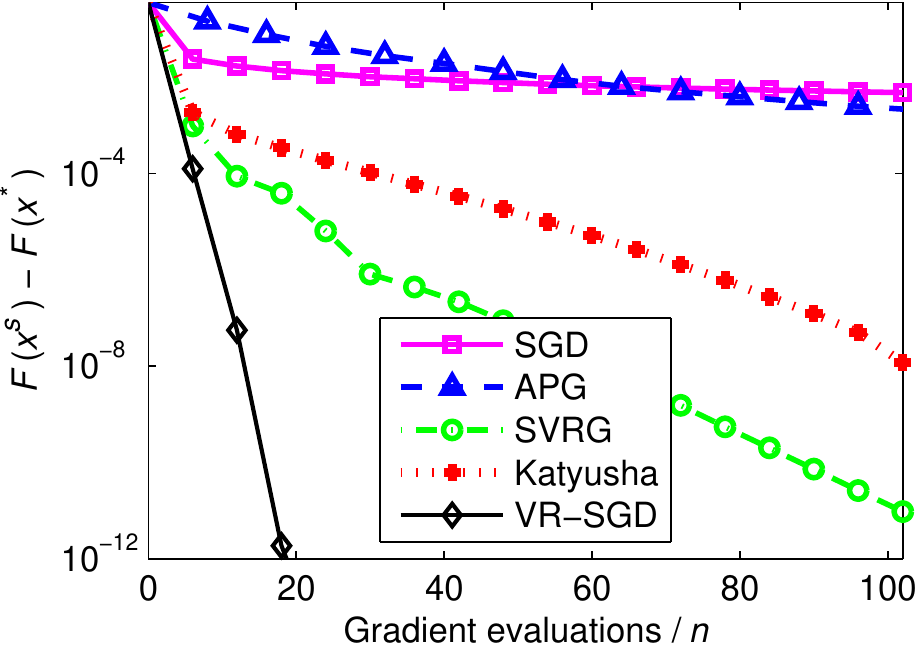}
\includegraphics[width=0.245\columnwidth]{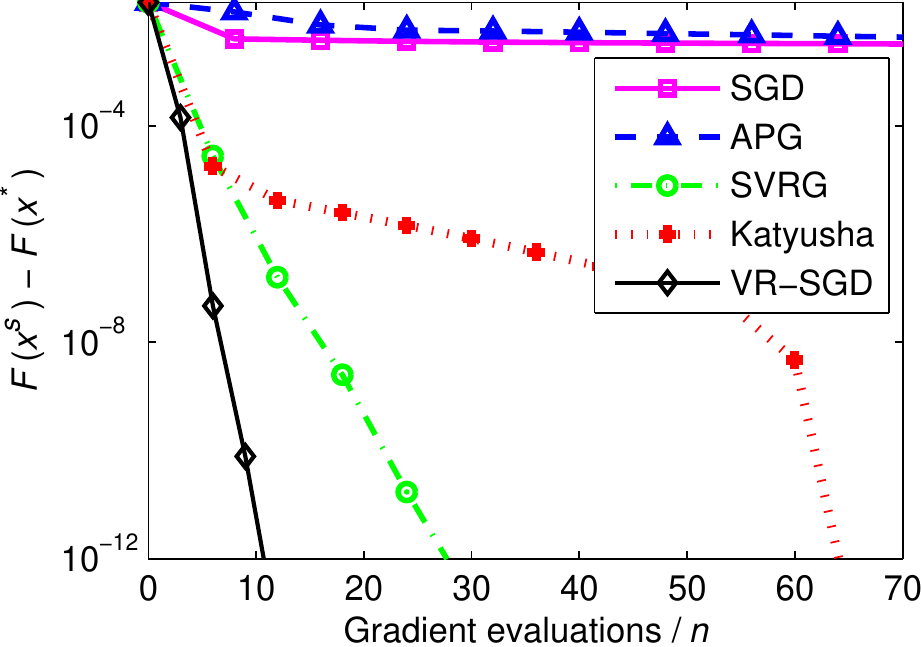}
\includegraphics[width=0.245\columnwidth]{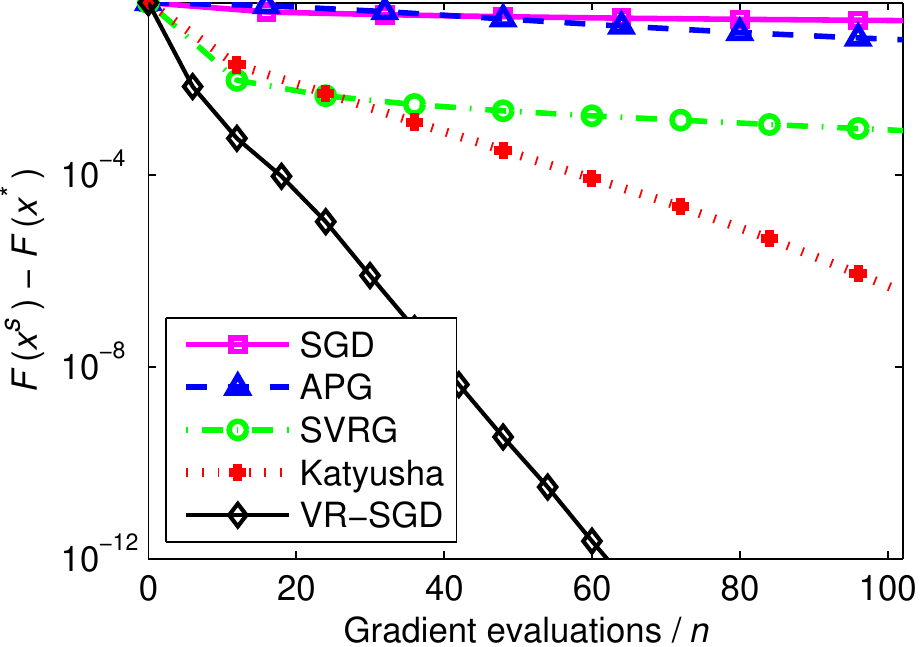}
\includegraphics[width=0.245\columnwidth]{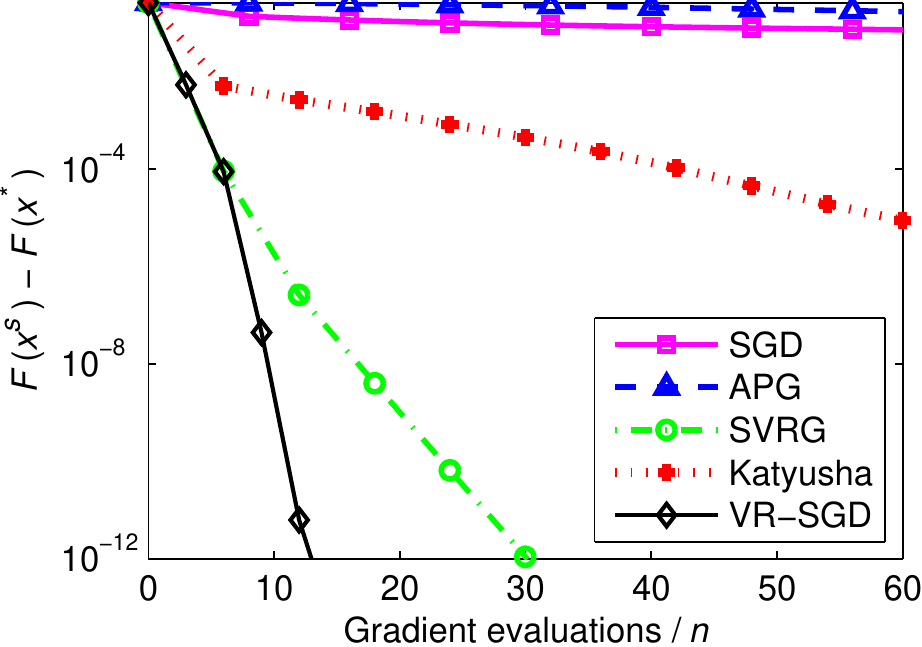}

\subfigure[Adult: $\lambda=10^{-4}$]{\includegraphics[width=0.245\columnwidth]{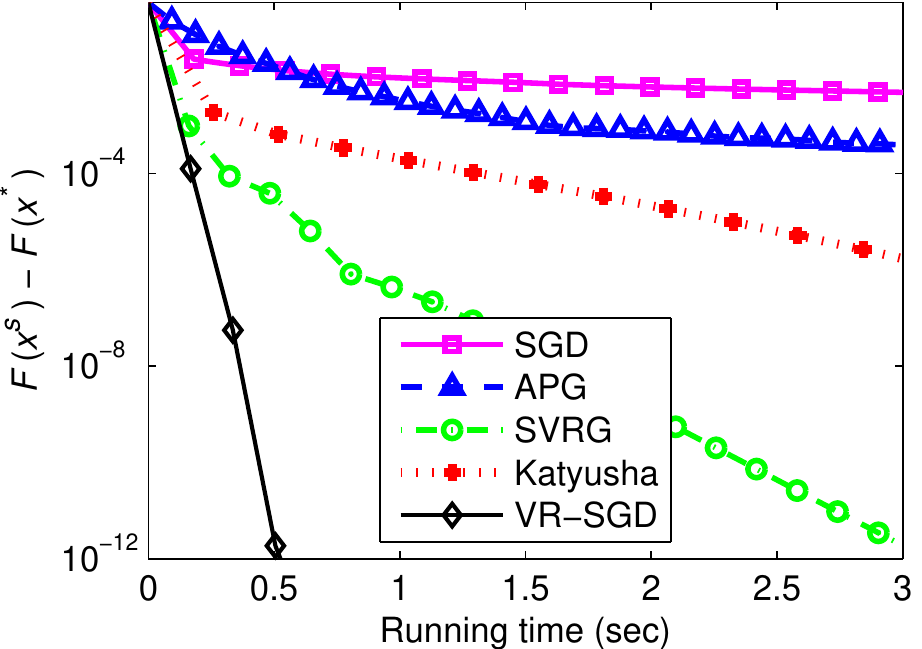}}
\subfigure[Covtype: $\lambda=10^{-4}$]{\includegraphics[width=0.245\columnwidth]{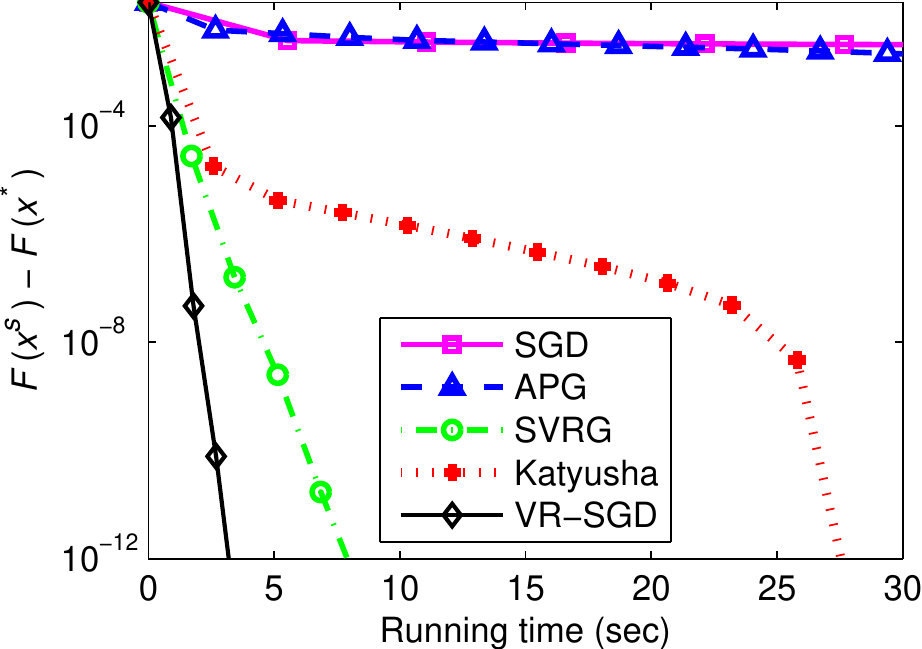}}
\subfigure[RCV1: $\lambda=10^{-4}$]{\includegraphics[width=0.245\columnwidth]{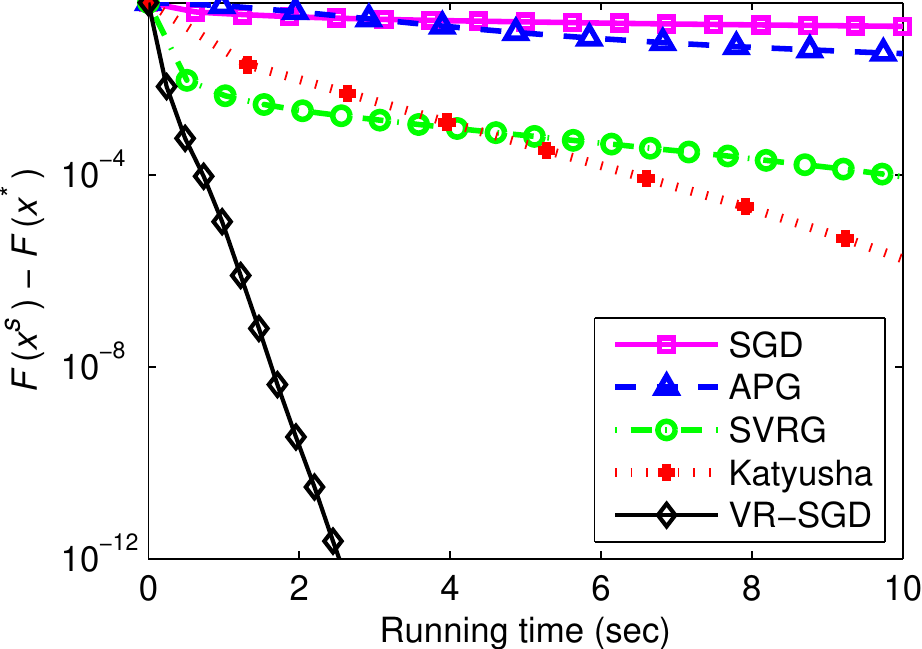}}
\subfigure[Epsilon: $\lambda=10^{-4}$]{\includegraphics[width=0.245\columnwidth]{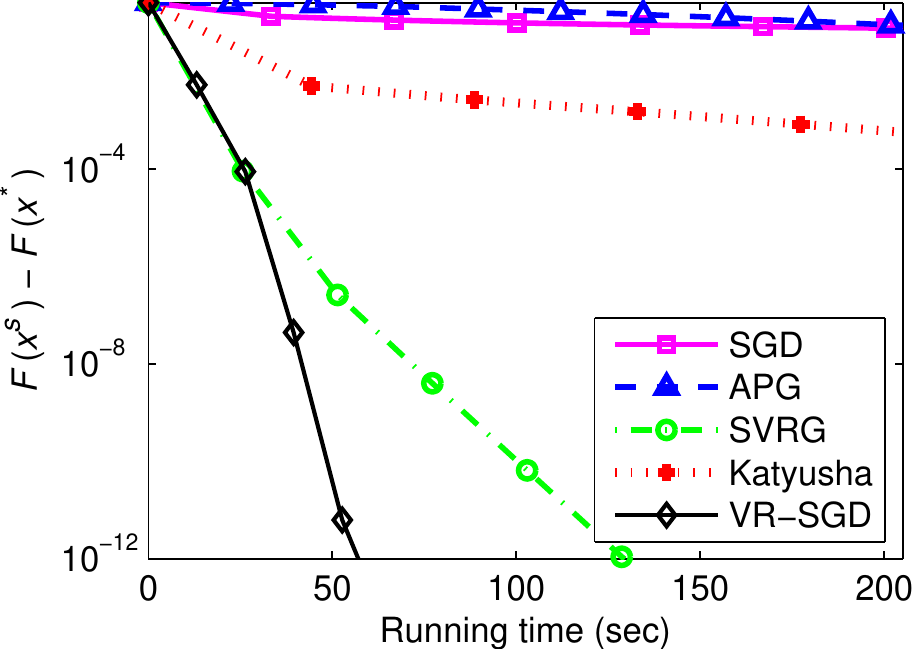}}
\vspace{1.6mm}

\includegraphics[width=0.245\columnwidth]{Fig723}
\includegraphics[width=0.245\columnwidth]{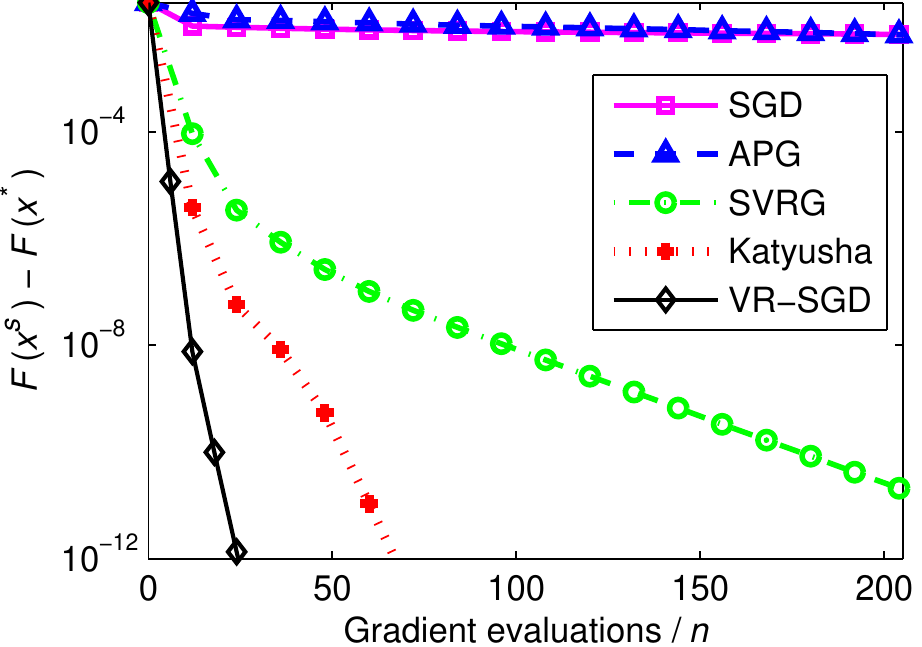}
\includegraphics[width=0.245\columnwidth]{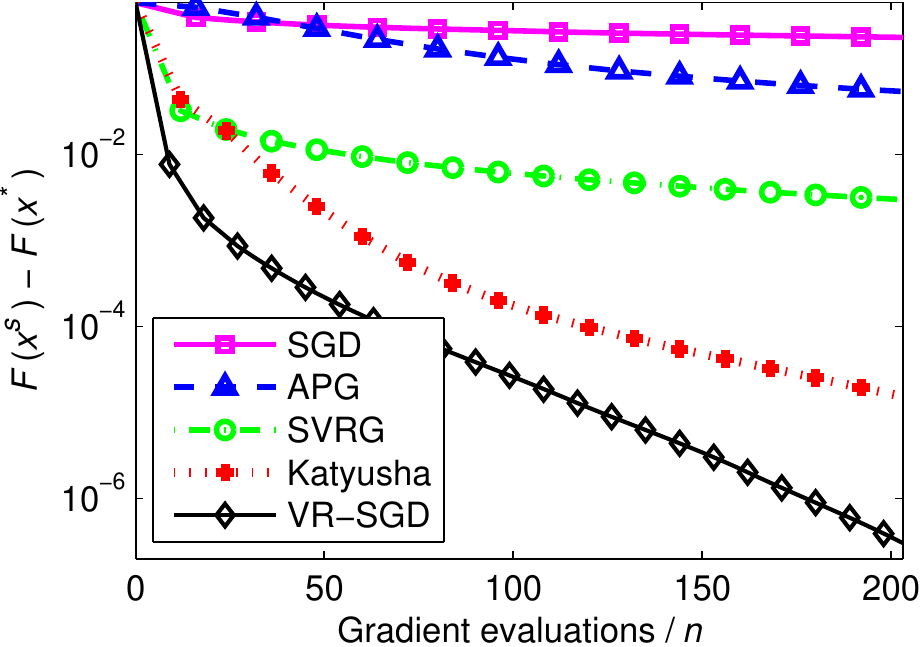}
\includegraphics[width=0.245\columnwidth]{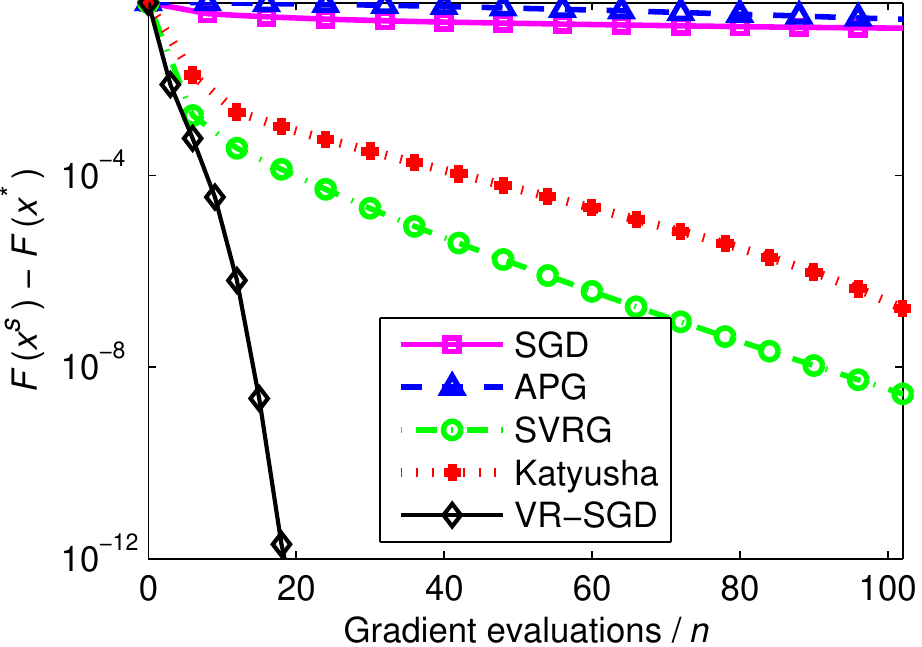}

\subfigure[Adult: $\lambda=10^{-5}$]{\includegraphics[width=0.245\columnwidth]{Fig724}}
\subfigure[Covtype: $\lambda=10^{-5}$]{\includegraphics[width=0.245\columnwidth]{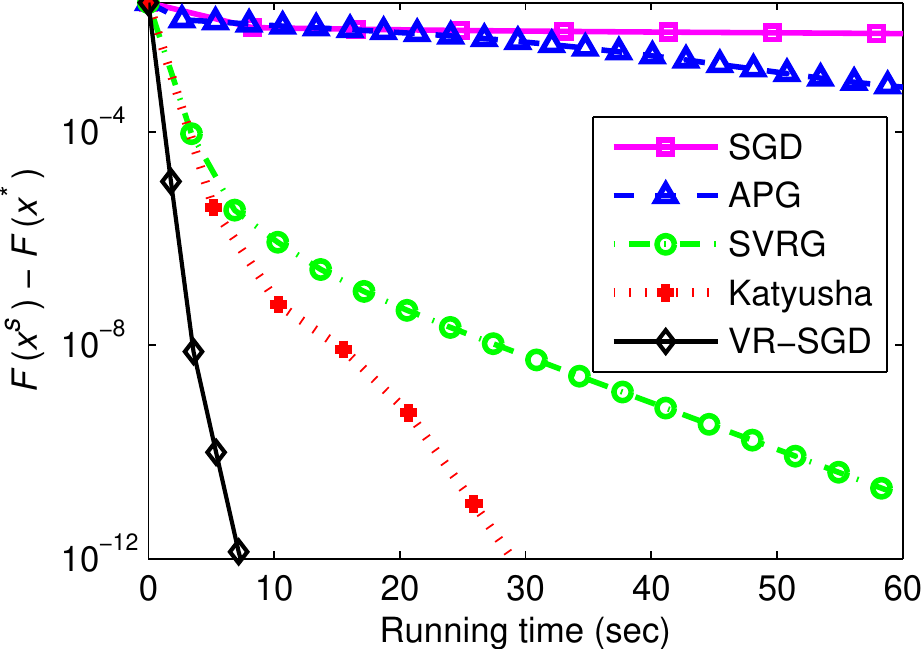}}
\subfigure[RCV1: $\lambda=10^{-5}$]{\includegraphics[width=0.245\columnwidth]{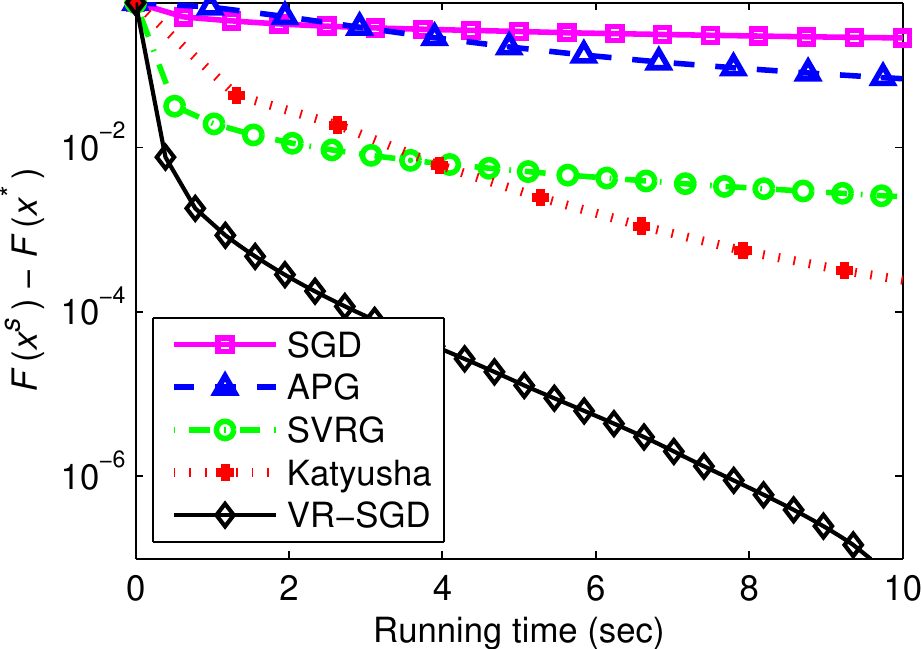}}
\subfigure[Epsilon: $\lambda=10^{-5}$]{\includegraphics[width=0.245\columnwidth]{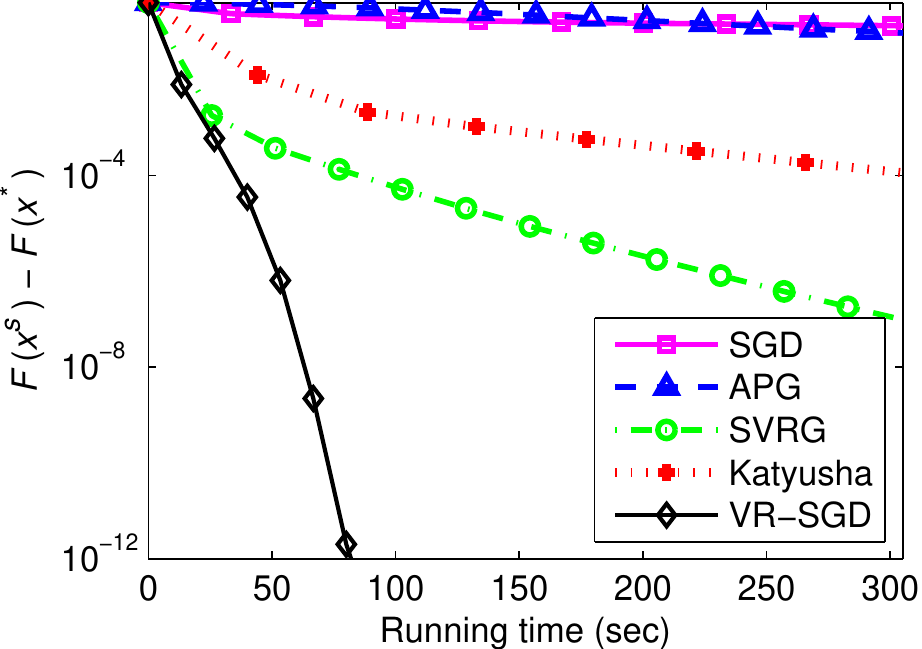}}
\caption{Comparison of SGD, APG~\cite{beck:fista}, SVRG, Katyusha~\cite{zhu:Katyusha}, and our VR-SGD method for solving $\ell_{1}$-norm regularized logistic regression problems. In each plot, the vertical axis shows the objective value minus the minimum, and the horizontal axis is the number of effective passes (top) or running time (bottom).}
\label{figs12}
\end{figure}

\begin{figure}[t]
\centering
\subfigure[Adult: $\lambda=10^{-4}$ (left) \;and\; $\lambda=10^{-5}$ (right)]{\includegraphics[width=0.246\columnwidth]{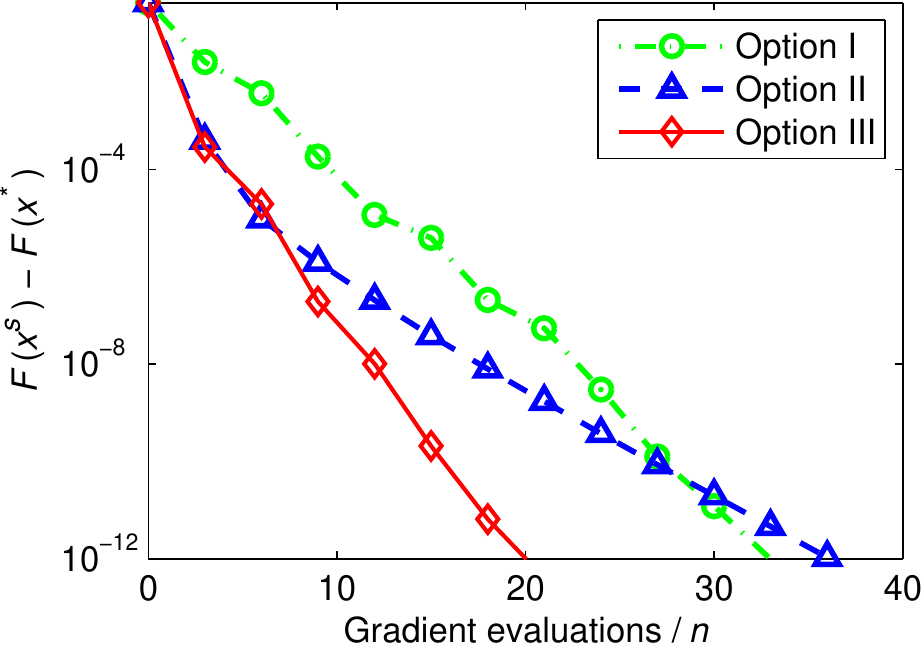}\:\includegraphics[width=0.246\columnwidth]{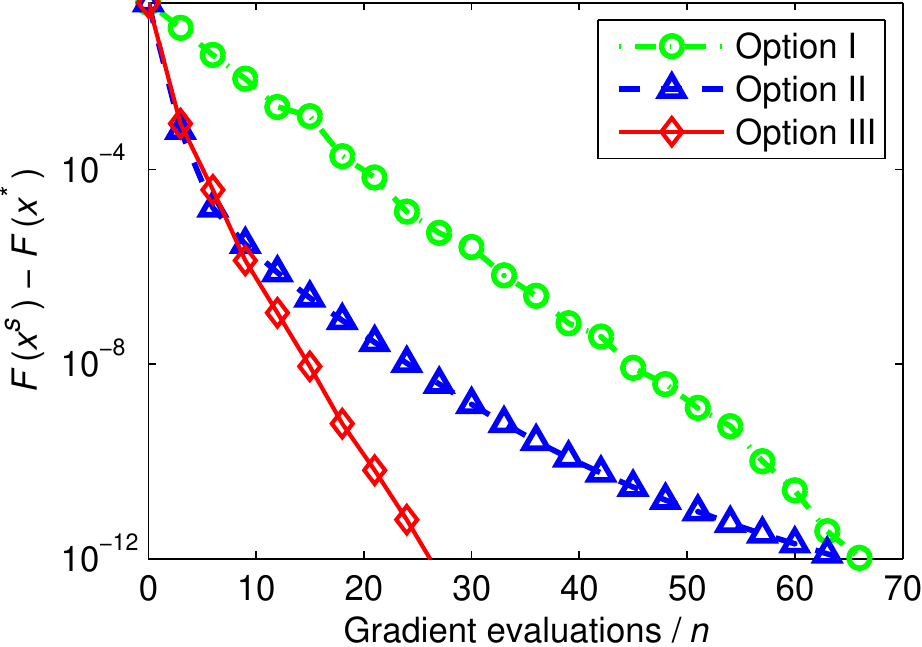}}
\subfigure[MNIST: $\lambda=10^{-5}$ (left) \;and\; $\lambda=10^{-6}$ (right)]{\includegraphics[width=0.246\columnwidth]{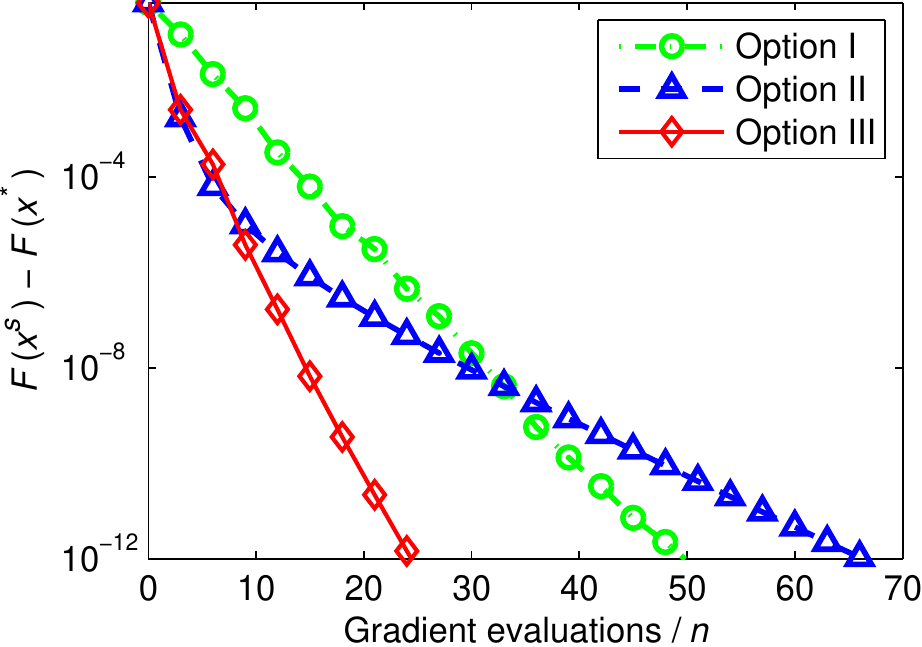}\:\includegraphics[width=0.246\columnwidth]{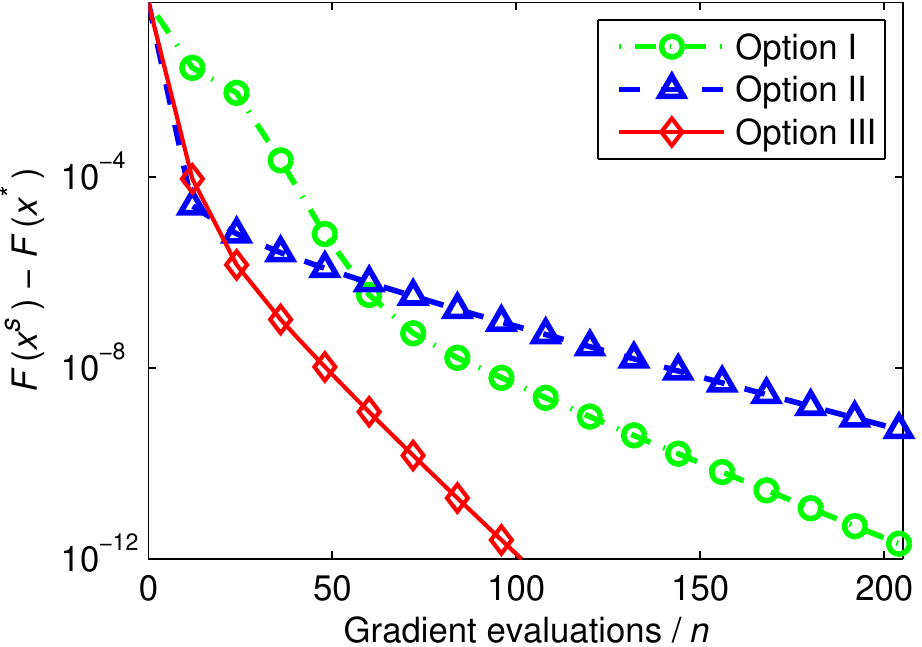}}

\subfigure[RCV1: $\lambda=10^{-4}$ (left) \;and\; $\lambda=10^{-5}$ (right)]{\includegraphics[width=0.246\columnwidth]{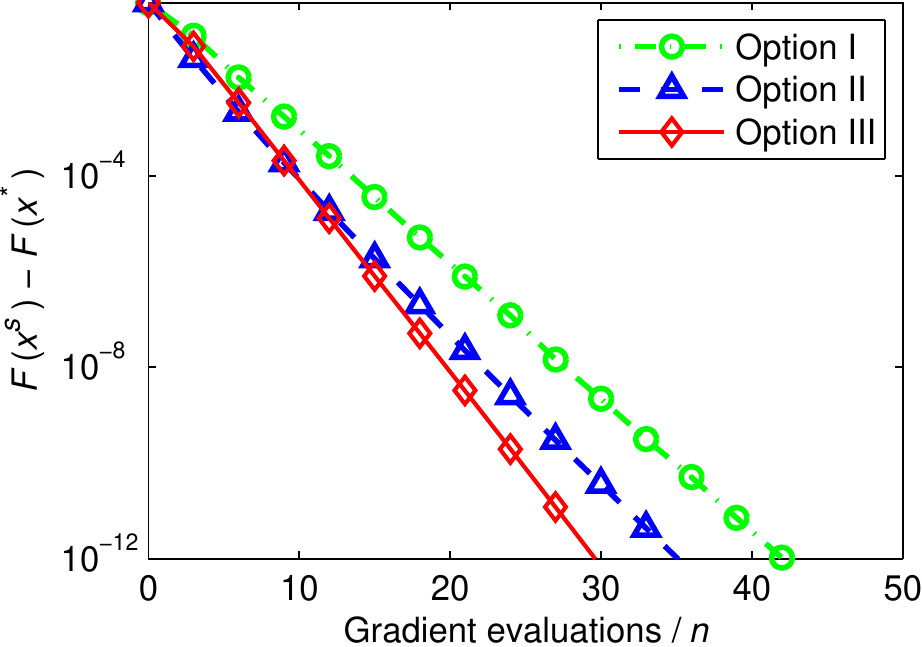}\:\includegraphics[width=0.246\columnwidth]{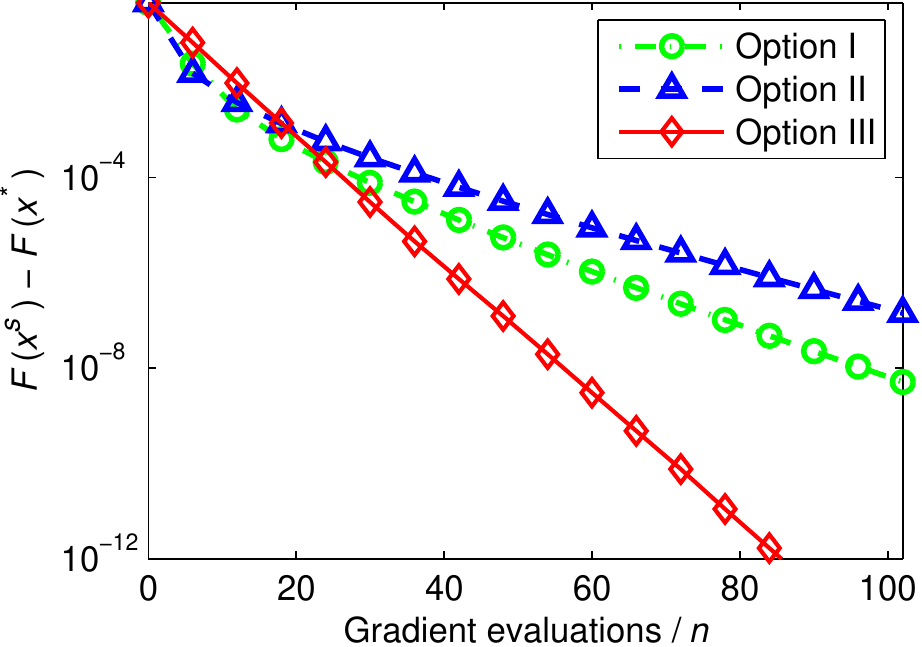}}
\subfigure[Epsilon: $\lambda=10^{-5}$ (left) \;and\; $\lambda=10^{-6}$ (right)]{\includegraphics[width=0.246\columnwidth]{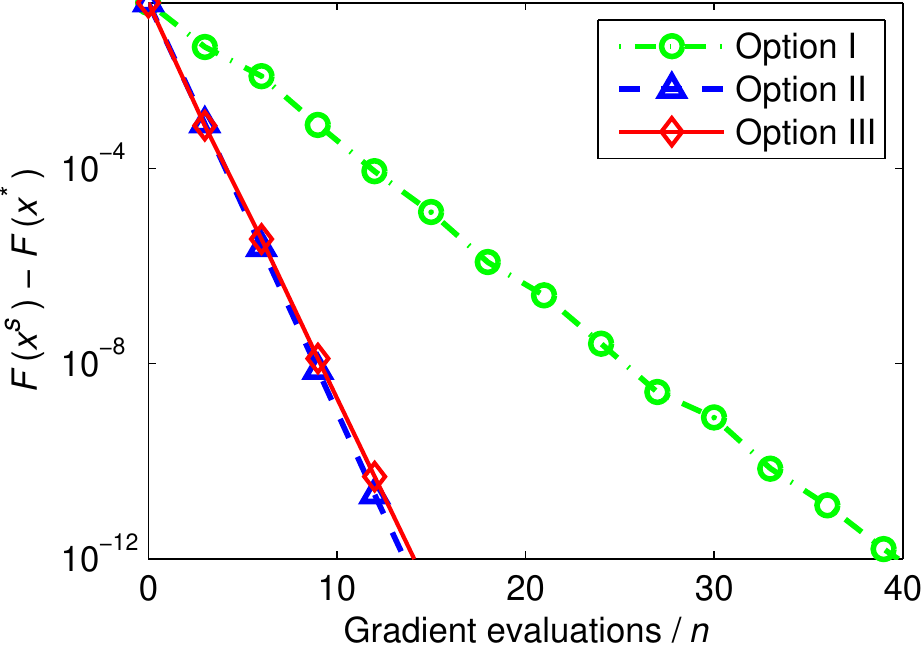}\:\includegraphics[width=0.246\columnwidth]{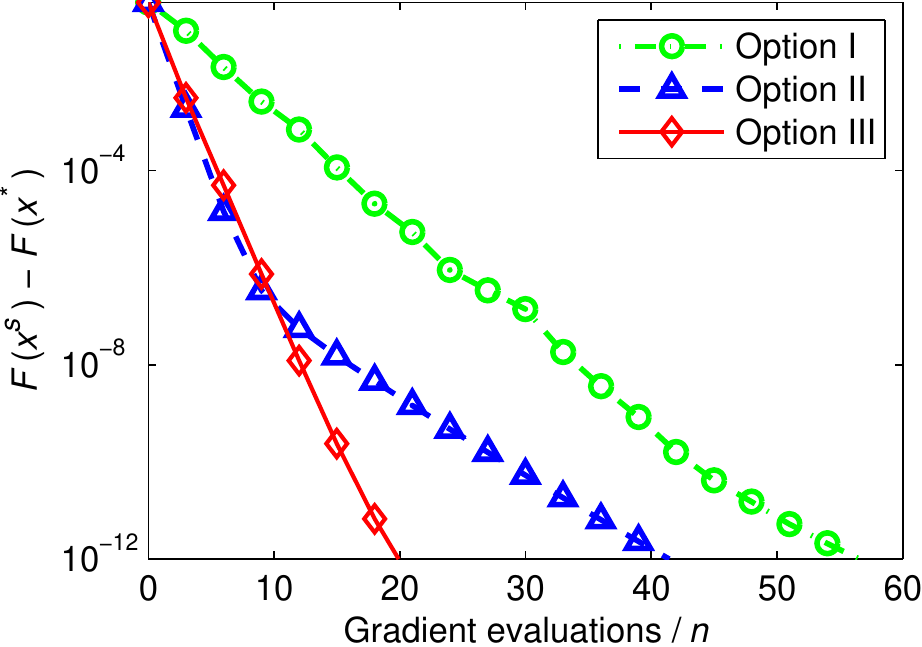}}
\caption{Comparison of the stochastic algorithms with Options I, II, and III for solving ridge regression problems with the regularizer $(\lambda/2)\|\cdot\|^{2}$. In each plot, the vertical axis shows the objective value minus the minimum, and the horizontal axis is the number of effective passes.}
\label{figs13}
\end{figure}

\begin{figure}[t]
\centering
\subfigure[Adult: $\lambda=10^{-4}$ (left) \;and\; $\lambda=10^{-5}$ (right)]{\includegraphics[width=0.246\columnwidth]{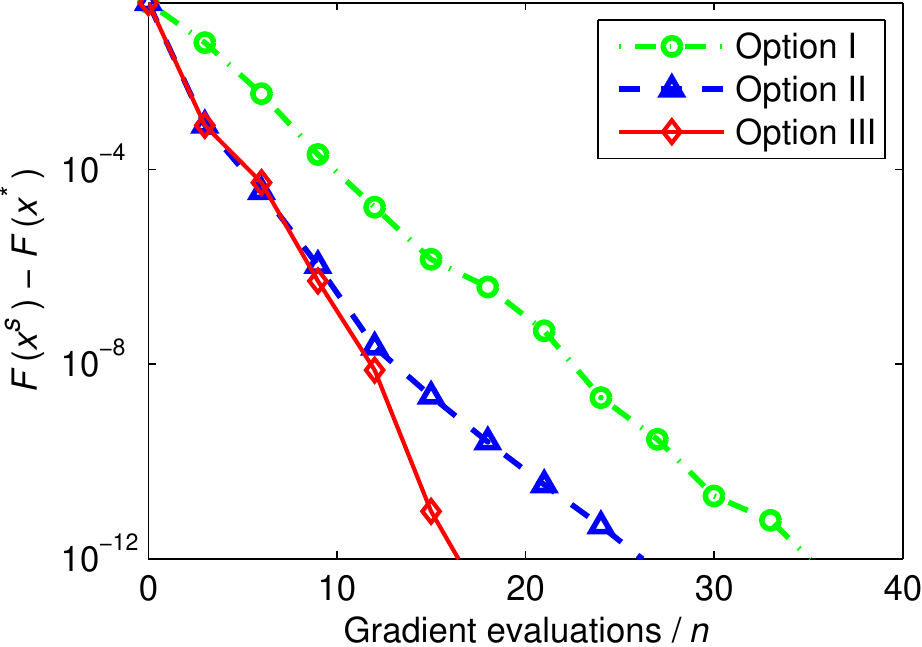}\:\includegraphics[width=0.246\columnwidth]{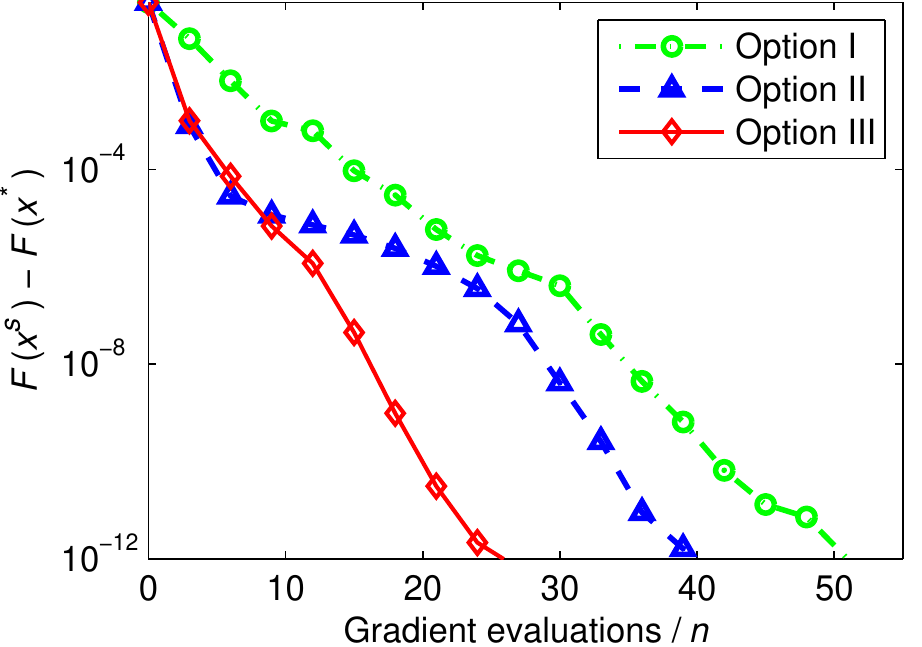}}
\subfigure[MNIST: $\lambda=10^{-4}$ (left) \;and\; $\lambda=10^{-5}$ (right)]{\includegraphics[width=0.246\columnwidth]{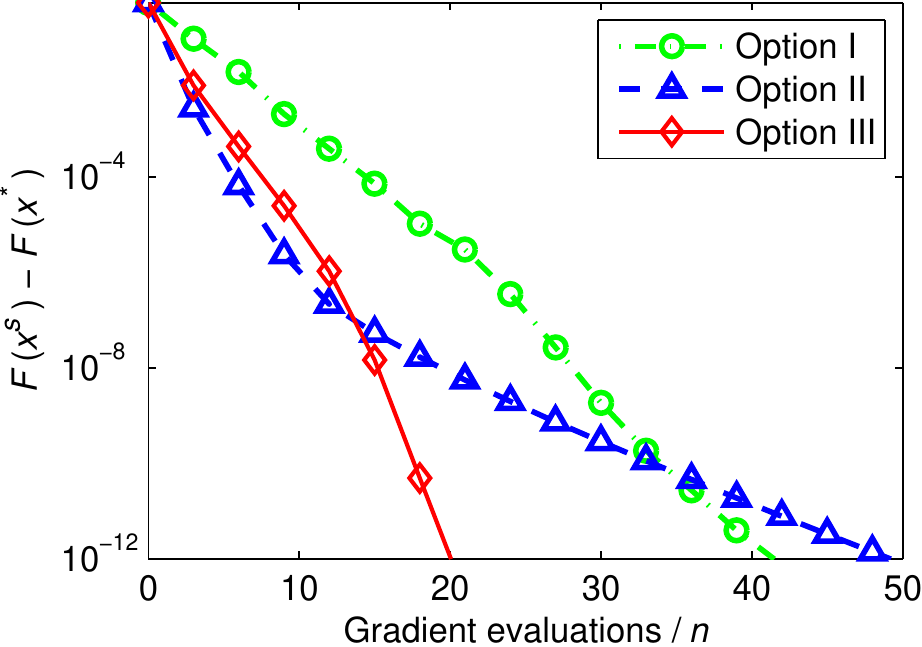}\:\includegraphics[width=0.246\columnwidth]{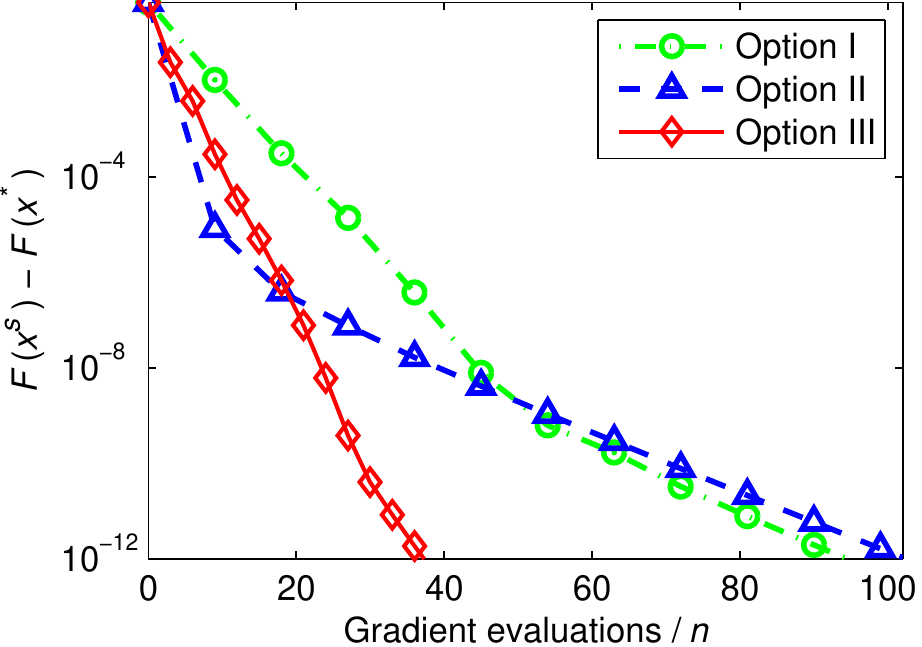}}

\subfigure[RCV1: $\lambda=10^{-4}$ (left) \;and\; $\lambda=10^{-5}$ (right)]{\includegraphics[width=0.246\columnwidth]{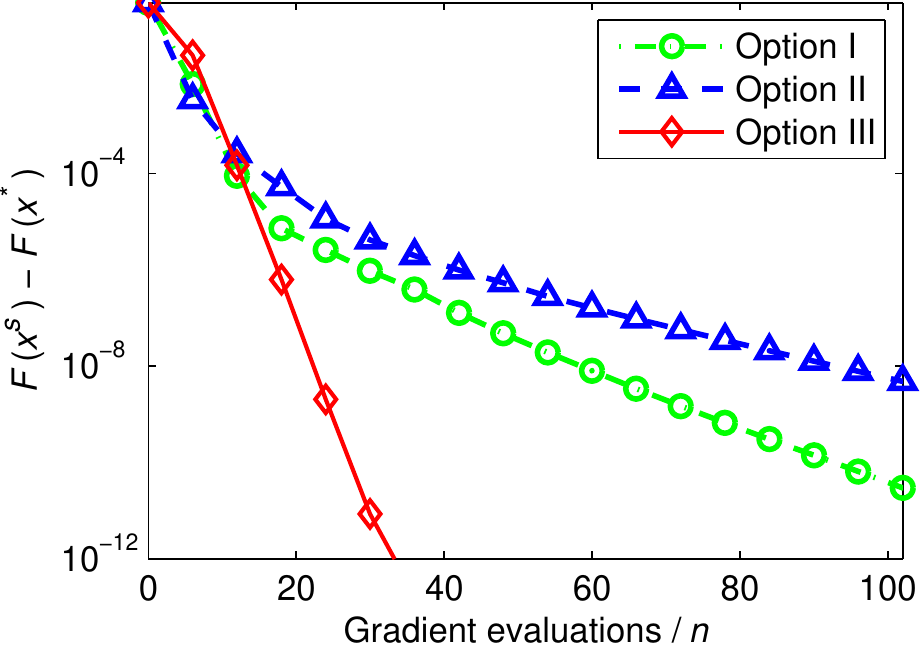}\:\includegraphics[width=0.246\columnwidth]{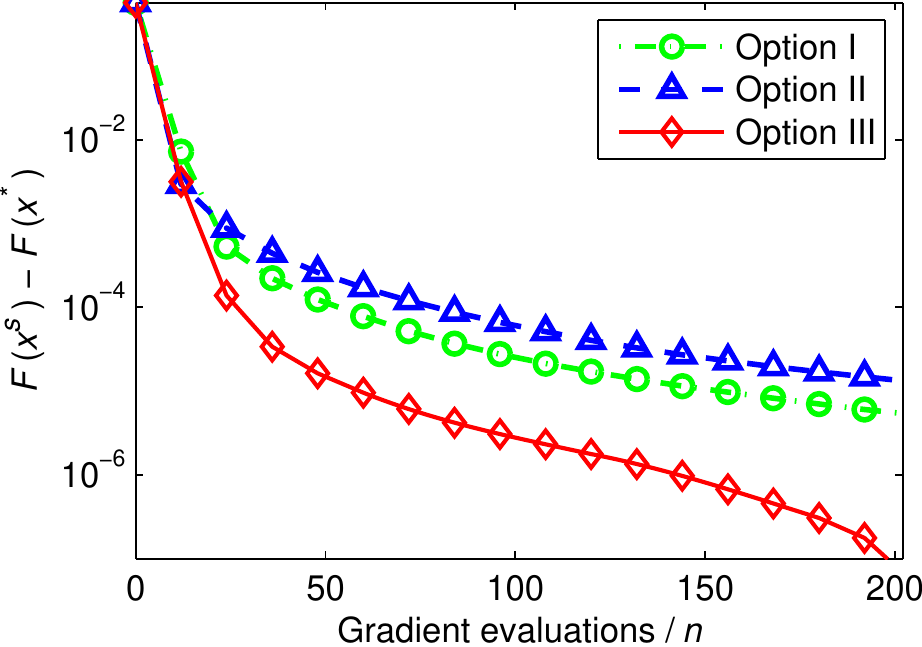}}
\subfigure[Epsilon: $\lambda=10^{-4}$ (left) \;and\; $\lambda=10^{-5}$ (right)]{\includegraphics[width=0.246\columnwidth]{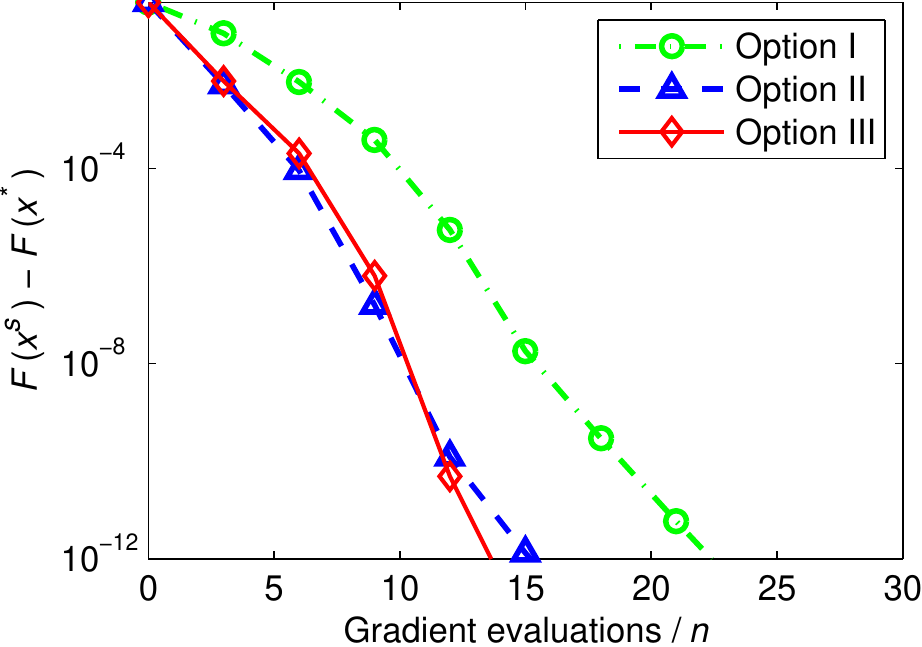}\:\includegraphics[width=0.246\columnwidth]{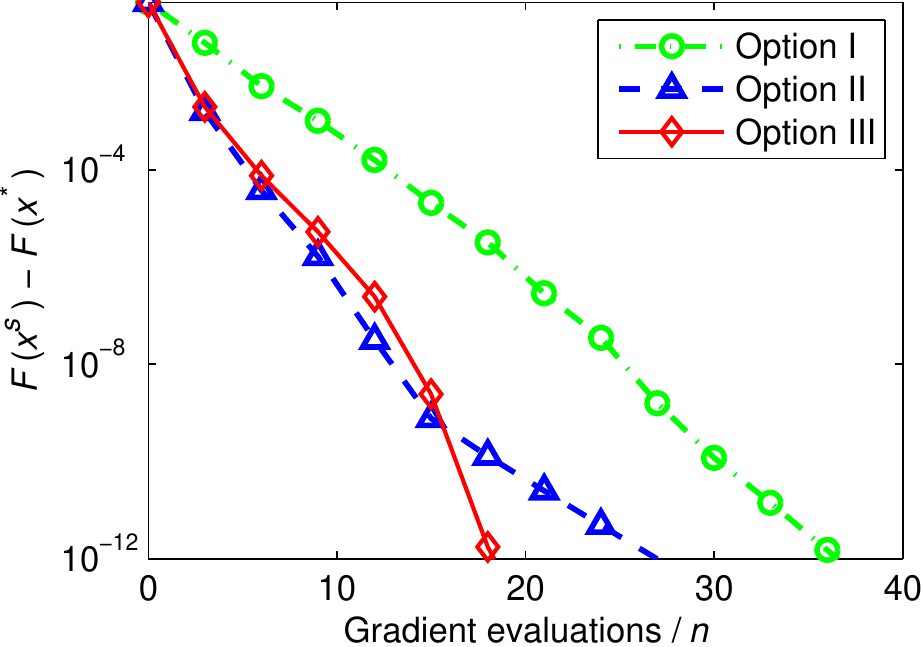}}
\caption{Comparison of the stochastic algorithms with Options I, II, and III for solving Lasso problems with the regularizer $\lambda\|\cdot\|_{1}$. In each plot, the vertical axis shows the objective value minus the minimum, and the horizontal axis is the number of effective passes.}
\label{figs14}
\end{figure}

\begin{figure}[!th]
\centering
\includegraphics[width=0.326\columnwidth]{Fig11}\,
\includegraphics[width=0.326\columnwidth]{Fig13}\,
\includegraphics[width=0.326\columnwidth]{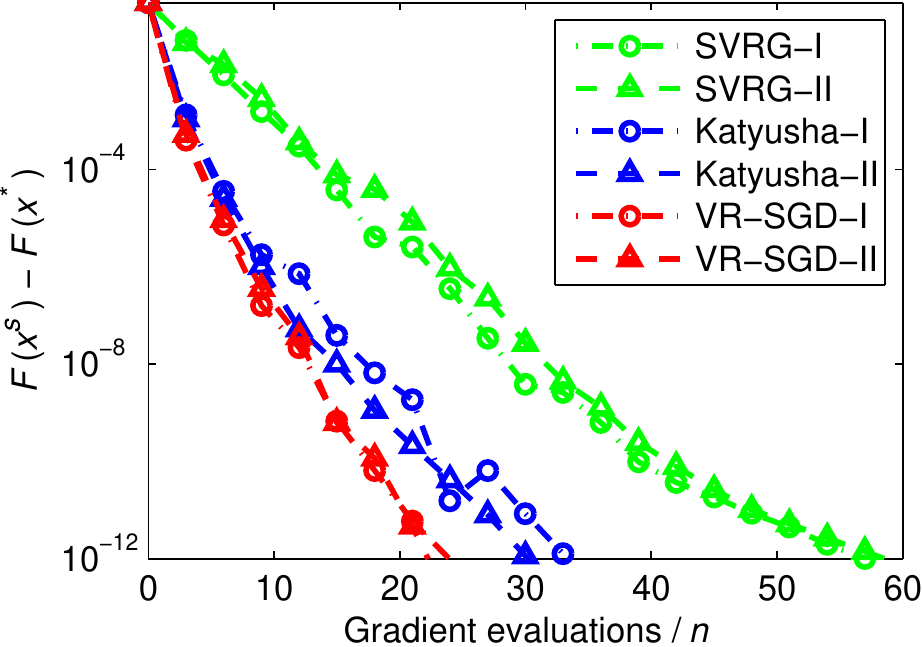}

\subfigure[$\lambda=10^{-3}$]{\includegraphics[width=0.326\columnwidth]{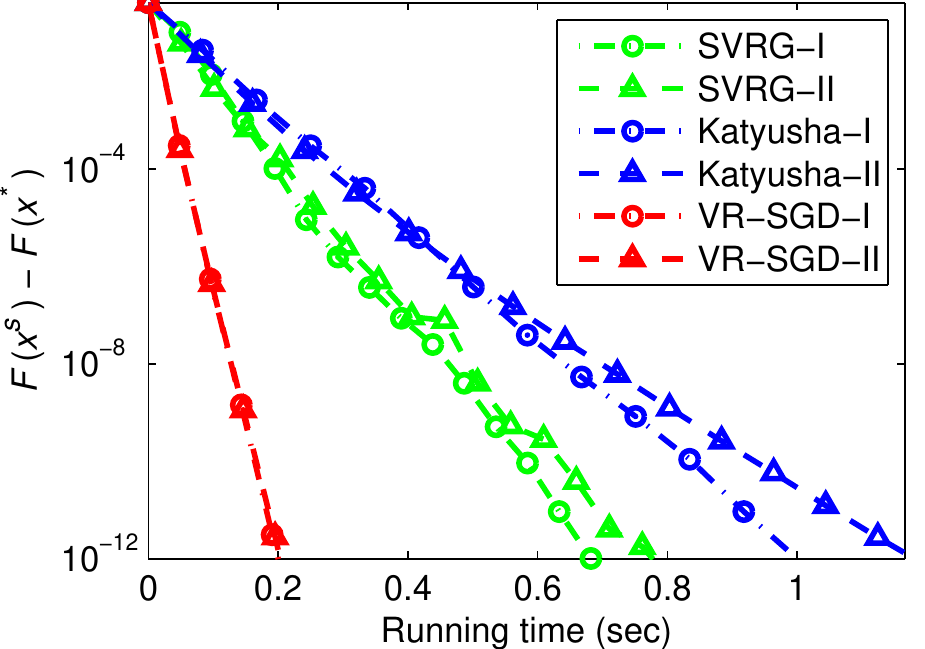}}\,
\subfigure[$\lambda=10^{-4}$]{\includegraphics[width=0.326\columnwidth]{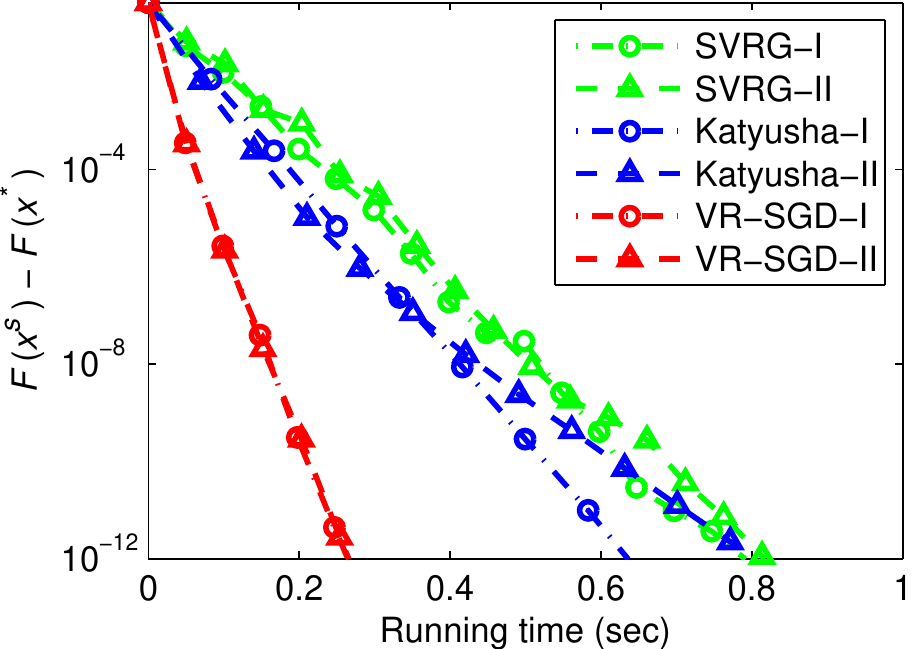}}\,
\subfigure[$\lambda=10^{-5}$]{\includegraphics[width=0.326\columnwidth]{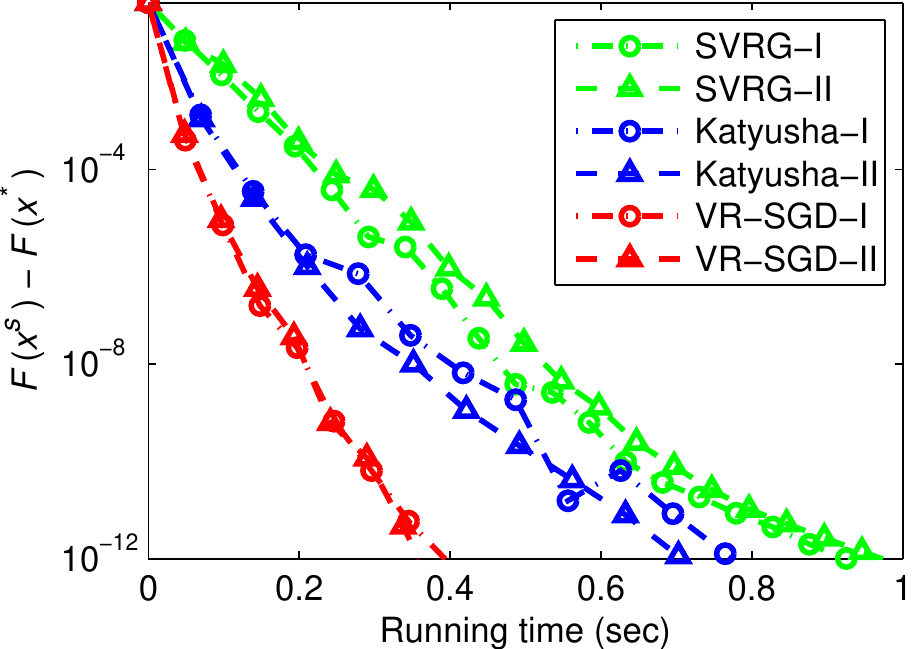}}
\vspace{1.6mm}

\includegraphics[width=0.326\columnwidth]{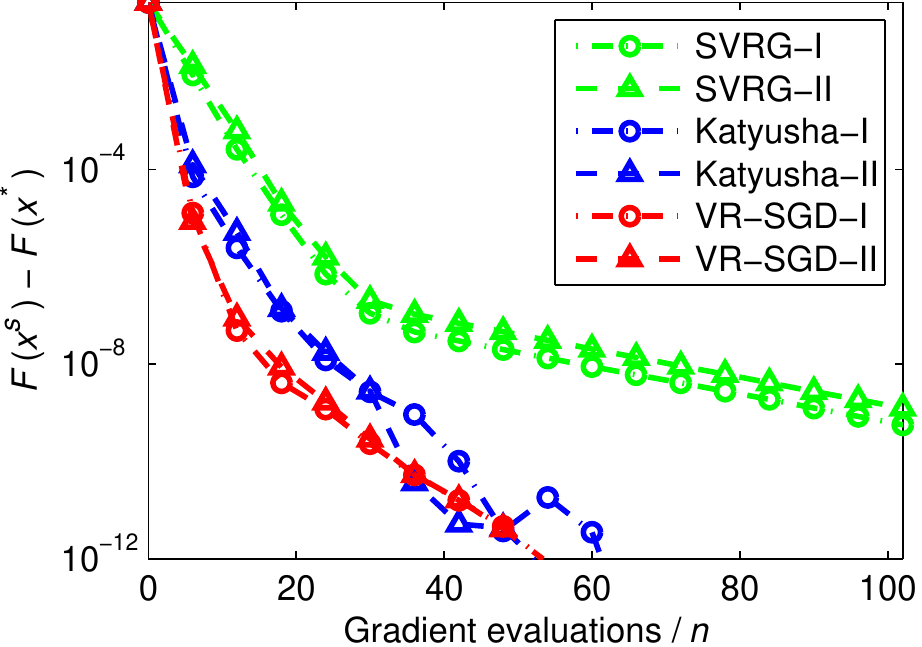}\,
\includegraphics[width=0.326\columnwidth]{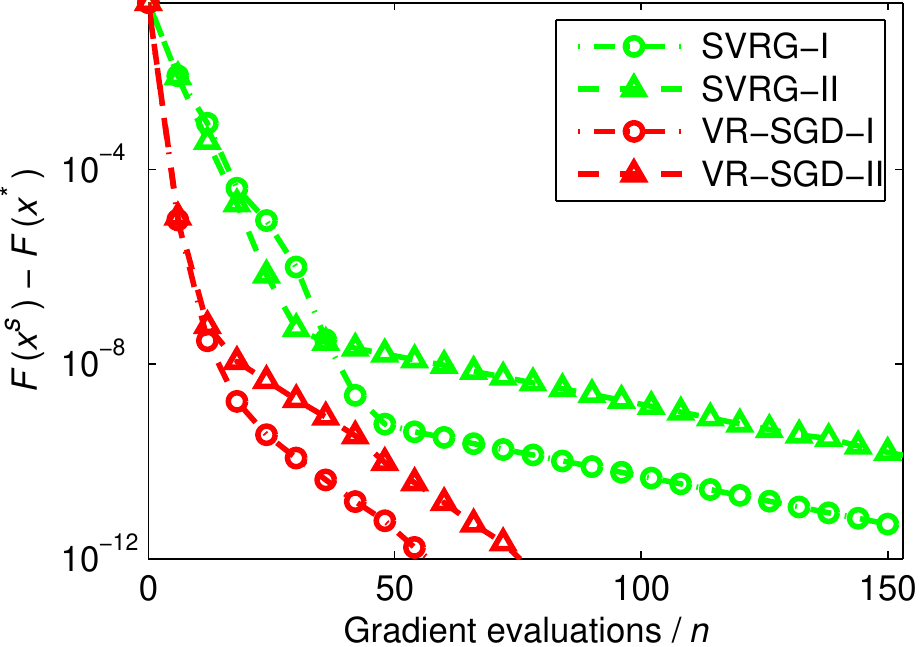}\,
\includegraphics[width=0.326\columnwidth]{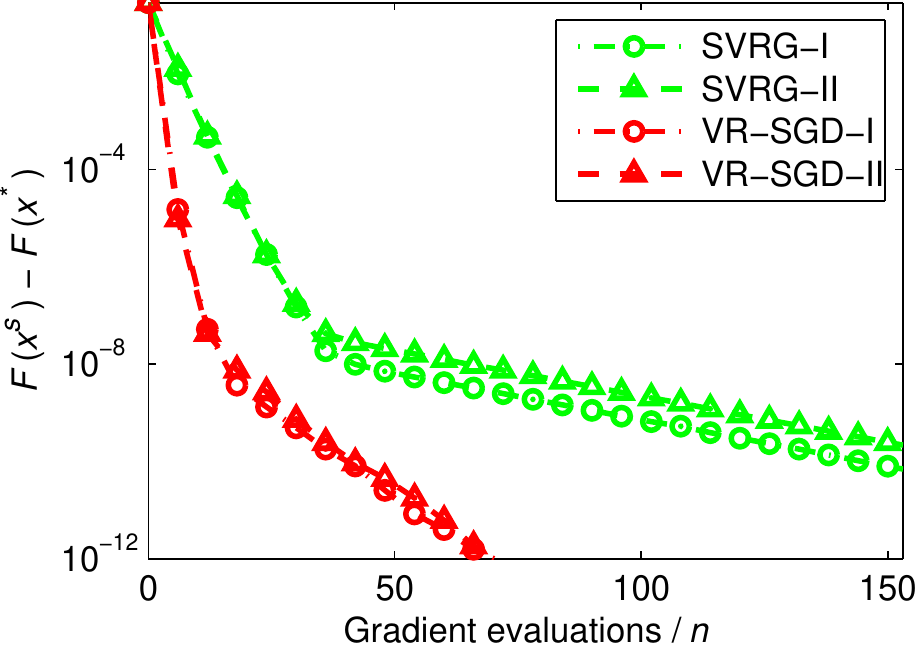}

\subfigure[$\lambda=10^{-6}$]{\includegraphics[width=0.326\columnwidth]{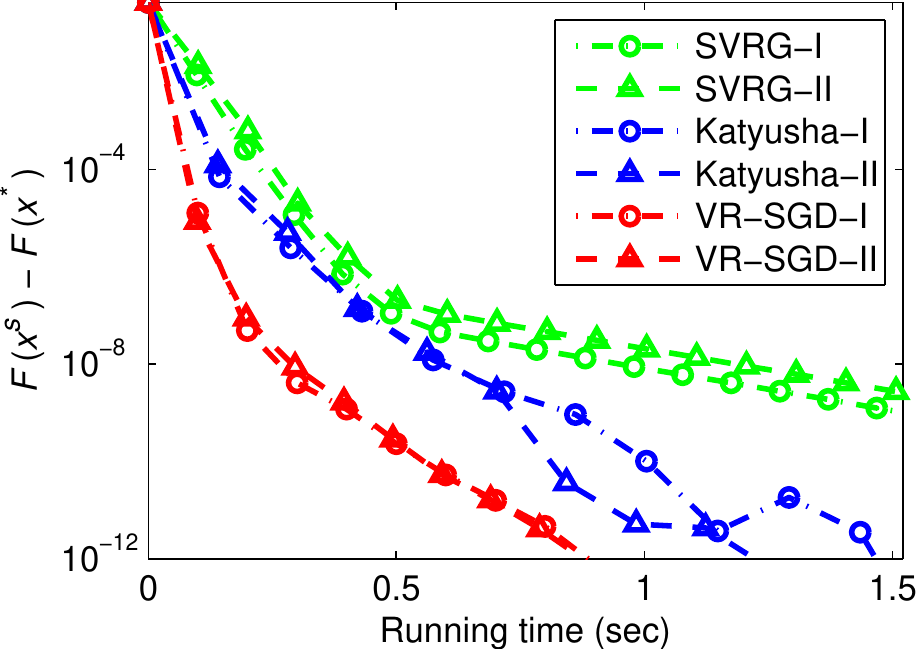}\label{figs4d}}\,
\subfigure[$\lambda=10^{-7}$]{\includegraphics[width=0.326\columnwidth]{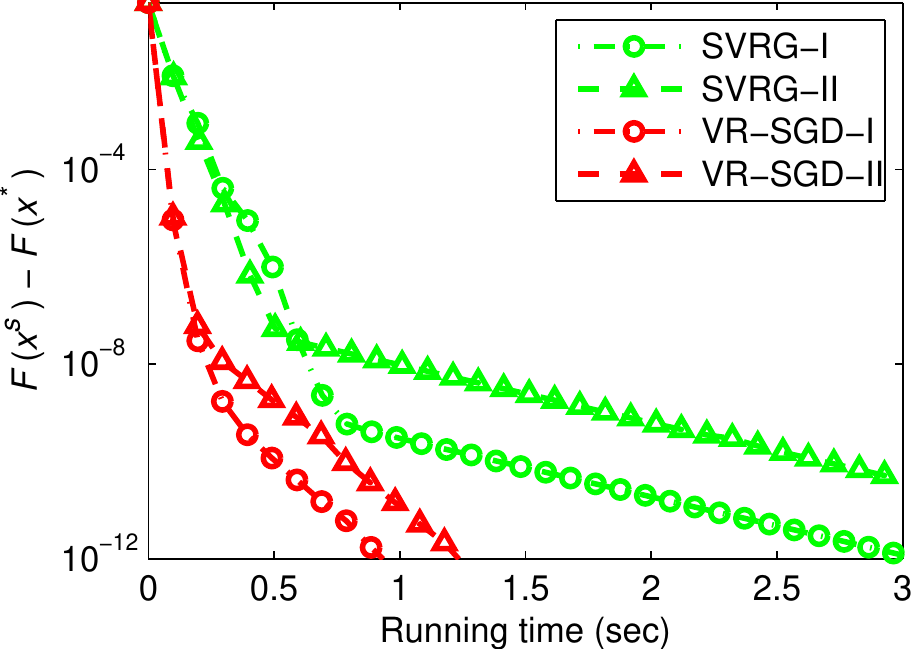}\label{figs4e}}\,
\subfigure[$\lambda=0$]{\includegraphics[width=0.326\columnwidth]{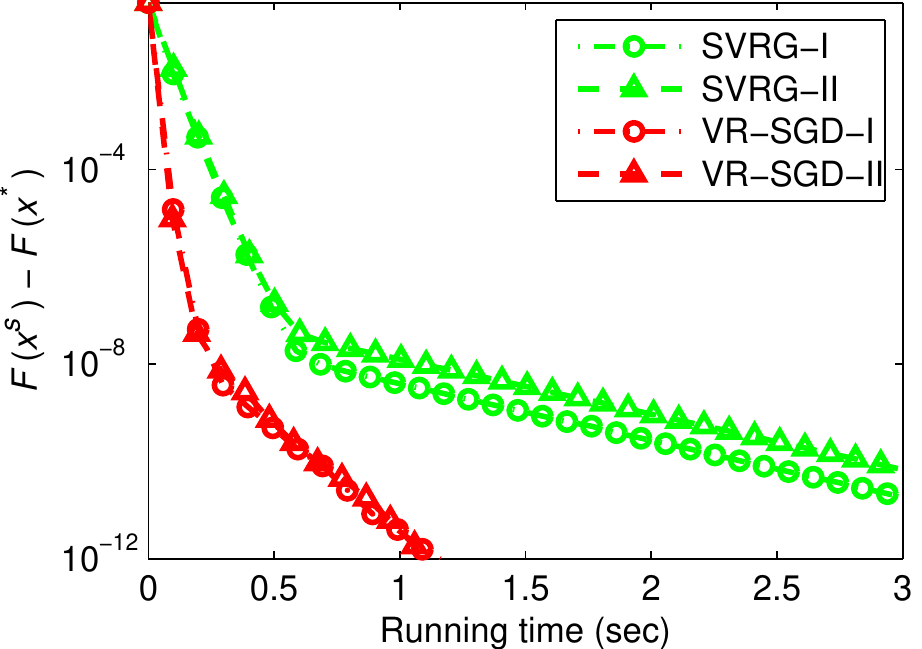}\label{figs4f}}
\caption{Comparison of SVRG~\cite{johnson:svrg}, Katyusha~\cite{zhu:Katyusha}, our VR-SGD method and their proximal versions for solving ridge regression problems with different regularization parameters on Adult. In each plot, the vertical axis shows the objective value minus the minimum, and the horizontal axis is the number of effective passes (top) or running time (bottom).}
\label{figs15}
\end{figure}

\begin{figure}[!th]
\centering
\includegraphics[width=0.326\columnwidth]{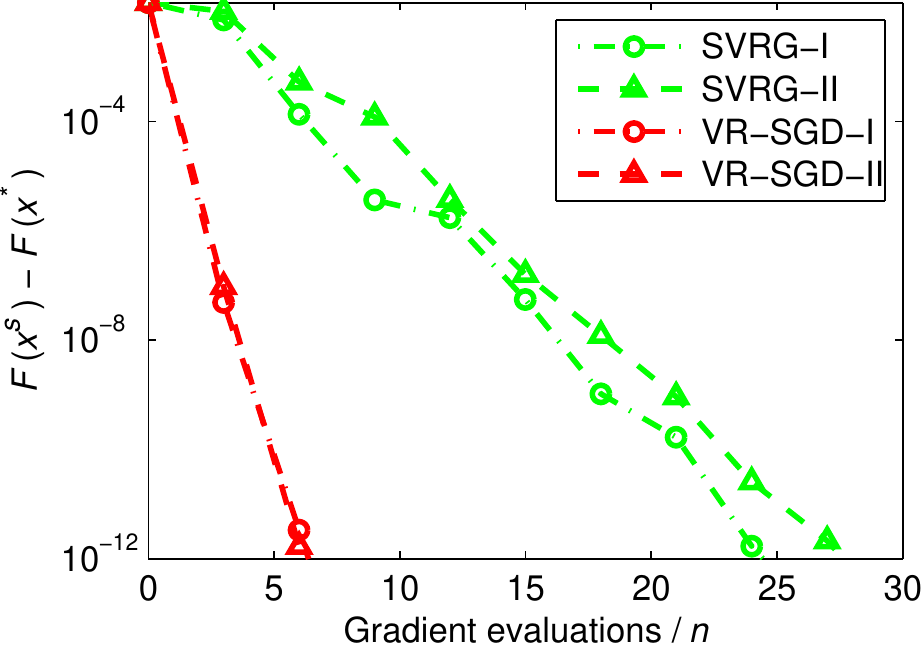}\,
\includegraphics[width=0.326\columnwidth]{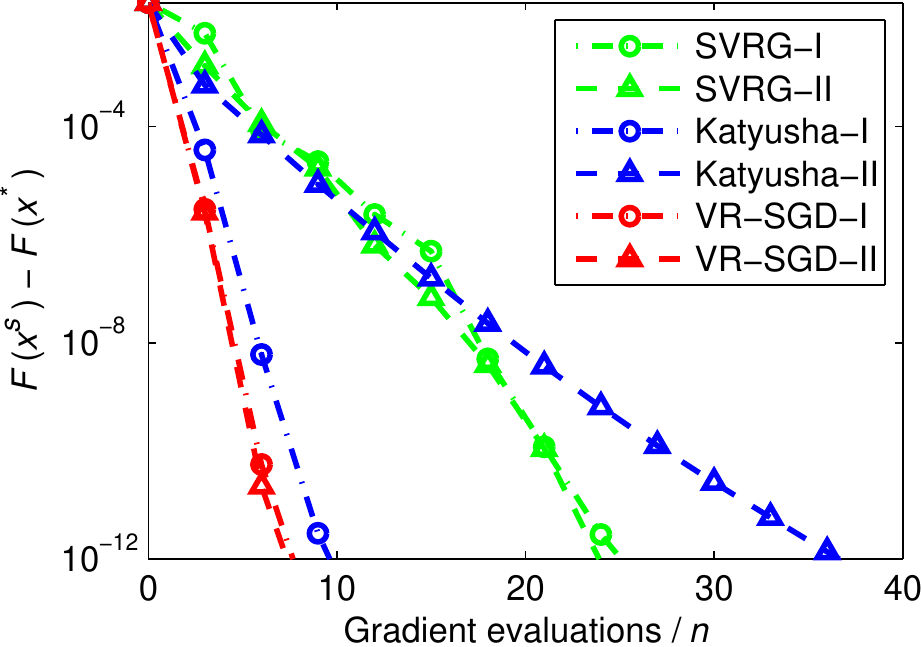}\,
\includegraphics[width=0.326\columnwidth]{Fig35}

\subfigure[$\lambda=10^{-3}$]{\includegraphics[width=0.326\columnwidth]{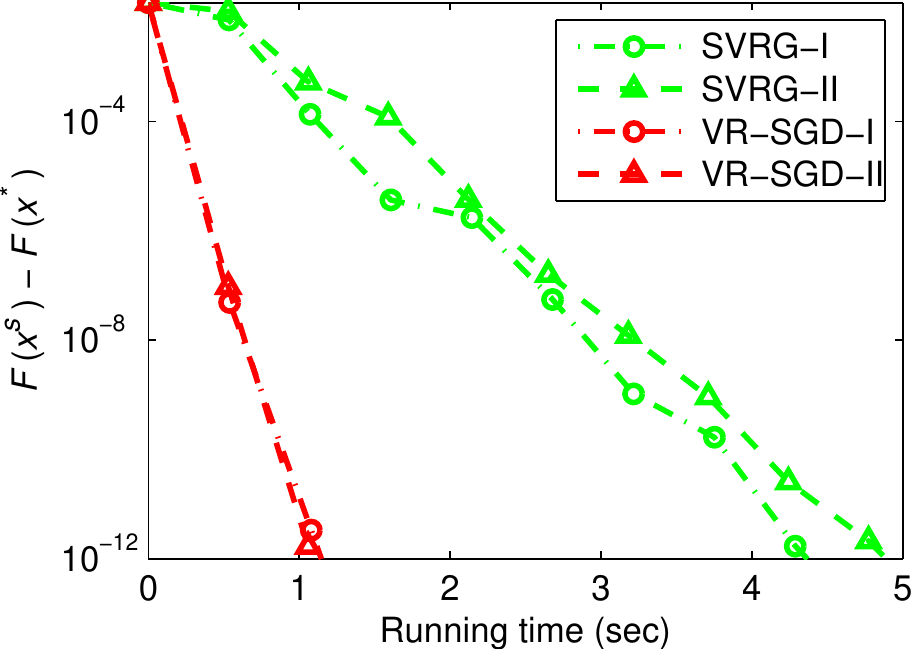}\label{figs5a}}\,
\subfigure[$\lambda=10^{-4}$]{\includegraphics[width=0.326\columnwidth]{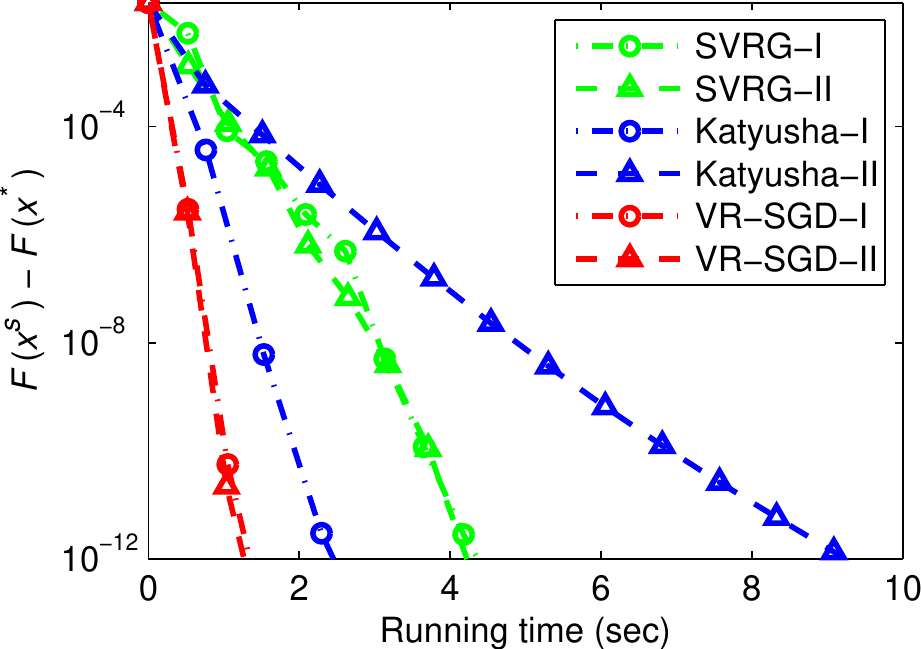}}\,
\subfigure[$\lambda=10^{-5}$]{\includegraphics[width=0.326\columnwidth]{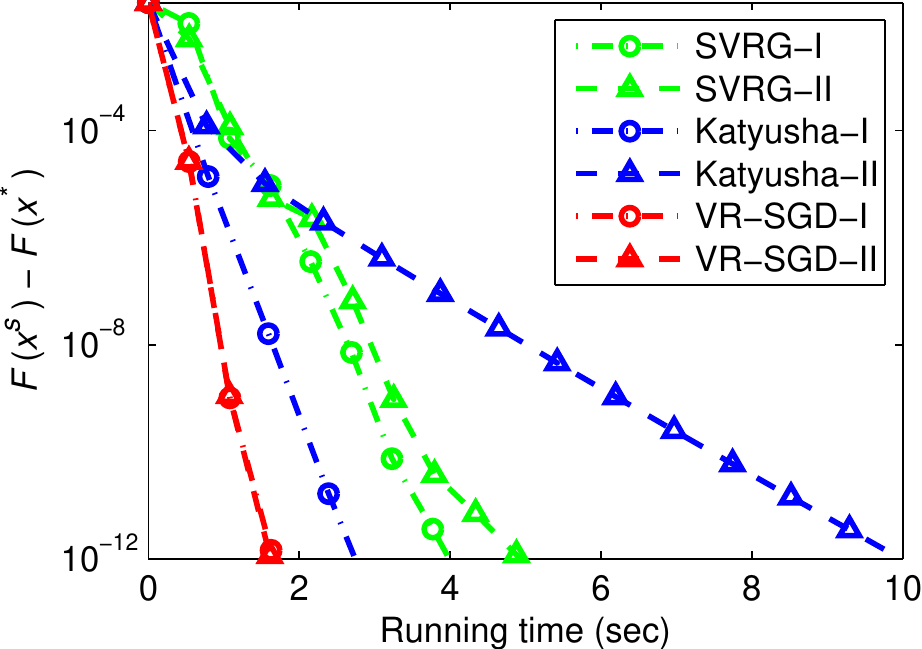}}
\vspace{1.6mm}

\includegraphics[width=0.326\columnwidth]{Fig37}\,
\includegraphics[width=0.326\columnwidth]{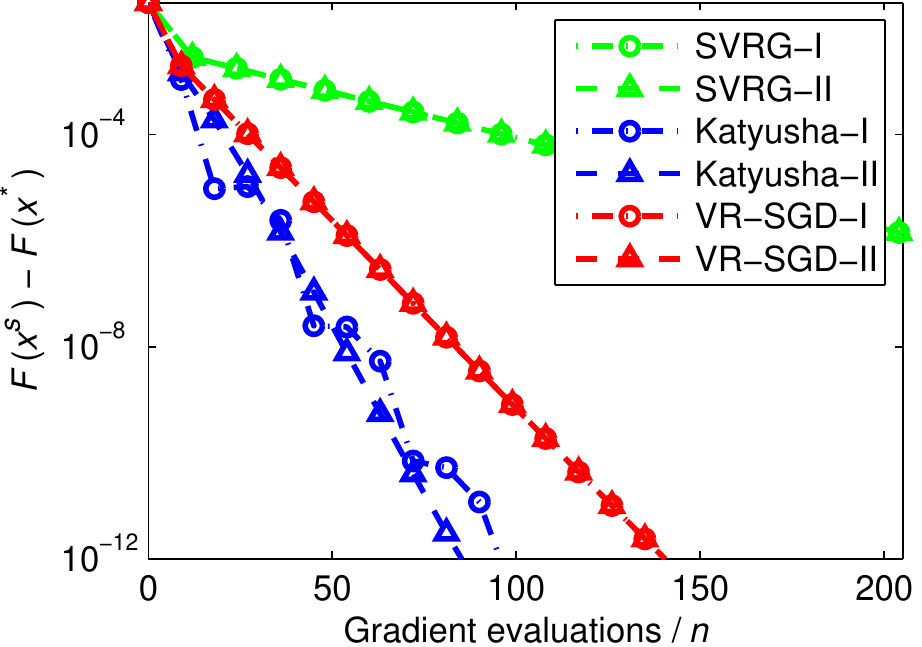}\,
\includegraphics[width=0.326\columnwidth]{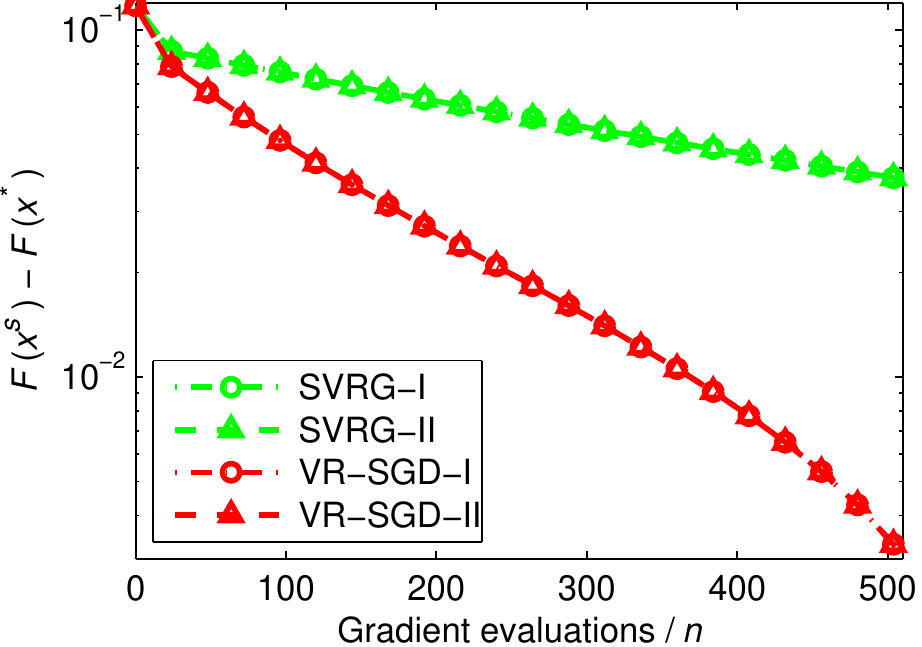}

\subfigure[$\lambda=10^{-6}$]{\includegraphics[width=0.326\columnwidth]{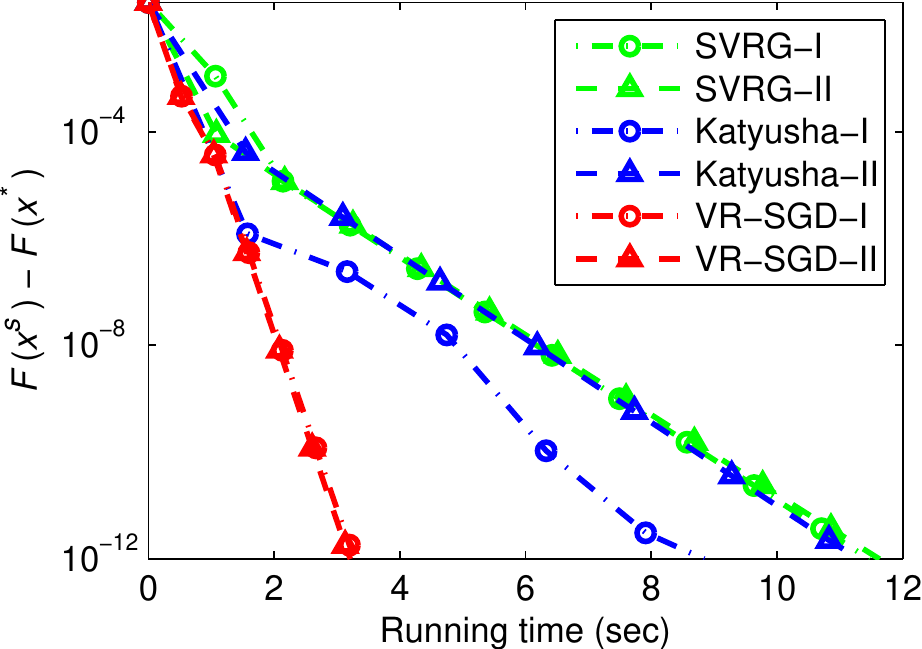}\label{figs5d}}\,
\subfigure[$\lambda=10^{-7}$]{\includegraphics[width=0.326\columnwidth]{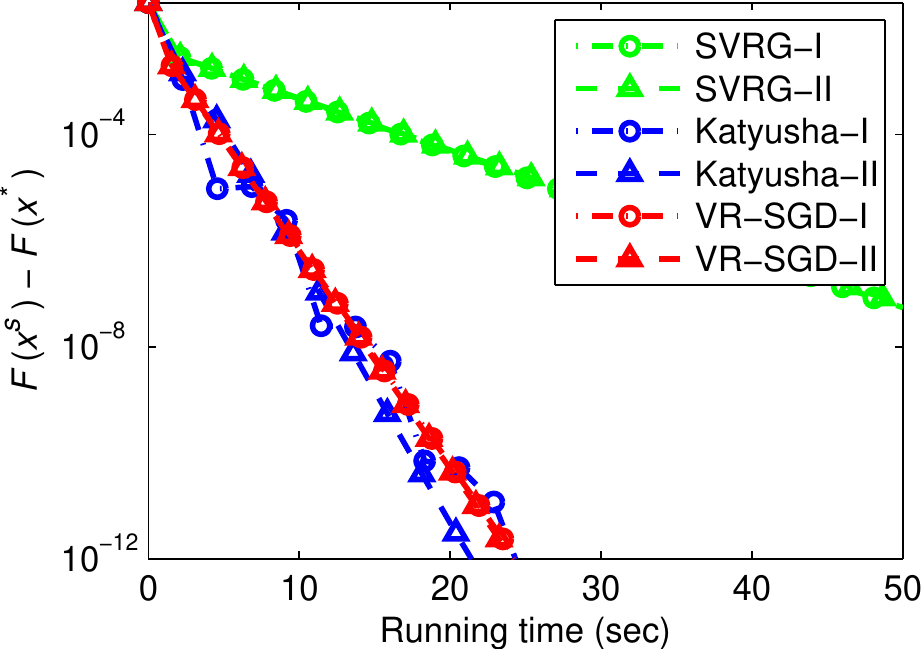}}\,
\subfigure[$\lambda=0$]{\includegraphics[width=0.326\columnwidth]{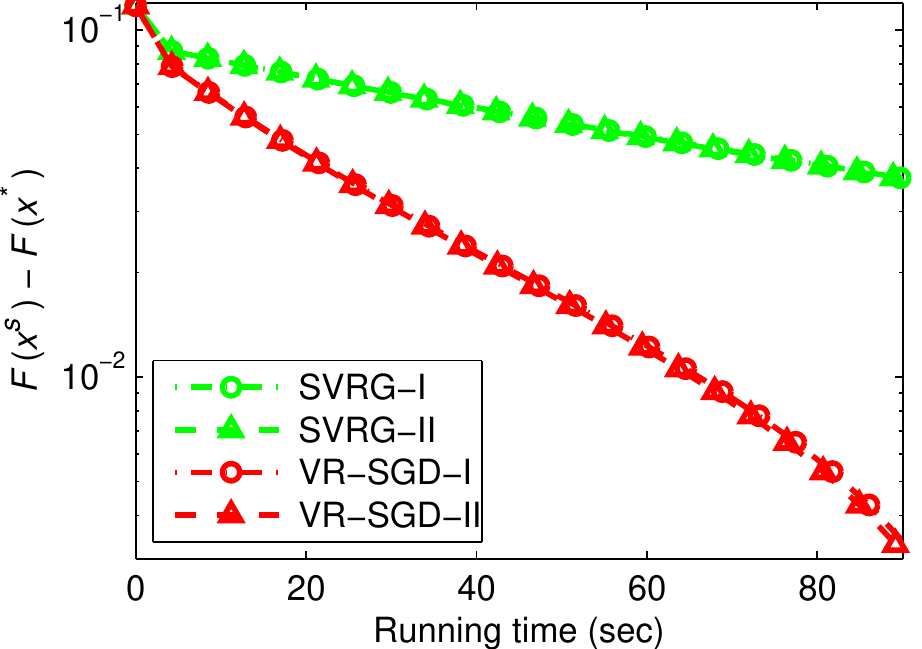}\label{figs5f}}
\caption{Comparison of SVRG~\cite{johnson:svrg}, Katyusha~\cite{zhu:Katyusha}, our VR-SGD method and their proximal versions for solving ridge regression problems with different regularization parameters on Covtype. In each plot, the vertical axis shows the objective value minus the minimum, and the horizontal axis is the number of effective passes (top) or running time (bottom).}
\label{figs16}
\end{figure}

\begin{figure}[t]
\centering
\subfigure[Adult: $\lambda=10^{-4}$ (left) \;and\; $\lambda=10^{-5}$ (right)]{\includegraphics[width=0.246\columnwidth]{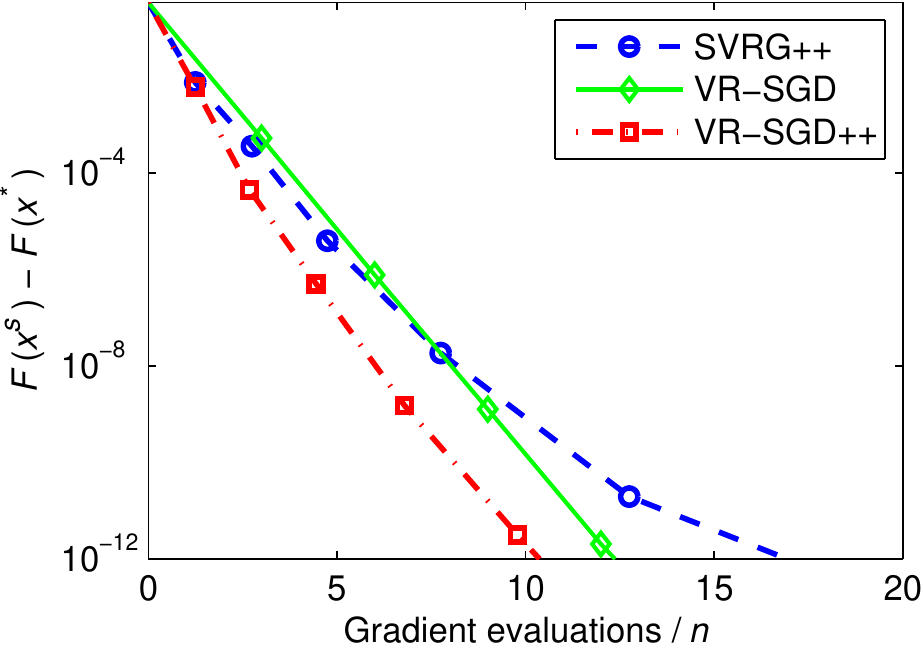}\:\includegraphics[width=0.246\columnwidth]{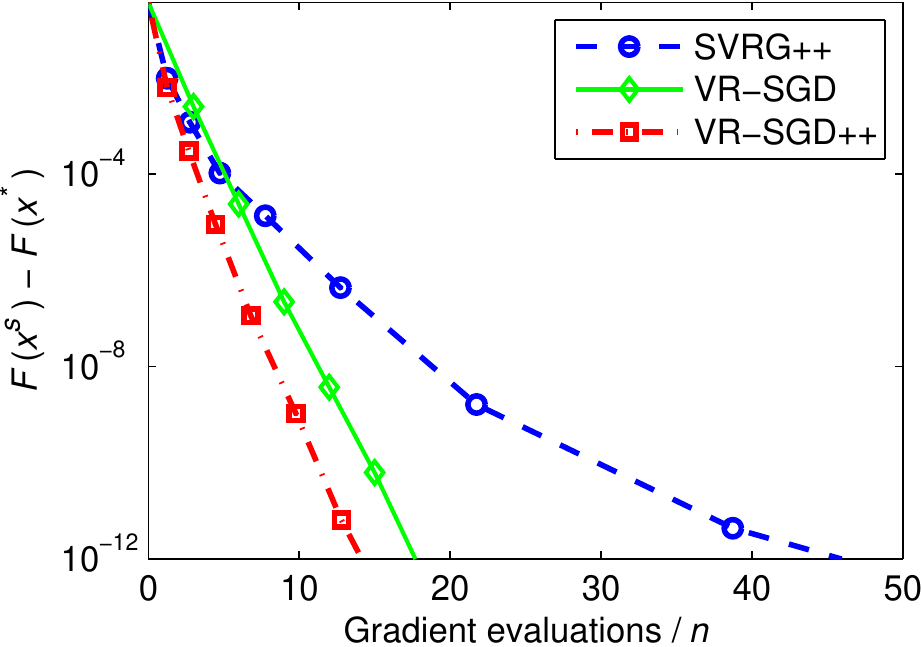}}
\subfigure[MNIST: $\lambda=10^{-4}$ (left) \;and\; $\lambda=10^{-5}$ (right)]{\includegraphics[width=0.246\columnwidth]{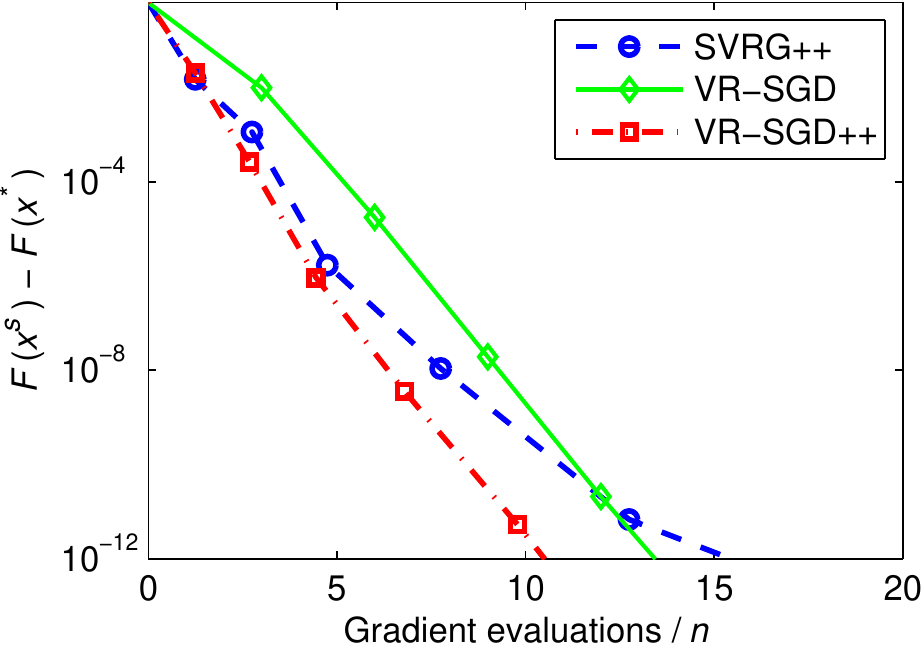}\:\includegraphics[width=0.246\columnwidth]{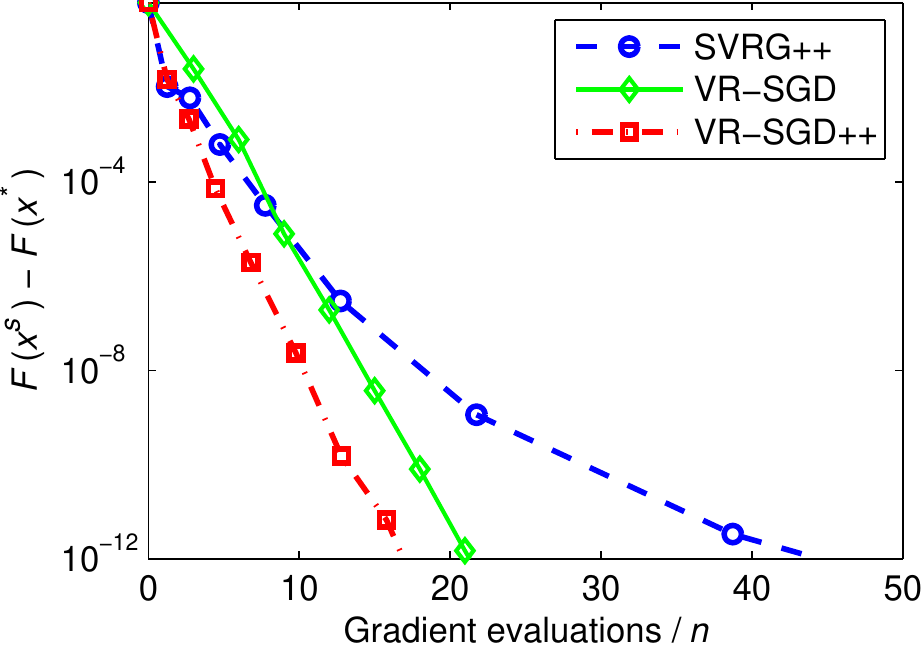}}
\vspace{1.6mm}

\subfigure[Covtype: $\lambda=10^{-5}$ (left) \;and\; $\lambda=10^{-6}$ (right)]{\includegraphics[width=0.246\columnwidth]{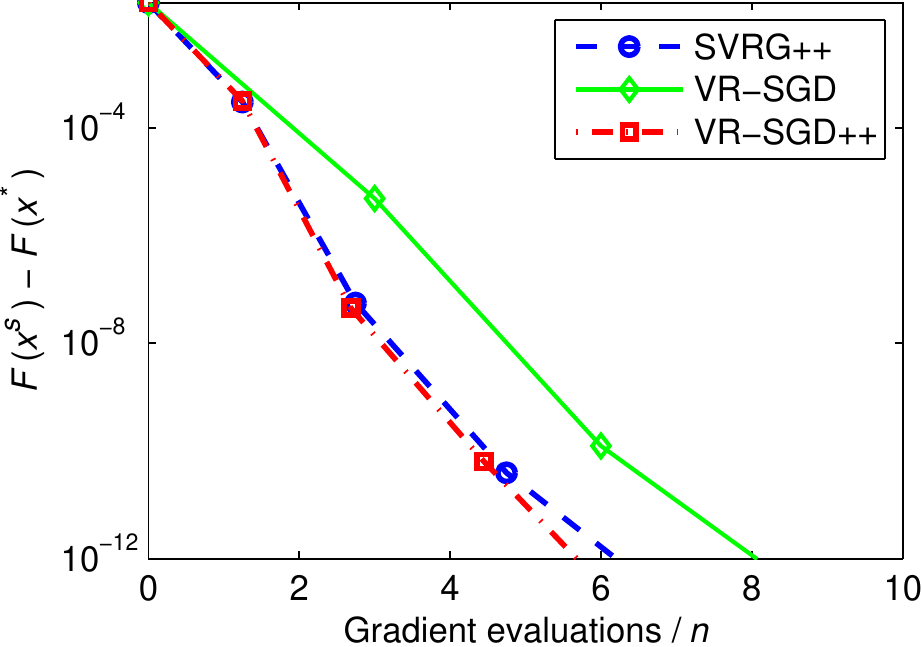}\:\includegraphics[width=0.246\columnwidth]{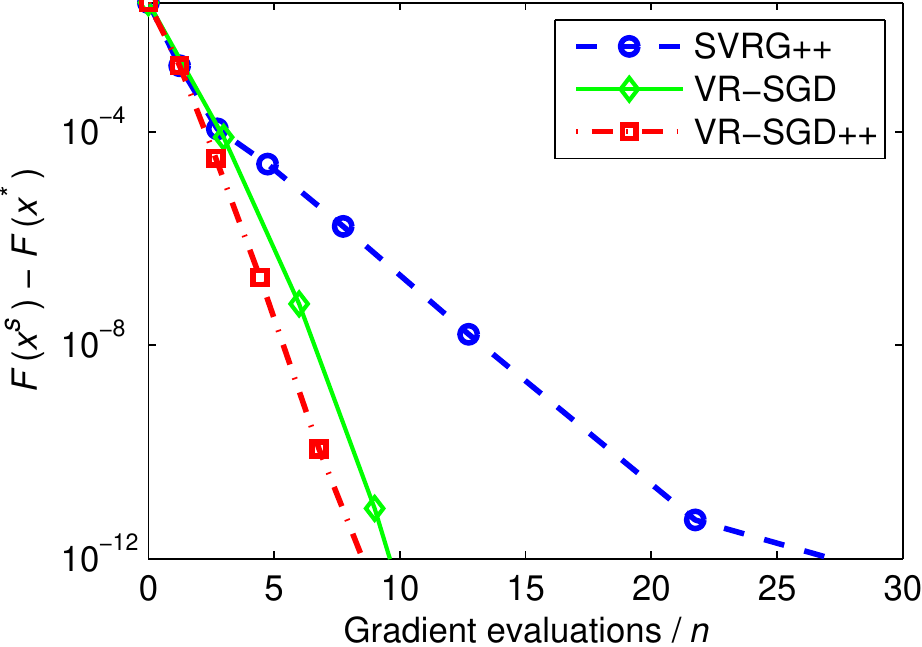}}
\subfigure[RCV1: $\lambda=10^{-4}$ (left) \;and\; $\lambda=10^{-5}$ (right)]{\includegraphics[width=0.246\columnwidth]{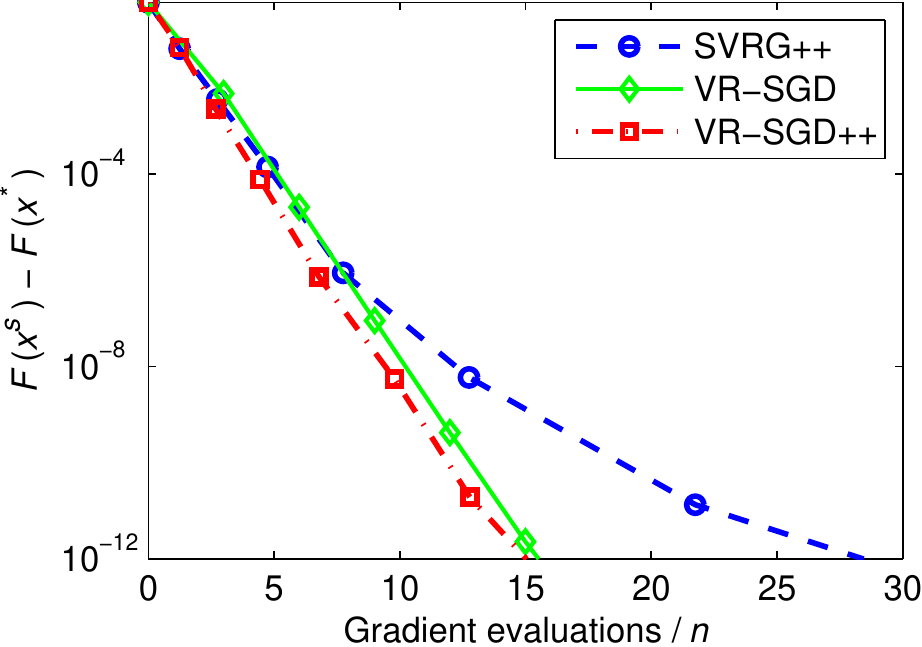}\:\includegraphics[width=0.246\columnwidth]{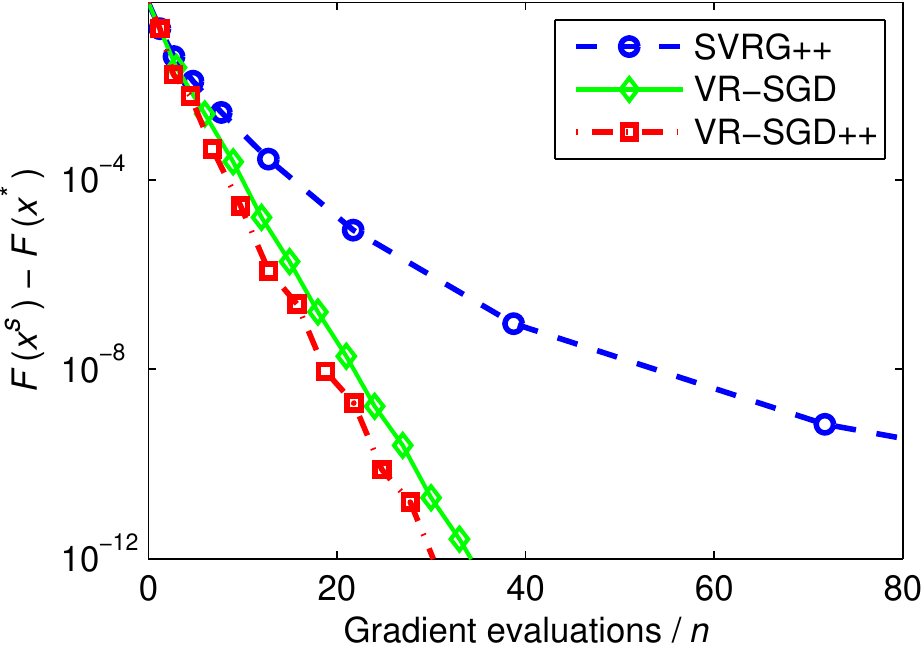}}
\caption{Comparison of SVRG++~\cite{zhu:univr}, VR-SGD and VR-SGD++ for solving logistic regression problems with the regularizer $(\lambda/2)\|\cdot\|^{2}$. In each plot, the vertical axis shows the objective value minus the minimum, and the horizontal axis is the number of effective passes.}
\label{figs31}
\end{figure}

\subsection*{More Results for Different Choices of Snapshot and Starting Points}
\label{sec52}
In this part, we presented more experimental results of the algorithms with the three choices (i.e., Options I, II and III in Table I in the main paper) for snapshot and starting points for solving ridge regression and Lasso problems, as shown in Figs.\ \ref{figs13} and~\ref{figs14}. All the results show that the setting Option III suggested in this paper (i.e., $\widetilde{x}^{s}\!=\frac{1}{m}\!\sum^{m}_{k=1}\!x^{s}_{k}$ and $x^{s+1}_{0}\!=x^{s}_{m}$) is a better choice than Options I and II for stochastic variance reduction optimization.

\subsection*{More Results for Common Stochastic Gradient and Prox-SG Updates}
In this part, we reported more results of SVRG~\cite{johnson:svrg}, Katyusha~\cite{zhu:Katyusha}, VR-SGD and their proximal variants in Figs.\ \ref{figs15} and~\ref{figs16}.

\subsection*{More Results for SVRG++ and VR-SGD++}
We also presented more experimental results of SVRG++~\cite{zhu:univr}, VR-SGD++ (i.e., VR-SGD with reducing the number of gradient calculations in early iterations, as shown in Algorithm~\ref{alg4}), and VR-SGD, as shown in Fig.\ \ref{figs31}. The results show that both VR-SGD++ and VR-SGD outperform SVRG++, which also means that if the epoch size is too large (due to the doubling-epoch technique used in~\cite{zhu:univr}), SVRG++ becomes slower and slower for later iterations.

\subsection*{More Results for Logistic Regression}
\label{sec53}
Figs.\ \ref{figs17}, \ref{figs18}, and~\ref{figs19} show the experimental results for $\ell_{2}$-norm (i.e., $\lambda_{2}\!=\!0$), $\ell_{1}$-norm (i.e., $\lambda_{1}\!=\!0$), and elastic net (i.e., $\lambda_{1}\!\neq\!0$ and $\lambda_{2}\!\neq\!0$) regularized logistic regression with different regularization parameters.

\subsection*{More Results for ERM with Non-Convex Sigmoid Loss}
Finally, we presented more experimental results (including training objective value and function suboptimality) in Figs.\ \ref{figs20} and~\ref{figs21}. Note that $x_{*}$ denotes the best solution obtained by running all those methods for a large number of iterations and multiple random initializations.

\begin{figure}[th]
\centering
\includegraphics[width=0.245\columnwidth]{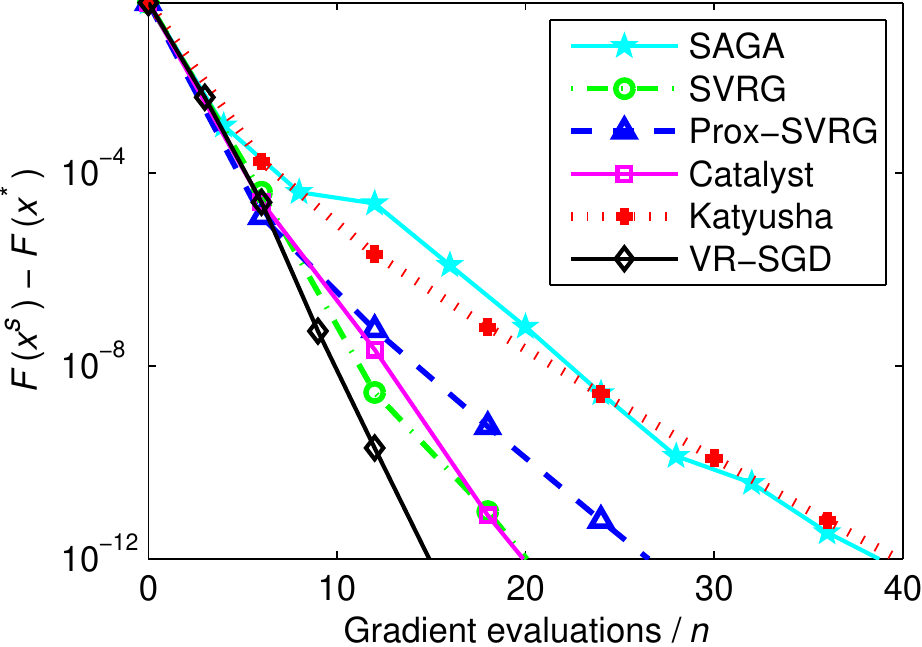}
\includegraphics[width=0.245\columnwidth]{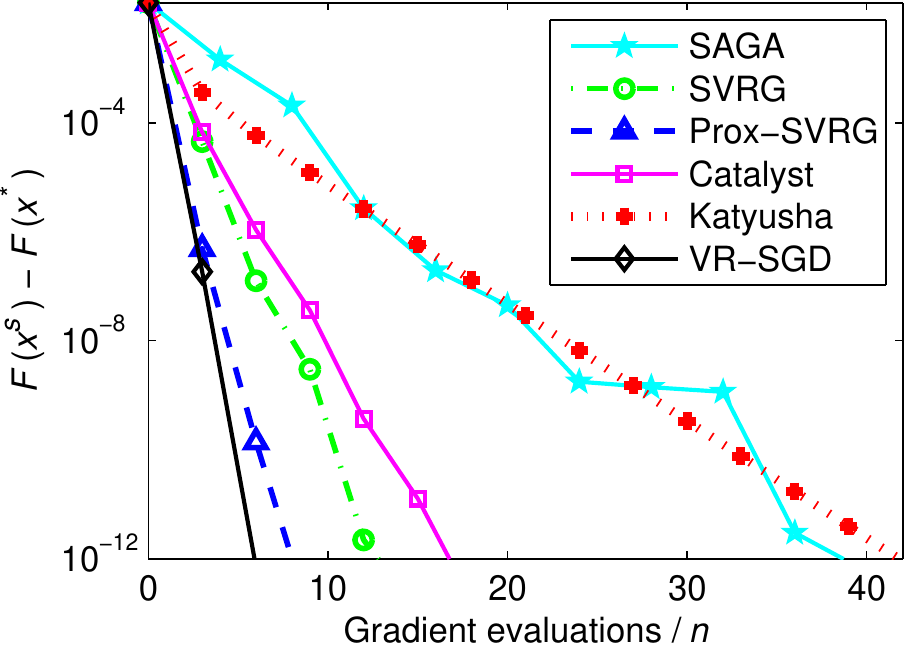}
\includegraphics[width=0.245\columnwidth]{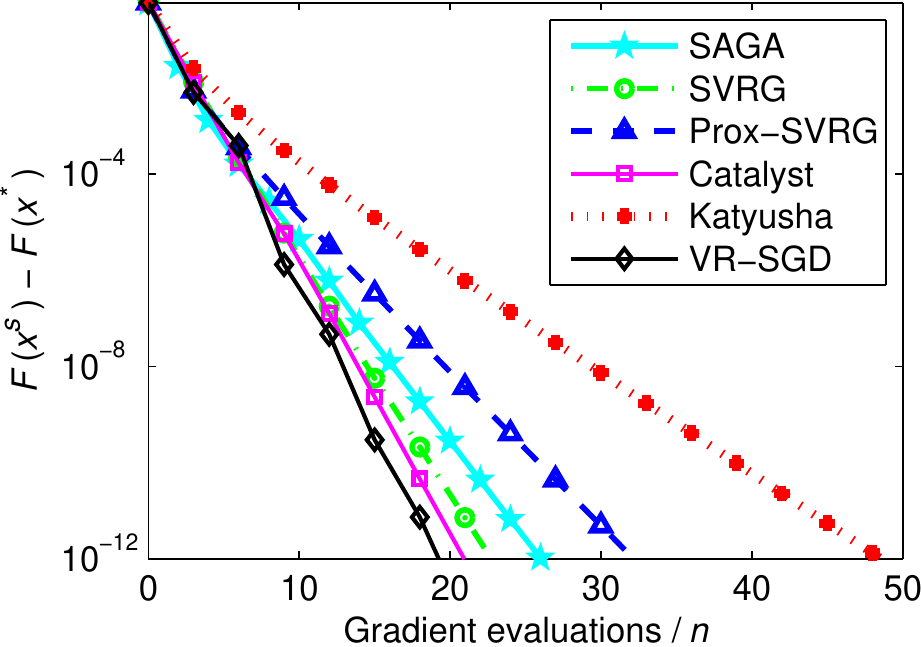}
\includegraphics[width=0.245\columnwidth]{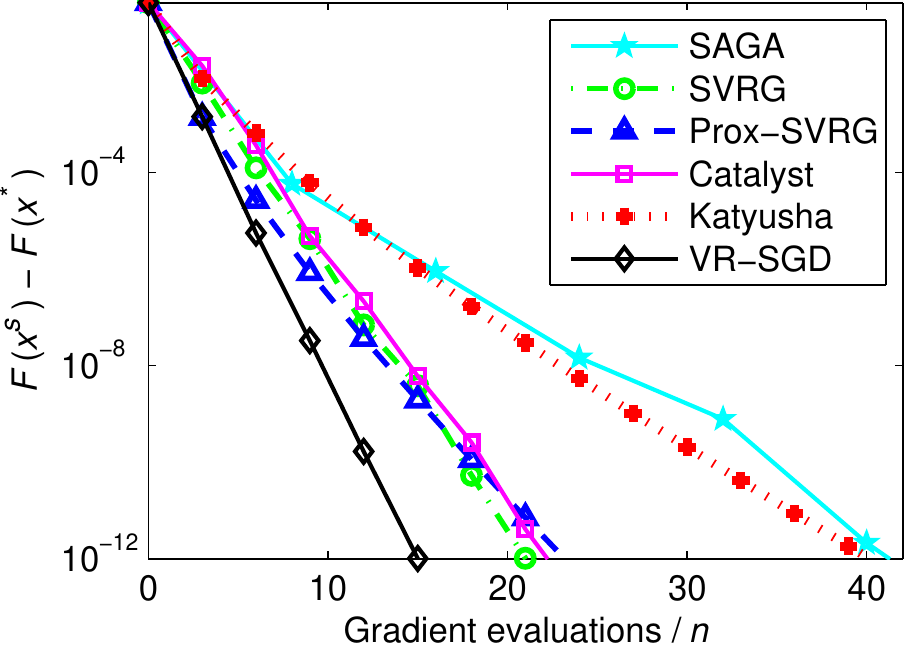}

\subfigure[Adult: $\lambda_{1}=10^{-4}$]{\includegraphics[width=0.245\columnwidth]{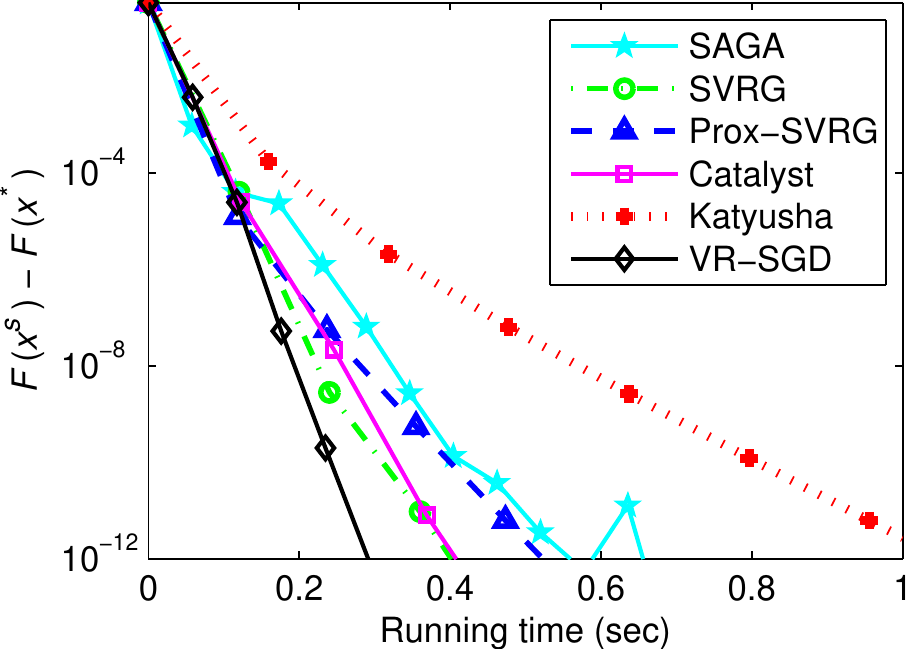}\label{figs1a}}
\subfigure[Covtype: $\lambda_{1}=10^{-4}$]{\includegraphics[width=0.245\columnwidth]{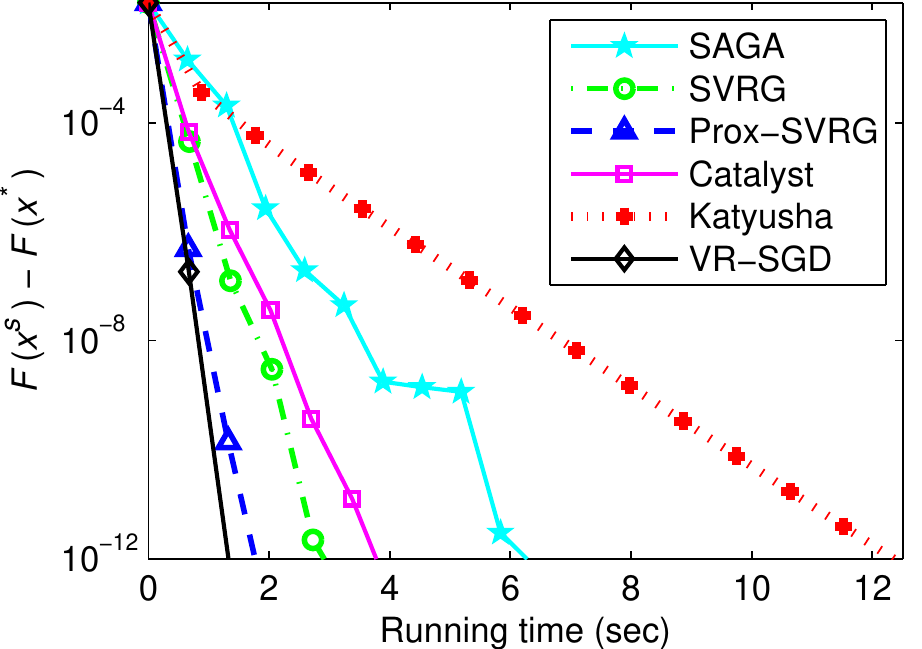}\label{figs1c}}
\subfigure[RCV1: $\lambda_{1}=10^{-4}$]{\includegraphics[width=0.245\columnwidth]{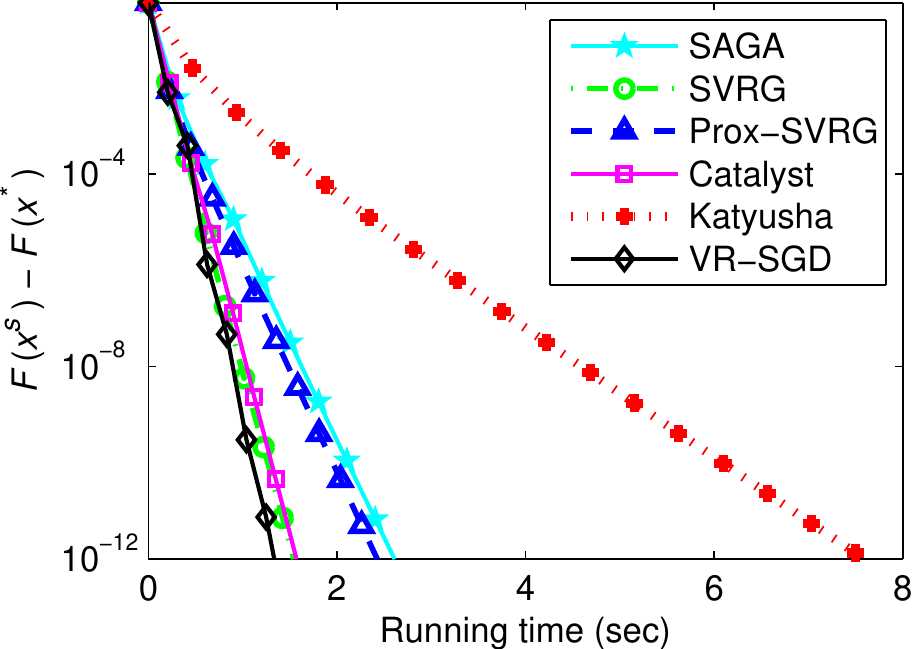}\label{figs1b}}
\subfigure[Epsilon: $\lambda_{1}=10^{-5}$]{\includegraphics[width=0.245\columnwidth]{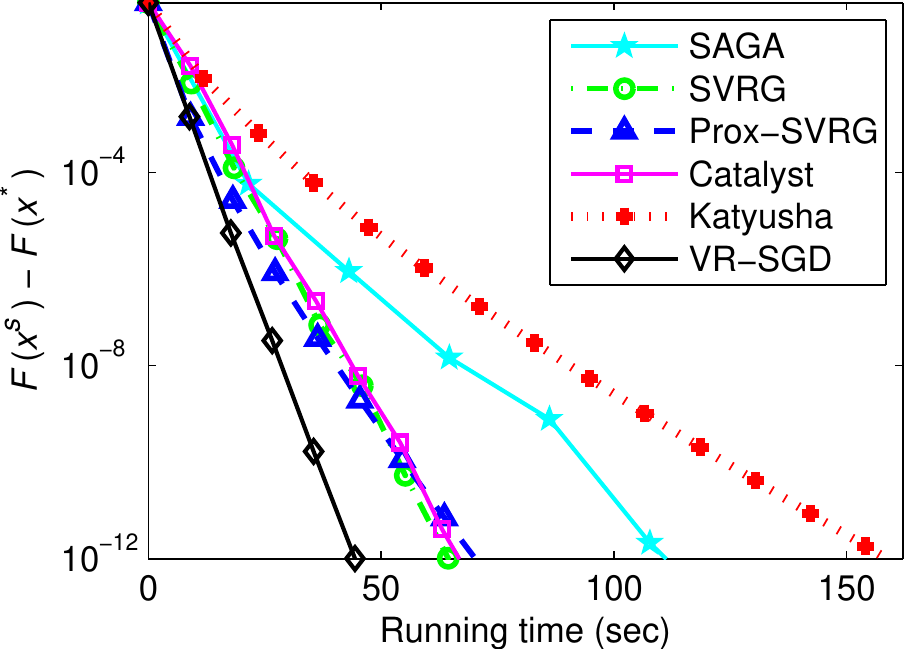}\label{figs1d}}
\vspace{1.6mm}

\includegraphics[width=0.245\columnwidth]{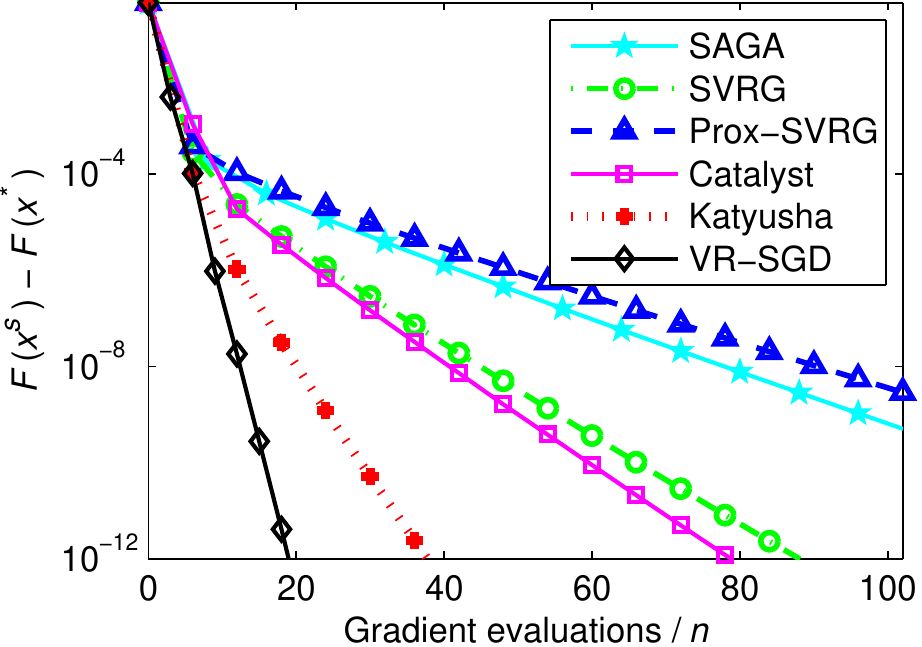}
\includegraphics[width=0.245\columnwidth]{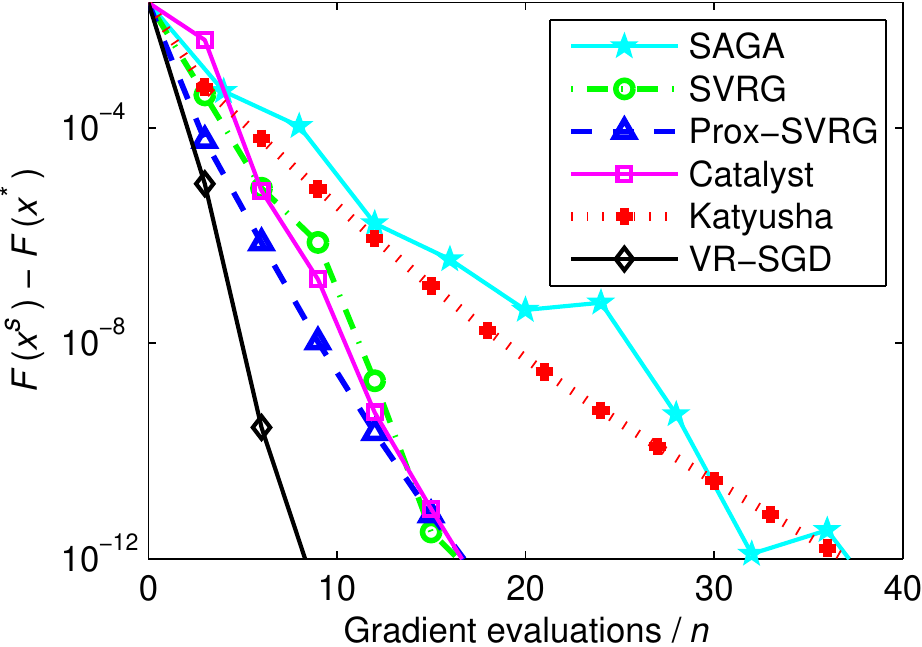}
\includegraphics[width=0.245\columnwidth]{Fig463}
\includegraphics[width=0.245\columnwidth]{Fig455}

\subfigure[Adult: $\lambda_{1}=10^{-5}$]{\includegraphics[width=0.245\columnwidth]{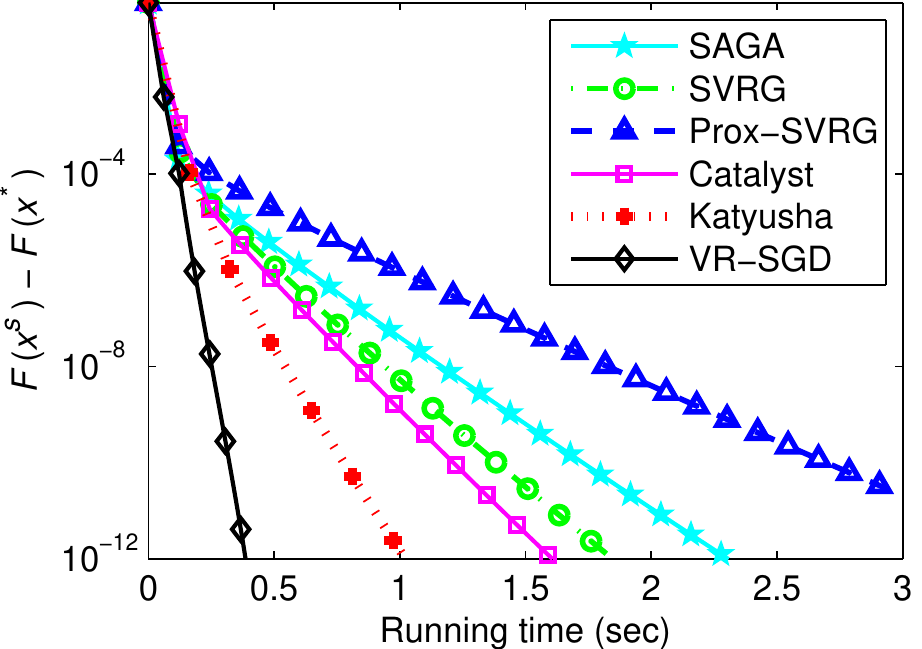}}
\subfigure[Covtype: $\lambda_{1}=10^{-5}$]{\includegraphics[width=0.245\columnwidth]{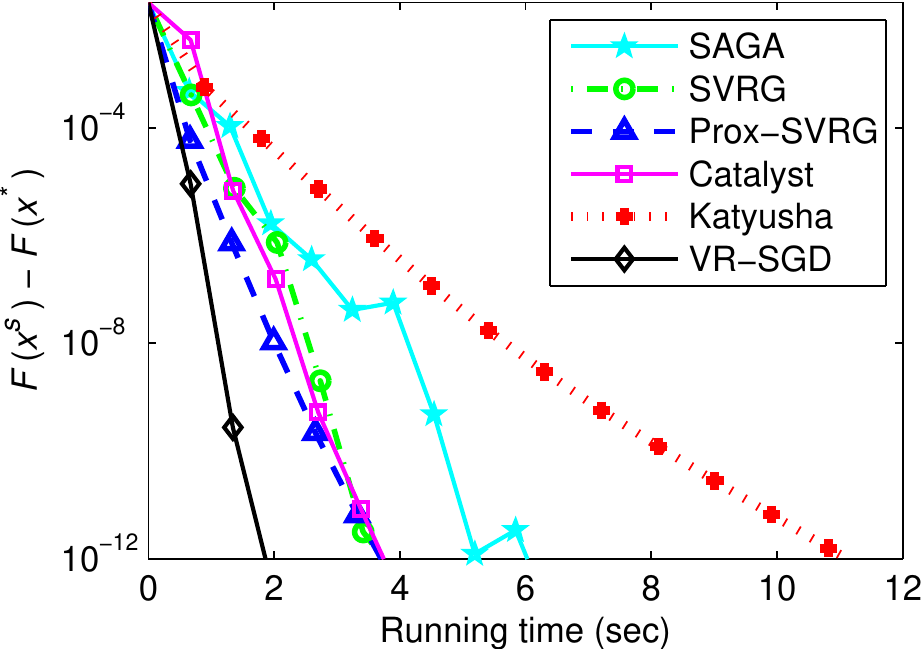}}
\subfigure[RCV1: $\lambda_{1}=10^{-5}$]{\includegraphics[width=0.245\columnwidth]{Fig464}}
\subfigure[Epsilon: $\lambda_{1}=10^{-6}$]{\includegraphics[width=0.245\columnwidth]{Fig456}\label{figs1h}}
\vspace{1.6mm}

\includegraphics[width=0.245\columnwidth]{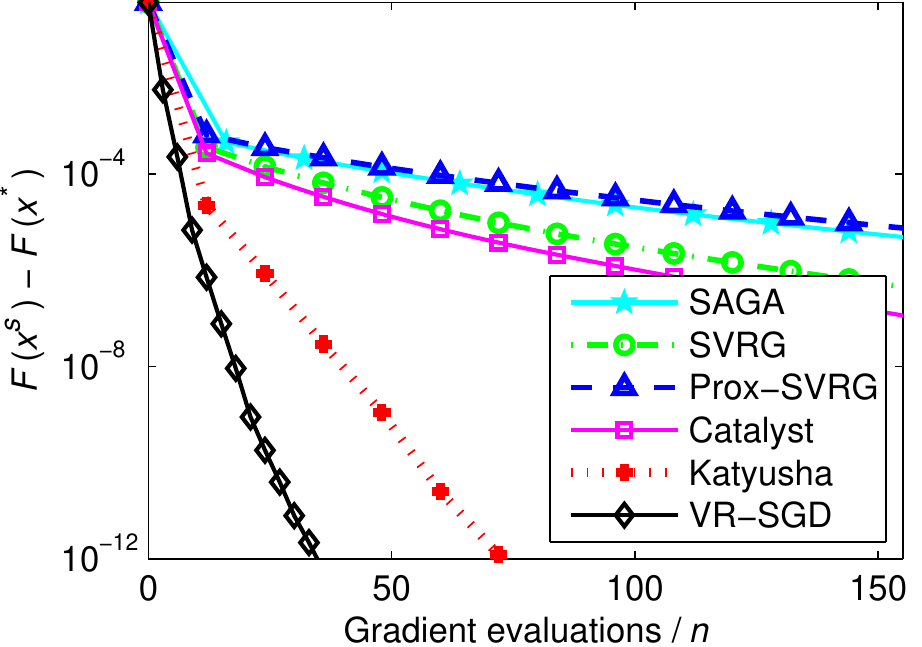}
\includegraphics[width=0.245\columnwidth]{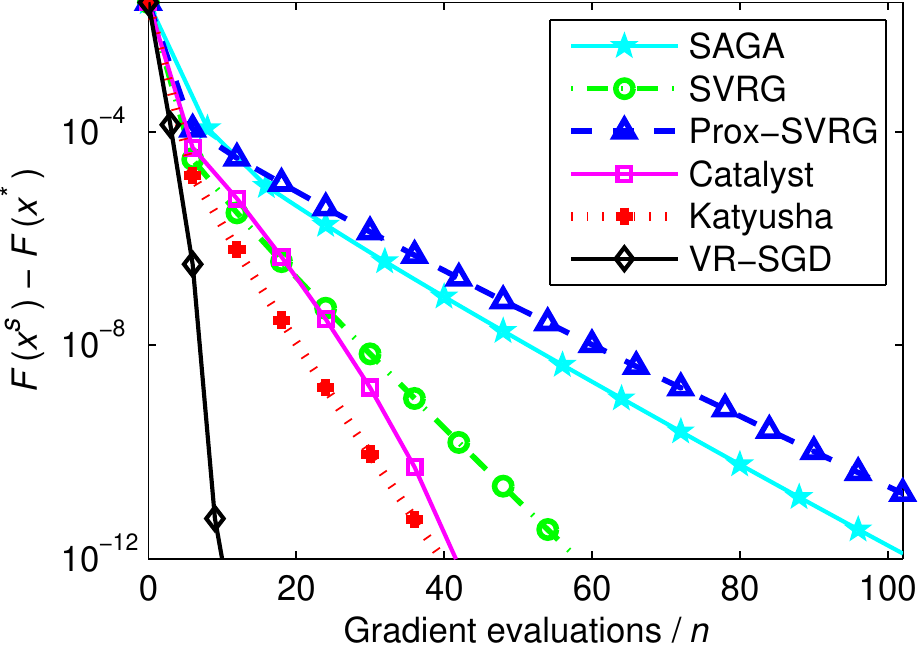}
\includegraphics[width=0.245\columnwidth]{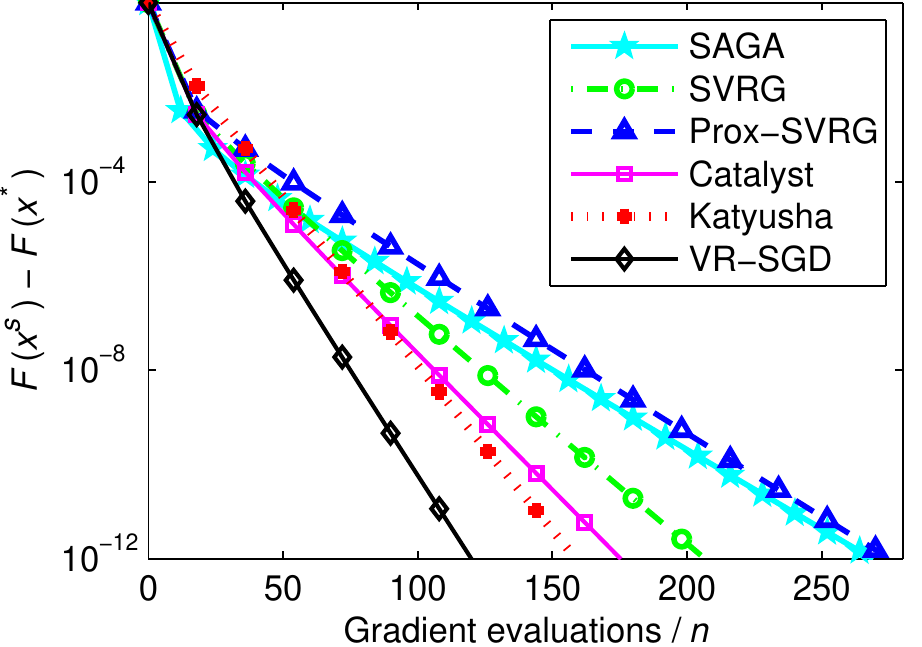}
\includegraphics[width=0.245\columnwidth]{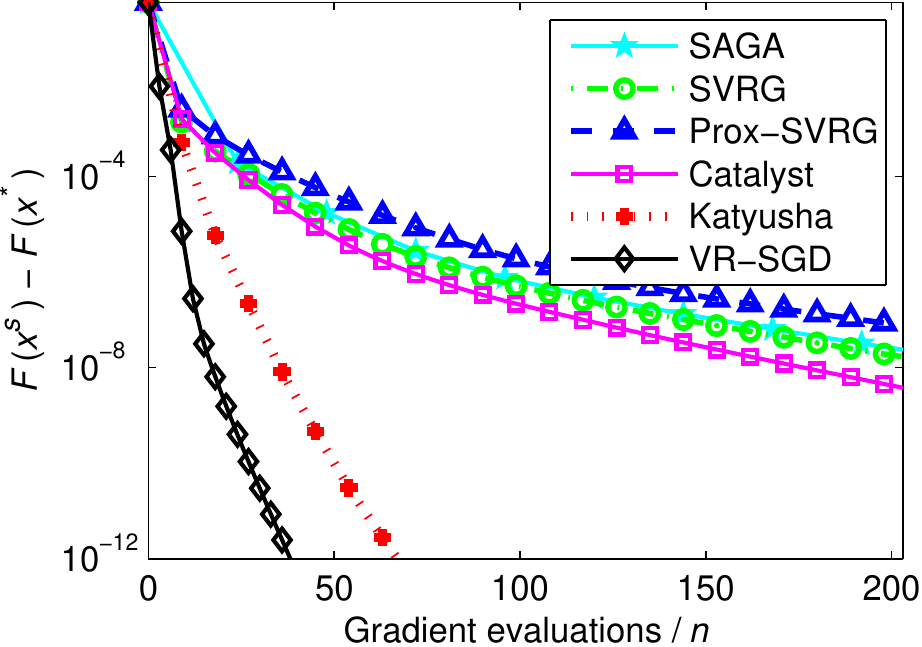}

\subfigure[Adult: $\lambda_{1}=10^{-6}$]{\includegraphics[width=0.245\columnwidth]{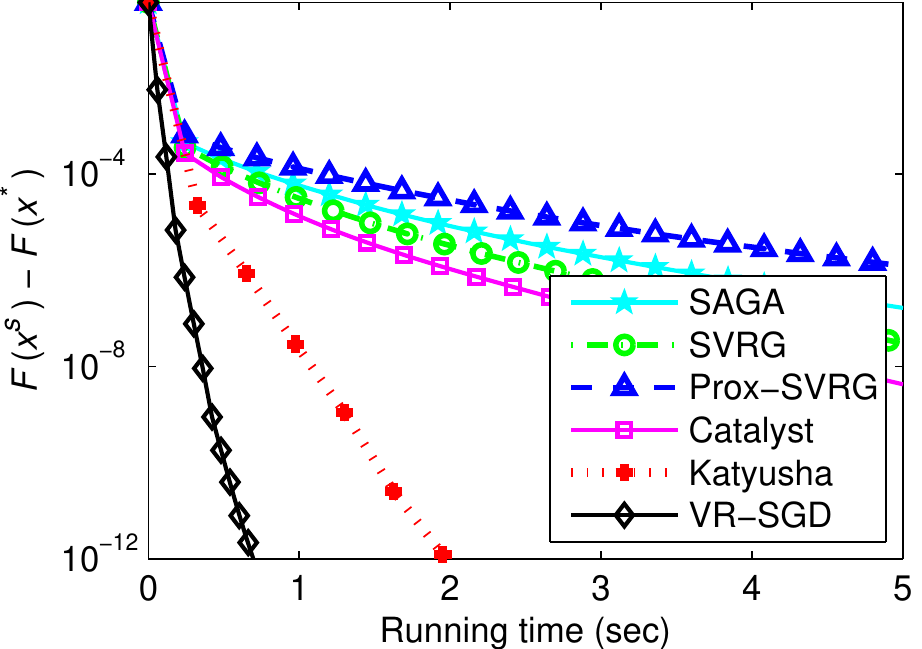}\label{figs1i}}
\subfigure[Covtype: $\lambda_{1}=10^{-6}$]{\includegraphics[width=0.245\columnwidth]{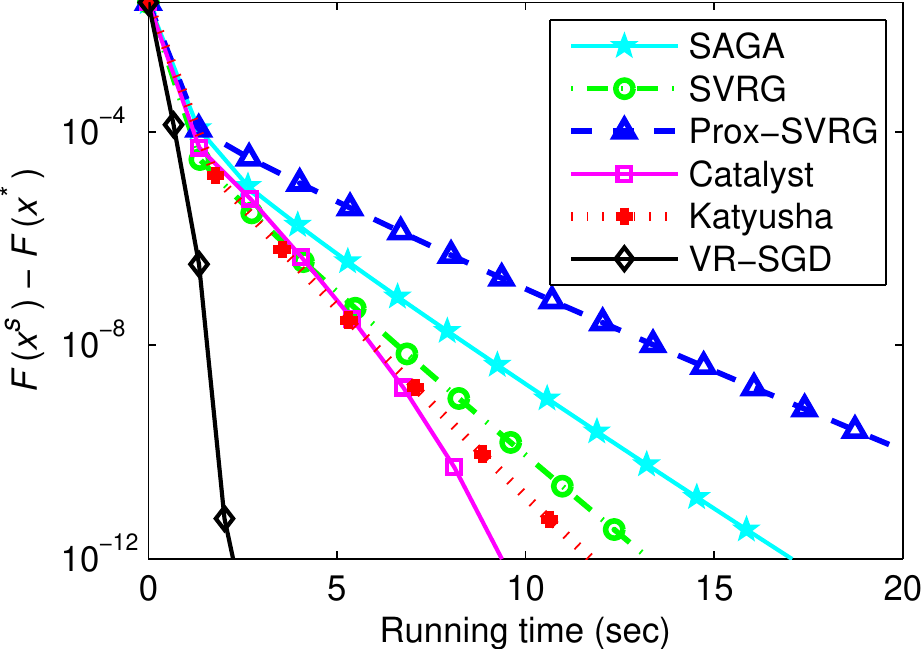}}
\subfigure[RCV1: $\lambda_{1}=10^{-6}$]{\includegraphics[width=0.245\columnwidth]{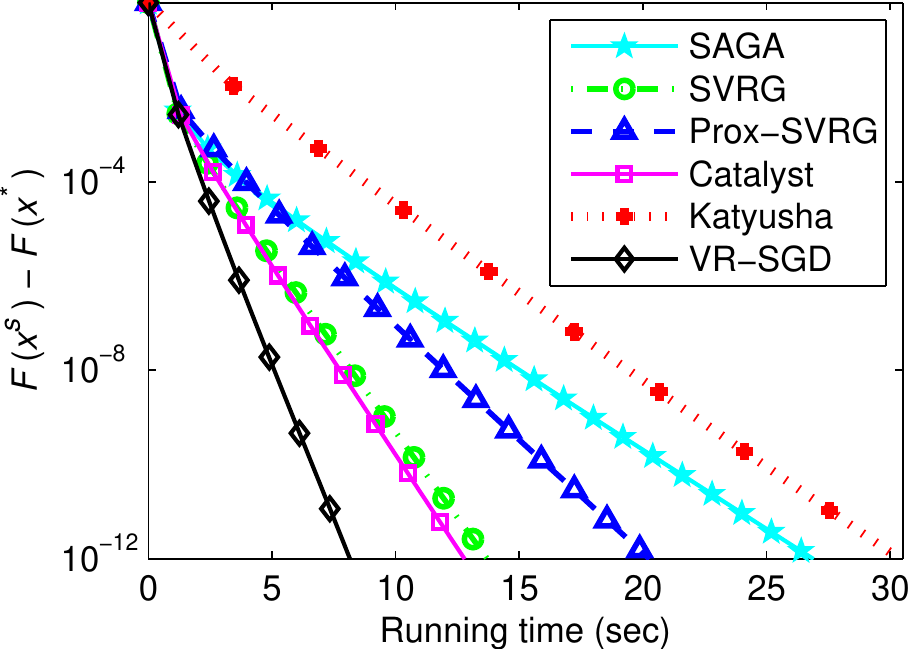}}
\subfigure[Epsilon: $\lambda_{1}=10^{-7}$]{\includegraphics[width=0.245\columnwidth]{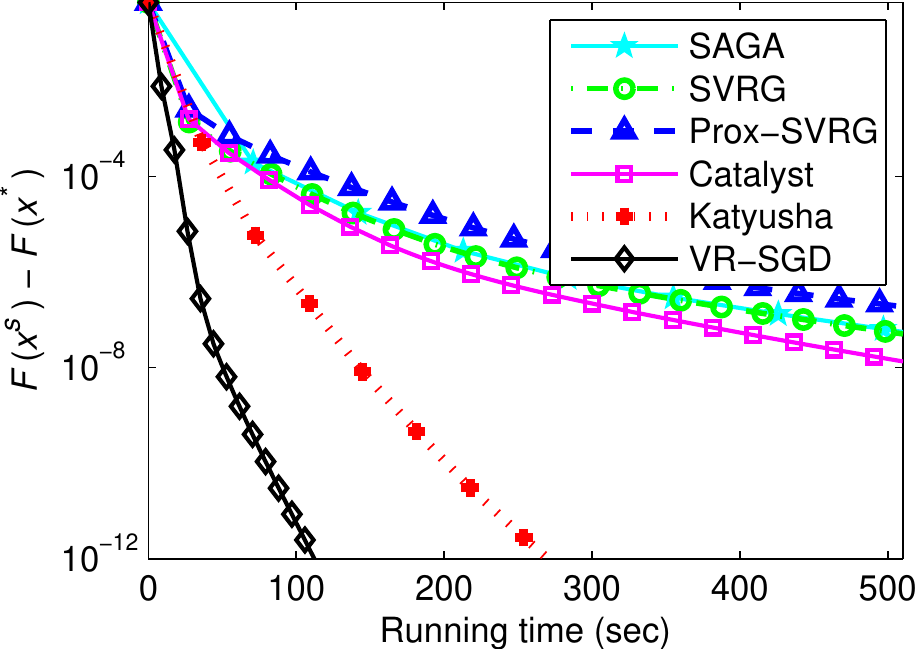}\label{figs1l}}
\caption{Comparison of SAGA~\cite{defazio:saga}, SVRG~\cite{johnson:svrg}, Prox-SVRG~\cite{xiao:prox-svrg}, Catalyst~\cite{lin:vrsg}, Katyusha~\cite{zhu:Katyusha}, and our VR-SGD method for solving $\ell_{2}$-norm regularized logistic regression problems (i.e., $\lambda_{2}=0$). In each plot, the vertical axis shows the objective value minus the minimum, and the horizontal axis is the number of effective passes (top) or running time (bottom).}
\label{figs17}
\end{figure}

\begin{figure}[th]
\centering
\includegraphics[width=0.245\columnwidth]{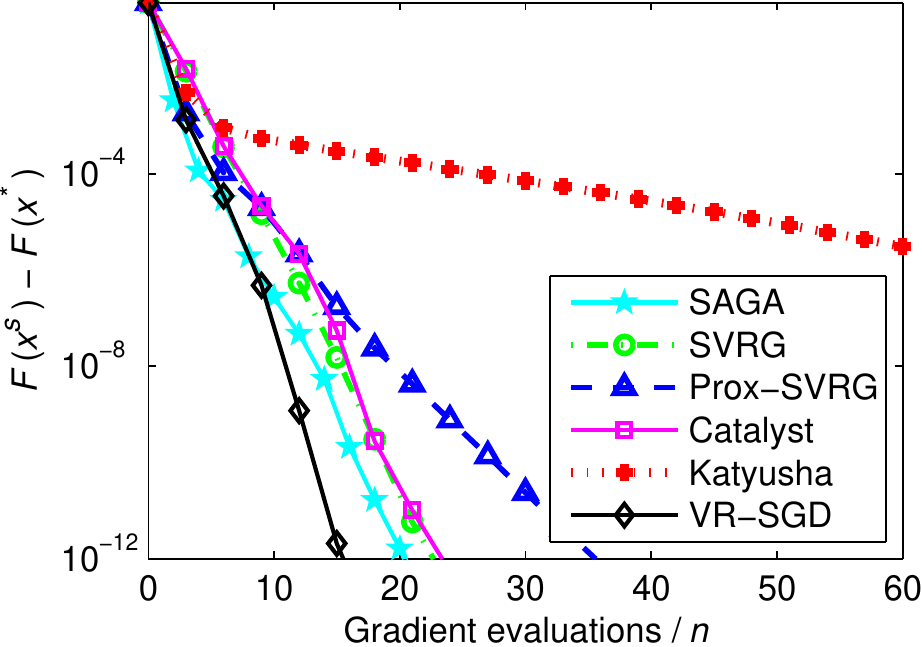}
\includegraphics[width=0.245\columnwidth]{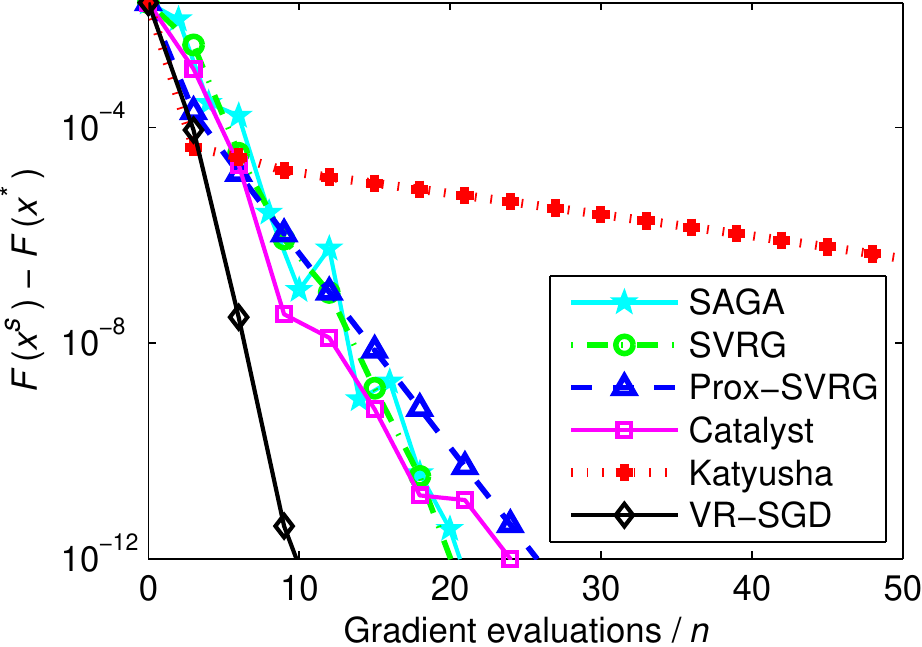}
\includegraphics[width=0.245\columnwidth]{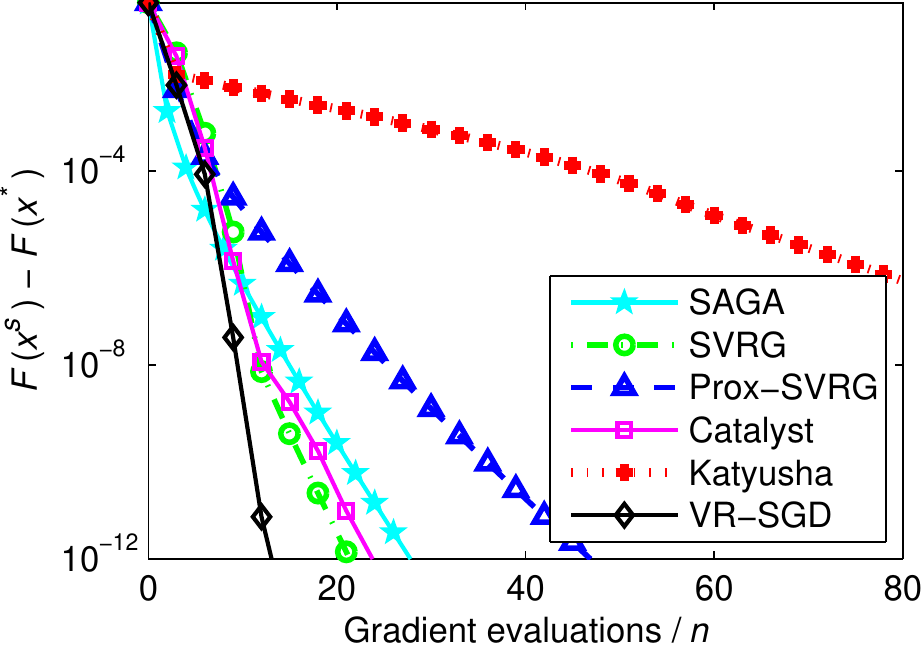}
\includegraphics[width=0.245\columnwidth]{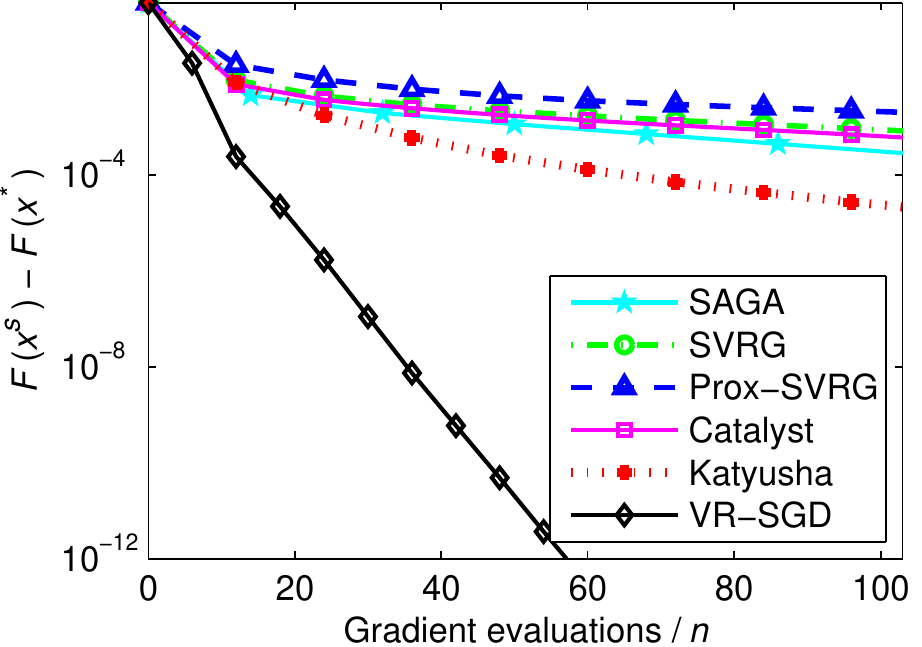}

\subfigure[$\lambda_{2}=10^{-4}$]{\includegraphics[width=0.245\columnwidth]{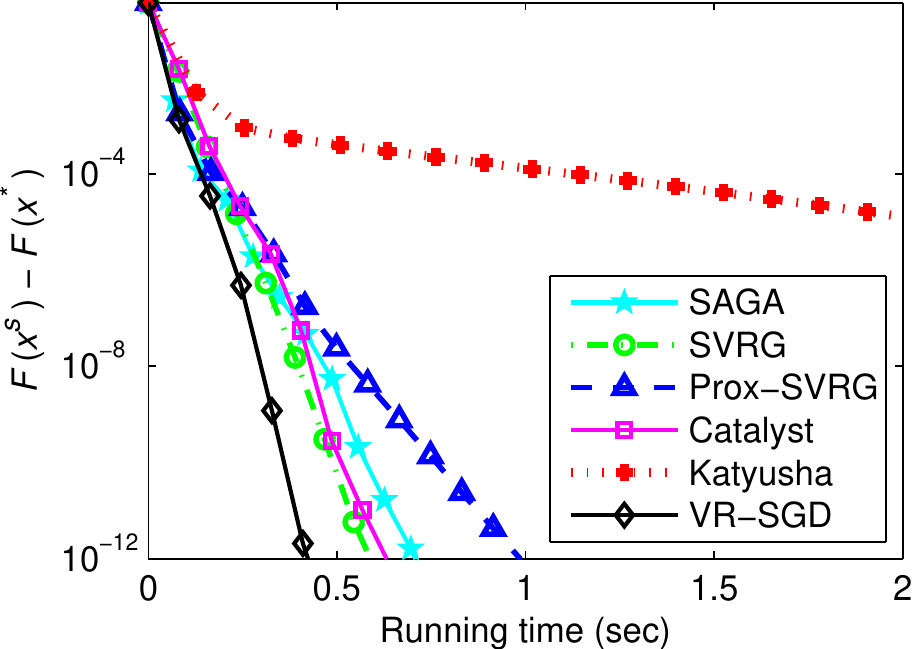}\:\includegraphics[width=0.245\columnwidth]{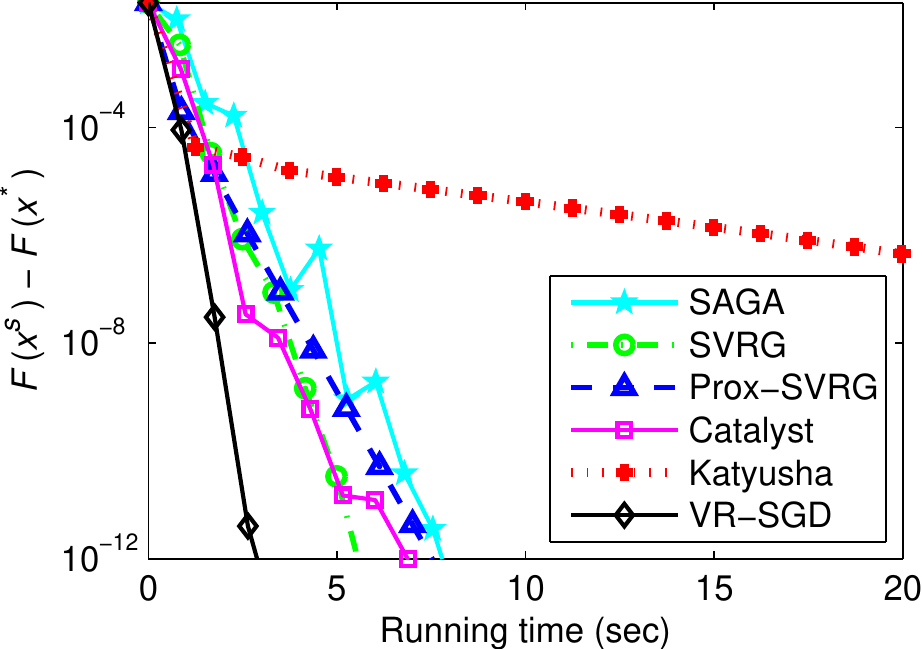}\:\includegraphics[width=0.245\columnwidth]{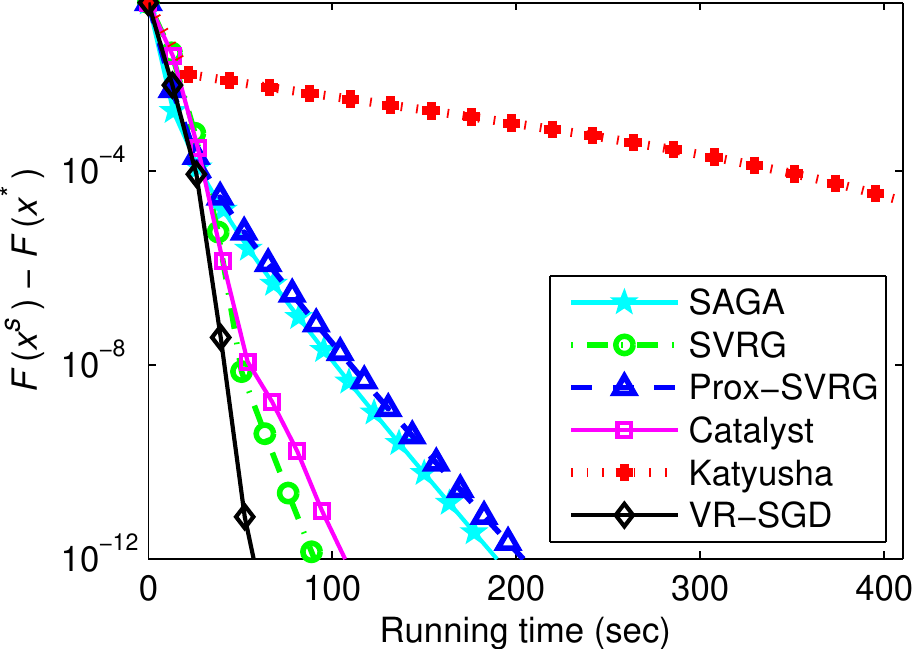}\label{figs2a}}
\subfigure[$\lambda_{2}=10^{-4}$]{\includegraphics[width=0.245\columnwidth]{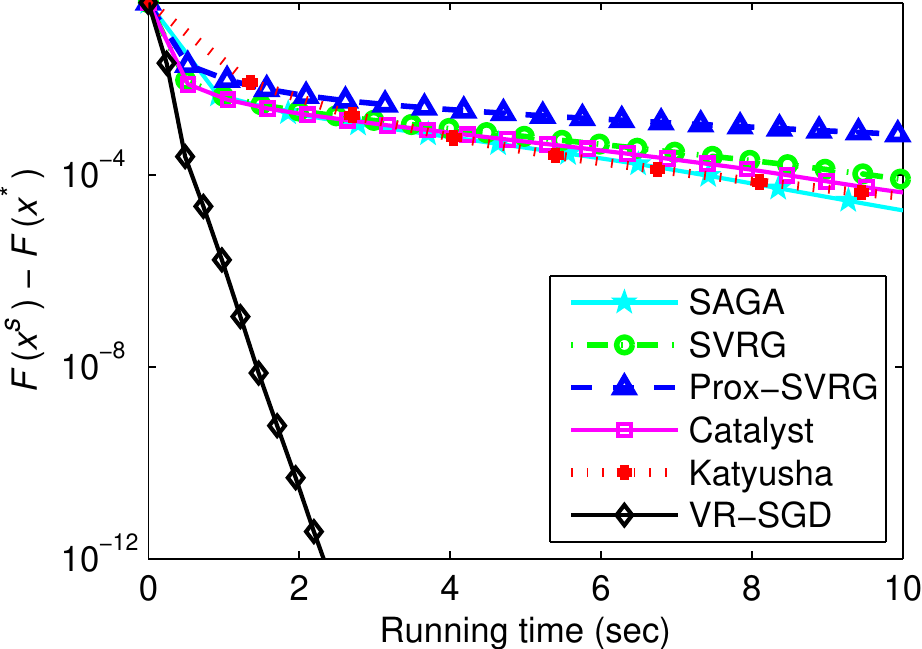}\label{figs2b}}
\vspace{1.6mm}

\includegraphics[width=0.245\columnwidth]{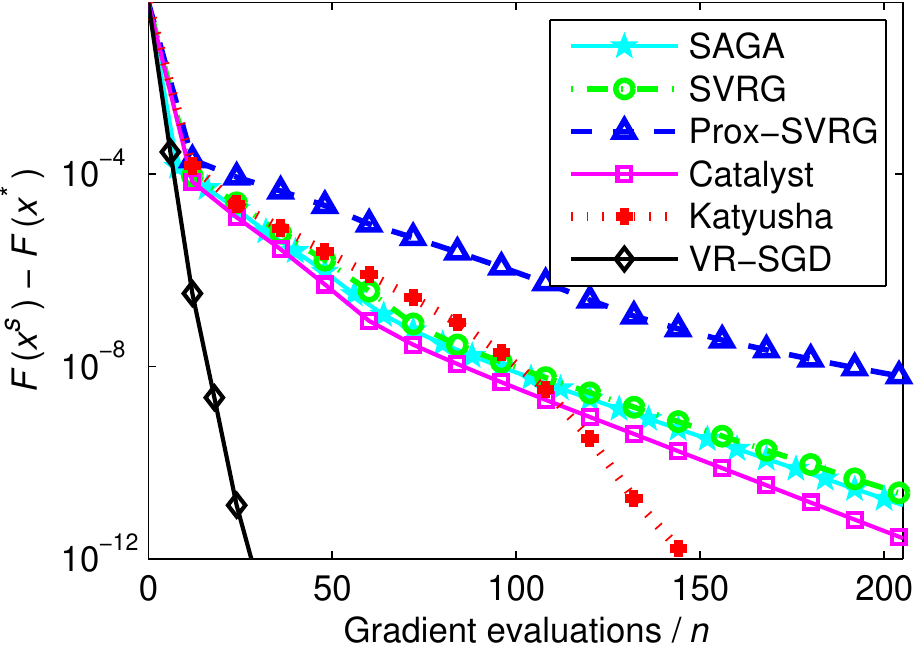}
\includegraphics[width=0.245\columnwidth]{Fig533}
\includegraphics[width=0.245\columnwidth]{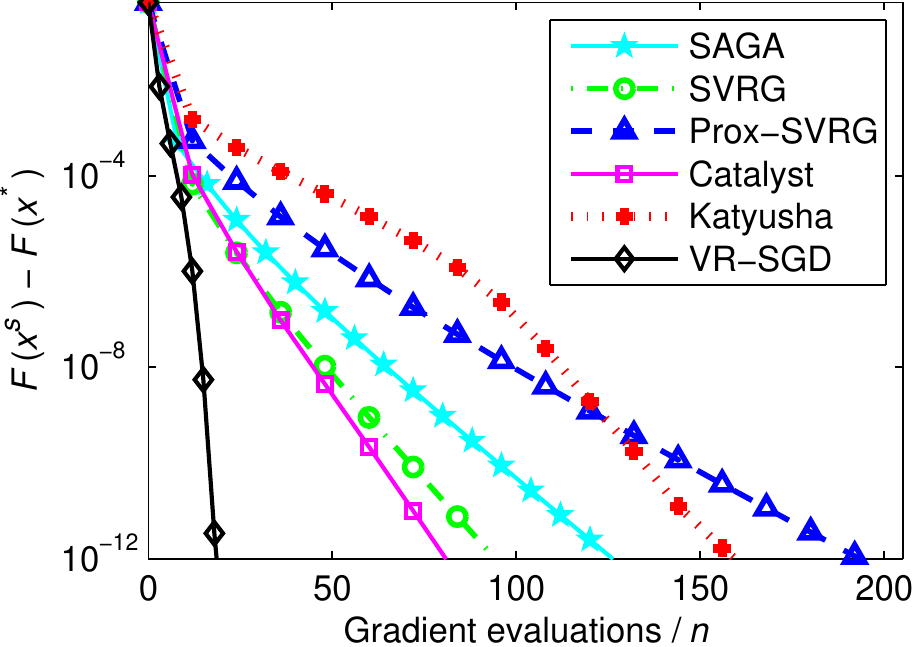}
\includegraphics[width=0.245\columnwidth]{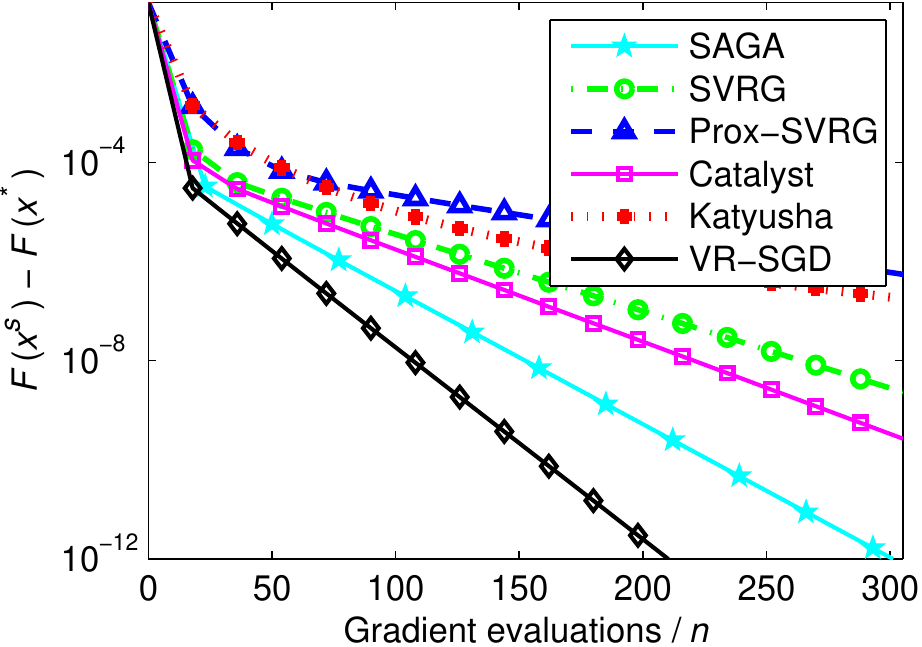}

\subfigure[$\lambda_{2}=10^{-5}$]{\includegraphics[width=0.245\columnwidth]{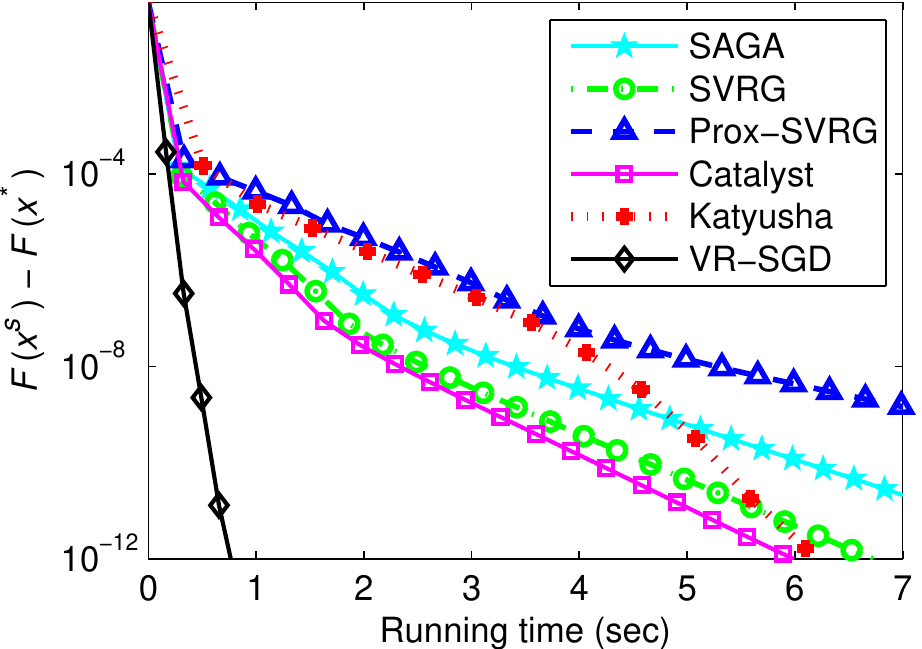}\:\includegraphics[width=0.245\columnwidth]{Fig534}\:\includegraphics[width=0.245\columnwidth]{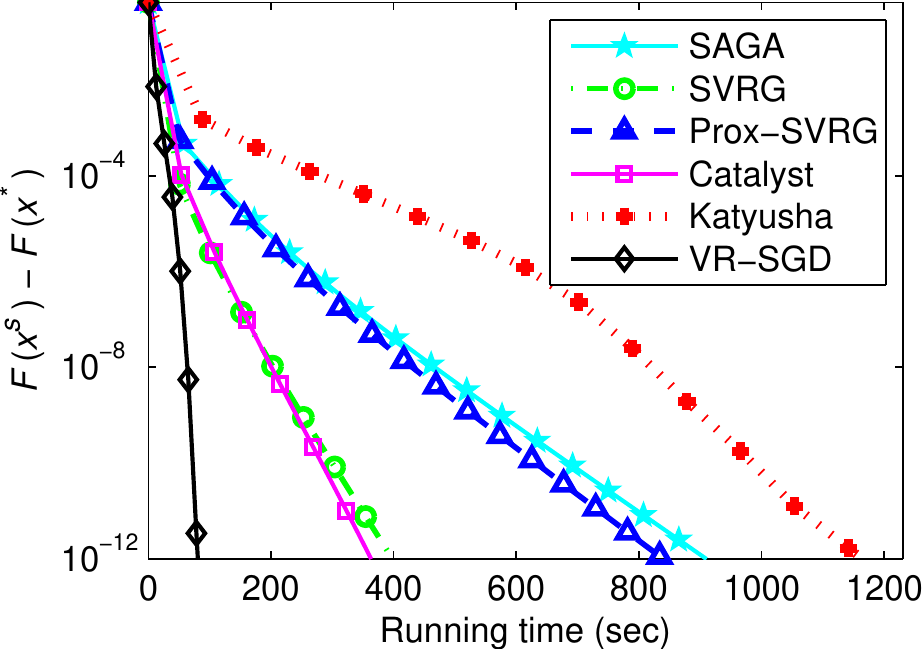}}
\subfigure[$\lambda_{2}=5\!*\!10^{-5}$]{\includegraphics[width=0.245\columnwidth]{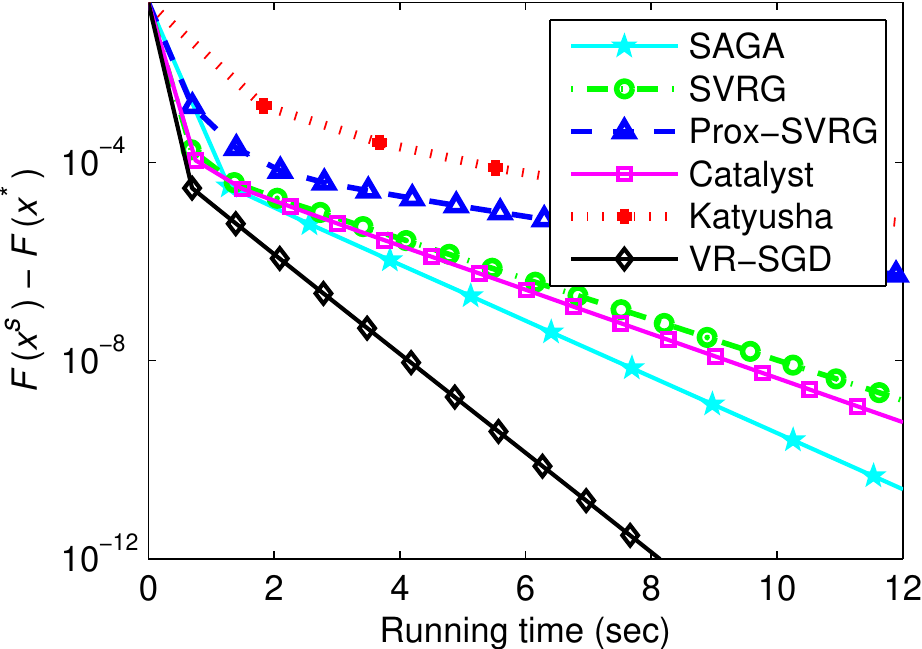}\label{figs2d}}
\vspace{1.6mm}

\includegraphics[width=0.245\columnwidth]{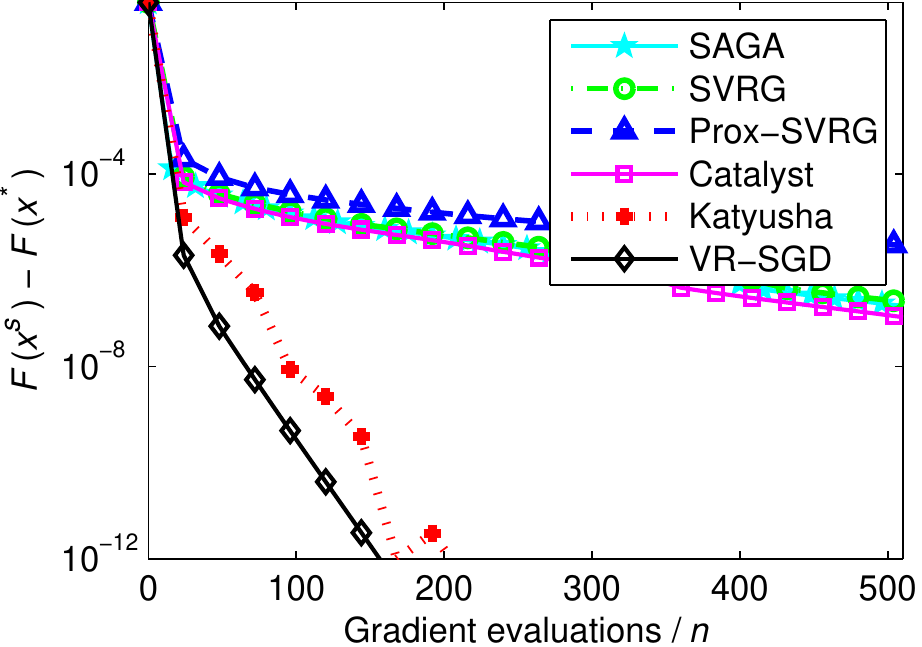}
\includegraphics[width=0.245\columnwidth]{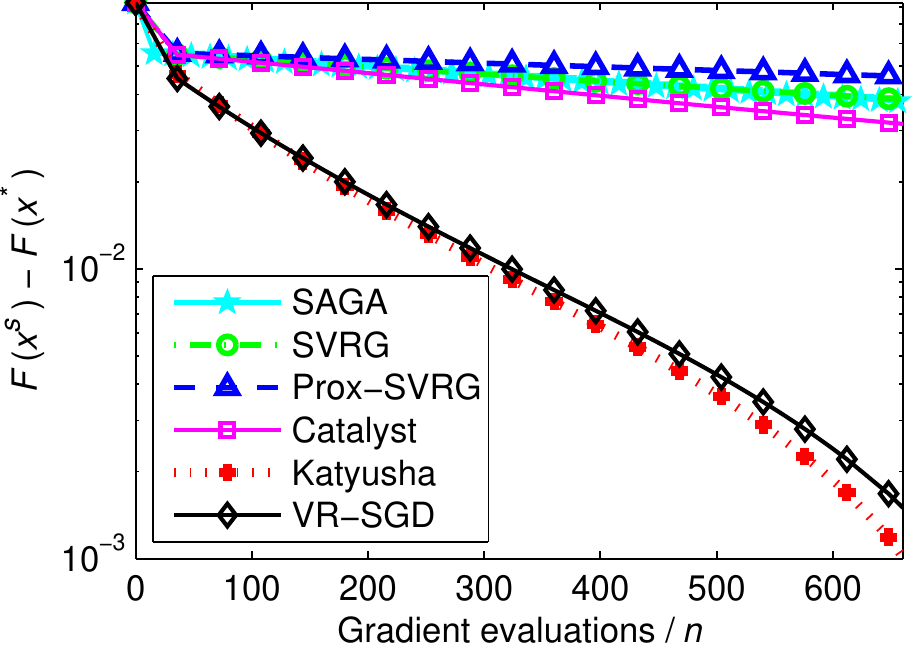}
\includegraphics[width=0.245\columnwidth]{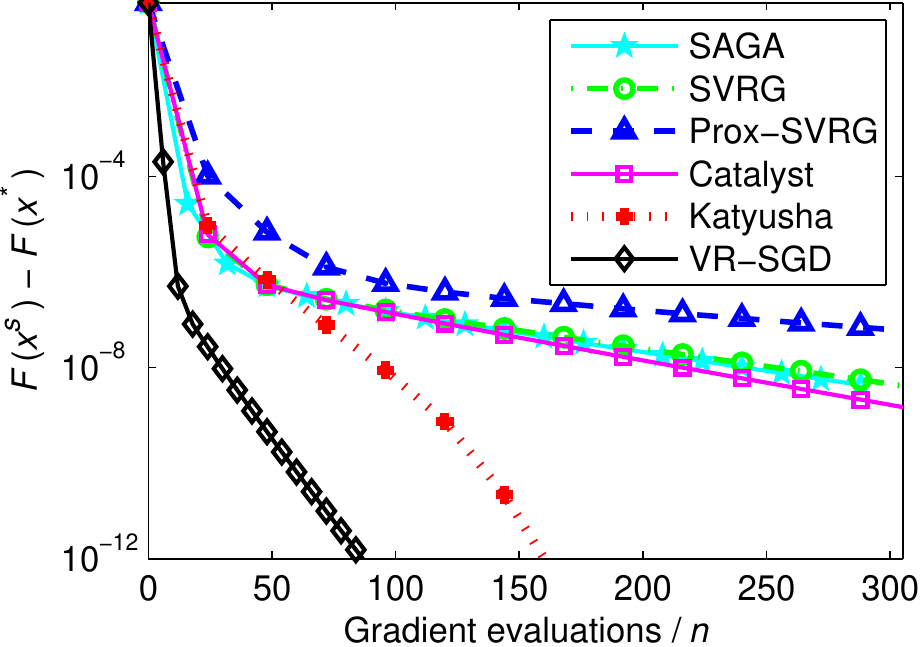}
\includegraphics[width=0.245\columnwidth]{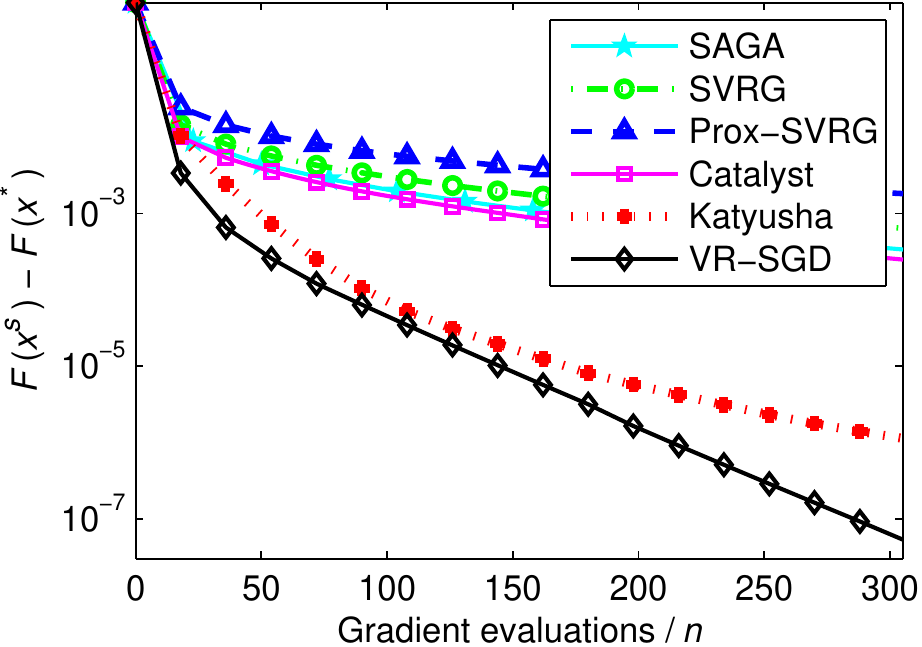}

\subfigure[$\lambda_{2}=10^{-6}$]{\includegraphics[width=0.245\columnwidth]{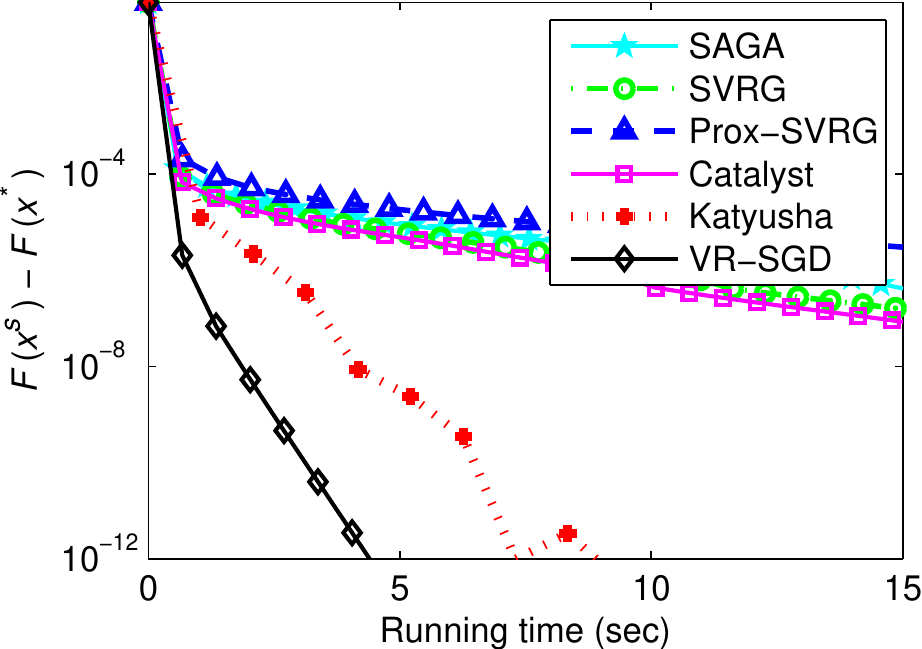}\,\includegraphics[width=0.245\columnwidth]{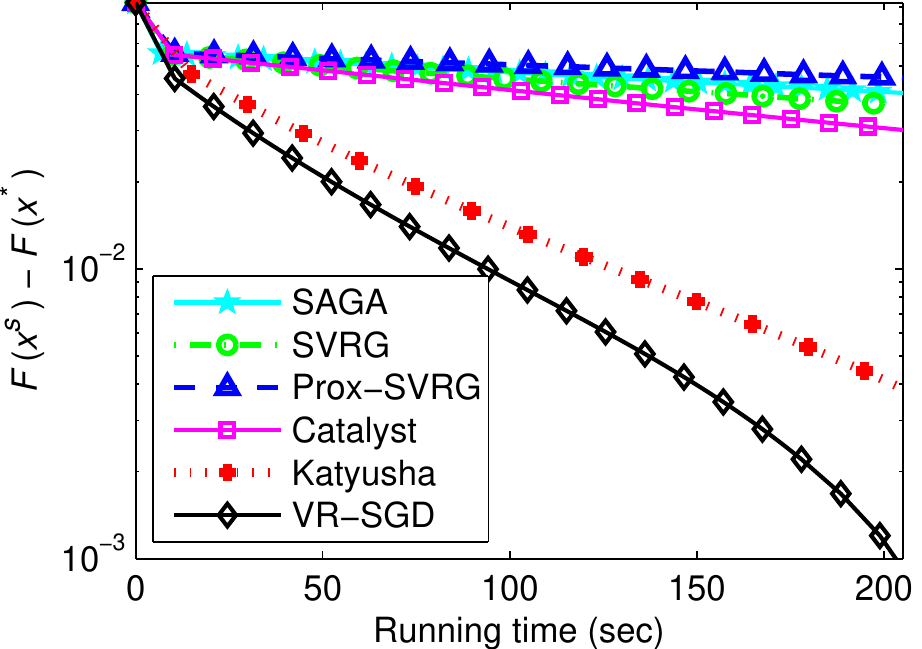}\,\includegraphics[width=0.245\columnwidth]{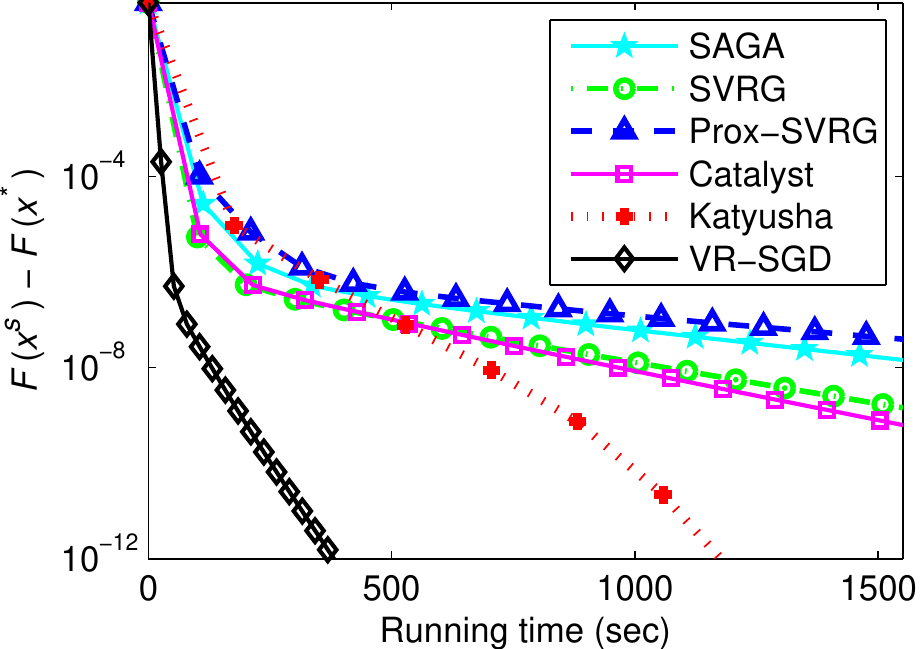}\label{figs2e}}
\subfigure[$\lambda_{2}=10^{-5}$]{\includegraphics[width=0.245\columnwidth]{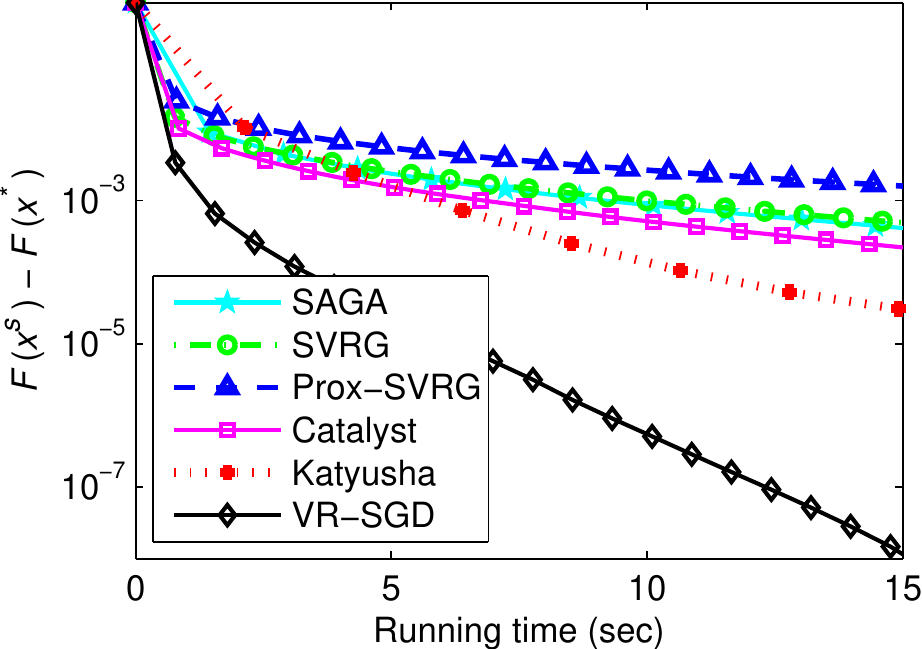}\label{figs2f}}
\caption{Comparison of SAGA~\cite{defazio:saga}, SVRG, Prox-SVRG~\cite{xiao:prox-svrg}, Catalyst~\cite{lin:vrsg}, Katyusha~\cite{zhu:Katyusha}, and our VR-SGD method for $\ell_{1}$-norm regularized logistic regression problems (i.e., $\lambda_{1}=0$) on the four data sets: Adult (the first column), Covtype (the sconced column), Epsilon (the third column), and RCV1 (the last column). In each plot, the vertical axis shows the objective value minus the minimum, and the horizontal axis is the number of effective passes (top) or running time (bottom).}
\label{figs18}
\end{figure}

\begin{figure}[th]
\centering
\includegraphics[width=0.245\columnwidth]{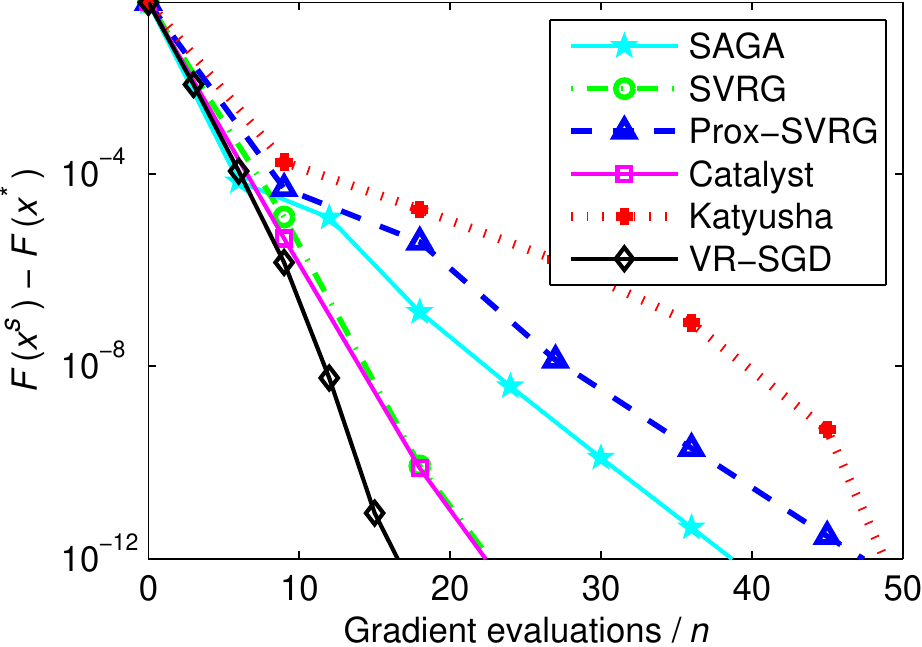}
\includegraphics[width=0.245\columnwidth]{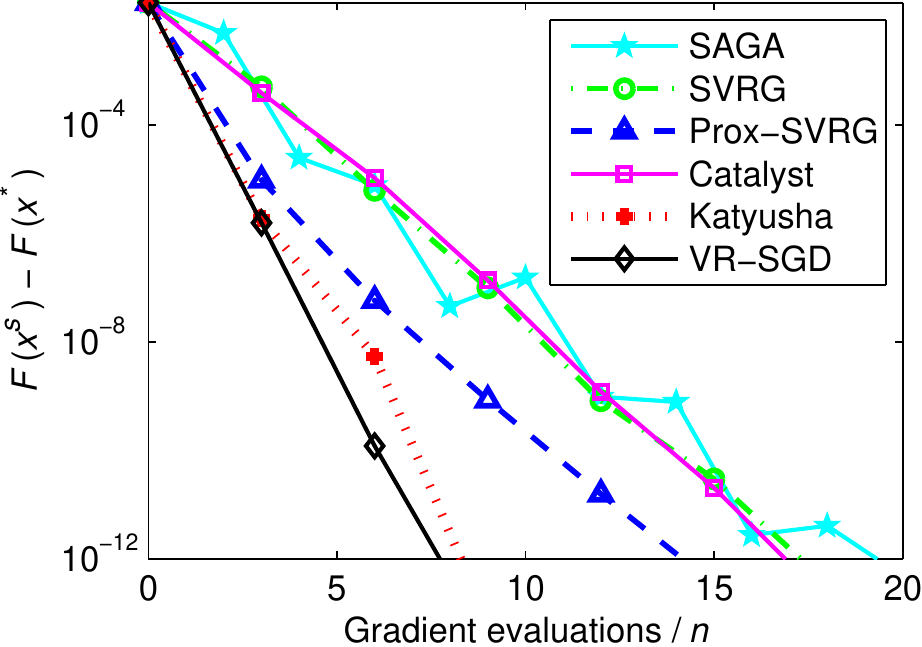}
\includegraphics[width=0.245\columnwidth]{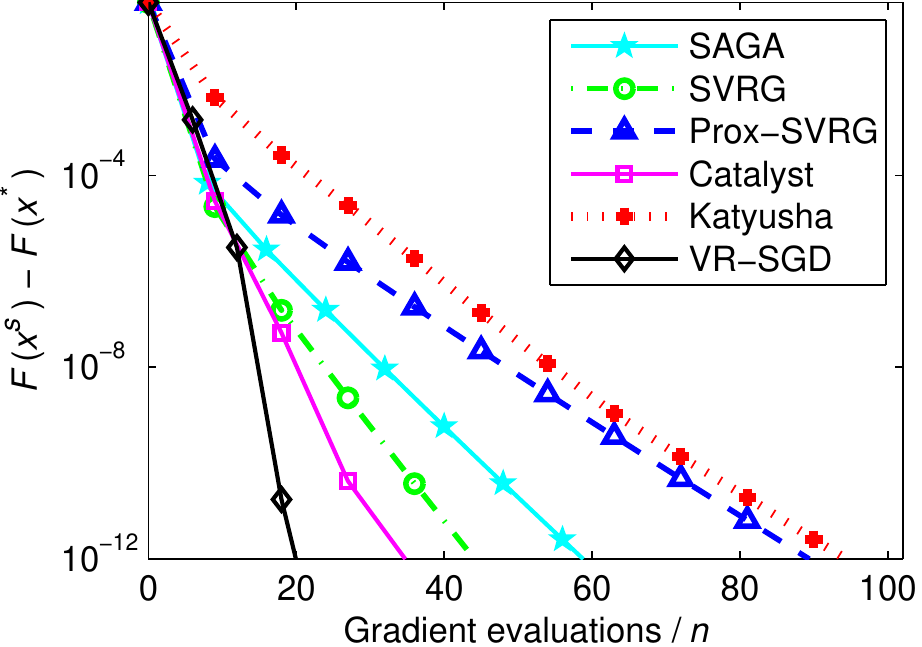}
\includegraphics[width=0.245\columnwidth]{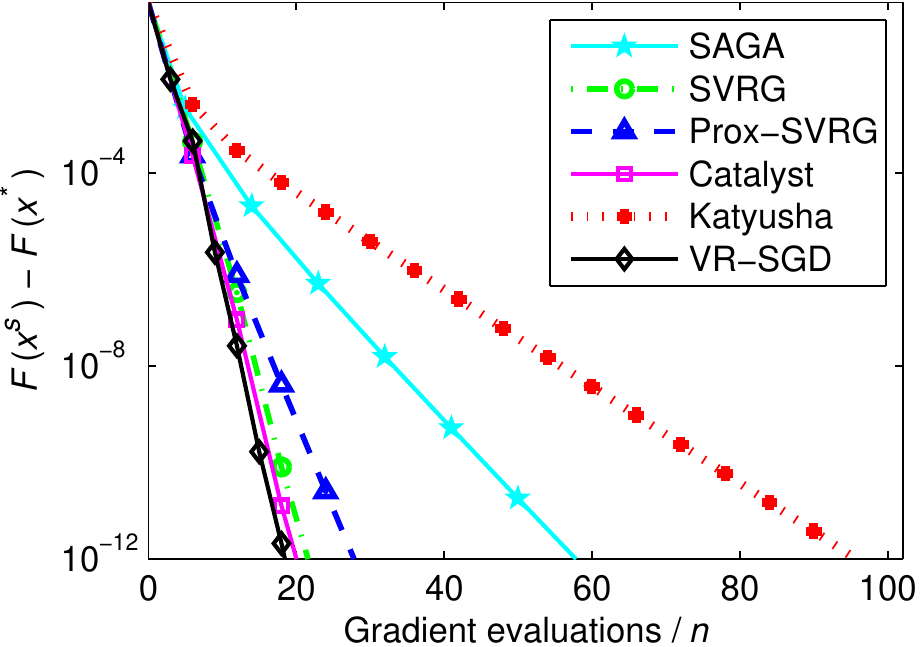}

\subfigure[$\lambda_{1}=10^{-5}$ \;and\; $\lambda_{2}=10^{-4}$]{\includegraphics[width=0.245\columnwidth]{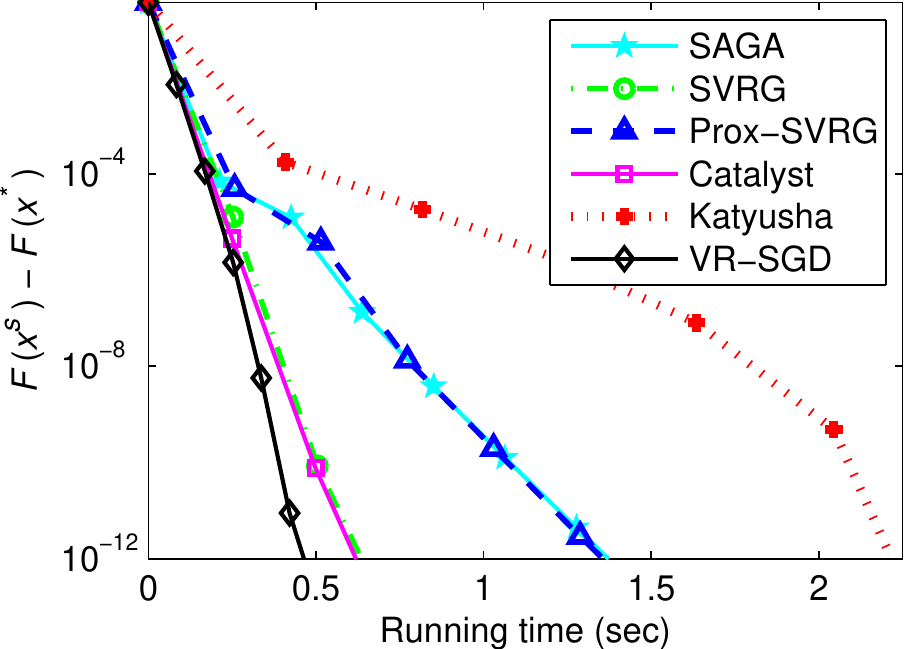}\:\includegraphics[width=0.245\columnwidth]{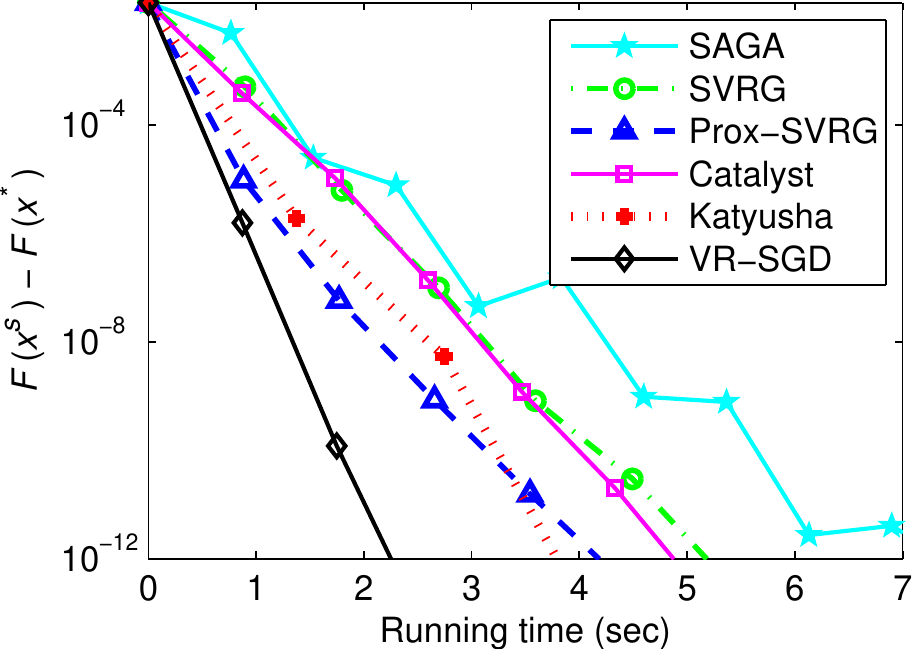}\:\includegraphics[width=0.245\columnwidth]{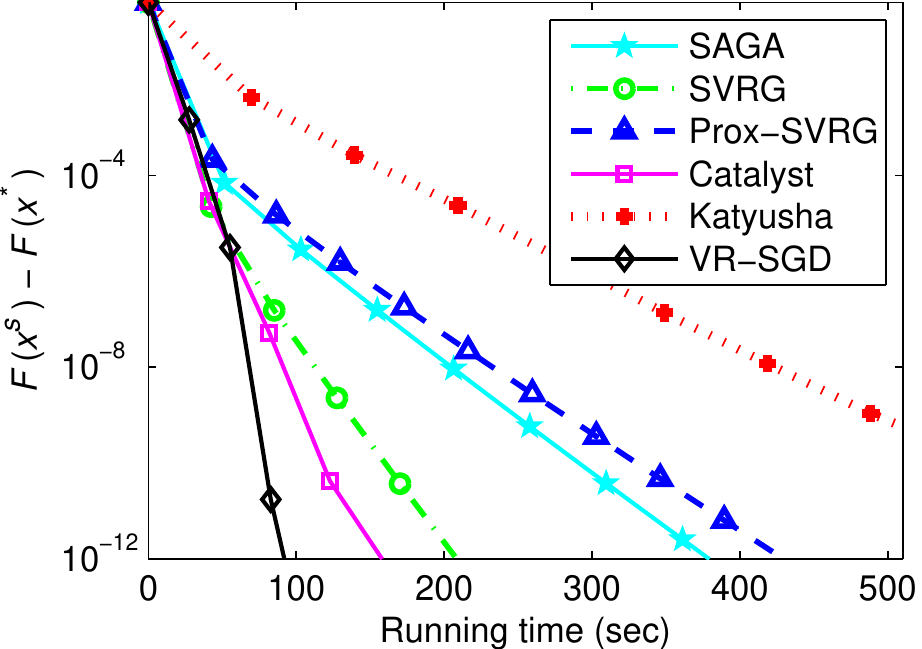}}\:\subfigure[$\lambda_{1}\!=\!10^{-4}$ and $\lambda_{2}\!=\!10^{-5}$]{\includegraphics[width=0.245\columnwidth]{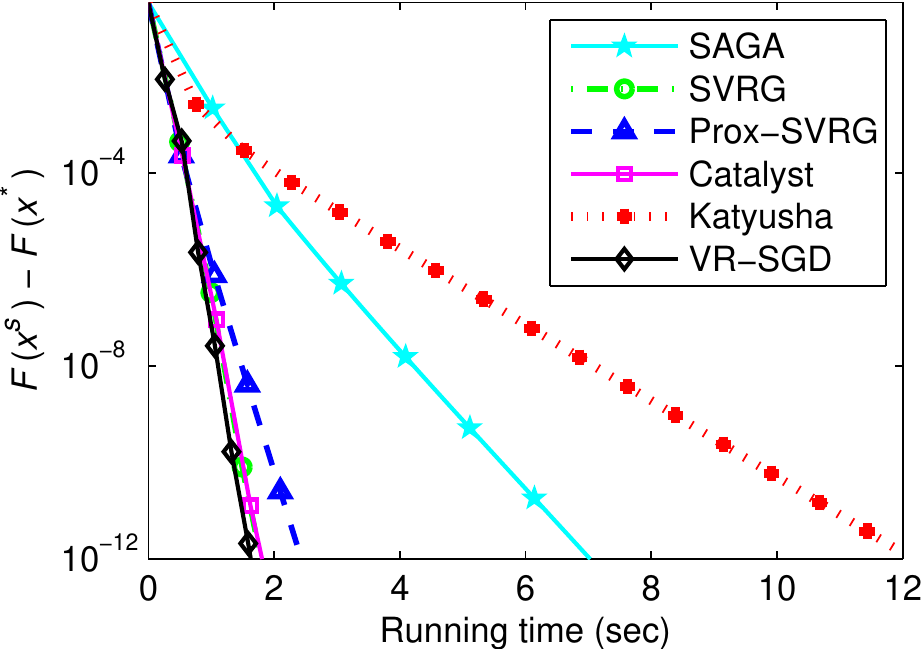}}
\vspace{1.6mm}

\includegraphics[width=0.245\columnwidth]{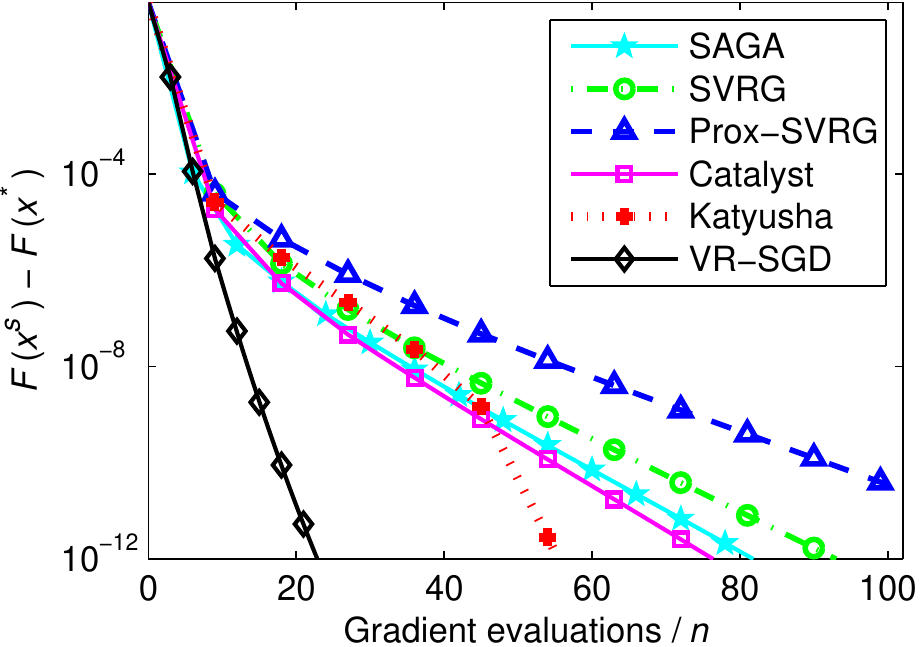}
\includegraphics[width=0.245\columnwidth]{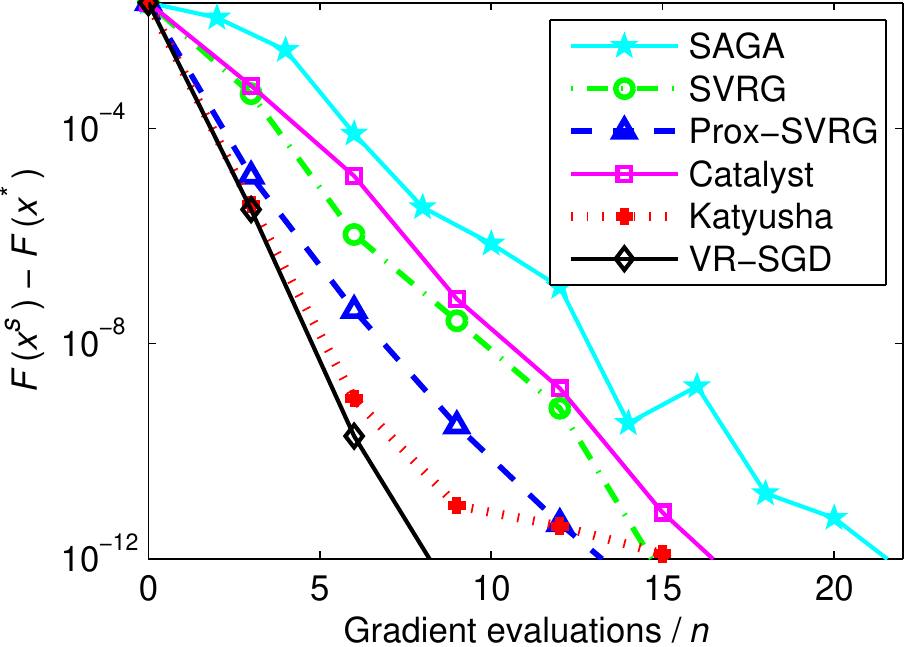}
\includegraphics[width=0.245\columnwidth]{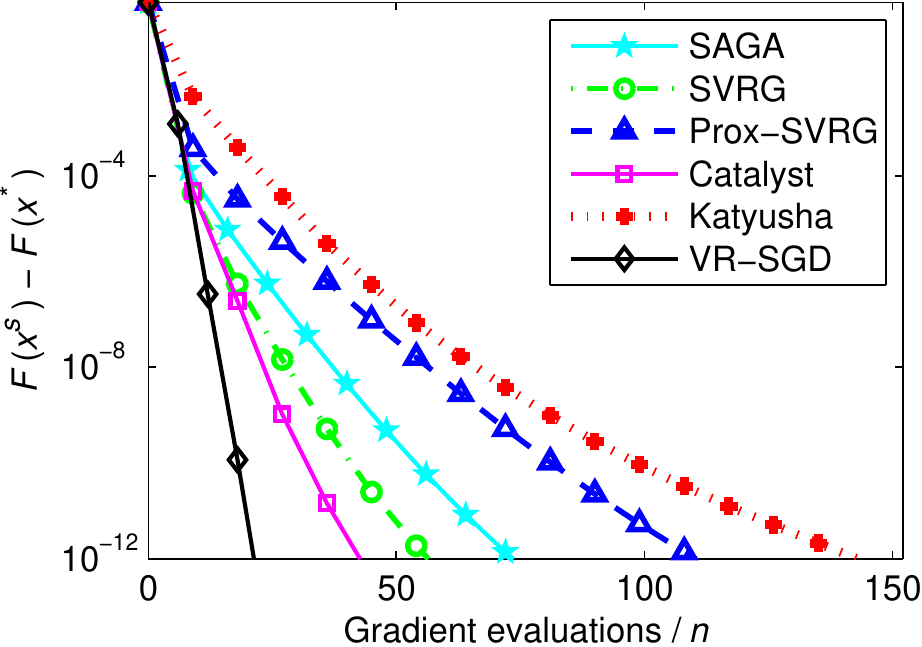}
\includegraphics[width=0.245\columnwidth]{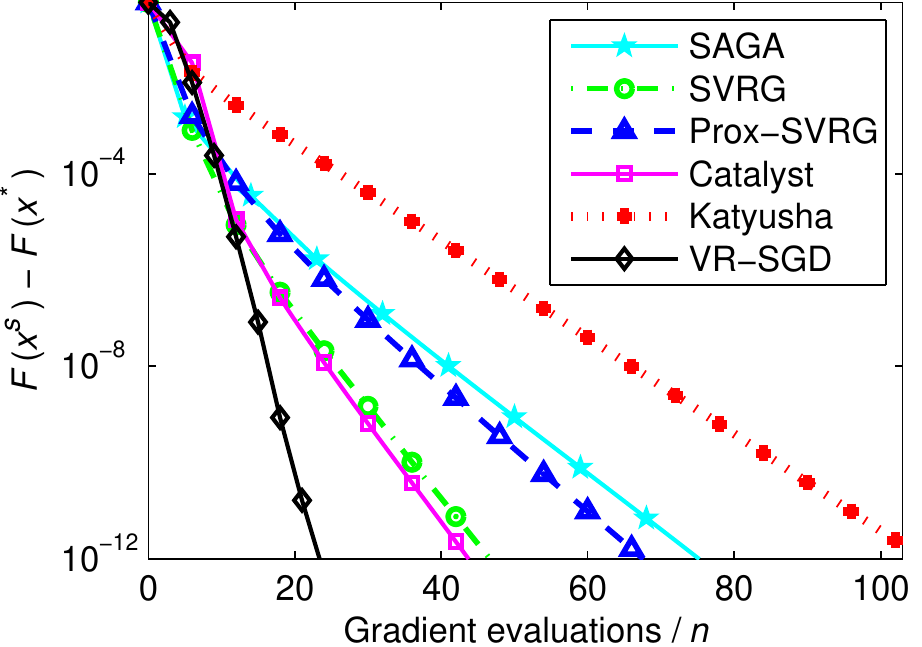}

\subfigure[$\lambda_{1}=10^{-5}$ \;and\; $\lambda_{2}=10^{-5}$]{\includegraphics[width=0.245\columnwidth]{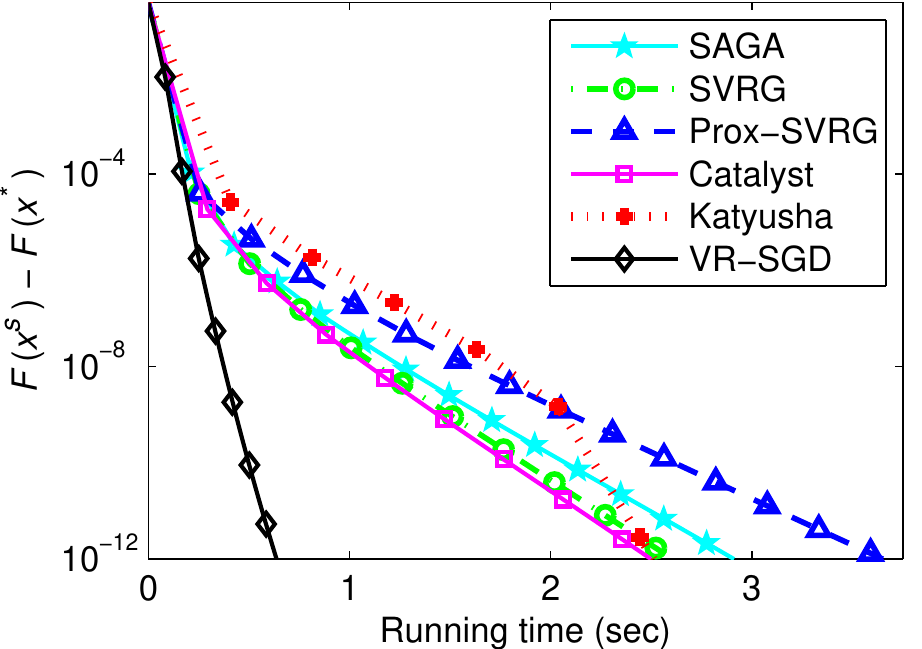}\:\includegraphics[width=0.245\columnwidth]{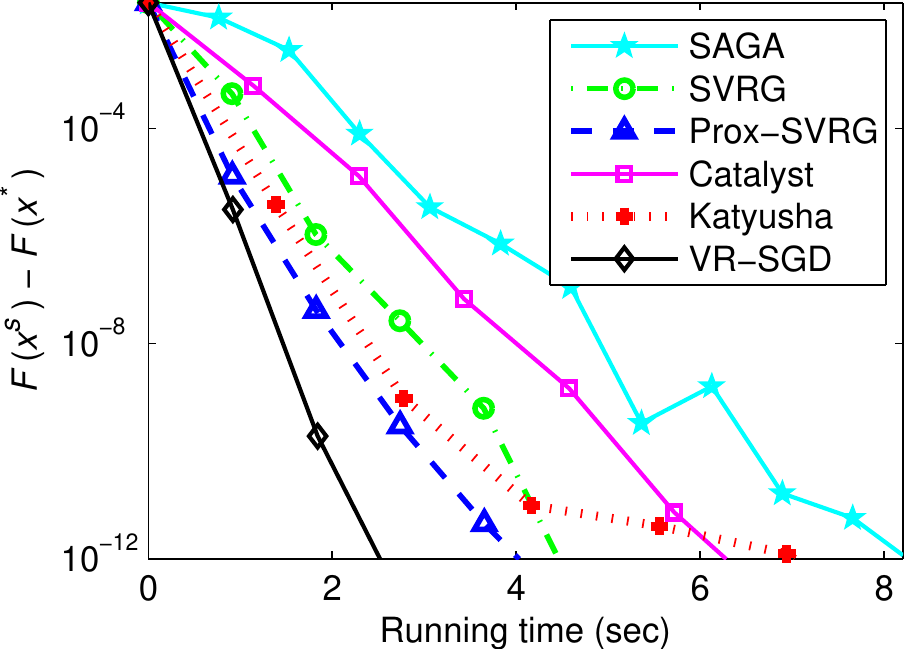}\:\includegraphics[width=0.245\columnwidth]{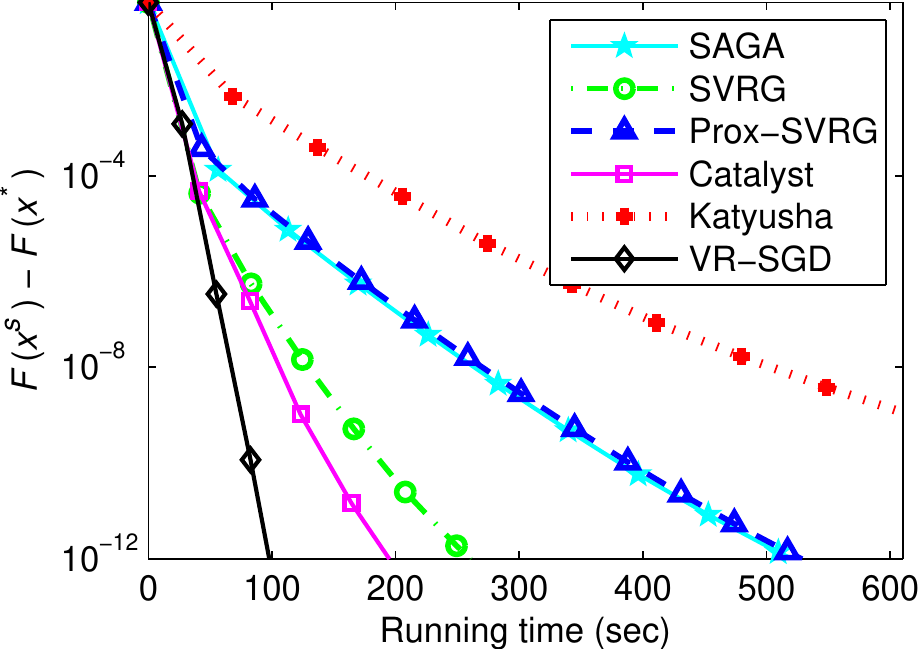}}\:\subfigure[$\lambda_{1}\!=\!10^{-5}$ and $\lambda_{2}\!=\!10^{-4}$]{\includegraphics[width=0.245\columnwidth]{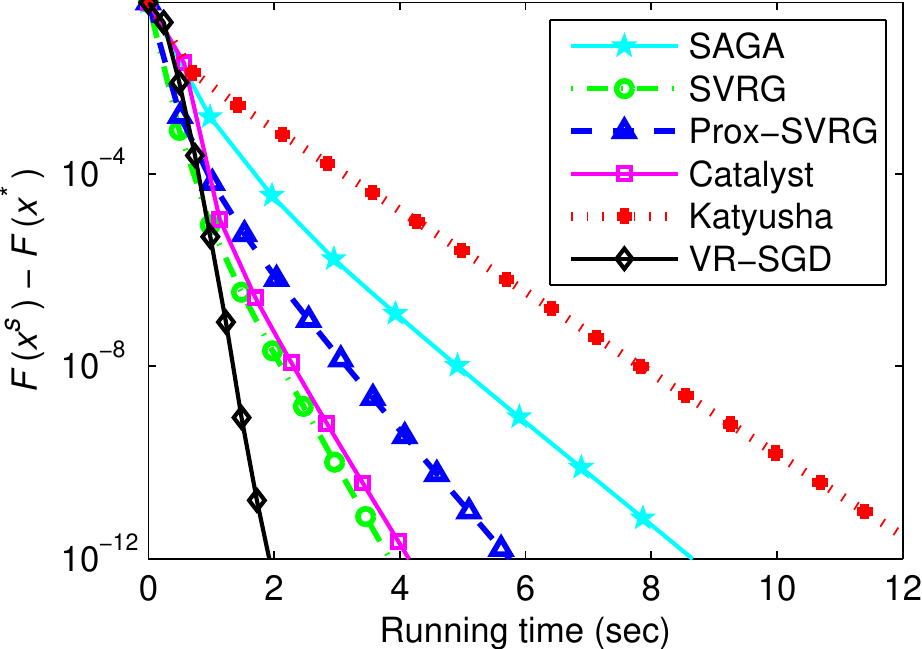}}
\vspace{1.6mm}

\includegraphics[width=0.245\columnwidth]{Fig617}
\includegraphics[width=0.245\columnwidth]{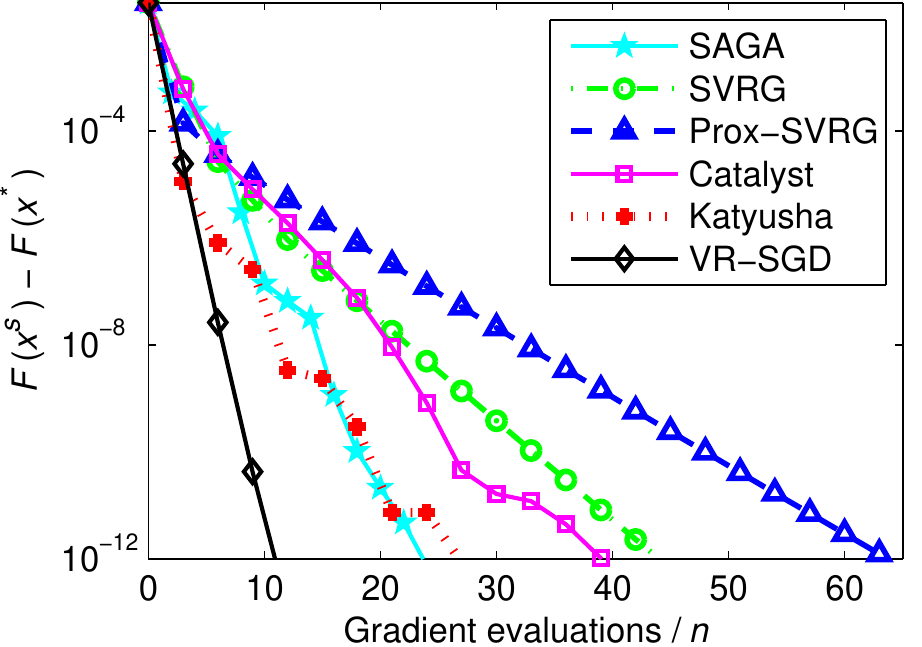}
\includegraphics[width=0.245\columnwidth]{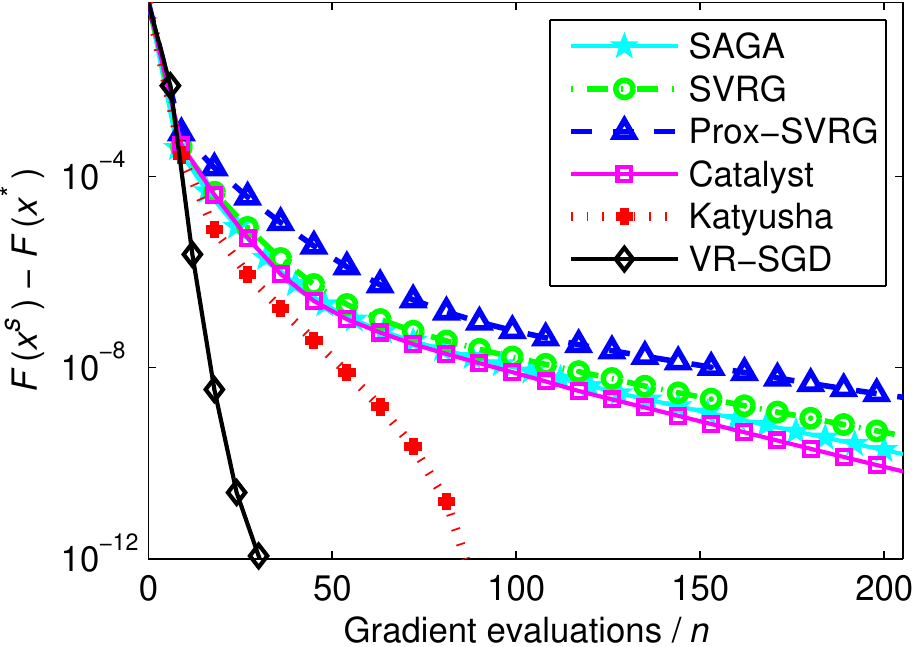}
\includegraphics[width=0.245\columnwidth]{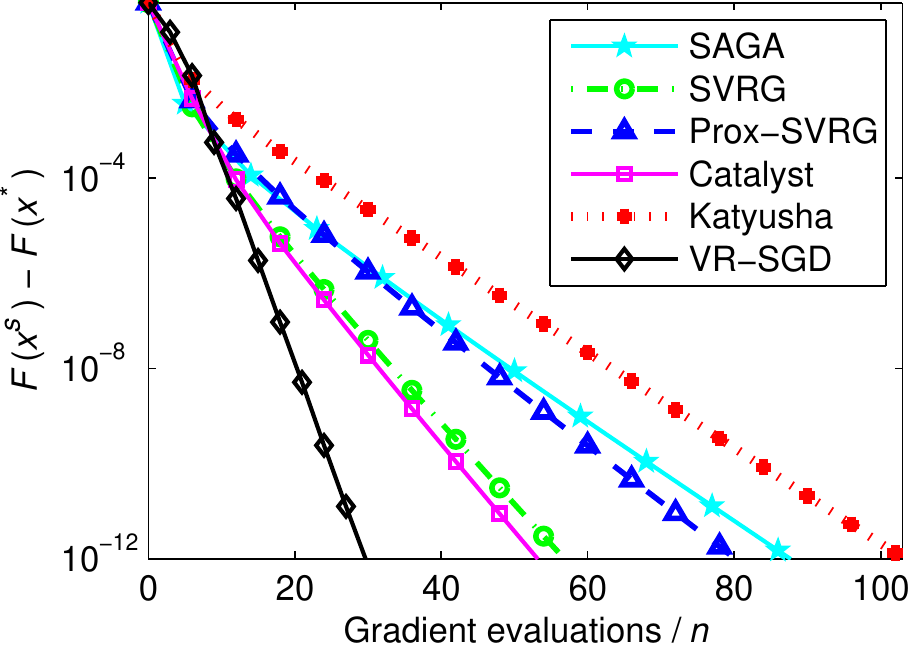}

\subfigure[$\lambda_{1}=10^{-6}$ \;and\; $\lambda_{2}=10^{-5}$]{\includegraphics[width=0.245\columnwidth]{Fig618}\:\includegraphics[width=0.245\columnwidth]{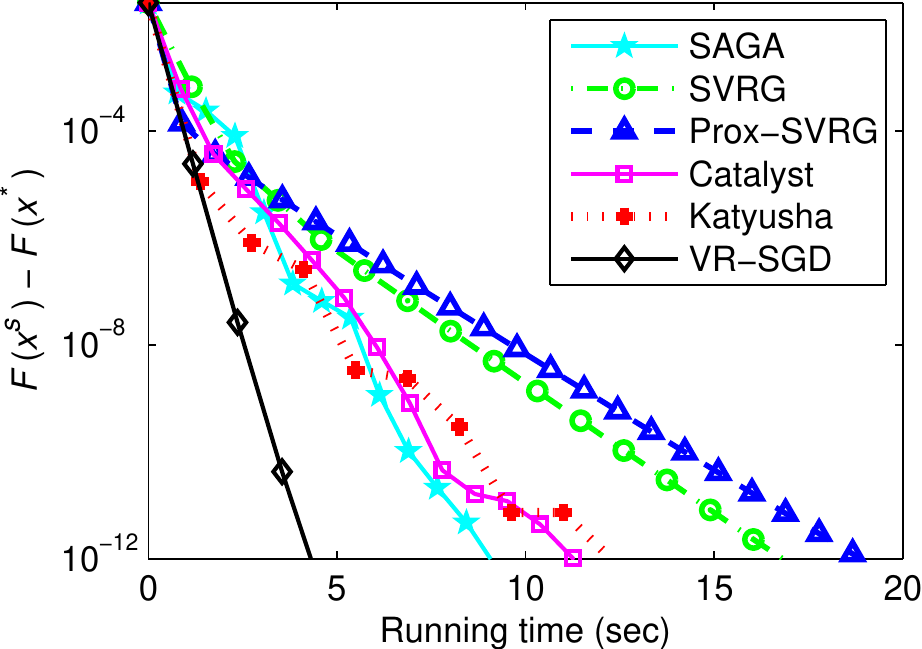}\:\includegraphics[width=0.245\columnwidth]{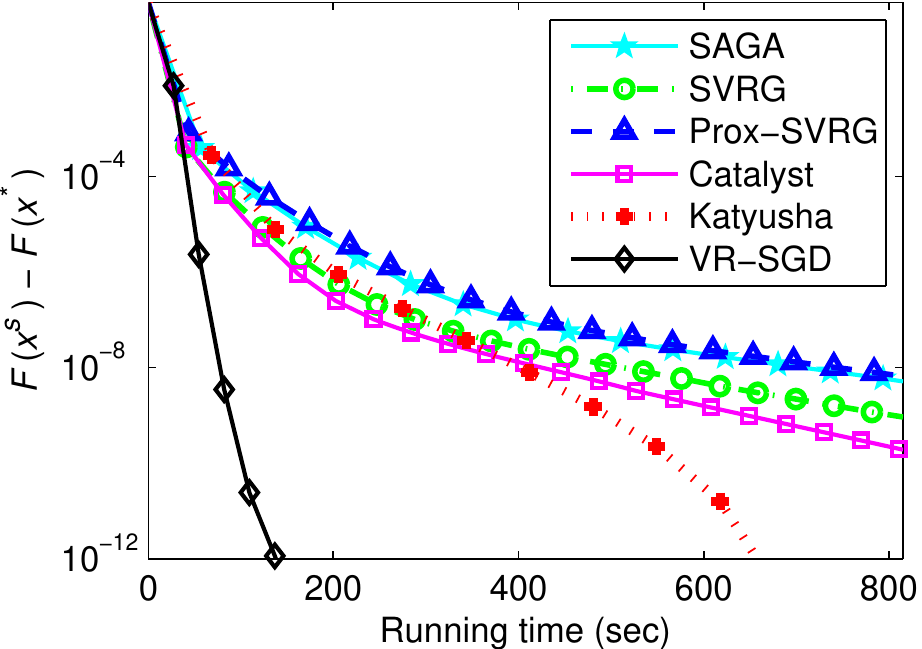}}\:\subfigure[$\lambda_{1}\!=\!10^{-5}$ and $\lambda_{2}\!=\!10^{-5}$]{\includegraphics[width=0.245\columnwidth]{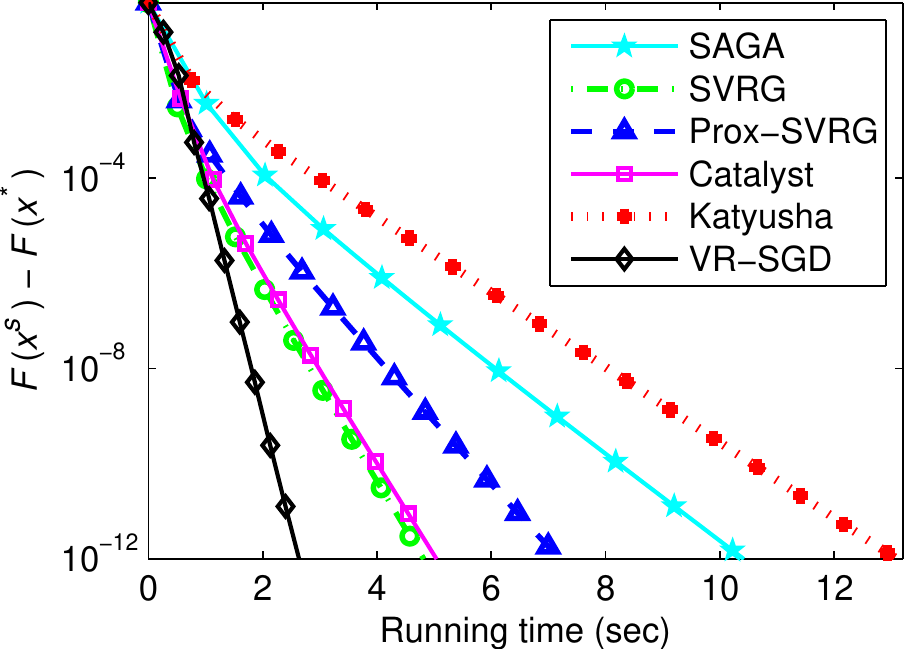}}
\caption{Comparison of SAGA~\cite{defazio:saga}, SVRG, Prox-SVRG~\cite{xiao:prox-svrg}, Catalyst~\cite{lin:vrsg}, Katyusha~\cite{zhu:Katyusha}, and our VR-SGD method for solving elastic net regularized logistic regression problems on the four data sets: Adult (the first column), Covtype (the sconced column), Epsilon (the third column), and RCV1 (the last column). In each plot, the vertical axis shows the objective value minus the minimum, and the horizontal axis is the number of effective passes (top) or running time (bottom).}
\label{figs19}
\end{figure}

\begin{figure}[th]
\centering
\includegraphics[width=0.245\columnwidth]{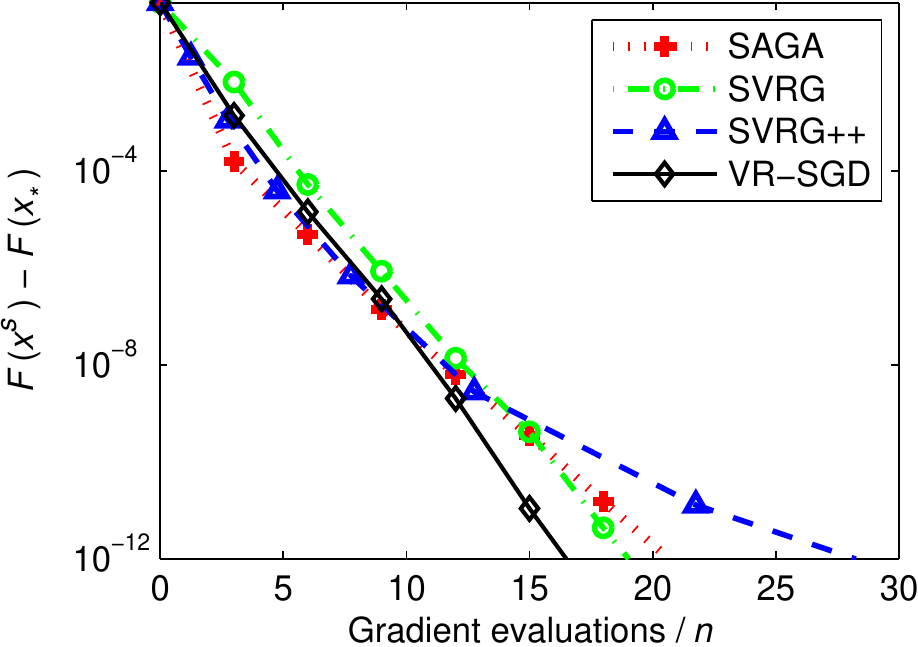}
\includegraphics[width=0.245\columnwidth]{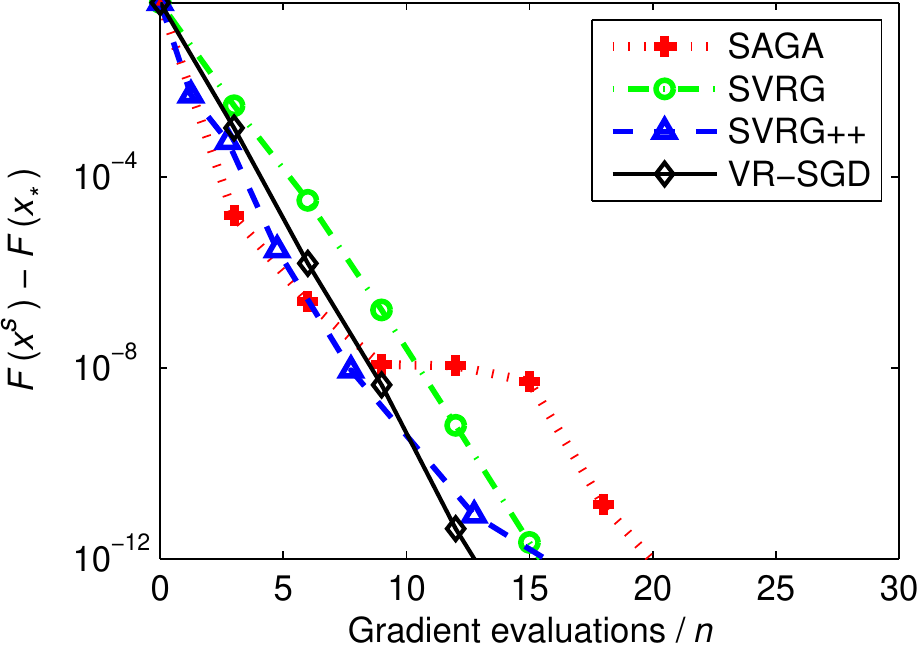}
\includegraphics[width=0.245\columnwidth]{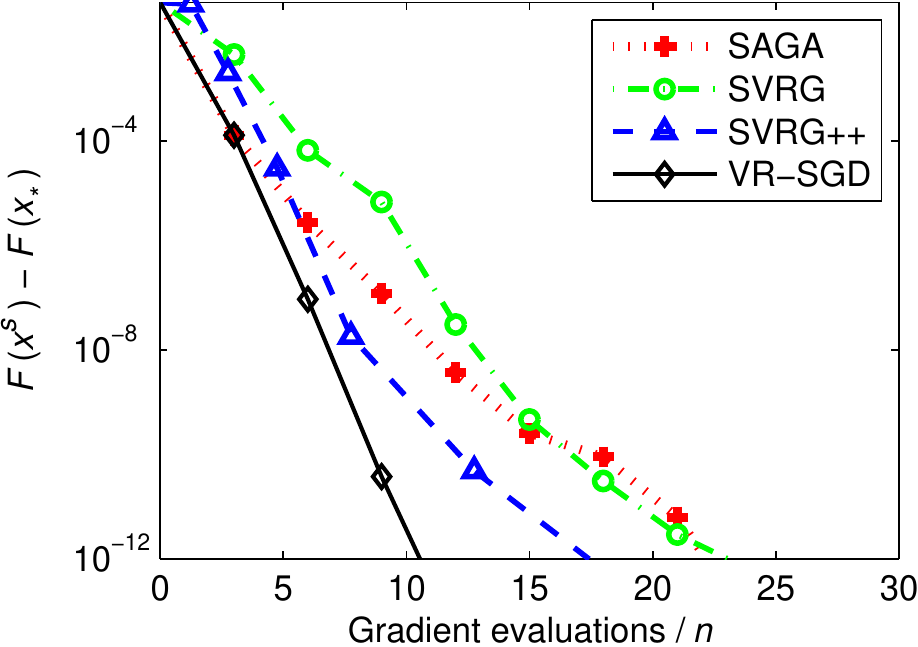}
\includegraphics[width=0.245\columnwidth]{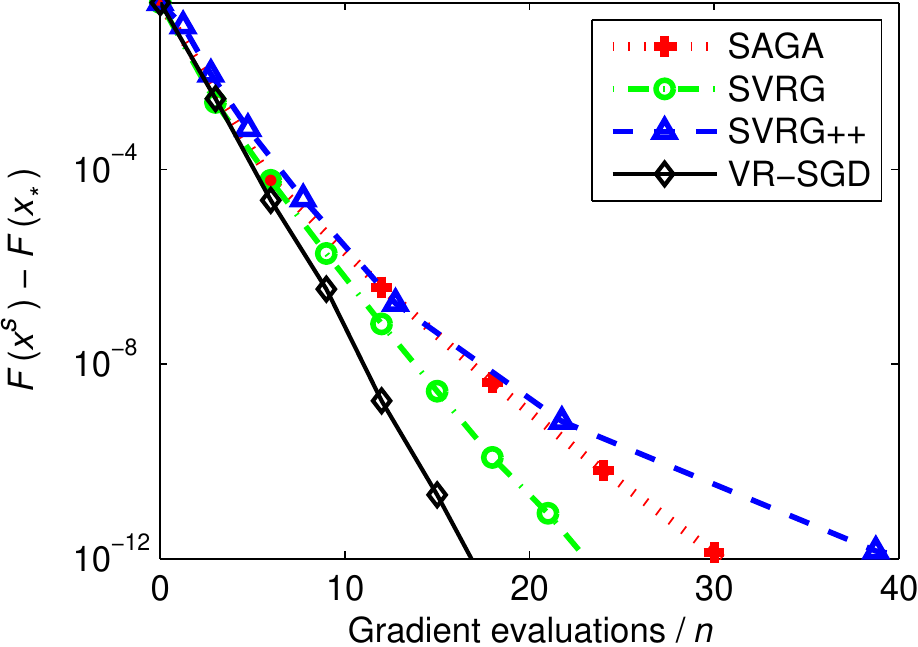}
\vspace{0.3mm}

\includegraphics[width=0.245\columnwidth]{Fig52}
\includegraphics[width=0.245\columnwidth]{Fig55}
\includegraphics[width=0.245\columnwidth]{Fig58}
\includegraphics[width=0.245\columnwidth]{Fig61}
\vspace{0.1mm}

\subfigure[Adult]{\includegraphics[width=0.245\columnwidth]{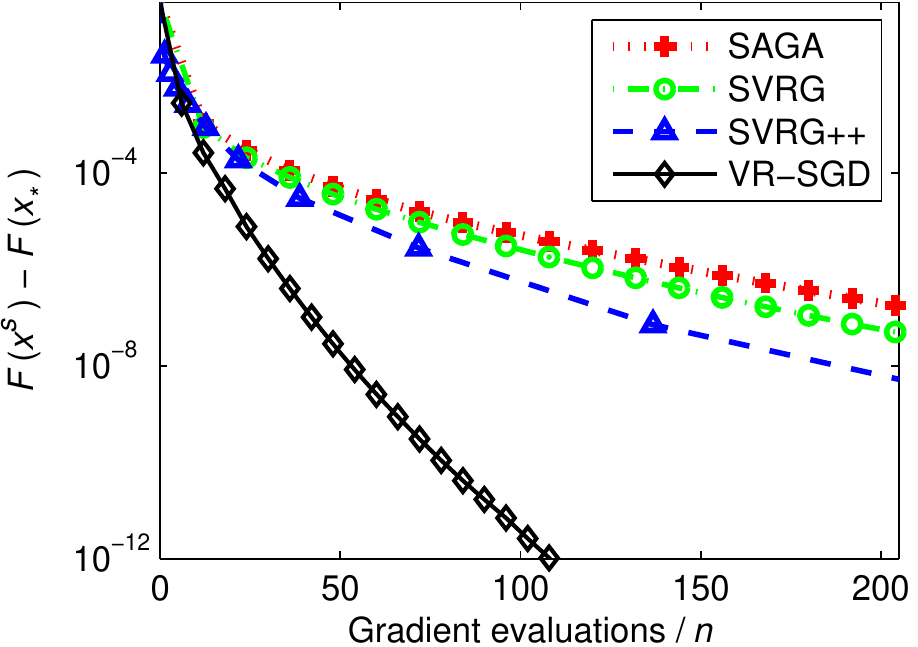}}
\subfigure[MNIST]{\includegraphics[width=0.245\columnwidth]{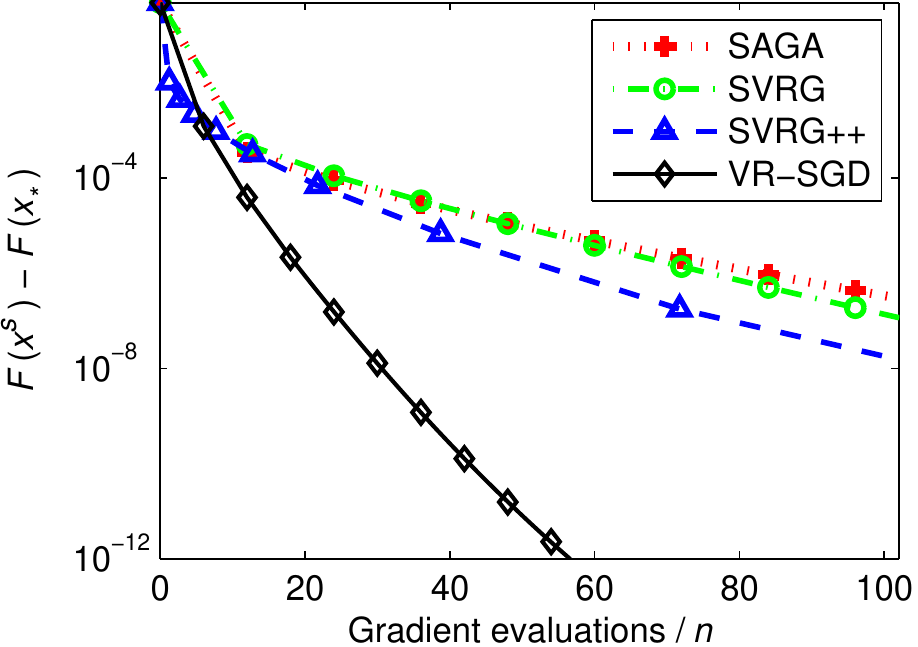}}
\subfigure[Covtype]{\includegraphics[width=0.245\columnwidth]{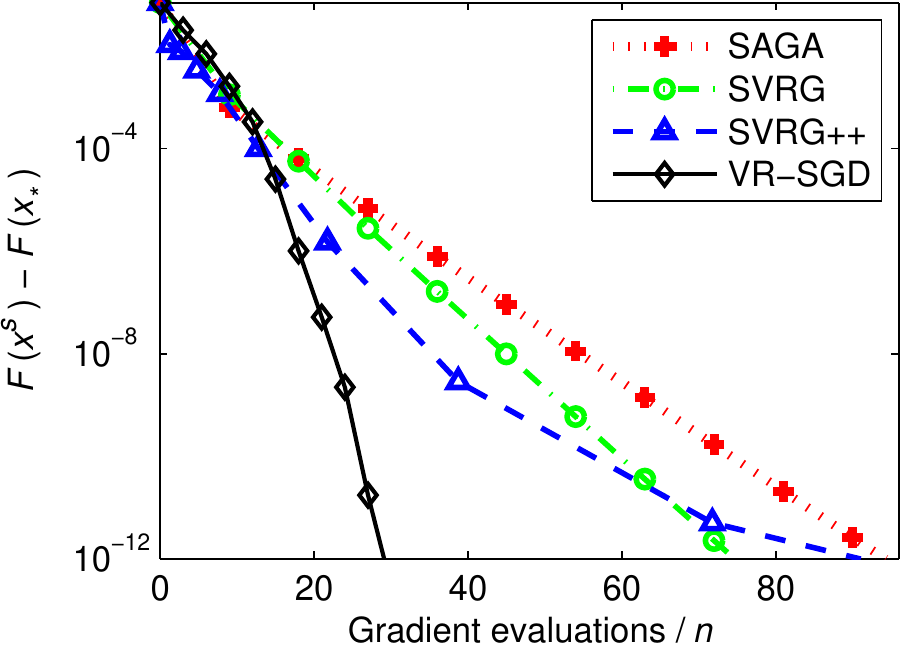}}
\subfigure[RCV1]{\includegraphics[width=0.245\columnwidth]{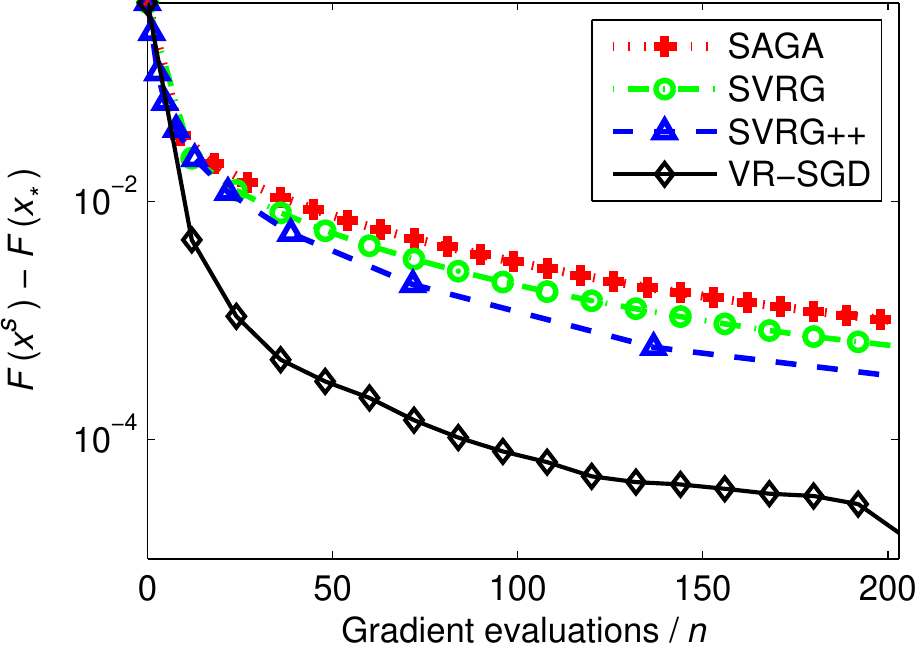}}
\caption{Comparison of SAGA~\cite{reddi:saga}, SVRG~\cite{zhu:vrnc}, SVRG++~\cite{zhu:univr}, and our VR-SGD method for solving non-convex ERM problems with sigmoid loss on the four data sets: $\lambda=10^{-4}$ (the first row), $\lambda=10^{-5}$ (the sconced row), and $\lambda=10^{-6}$ (the last row). Note that $x_{*}$ denotes the best solution obtained by running all those methods for a large number of iterations and multiple random initializations.}
\label{figs20}
\end{figure}

\begin{figure}[th]
\centering
\includegraphics[width=0.245\columnwidth]{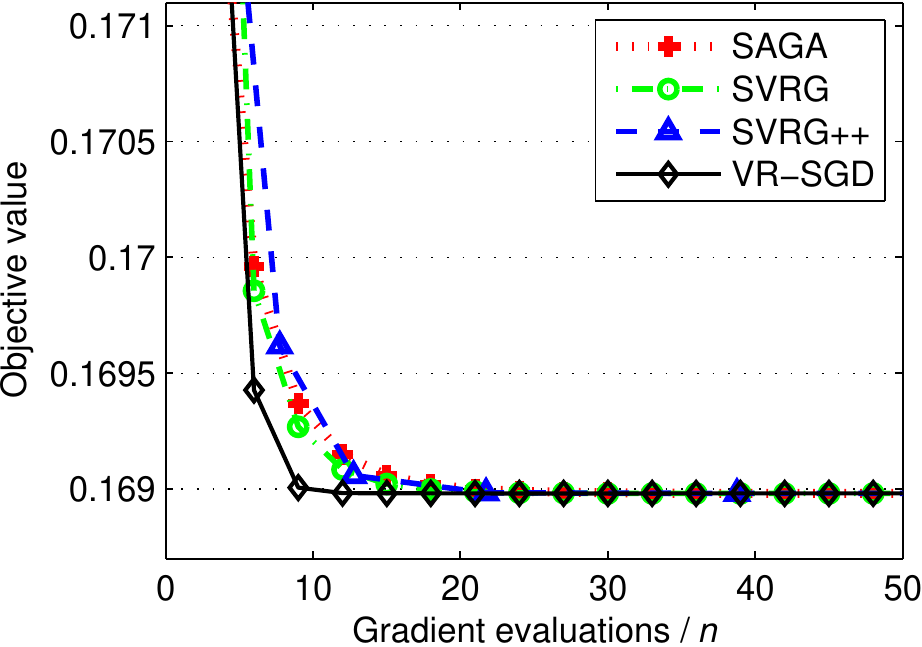}
\includegraphics[width=0.245\columnwidth]{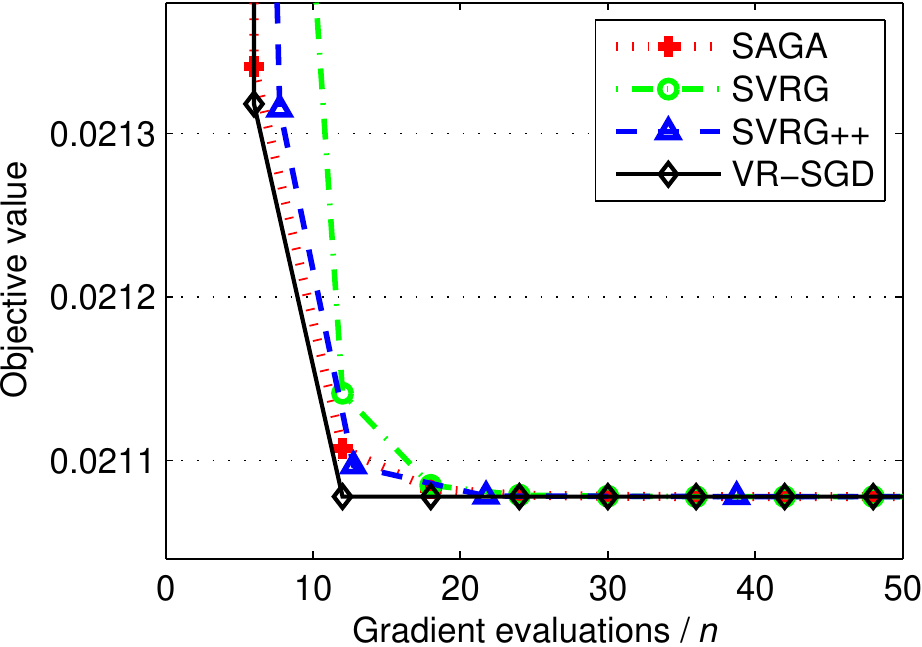}
\includegraphics[width=0.245\columnwidth]{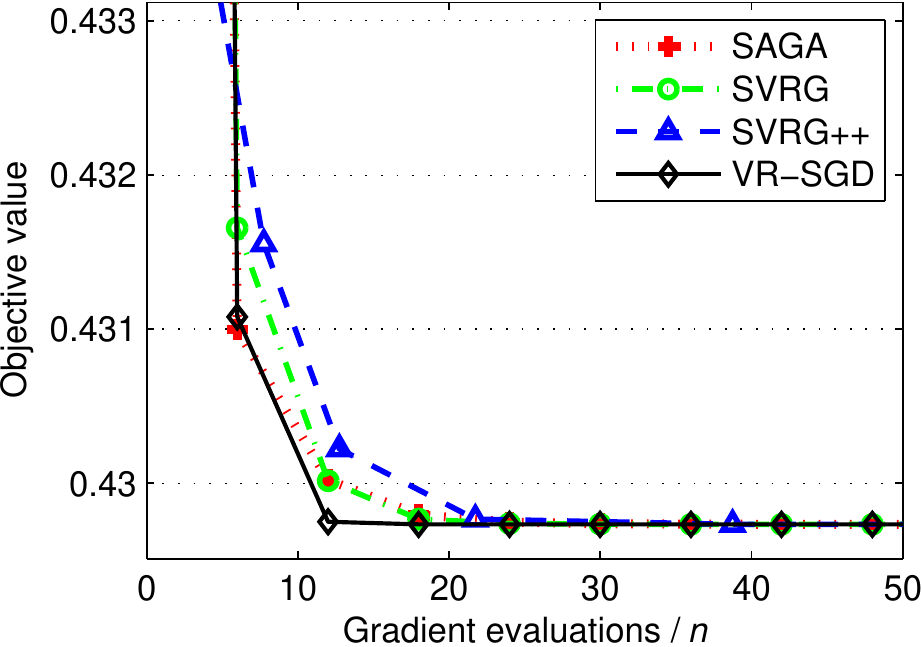}
\includegraphics[width=0.245\columnwidth]{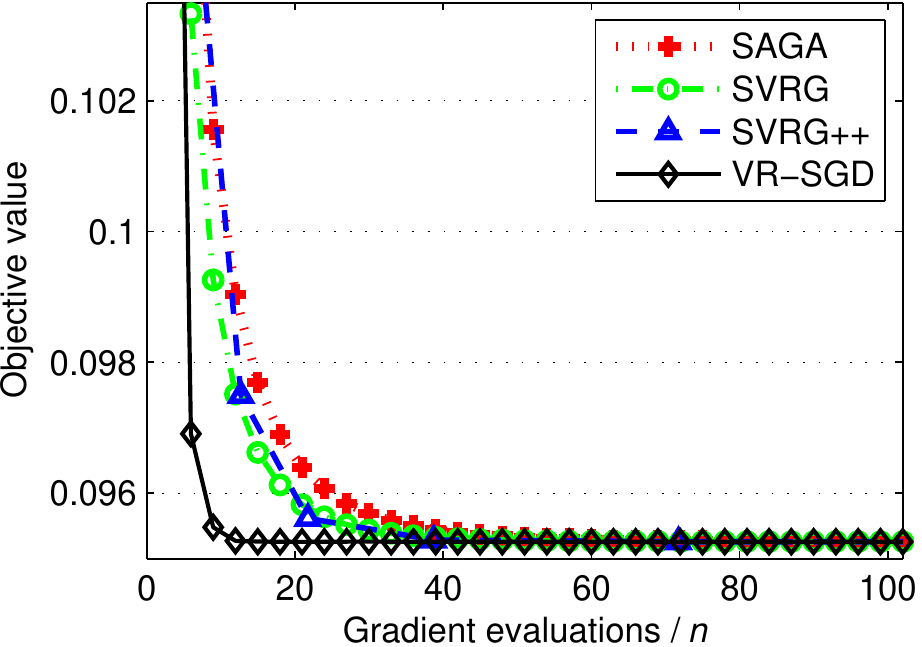}
\vspace{0.1mm}

\includegraphics[width=0.245\columnwidth]{Fig72}
\includegraphics[width=0.245\columnwidth]{Fig75}
\includegraphics[width=0.245\columnwidth]{Fig78}
\includegraphics[width=0.245\columnwidth]{Fig81}
\vspace{0.1mm}

\subfigure[Adult]{\includegraphics[width=0.245\columnwidth]{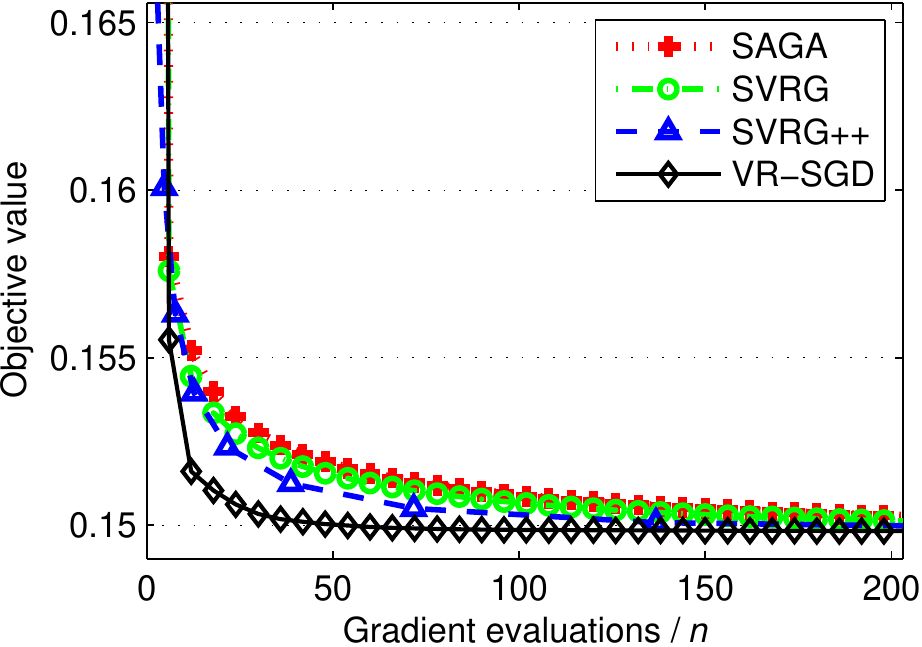}}
\subfigure[MNIST]{\includegraphics[width=0.245\columnwidth]{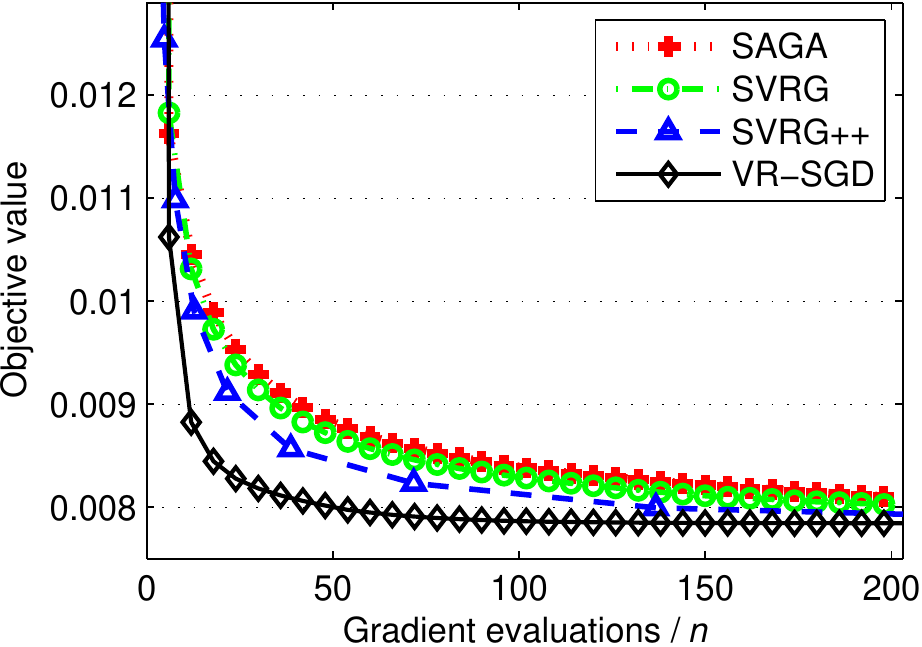}}
\subfigure[Covtype]{\includegraphics[width=0.245\columnwidth]{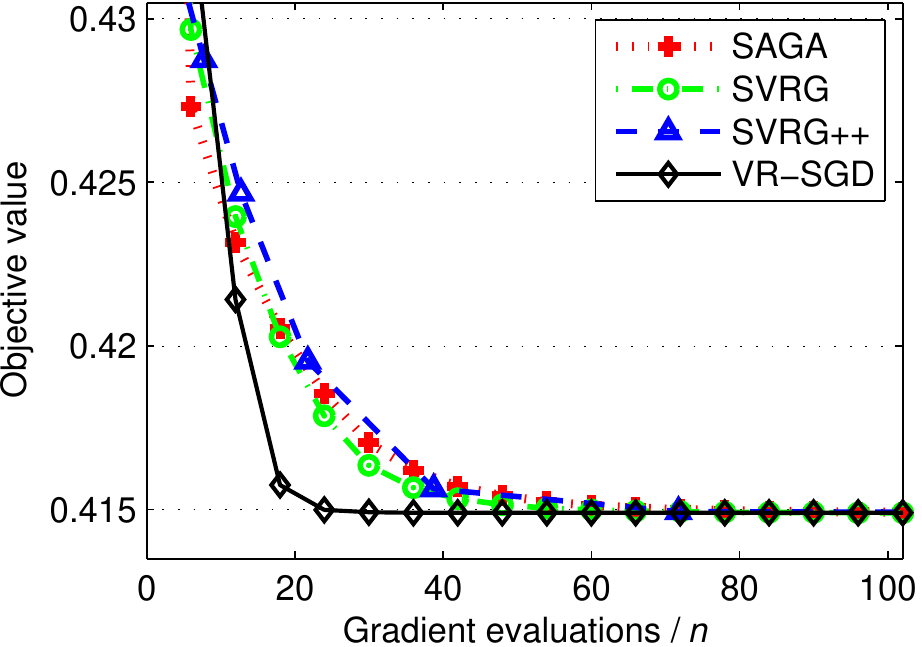}}
\subfigure[RCV1]{\includegraphics[width=0.245\columnwidth]{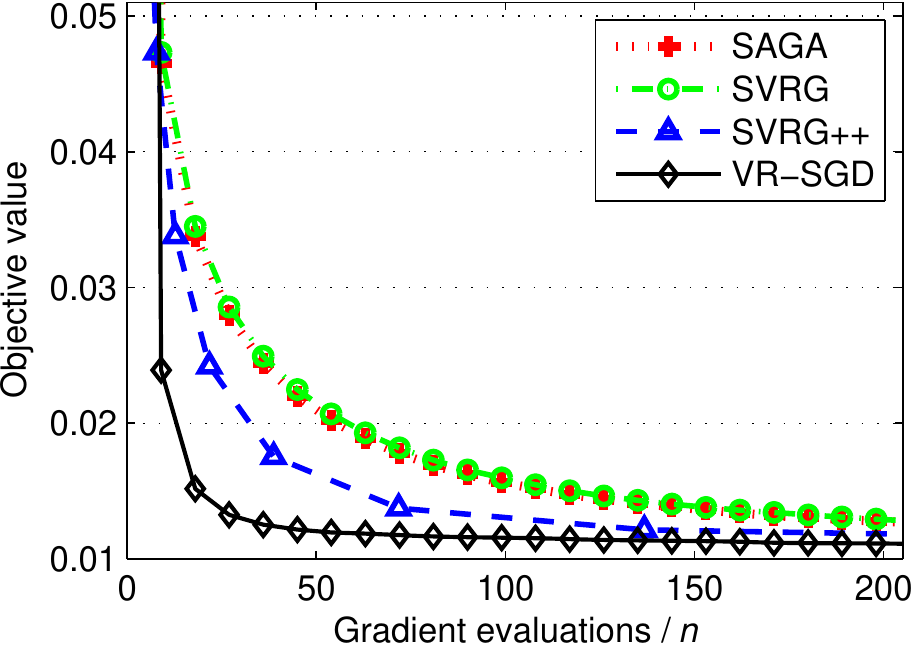}}
\caption{Comparison of SAGA~\cite{reddi:saga}, SVRG~\cite{zhu:vrnc}, SVRG++~\cite{zhu:univr}, and our VR-SGD method for solving non-convex ERM problems with sigmoid loss on the four data sets: $\lambda=10^{-5}$ (the first row), $\lambda=10^{-6}$ (the sconced row), and $\lambda=10^{-7}$ (the last row).}
\label{figs21}
\end{figure}

%

\end{document}